\documentclass{article}
\usepackage{spconf}
\usepackage{graphicx}
\usepackage[linesnumbered,ruled,vlined]{algorithm2e}
\usepackage{amsmath}
\usepackage{amssymb}
\usepackage{multirow}
\usepackage{lipsum}
\usepackage{booktabs}
\usepackage{makecell}
\usepackage{multicol}
\usepackage{hyperref}
\usepackage{microtype}
\usepackage{xcolor}

\title{Spherical Dense Text-to-Image Synthesis}
\name{Timon Winter$^{\star}$, Stanislav Frolov$^{\star,\dagger}$, Brian Bernhard Moser$^{\star,\dagger}$, Andreas Dengel$^{\star,\dagger}$
}
\address{
$^{\dagger}$RPTU Kaiserslautern-Landau, Germany \\
$^{\star}$German Research Center for Artificial Intelligence, Germany 
}

\begin{document}
%
\maketitle

\begin{abstract}
Recent advancements in text-to-image (T2I) have improved synthesis results, but challenges remain in layout control and generating omnidirectional panoramic images.
Dense T2I (DT2I) and spherical T2I (ST2I) models address these issues, but so far no unified approach exists.
Trivial approaches, like prompting a DT2I model to generate panoramas can not generate proper spherical distortions and seamless transitions at the borders.
Our work shows that spherical dense text-to-image (SDT2I) can be achieved by integrating training-free DT2I approaches into finetuned panorama models.
Specifically, we propose MultiStitchDiffusion (MSTD) and MultiPanFusion (MPF) by integrating MultiDiffusion into StitchDiffusion and PanFusion, respectively.
Since no benchmark for SDT2I exists, we further construct Dense-Synthetic-View (DSynView), a new synthetic dataset containing spherical layouts to evaluate our models.
Our results show that MSTD outperforms MPF across image quality as well as prompt- and layout adherence.
MultiPanFusion generates more diverse images but struggles to synthesize flawless foreground objects.
We propose bootstrap-coupling and turning off equirectangular perspective-projection attention in the foreground as an improvement of MPF.
\href{https://github.com/sdt2i/spherical-dense-text-to-image}{\underline{Link to code}}
\end{abstract}


\section{Introduction}

\begin{figure}[ht]
  \centering
  \includegraphics[width=1\linewidth]{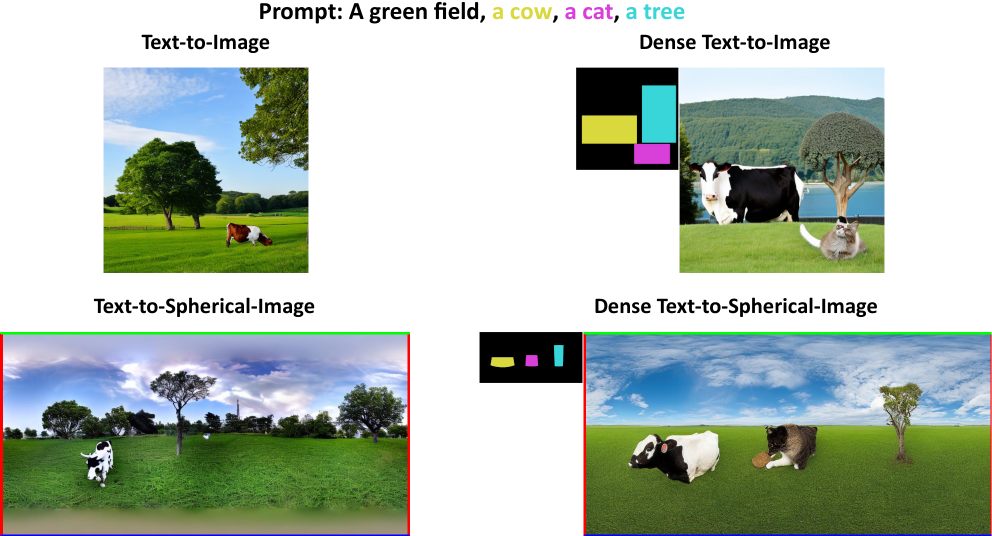}
  \caption{
    Task comparison:
    Traditional text-to-image (top-left) generates images based on a single global prompt.
    Dense text-to-image (top-right) introduces masks to control the spatial layout.
    Spherical text-to-image (bottom left) synthesizes 360x180-degree panoramas with seamless transition over the border and distortions at the poles.
    In this work, we integrate both approaches to enable controllable spherical dense text-to-image synthesis (bottom-right).
    }
  \label{fig:task_overview}
\end{figure}

Text-to-image (T2I) has gained significant traction recently, with advancements like StableDiffusion \cite{rombach2022highresolution} driving progress \cite{zhang2024texttoimagediffusionmodelsgenerative}. However, user demands have also increased with longer, more complex prompts. Traditional models often fail to handle detailed prompts, misaligning object properties, positional relations, or ignoring details entirely.
To address these issues, dense text-to-image (DT2I) models \cite{frolov2022dt2i, bartal2023multidiffusion, kim2023dense} were introduced, tackling these challenges directly. Many approaches enhance user control by allowing explicit layout input, such as masks \cite{Avrahami_2023, couairon2023zeroshot}. \autoref{fig:task_overview} illustrates examples of various DT2I models.

At the same time, users also often seek to customize the image dimensions. Cylindrical panoramas must seamlessly transition between left and right borders, while spherical panoramas require equirectangular projection (ERP) to produce distortion-free 360x180-degree images when mapped onto a sphere. Spherical text-to-image (ST2I) models \cite{tang2023mvdiffusionenablingholisticmultiview, wang2023customizing360degreepanoramastexttoimage, zhang2024tamingstablediffusiontext} address these needs.

Despite these advancements, no existing model combines DT2I and ST2I capabilities.
This work addresses this gap by proposing the first spherical dense text-to-image (SDT2I), see \autoref{fig:task_overview}.  
We hypothesize that SDT2I can be achieved by integrating a pre-trained spherical image model with a DT2I approach as a ``plug \& play'' component. Using StitchDiffusion \cite{wang2023customizing360degreepanoramastexttoimage} and PanFusion \cite{zhang2024tamingstablediffusiontext} as our spherical image backbones and MultiDiffusion (MD) \cite{bartal2023multidiffusion} as the DT2I plugin, we extend these models to support layout guidance through free-hand masks.  
Additionally, we improve condition adherence and image quality by introducing a bootstrap coupling mechanism.
To evaluate our models, we create Dense-Synthetic-View (DSynView), a new synthetic dataset to evaluate our models by generating a diverse set of spherical panoramic images with layout annotations.
In summary, our contributions are as follows:
(1) We integrate MultiDiffusion (MD) \cite{bartal2023multidiffusion} into StitchDiffusion \cite{wang2023customizing360degreepanoramastexttoimage} and PanFusion \cite{zhang2024tamingstablediffusiontext}, creating MultiStitchDiffusion (MSTD) and MultiPanFusion (MPF), the first SDT2I models.  
(2) These models are evaluated on image quality, diversity, and text/layout adherence, showing MSTD matches its baselines.  
(3) We introduce Dense-Synthetic-View (DSynView), a synthetic dataset with diverse masks and prompts for SDT2I.  
(4) A naive MPF implementation underperforms MSTD, so we enhance MPF with bootstrap-coupling and improved foreground attention for higher quality.

\section{Background}
\label{sec:related}

\subsection{Diffusion Models}
\label{sec:diffusion}

The core idea of diffusion models \cite{sohldickstein2015deep} is to split the generative process into $T$ timesteps and learn to generate images through a sequence of operations:
\begin{align}
  I_T, I_{T-1}, ..., I_0 \quad \text{s.t.} \quad I_{t-1} = \theta(I_t|y),
  \label{eq:diffusion}
\end{align}

\noindent
where $y$ is the input embedding, $I_t$ is the output image at timestep $t$ and $\theta$ is the diffusion model.
Training and sampling is organized in a \textit{forward-process} and a \textit{reverse-process}.
The forward process is only applied at training time.
Taking one training image as data $x$, Gaussian noise $\epsilon$ is added to the image according to the noise level of each timestep.
The model is optimized to predict a noise residual $\epsilon_\theta$ which can be subtracted from the diffused image to match the denoised ground truth data.
In the reverse process, $I_T$ is initialized as a Gaussian noise distribution.
The predictive model is now used to predict a noise residual at timestep $t$ using prompt $y$.
This residual is subtracted from $I_t$, receiving $I_{t-1}$, which serves as an input at timestep $t-1$.
Thus, the initial noise is gradually removed from the image until after $T$ steps, a clean output image remains.

\subsection{Latent Diffusion Models (LDM)}
\label{sec:stablediffusion}

LDMs \cite{rombach2022highresolution} address the high computational cost of traditional diffusion models by operating in a reduced latent space instead of pixel space. Using a pre-trained variational auto-encoder \cite{kingma2022autoencodingvariationalbayes}, an encoder $\mathcal{E}$ maps an image $x$ to a latent representation $z = \mathcal{E}(x)$, while a decoder $\mathcal{D}$ reconstructs it: $\Tilde{x} = \mathcal{D}(z) \approx x$.
LDMs also enhance sample quality by abstracting irrelevant details in the latent space. A cross-attention mechanisms \cite{vaswani2023attention} links input tokens to image patches to enable text-to-image synthesis.
Stable Diffusion (SD), built on the LDM framework, has become the most widely used open-source model for image generation.

\subsection{Dense Text-to-Image (DT2I)}
\label{sec:dt2i}

In DT2I, a global prompt is extended with tuples of local prompts and layouts that define object placement, typically as rectangular boxes or free-hand masks. MultiDiffusion (MD) \cite{bartal2023multidiffusion} introduced a training-free framework for DT2I by fusing multiple diffusion paths.
Mores specifically, an image is generated in a sliding window fashion, ensuring consistency across overlapping regions by averaging the denoising predictions.
For panoramas, overlaps are defined by sliding windows, while for DT2I, they arise from user-defined binary masks.
To align synthesized objects with input masks, the authors proposed a bootstrapping phase, where the model focuses on mask areas during early diffusion steps, directing attention away from the background. This improves object placement accuracy and adherence to layouts.

\subsection{Spherical Text-to-Image (ST2I)}
\label{sec:3_2_1}

As the demand for immersive experiences grows, moving from traditional 2D image synthesis to spherical image generation has become increasingly important.
StitchDiffusion \cite{wang2023customizing360degreepanoramastexttoimage} builds on MD \cite{bartal2023multidiffusion} to create seamless cylindrical panoramas.
While MD smooths horizontal transitions, it lacks a full cyclic connection. StitchDiffusion introduces a ``stitch block'' that merges the left and right image edges during denoising, ensuring a continuous panorama.
For spherical panoramas, ERP distortions are added by fine-tuning SD with LoRA \cite{hu2021lora} layers on 120 panoramic images.
PanFusion \cite{zhang2024tamingstablediffusiontext} addresses ST2I generation by splitting the process into two branches: a panorama branch for global coherence and a perspective branch leveraging SD's generative strength for individual images. 
Communication between branches is enabled by equirectangular-perspective projection attention (EPPA), enhanced with spherical positional encoding and an attention mask to align feature maps from both branches.
PanFusion adapts SD to 512x1024 resolution using LoRA \cite{hu2021lora} layers and trains on the Matterport3D dataset \cite{chang2017matterport3dlearningrgbddata}. Circular padding ensures loop consistency, while perspective views are generated via icosahedron tessellation.

\section{Methods}
\label{sec:methods}

Our goal is to develop the first spherical dense text-to-image (SDT2I) models by integrating training-free DT2I
approaches into finetuned panorama models.
To that end, we propose MultiStitchDiffusion (MSTD) and MultiPanFusion (MPF) in the next sections.

\subsection{MultiStitchDiffusion (MSTD)}
\label{sec:mstd}

As our first method, we integrate StitchDiffusion \cite{wang2023customizing360degreepanoramastexttoimage} with the region-based variant of MD \cite{bartal2023multidiffusion}.
The input includes a global text prompt and $N$ masks with corresponding local prompts. 
Following MD, we preprocess the masks and extend the edges cyclically to handle the stitching steps. StitchDiffusion’s LoRA is activated by the trigger word ``360-degree panoramic image'', with an option to toggle it separately for background and foreground objects.
Inference is split into multiple processes, each handling one mask and prompt.
After every denoising step, latents are merged as described in \autoref{sec:dt2i}.

\subsection{MultiPanFusion (MPF)}
\label{sec:mpf}

As our second method, we use PanFusion \cite{zhang2024tamingstablediffusiontext} as our base model and integrate MD into the inference process.
The modification follows a similar approach to our StitchDiffusion integration, with computations on latents repeated multiple times for each denoising path. To maintain alignment, masks are rotated alongside latents during each denoising step. After denoising, latents are merged as in MD.
Since PanFusion contains two branches (see \autoref{sec:dt2i}), there are multiple possible implementations.
Initially, we integrate MD into the panorama branch only. To improve consistency during denoising, we extend MD to the perspective branch by projecting masks from ERP to perspective format and applying MD similarly to the panorama branch.

\subsubsection{Eliminating Bootstrapping-Flaws}
During evaluation, we observed artifacts like blurring and pixelation around the foreground objects. These issues likely stem from the naive application of MD’s bootstrapping phase, which assumes that mask backgrounds can be replaced with constant colors before denoising, as these areas are ignored during latent merging.  
In PanFusion’s dual-branch approach, this assumption fails. Using different background colors for the panorama and perspective branches or disabling MD in one branch creates conflicts in the EPPA module, leading to unpredictable results. Additionally, the Gaussian-smoothed EPP attention mask can spread background colors around objects over multiple denoising steps.  
To address this, we propose two improvements:  
(1) Disabling the EPPA module for foreground objects during the bootstrapping phase to eliminate conflicts.  
(2) Coupling bootstrapping backgrounds across branches by assigning the same random color to both panorama and perspective branches and ensuring consistent background colors for objects at each timestep.

\section{Experiments}
\label{sec:experiments}

\subsection{Dataset \& Metrics }
\label{sec:dataset}

Since no benchmark exists for spherical dense text-to-image synthesis, we create Dense-Synthetic-View (DSynView), a new synthetic dataset containing spherical layouts to evaluate our models.
In DSynView, we combined three text-prompts used for conditioning the background with two different sets of up to three fitting foreground prompts associated with self-created masks.

\begin{table}[ht]
    \small
    \centering
    \begin{tabular}{@{}clll@{}}
    \toprule
    foreground prompts & small & medium & large\\
    \midrule
    \multirow{2}{*}{\makecell{\textcolor{red}{table}, \textcolor{red}{bed},\\ \textcolor{green}{cow}, \textcolor{green}{sheep},\\ \textcolor{blue}{bus}, \textcolor{blue}{car}}}
    & \includegraphics[scale=0.15]{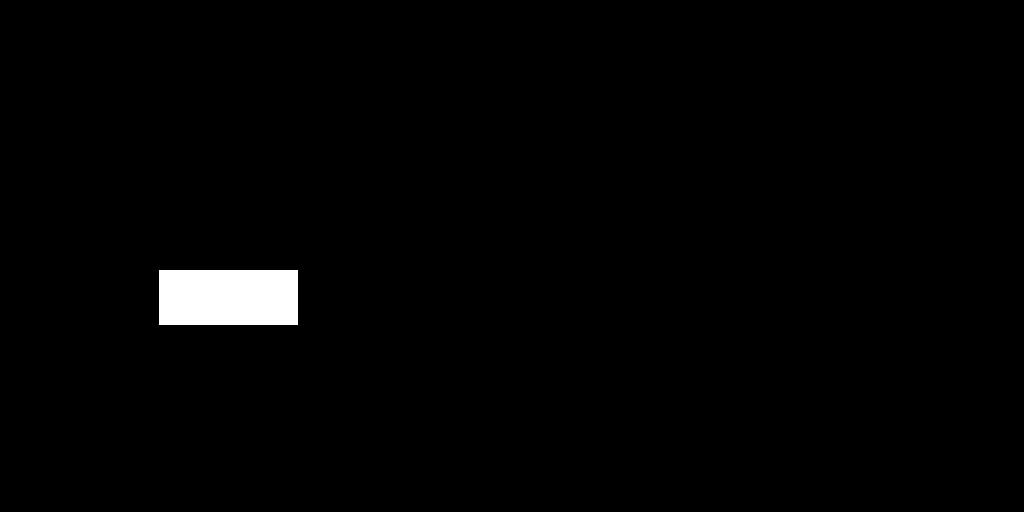} 
    & \includegraphics[scale=0.15]{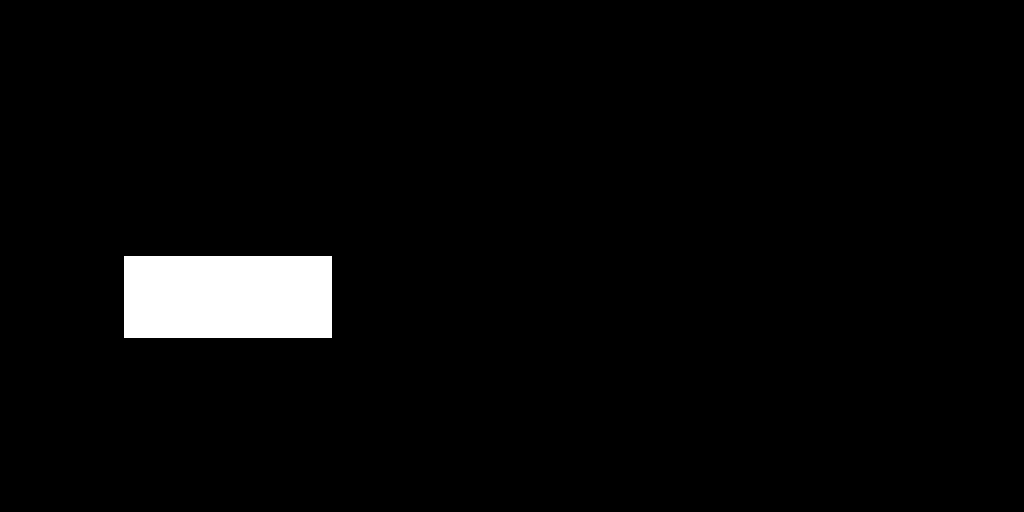} 
    & \includegraphics[scale=0.15]{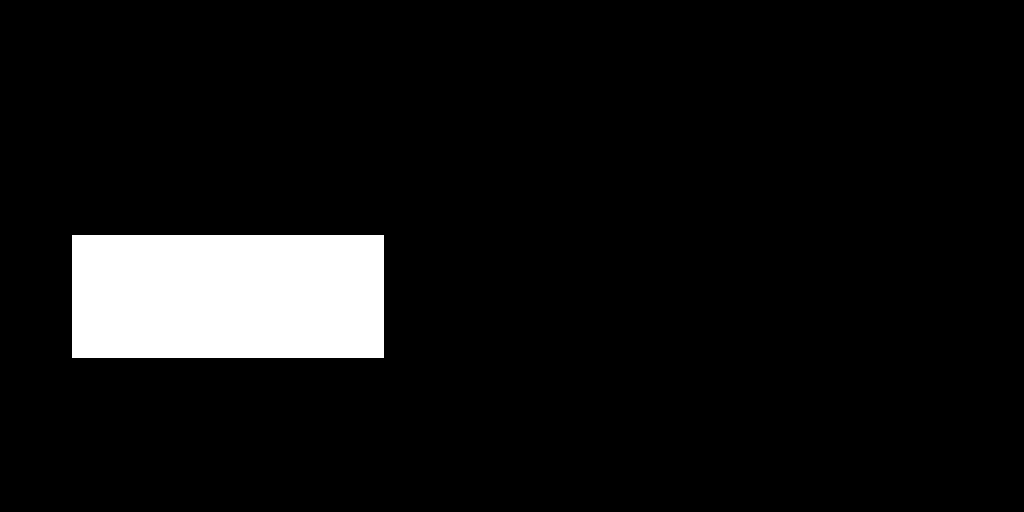} \\
    & \includegraphics[scale=0.03625]{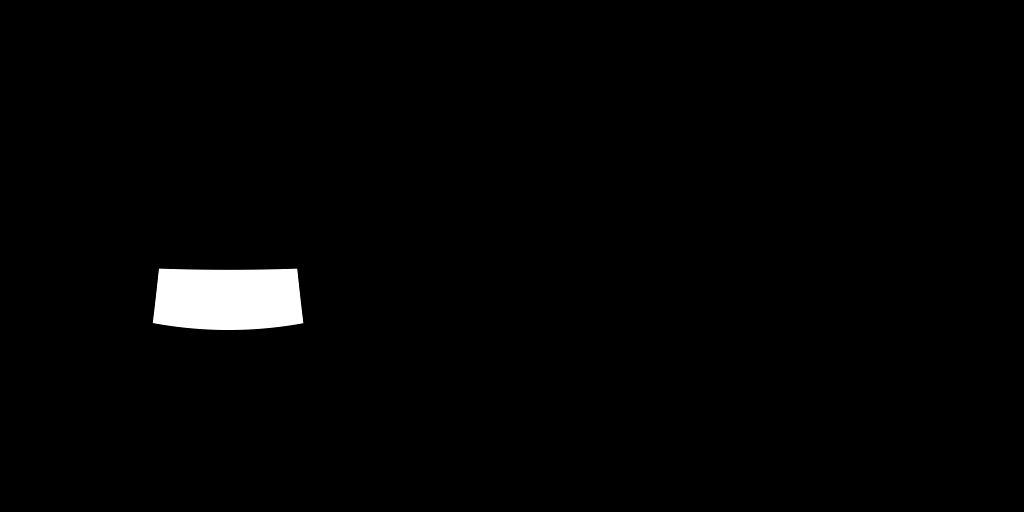} 
    & \includegraphics[scale=0.03625]{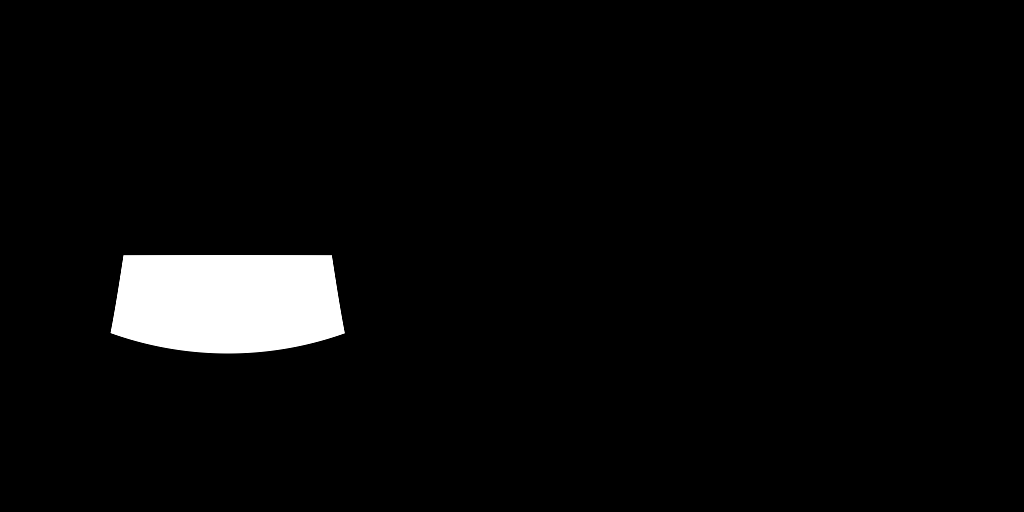} 
    & \includegraphics[scale=0.03625]{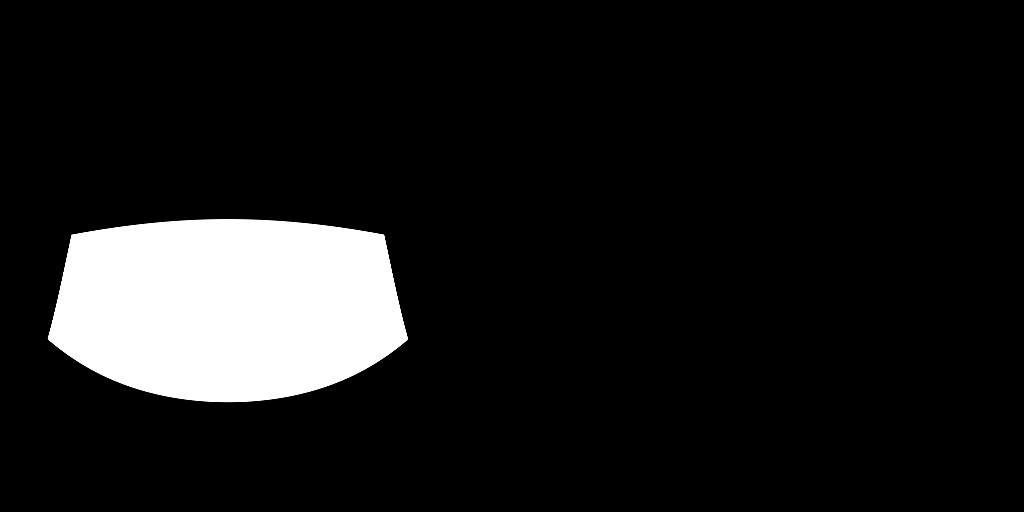} \\
    \midrule
    \multirow{2}{*}{\makecell{\textcolor{red}{television}, \textcolor{red}{potted plant},\\ \textcolor{green}{cat}, \textcolor{green}{pond},\\ \textcolor{blue}{sign}, \textcolor{blue}{bicycle}}} 
    & \includegraphics[scale=0.15]{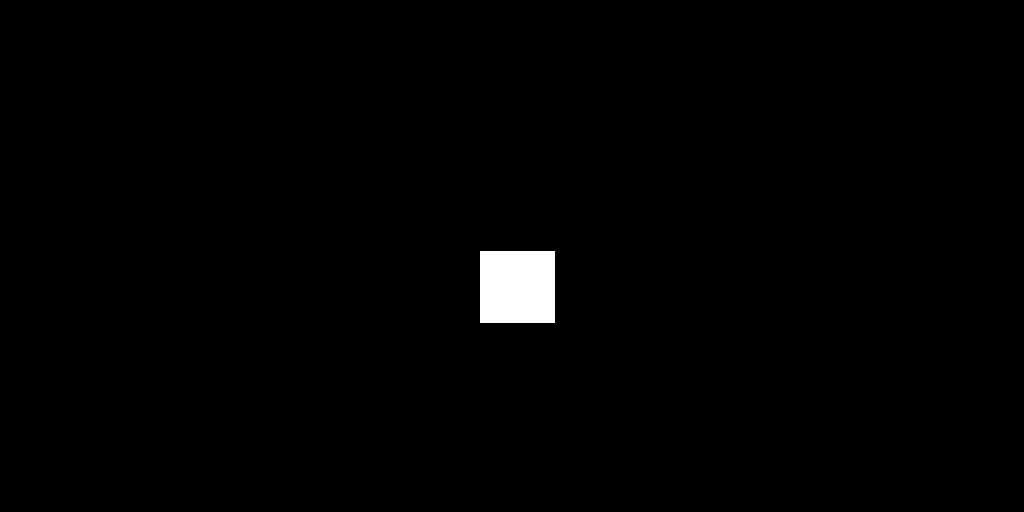} 
    & \includegraphics[scale=0.15]{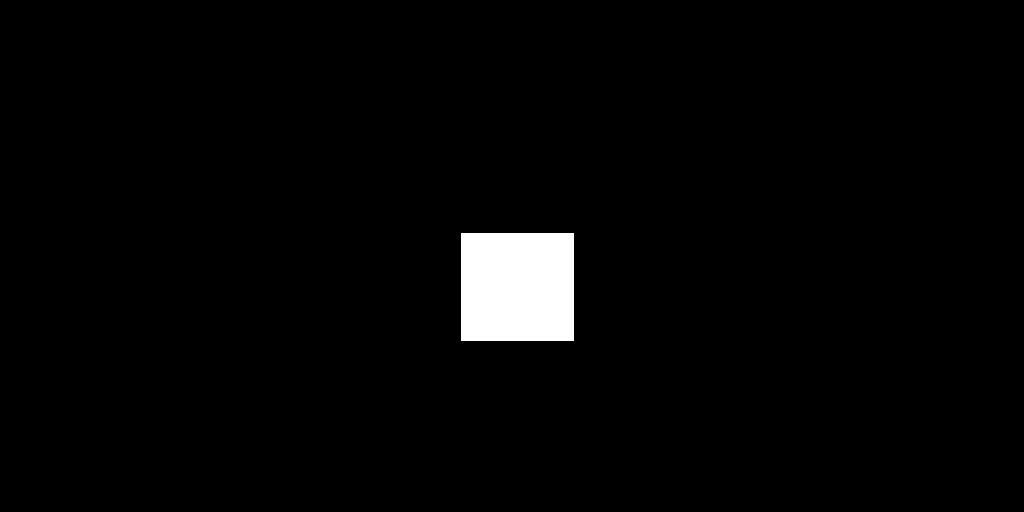} 
    & \includegraphics[scale=0.15]{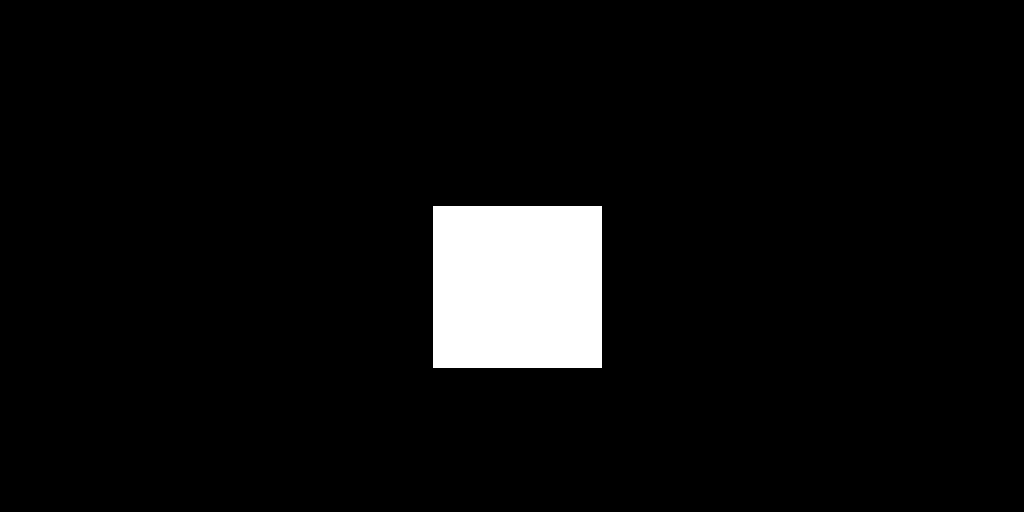} \\
    & \includegraphics[scale=0.03625]{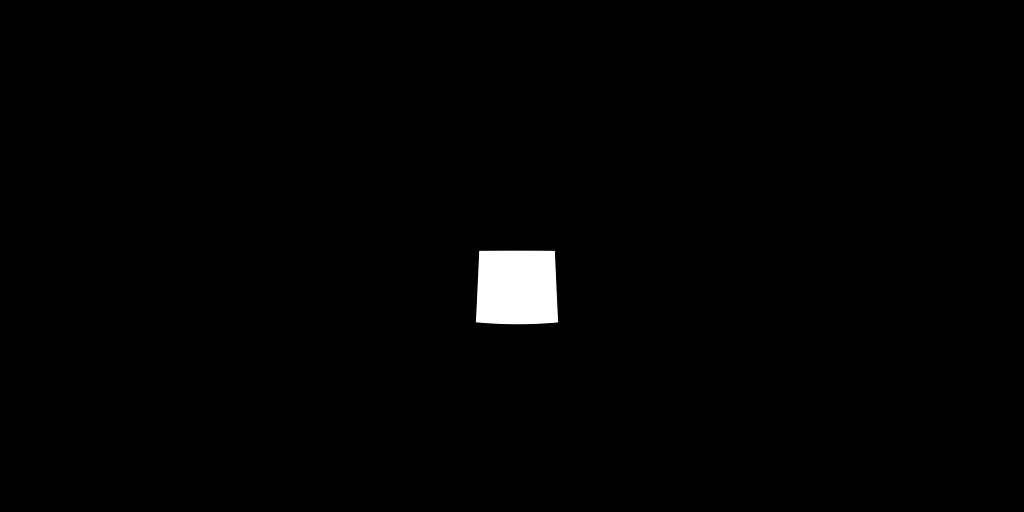} 
    & \includegraphics[scale=0.03625]{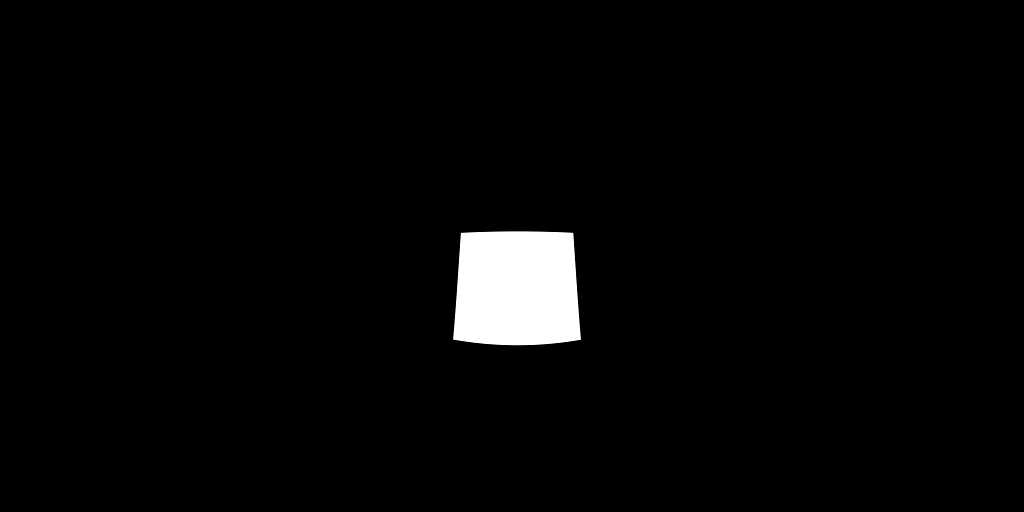} 
    & \includegraphics[scale=0.03625]{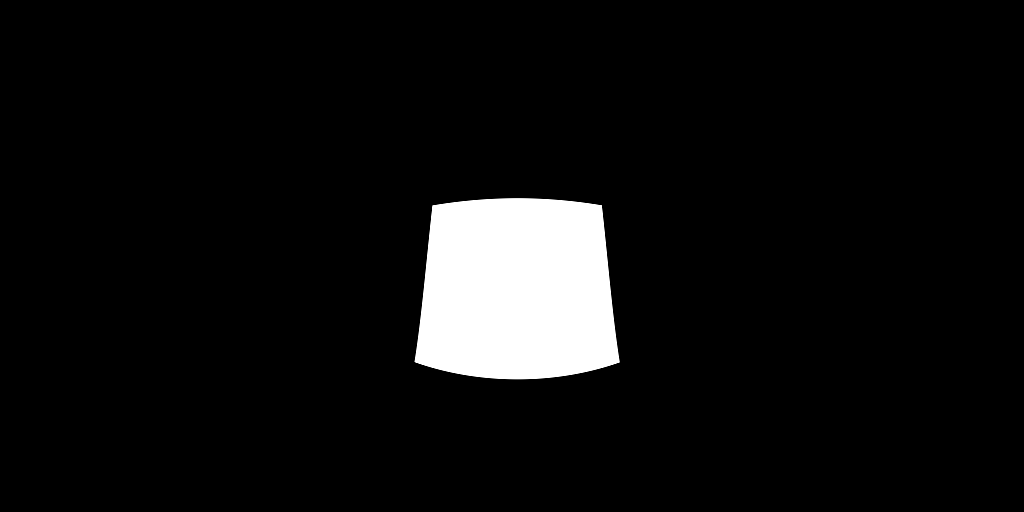} \\
    \midrule
    \multirow{2}{*}{\makecell{\textcolor{red}{wardrobe}, \textcolor{red}{door},\\ \textcolor{green}{tree}, \textcolor{green}{windmill},\\ \textcolor{blue}{building}, \textcolor{blue}{traffic light}}} 
    & \includegraphics[scale=0.15]{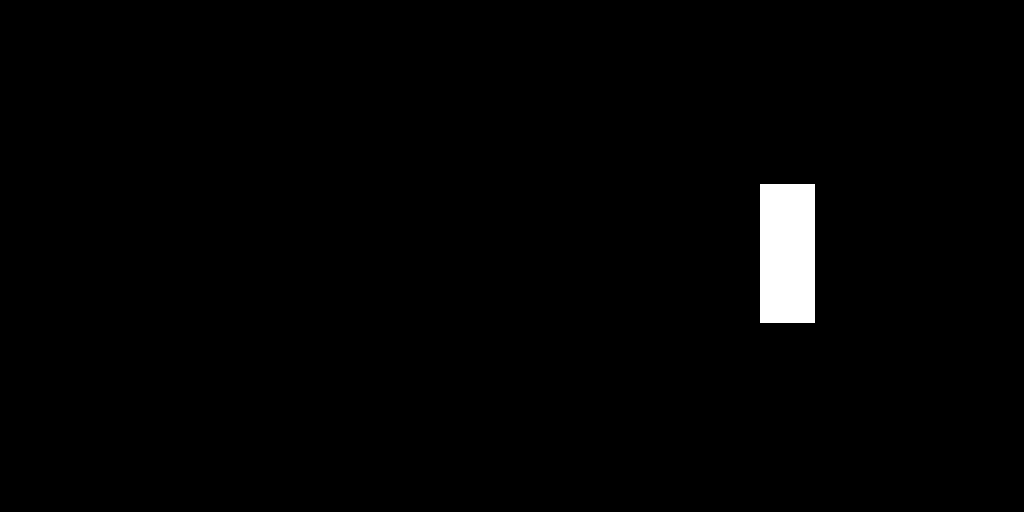} 
    & \includegraphics[scale=0.15]{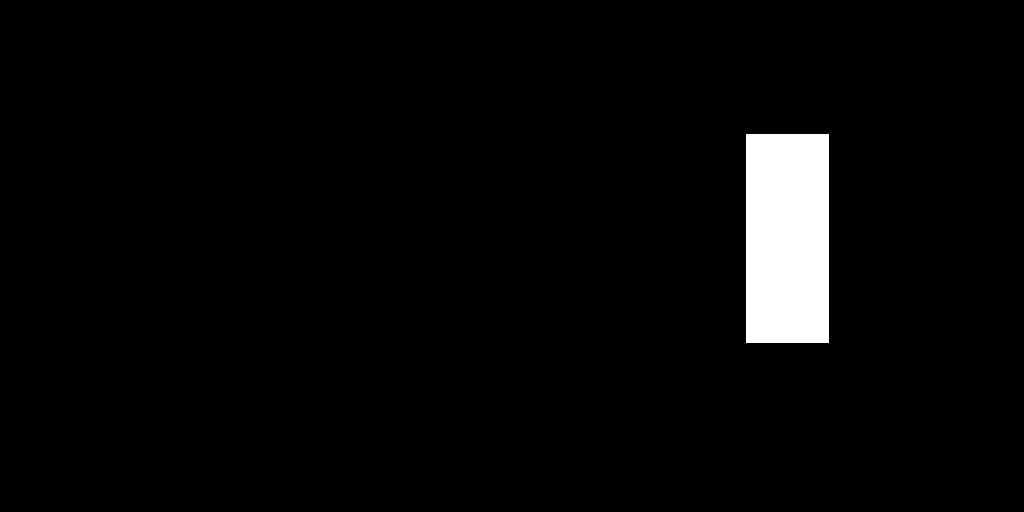} 
    & \includegraphics[scale=0.15]{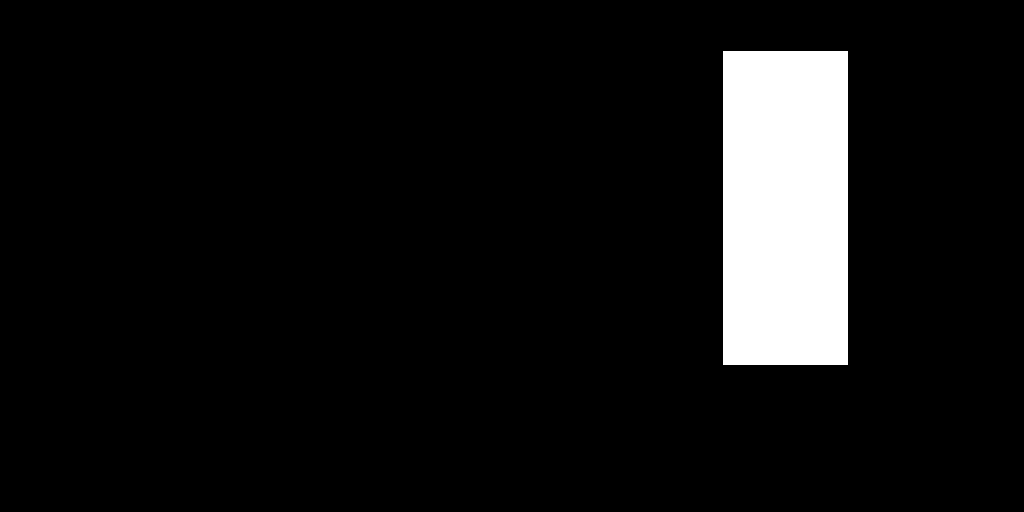} \\
    & \includegraphics[scale=0.03625]{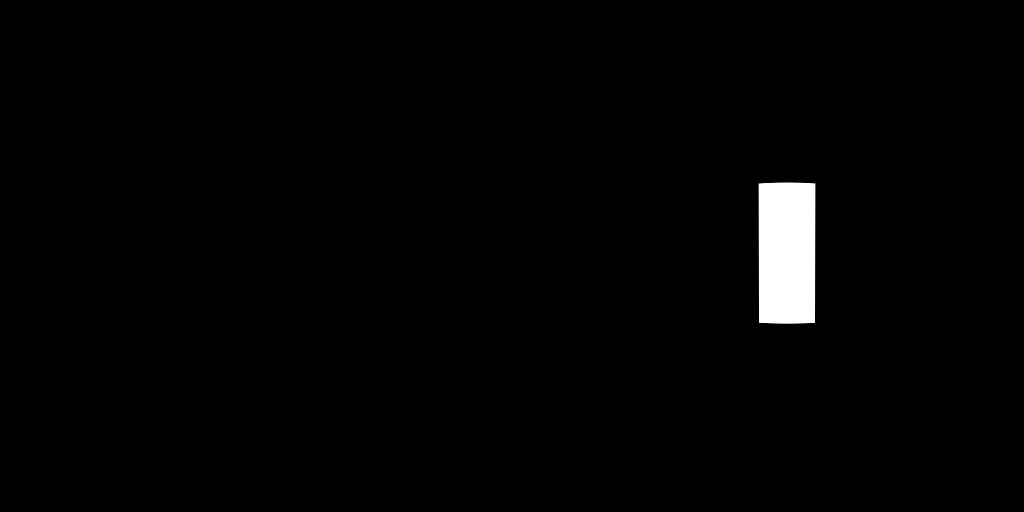} 
    & \includegraphics[scale=0.03625]{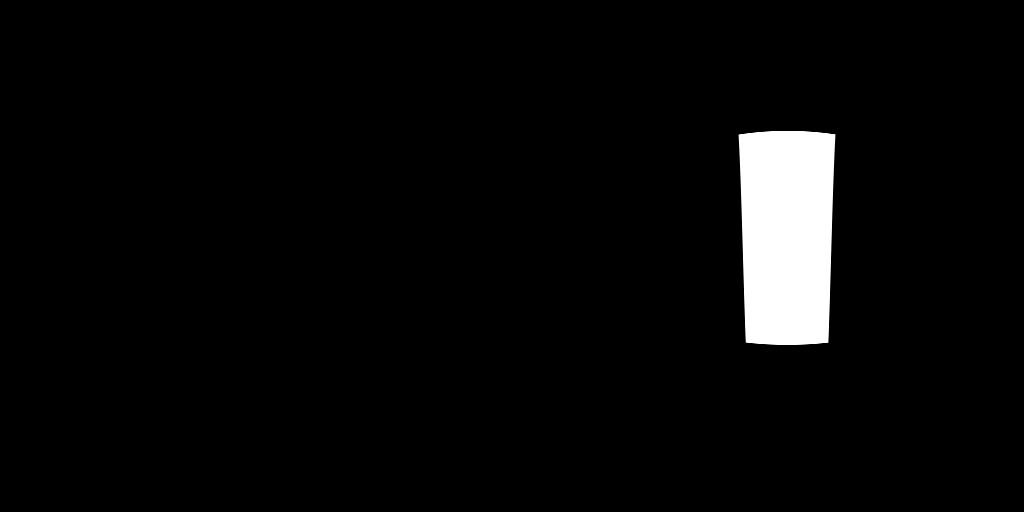} 
    & \includegraphics[scale=0.03625]{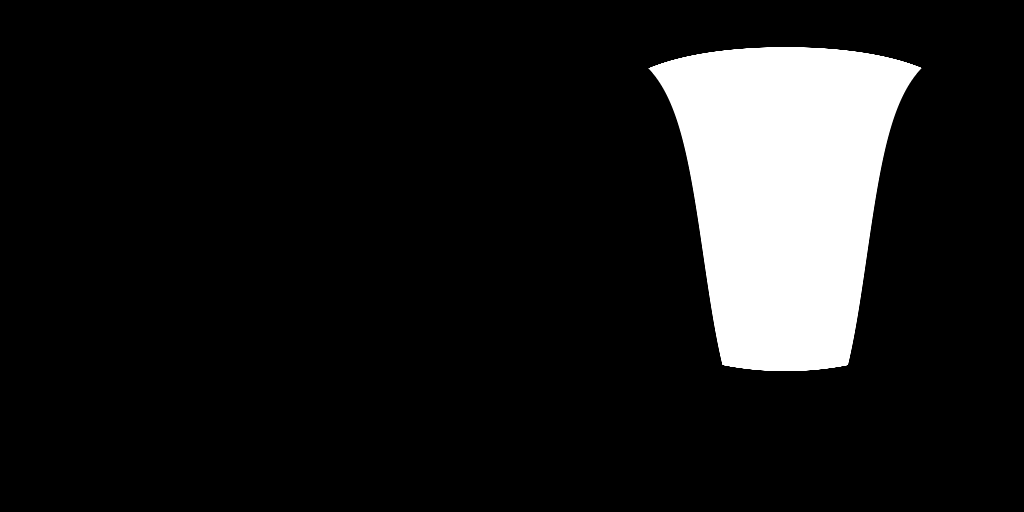} \\
    \bottomrule
    \end{tabular}
    \caption{Normal and ERP-re-projected masks with associated foreground prompts, assigned to \textcolor{red}{an indoor room}, \textcolor{green}{a green field}, and \textcolor{blue}{a busy street} as background prompts}
    \label{tab:settings_masks}
\end{table}

We evaluate six scenes combined with 168 seeds, resulting in 1,008 panoramas and 3,024 perspective images for testing. \autoref{tab:settings_masks} summarizes the conditions.  
To examine the impact of three mask sizes and ERP re-projection, we generate ERP-reprojected masks by first determining the mask's center in planar coordinates, converting it to spherical coordinates, and projecting the corresponding perspective view with a 120° FoV. The target perspective mask is then used to construct a bounding box, which is re-projected onto an empty ERP image to create the final mask.
As shown in \autoref{tab:settings_masks}, we also evaluate how the model performs with different aspect ratios, such as long, tall, or square masks, paired with suitable prompts like a bed, door, or sign.  
To create reference 2D images for evaluating generated outputs using FID \cite{heusel2017gans}, ImageReward \cite{xu2024imagereward} and CLIP-Score \cite{clipscore}, we use MD to generate perspective views of foreground objects based on prompts and background scenes.
The target perspective masks are derived from ERP masks, ensuring the object is centered and fully visible. With 18 prompts and 168 seeds, the dataset includes 18,144 reference images.

\subsection{Experimental Setup}
\label{sec:setup}

We configure MultiStitchDiffusion (MSTD) by following StitchDiffusion \cite{wang2023customizing360degreepanoramastexttoimage}.
We systematically vary key hyperparameters as shown in \autoref{tab:hyperparams}, using a default setting and modifying at most one parameter at a time.
We also run a baseline by disabling StitchDiffusion entirely (i.e., pure MultiDiffusion).
For MultiPanFusion (MPF), we use the setup of PanFusion \cite{zhang2024tamingstablediffusiontext} and evaluate the same hyperparameters.
We consider MD applied in the panorama branch (md\_pano), perspective branch (md\_pers), or both (md\_both).
Bootstrap-coupling can ensure the same background color across branches (branches-coupling) and the same color for each object (objects-coupling), while we optionally disable EPPA for foreground objects.

\begin{table}[!t]
    \small
    \centering
    \begin{tabular}{@{}lcc@{}}
    \toprule
    parameter & value range & default \\ & & (MSTD / MPF) \\
    \midrule
    bootstrapping & \{1, 5, 10, \dots, 50\} & 20 \\
    stride (MSTD) & \{4, 8, 16, 32\} & 8 \\
    mask size & \{\text{S, M, L}\} & M \\
    mask type & \{\text{regular, ERP-reproj.}\} & ERP-reproj. \\
    mask indices & \(\mathcal{P}\{0, 1, 2\}\) & \{0, 1, 2\} \\
    LoRA enabled & \{\text{yes, bg-only, no}\} & bg-only / yes \\
    bootstrap-cpl. & \{\text{branches, objects, none}\} & none / branches \\
    noise-cpl. & \{\text{yes, no}\} & yes \\
    global prompt & \{\text{yes, no}\} & no \\
    fg. EPPA (MPF) & \{\text{yes, no}\} & yes \\
    \bottomrule
    \end{tabular}
    \caption{Key hyperparameters used to benchmark MultiStitchDiffusion (MSTD) and MultiPanFusion (MPF).}
    \label{tab:hyperparams}
\end{table}

\subsection{Quantitative Results}
\label{sec:quantitative_results}
In this section, we report a selection of the most informative quantitative results.
Full benchmark results are included in the \href{https://sigport.org/documents/spherical-dense-text-image-synthesis}{supplementary material.}
Our main results are in \autoref{tab:performance}.
When comparing MSTD to the baselines, we observe that most scores are roughly the same.
MPF performs worse than MSTD in all metrics.
Furthermore, an IoU of 0.05 at \textit{md\_pers} shows the insufficiency of applying MD at the perspective branch only for synthesizing foreground objects.
Even when combined with MD applied at the panorama branch, the scores do not change significantly.
Notably, our model reaches roughly the same image quality while generating full spherical images.

\begin{table}[ht]
    \small
    \centering
    \begin{tabular}{@{}llccccc@{}}
    \toprule
     & task & IoU$\uparrow$ & IR$\uparrow$ & FID$\downarrow$ \\
    \midrule
    MD (original) & DT2I & 0.57 & -0.47 & 82.61\\
    MD (with pano-LoRA) & DT2I & \underline{0.66} & \underline{-0.42} & \textbf{60.99}\\
    \midrule
    MSTD (ours) & SDT2I & \textbf{0.67} & \textbf{-0.37} & \underline{61.12}\\
    \midrule
    MPF (md\_pano) (ours) & SDT2I & 0.45 & -1.19 & 84.60\\
    MPF (md\_pers) (ours) & SDT2I & 0.05 & -1.79 & 107.00\\
    MPF (md\_both) (ours) & SDT2I & 0.44 & -1.22 & 84.82\\
    \bottomrule
    \end{tabular}
    \caption{Performance of our MSTD and MPF models compared to the MD baseline. ``with pano-LoRA'' means MD is adapted to the ERP-distorted style.}
    \label{tab:performance}
\end{table}

\begin{figure}[t]
    \centering
    \hfill
    \includegraphics[width=0.48\linewidth]{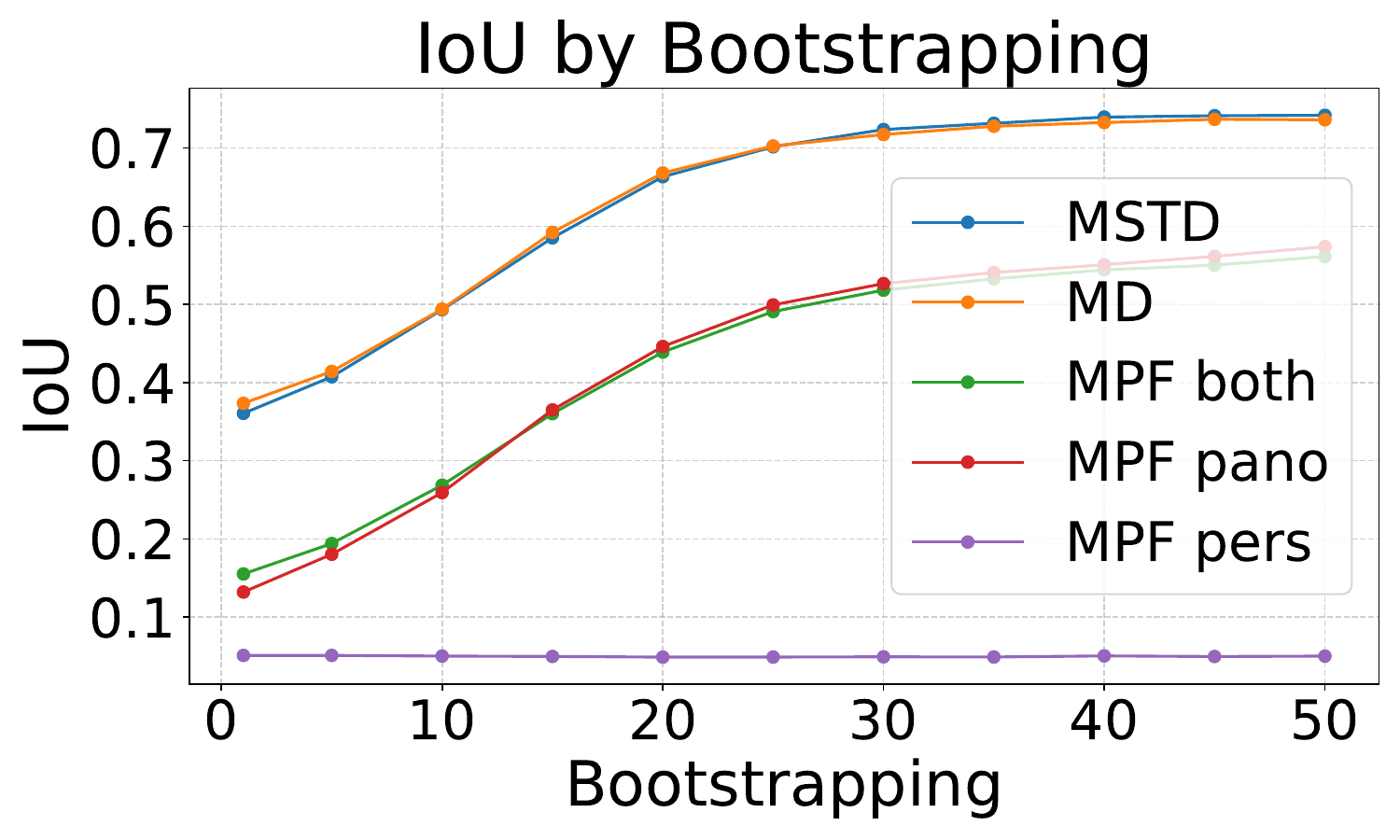}
    \includegraphics[width=0.48\linewidth]{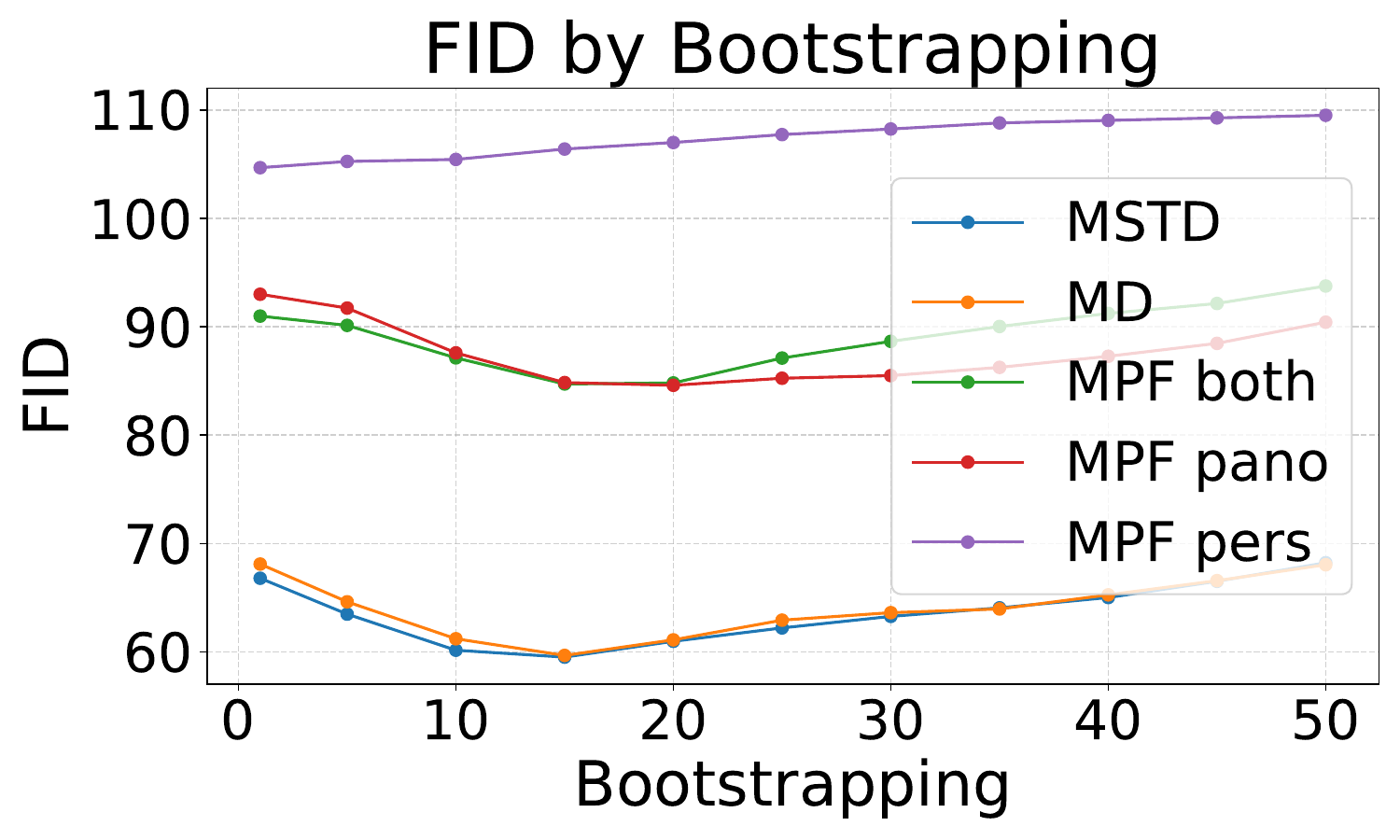}
    \caption{The influence of bootstrapping on our metrics for every approach, showing a functional relationship which is non-monotonous at FID.}
    \label{fig:plots_bootstrapping}
\end{figure}

\begin{table}[ht]
    \small
    \centering
    \begin{tabular}{@{}lccccc@{}}
    \toprule
     bootstrap-coupling & IoU$\uparrow$ & IR$\uparrow$ & FID$\downarrow$ \\
    \midrule
    no  & 0.43 & -1.26 & 85.73\\
    yes & \textbf{0.44} & \textbf{-1.22} & \textbf{84.82}\\
    \bottomrule
    \end{tabular}
    \caption{Bootstrap-coupling at PanFusion with MultiDiffusion applied at both branches is better acroos all metrics.}
    \label{tab:bootstrap_coupling}
\end{table}

\begin{table}[ht]
    \small
    \centering
    \begin{tabular}{@{}lccc@{}}
    \toprule
    Method & IoU$\uparrow$ & IR$\uparrow$ & FID$\downarrow$ \\
    \midrule
    MSTD (no global prompt) & \textbf{0.67} & -0.37 & 61.12 \\
    MSTD (global prompt) & 0.59 & \textbf{-0.35} & \textbf{57.03} \\
    \midrule
    MPF (md\_pano, no global prompt) & \textbf{0.45} & \textbf{-1.19} & 84.60 \\
    MPF (md\_pano, global prompt) & 0.43 & -1.25 & \textbf{74.84} \\
    \midrule
    MPF (md\_pers, no global prompt) & 0.05 & -1.79 & 107.00 \\
    MPF (md\_pers, global prompt) & \textbf{0.08} & \textbf{-1.60} & \textbf{81.94} \\
    \midrule
    MPF (md\_both, no global prompt) & \textbf{0.44} & \textbf{-1.22} & 84.82 \\
    MPF (md\_both, global prompt) & 0.41 & -1.32 & \textbf{77.20} \\
    \bottomrule
    \end{tabular}
    \caption{Influence of the inclusion of foreground objects in the background prompt. Global prompts lead to increased image quality but slightly lower mask fidelity.}
    \label{tab:global_prompt}
\end{table}

\begin{table}[ht]
    \small
    \centering
    \begin{tabular}{@{}lccc@{}}
    \toprule
    Method & IoU$\uparrow$ & IR$\uparrow$ & FID$\downarrow$ \\
    \midrule
    MPF md\_pano (FG + BG EPPA) & 0.45 & -1.19 & 84.60 \\
    MPF md\_pano (BG EPPA) & \textbf{0.53} & \textbf{-1.09} & \textbf{80.61} \\
    \midrule
    MPF md\_pers (FG + BG EPPA) & \textbf{0.05} & \textbf{-1.79} & 107.00 \\
    MPF md\_pers (BG EPPA) & \textbf{0.05} & \textbf{-1.79} & \textbf{106.68} \\
    \midrule
    MPF md\_both (FG + BG EPPA) & 0.44 & -1.22 & 84.82 \\
    MPF md\_both (BG EPPA) & \textbf{0.52} & \textbf{-1.10} & \textbf{82.21} \\
    \bottomrule
    \end{tabular}
    \caption{Influence of not applying EPPA at foreground layers, showing an overall improvement at mask adherence and image quality with MultiPanFusion.}
    \label{tab:eppa}
\end{table}

\autoref{fig:plots_bootstrapping} shows that as the bootstrapping parameter increases from 0 to its maximum in increments of 5, IoU steadily improves until plateauing around 30, while FID reaches its optimal value at a bootstrapping of 15.
\autoref{tab:bootstrap_coupling} shows that bootstrap coupling between both branches in MPF leads to a small quantitative improvement in all scores.
We anticipated greater improvement as this technique largely resolves visual artifacts.
\autoref{tab:global_prompt} examines the effect of including local prompts in the background prompt. While this negatively impacts IoU in all methods except \textit{md\_pers}, it simultaneously lowers FID, making optimization challenging. Further qualitative analysis is needed for better insights.
In \autoref{tab:eppa}, we analyze the influence of EPPA.
Against our expectations, the ignorance of EPPA at foreground objects has a positive and more considerable impact than bootstrap-coupling.
IoU, Image-Reward, and FID have better scores when using this variant.

\subsection{Qualitative Results}
\label{sec:qualitative_results}

\begin{figure}[ht]
    \centering
    \includegraphics[width=1\linewidth]{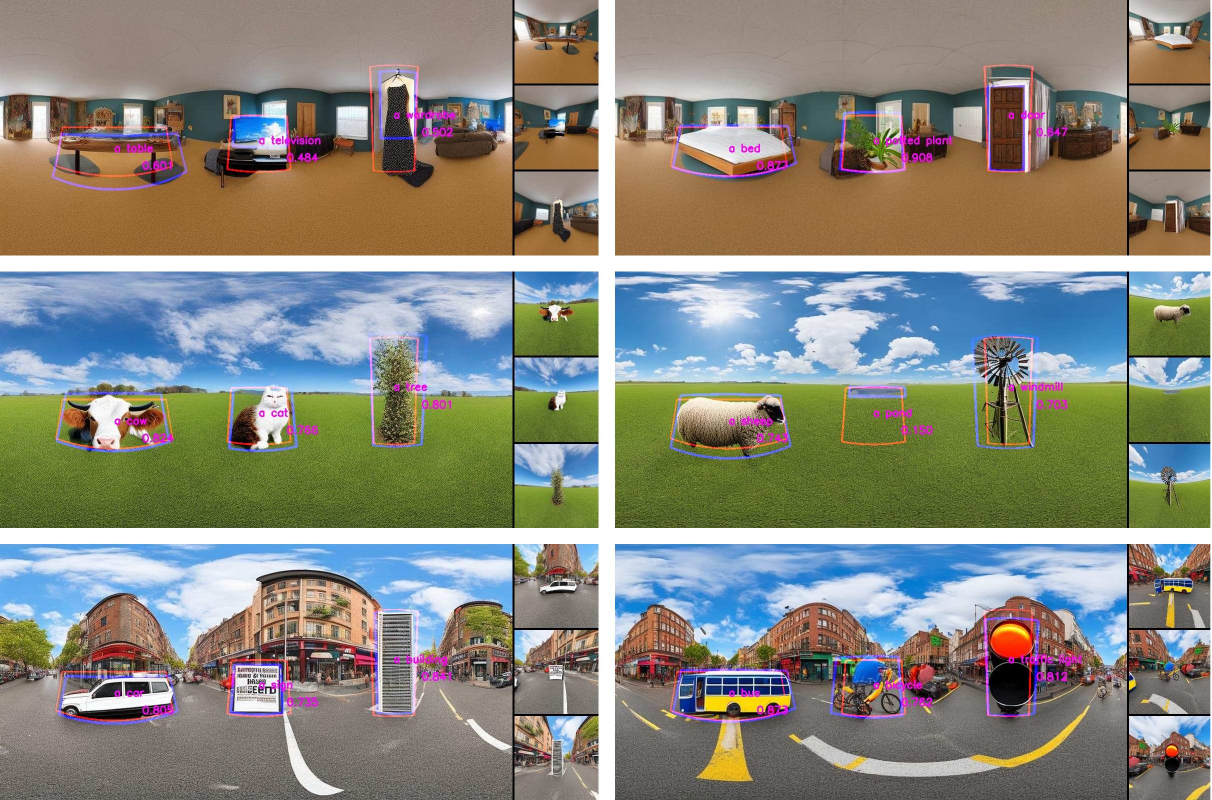}
    \caption{Prompt and mask adherence of MultiStitchDiffusion; showing a high IoU but some objects, e.g. the building or the wardrobe, not blending well into the background.}
    \label{fig:stitch_mask_prompt}
\end{figure}

MSTD generally handles prompt adherence well, synthesizing most requested objects but occasionally misinterpreting specific items (e.g., partial traffic lights). Mask alignment is decent (IoU around 2/3), yet bounding boxes often lack precision. Backgrounds show limited variety, frequently resulting in incomplete or duplicated objects. Bootstrapping strongly influences performance: higher values improve alignment but can distort shapes and hamper blending. Medium mask sizes strike a balance between neglect (small) and oversizing (large). LoRA for foreground objects tends to add distortions, while projecting masks in an ERP-aligned manner avoids unnatural object shapes.

\begin{figure}[!ht]
    \centering
    \includegraphics[width=1\linewidth]{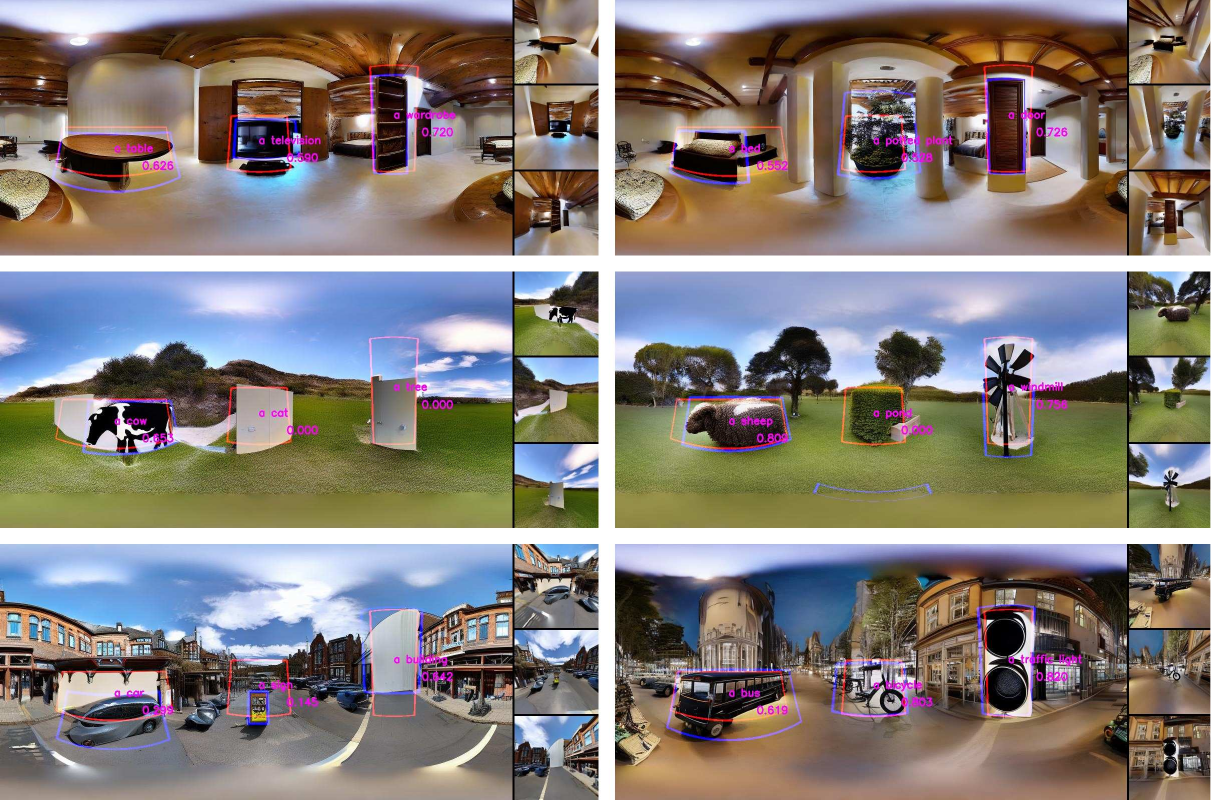}
    \caption{Prompt and mask adherence of MultiPanFusion being satisfactory in some examples. The middle-left image, however, shows a failure case.}
    \label{fig:panfusion_mask_prompt}
\end{figure}

MPF sometimes misses or misinterprets objects, but it provides richer, more diverse backgrounds than MSTD. Although objects typically align with their masks, some appear over-fitted, producing artifacts around the mask boundary. Applying MD to both panorama and perspective branches reduces missing objects but can still introduce conflicts. Disabling foreground EPPA sometimes clarifies objects, while bootstrapping affects object size and shape, with an ideal range around 10–20. Coupling the bootstrapping background across branches helps slightly but is not a major factor. Mask size and global prompt usage have similar impacts as in MSTD, and other parameters show minimal effects.

\begin{figure}[!ht]
    \centering
    \includegraphics[width=1\linewidth]{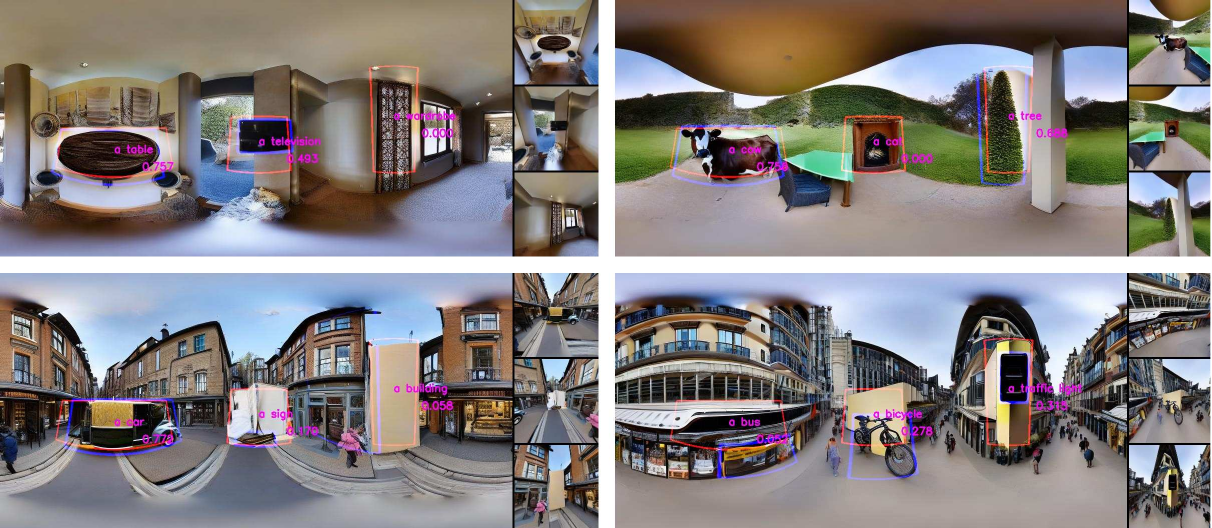}
    \caption{Various failure-cases of MultiPanFusion, visualizing noise around the objects, unfitting ceilings, wrong objects, and foreground elements merging with the background.}
    \label{fig:panfusion_fails}
\end{figure}

\textit{Failure Case Analysis:} The weaker quantitative results of MPF might stem from several recurring issues. First, foreground objects (e.g., a rectangular mask at high bootstrapping) can blend awkwardly with the background, occasionally creating indoor elements in outdoor scenes. This is likely tied to the training data of PanFusion, which is primarily indoors, causing outdoor prompts to appear as if viewed from inside a room. Second, MPF sometimes fails to integrate foreground objects into scene layouts, merging cars with buildings or introducing other distortions, see \autoref{fig:panfusion_fails}. Lastly, artifacts like blurriness and pixelation persist around masked regions, even with our improvements. These artifacts raise FID and CMMD and reduce scores for prompt-adherence (IoU, CLIP-score, and Image-Reward).

\section{Conclusion \& Future Work}
\label{sec:conclusion}

We introduced MultiStitchDiffusion (MSTD) and MultiPanFusion (MPF), the first approaches for spherical dense text-to-image (SDT2I) synthesis, along with a benchmark for evaluation. MSTD preserves prompt- and mask-adherence from MultiDiffusion (MD) while producing coherent panoramas with StitchDiffusion. MPF builds on PanFusion’s dual-branch architecture, offering increased scene variety but showing more artifacts and misinterpretations. Our experiments highlight the critical role of hyperparameters (e.g., bootstrapping, mask size, and LoRA usage) and suggest improvements like bootstrap-coupling and reduced use of EPPA.
However, both approaches still face challenges: MSTD’s repetitive backgrounds and MPF’s indoor biases, object merges, and blurry artifacts can limit real-world applicability.
Furthermore, our evaluation relies on MD-generated references and 18 prompts, which may limit generalizability. Future work should explore richer data sets with real panorama references. Despite these limitations, our methods represent a significant step toward flexible, text-guided 360° image synthesis.

\noindent
\textbf{Acknowledgments:}
This work was supported by the BMBF project SustainML (Grant 101070408).

\bibliographystyle{IEEEbib}
\bibliography{strings,refs}

\newpage
\section*{Appendix}
\label{sec:appendix}
\setcounter{section}{1}
\setcounter{table}{0}
Here we report a selection of materials that provides further insight into our experiments.
We start by showing an extract of reference images from DSynView on which the quality assessment was conducted.
Next, we present a greater variety of qualitative results that show the influence of different parameters.
To improve visibility, we subsequently show the extracted perspective images at larger size.
Furthermore, we present qualitative results from experiments testing the limitations of MultiPanFusion.
Finally, we show the full tables of quantitative results with MSTD and MPF, including all assessed parameters as well as two additional metrics, CLIP-score and CLIP Maximum Mean Discrepancy (CMMD)\,\cite{jayasumana2024rethinkingfidbetterevaluation}.
We visualize the influence of bootstrapping and mask size in the same way we did in \autoref{fig:plots_bootstrapping}.

\begin{figure}[ht]
    \centering
    \includegraphics[width=0.15\linewidth]{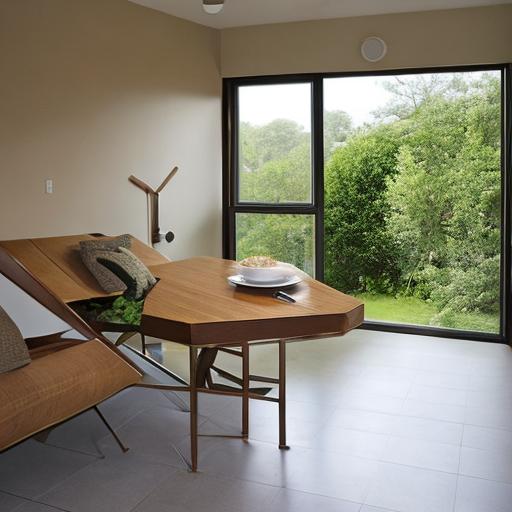}
    \includegraphics[width=0.15\linewidth]{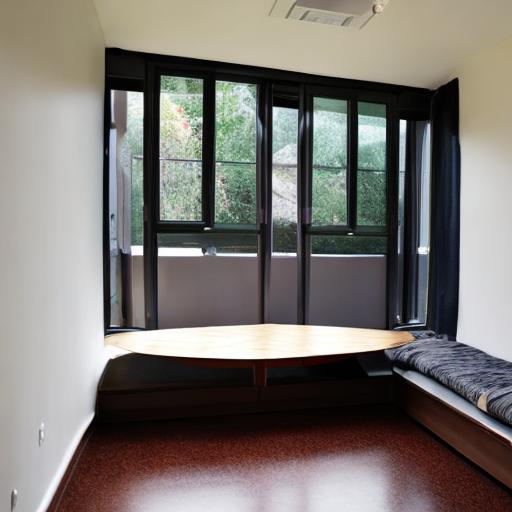}
    \includegraphics[width=0.15\linewidth]{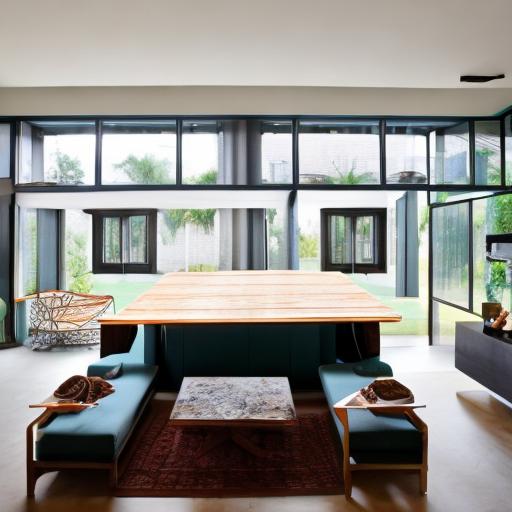}
    \includegraphics[width=0.15\linewidth]{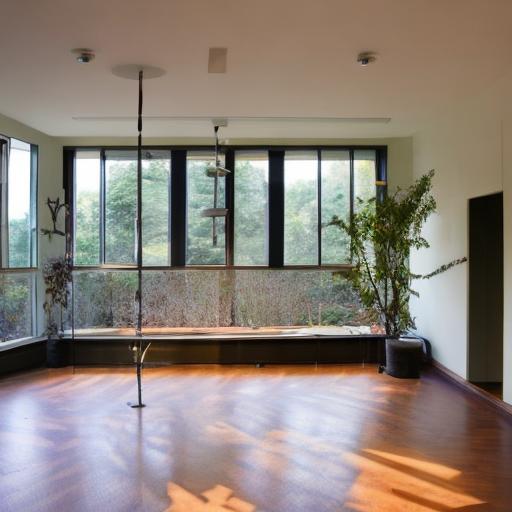}
    \includegraphics[width=0.15\linewidth]{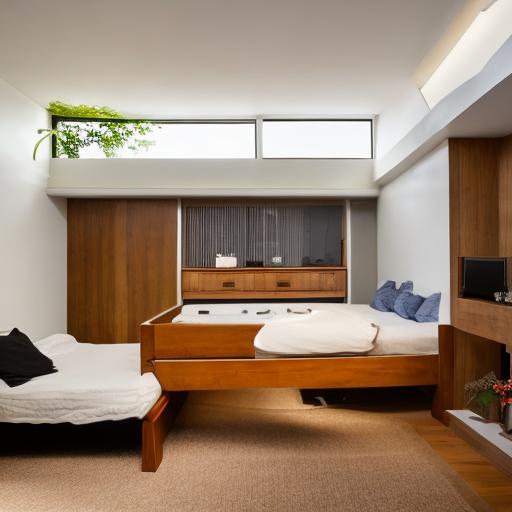}
    \includegraphics[width=0.15\linewidth]{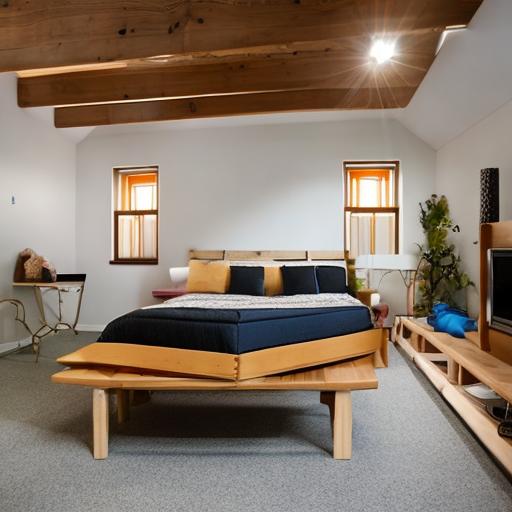}
    \hfill
    \includegraphics[width=0.15\linewidth]{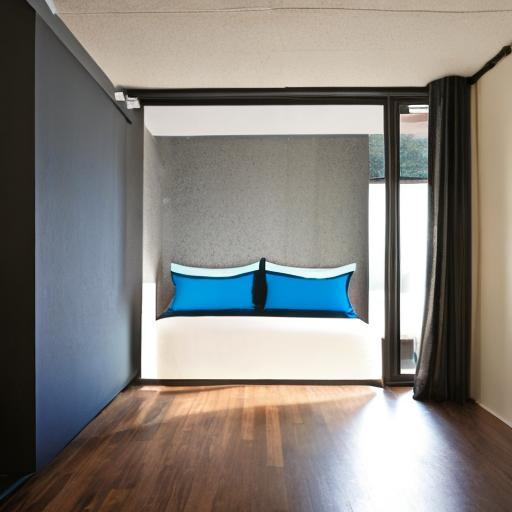}
    \includegraphics[width=0.15\linewidth]{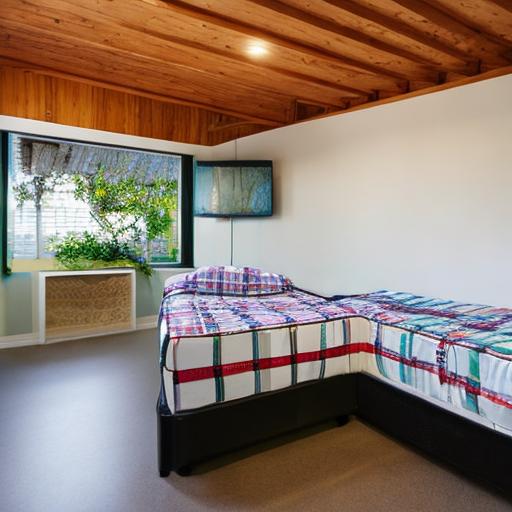}
    \includegraphics[width=0.15\linewidth]{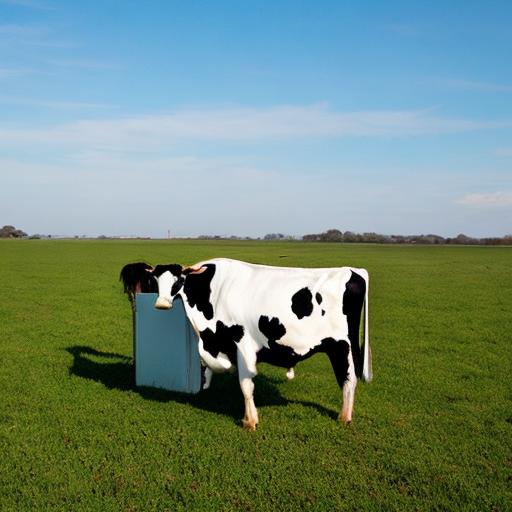}
    \includegraphics[width=0.15\linewidth]{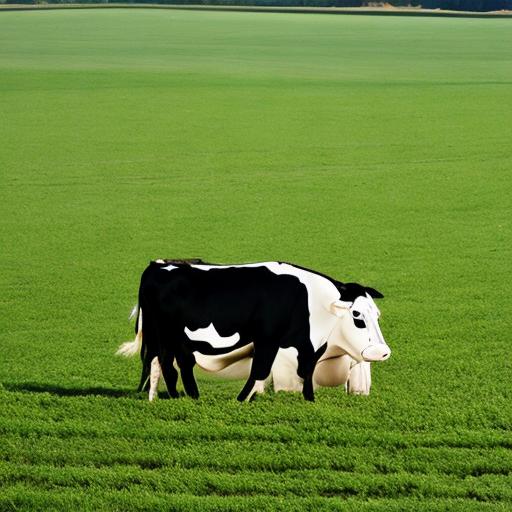}
    \includegraphics[width=0.15\linewidth]{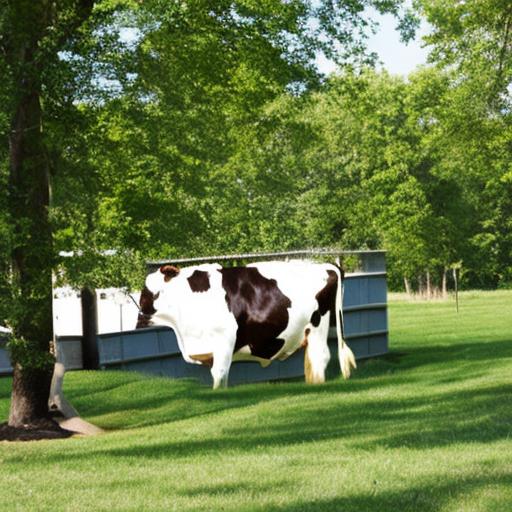}
    \includegraphics[width=0.15\linewidth]{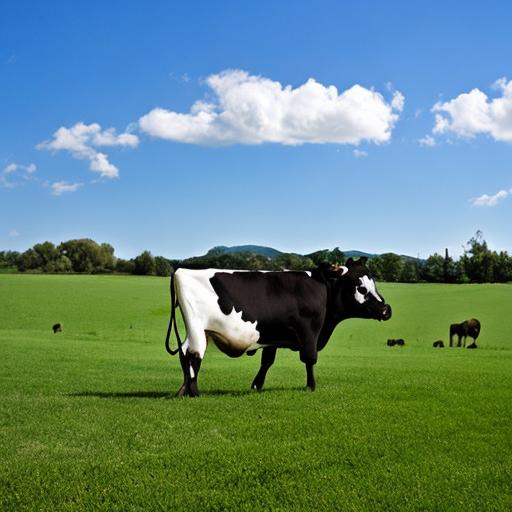}
    \includegraphics[width=0.15\linewidth]{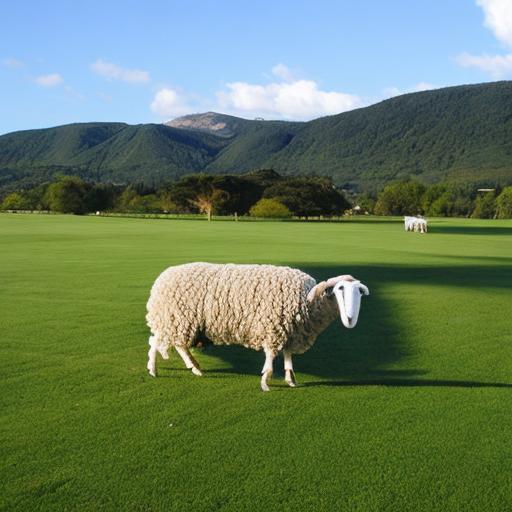}
    \includegraphics[width=0.15\linewidth]{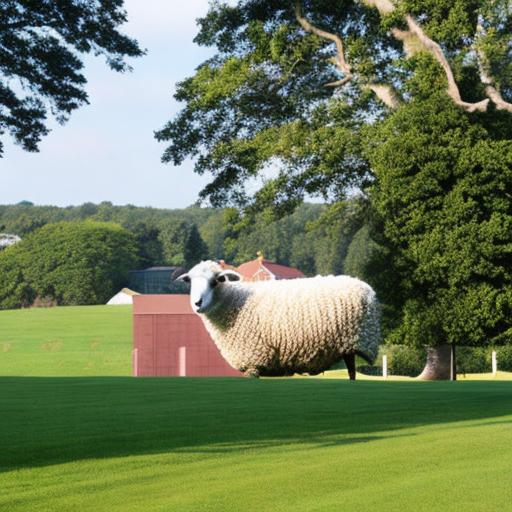}
    \includegraphics[width=0.15\linewidth]{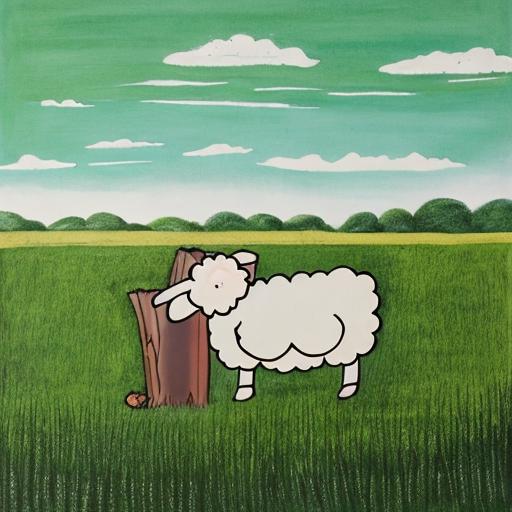}
    \includegraphics[width=0.15\linewidth]{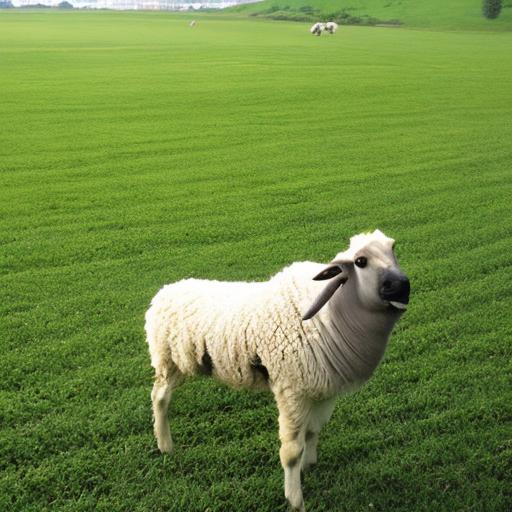}
    \includegraphics[width=0.15\linewidth]{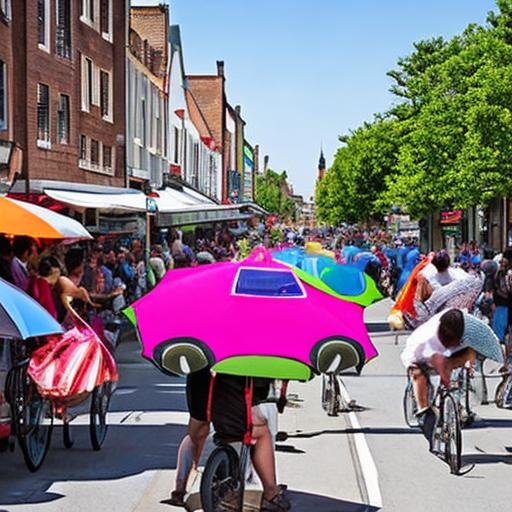}
    \includegraphics[width=0.15\linewidth]{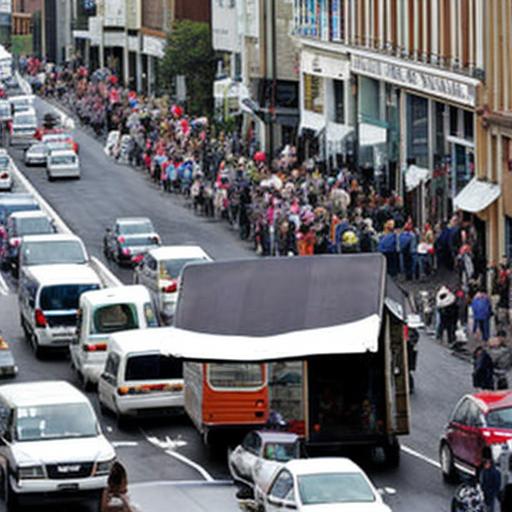}
    \includegraphics[width=0.15\linewidth]{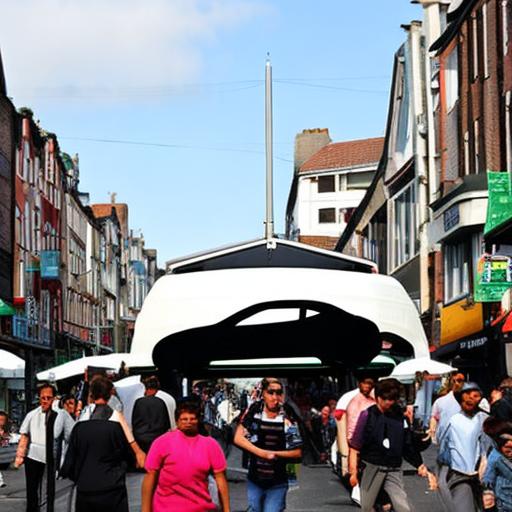}
    \includegraphics[width=0.15\linewidth]{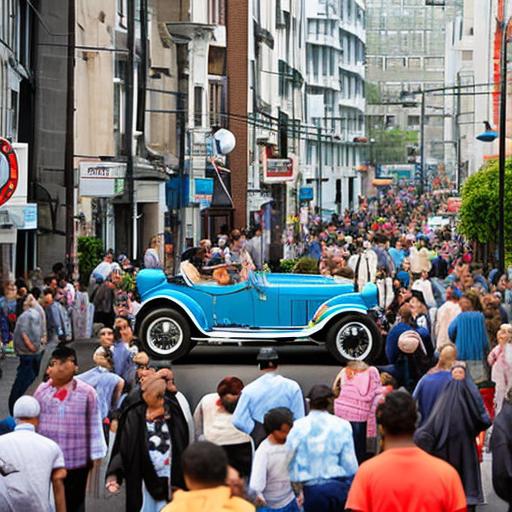}
    \includegraphics[width=0.15\linewidth]{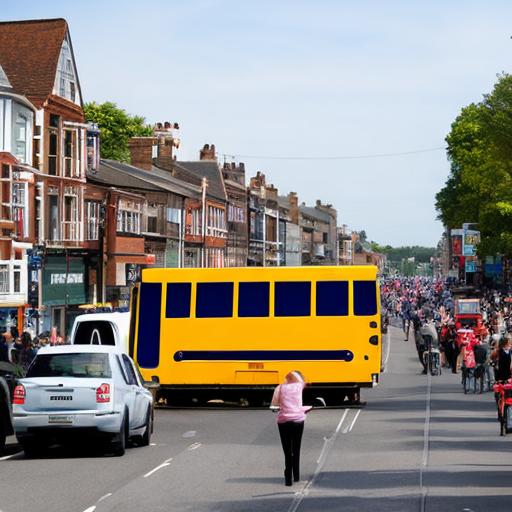}
    \includegraphics[width=0.15\linewidth]{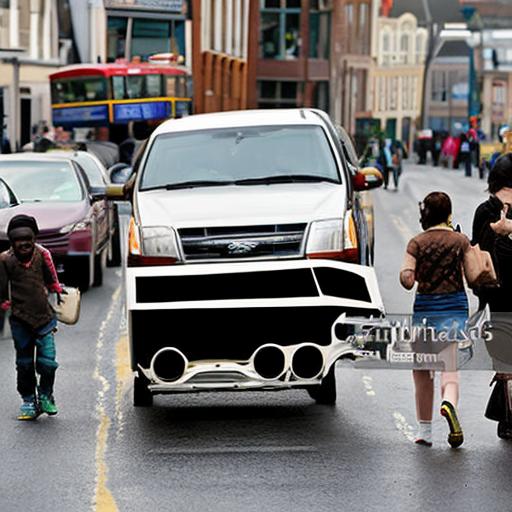}
    \includegraphics[width=0.15\linewidth]{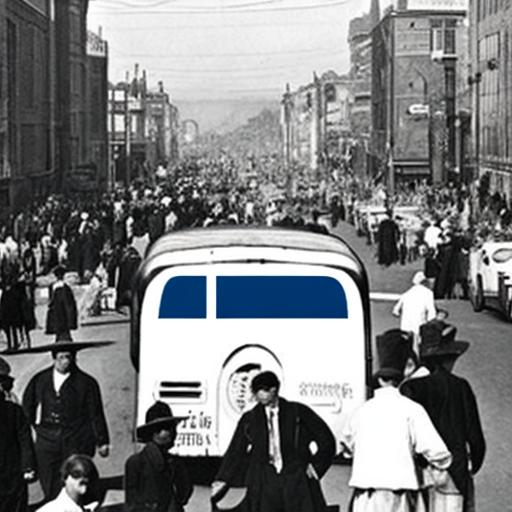}
    \includegraphics[width=0.15\linewidth]{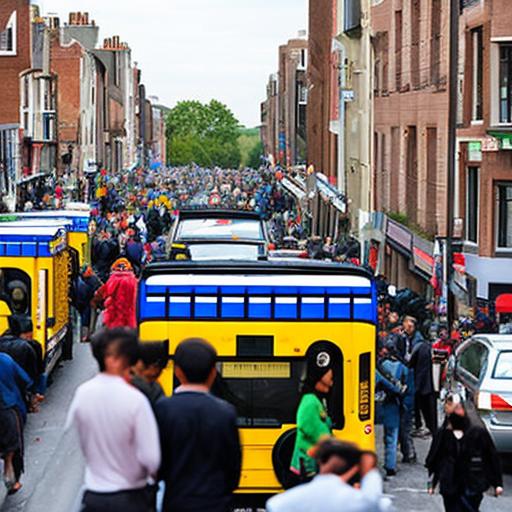}
    \includegraphics[width=0.15\linewidth]{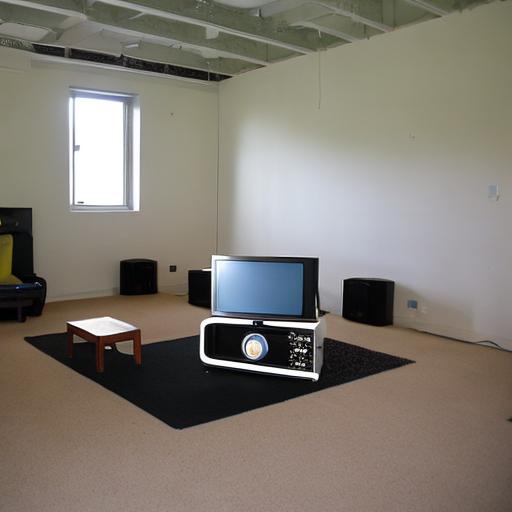}
    \includegraphics[width=0.15\linewidth]{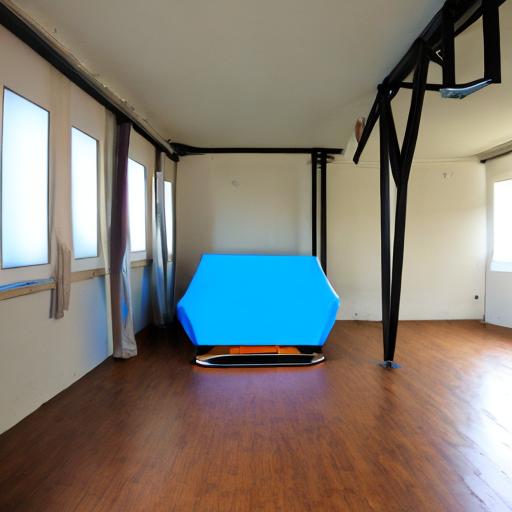}
    \includegraphics[width=0.15\linewidth]{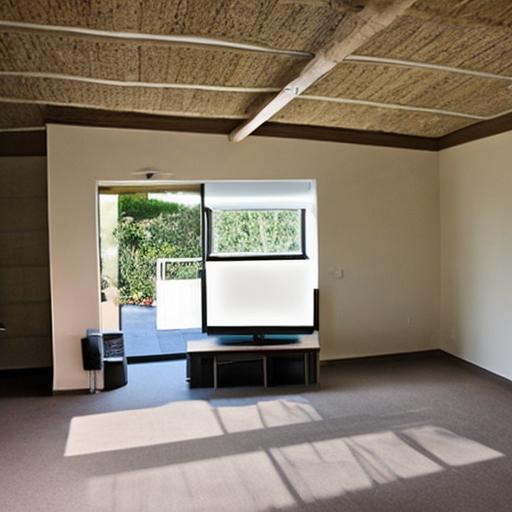}
    \includegraphics[width=0.15\linewidth]{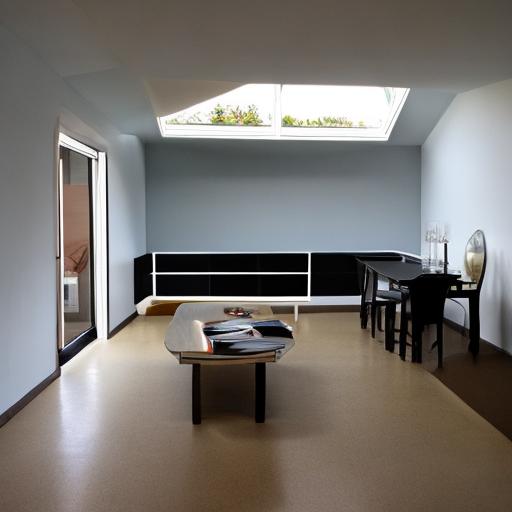}
    \includegraphics[width=0.15\linewidth]{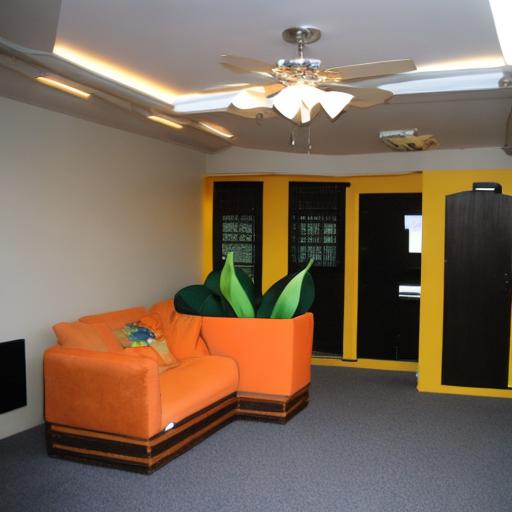}
    \includegraphics[width=0.15\linewidth]{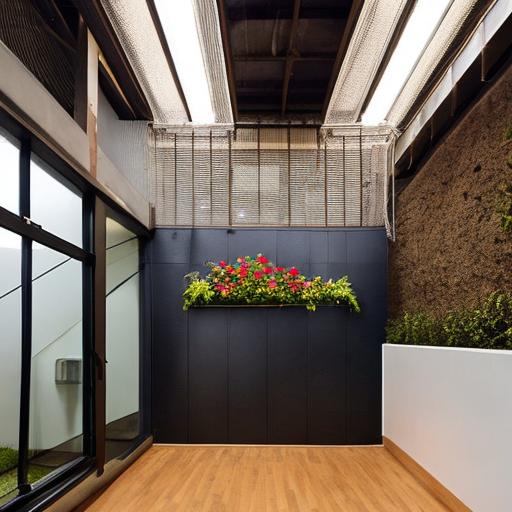}
    \includegraphics[width=0.15\linewidth]{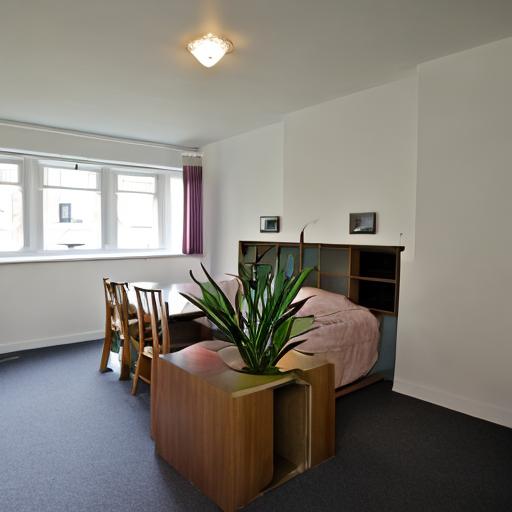}
    \includegraphics[width=0.15\linewidth]{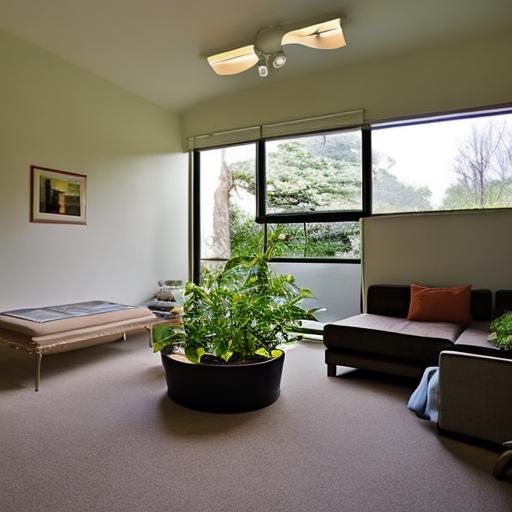}
    \includegraphics[width=0.15\linewidth]{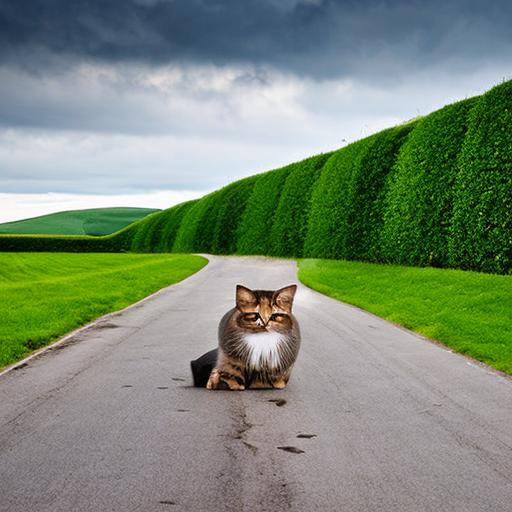}
    \includegraphics[width=0.15\linewidth]{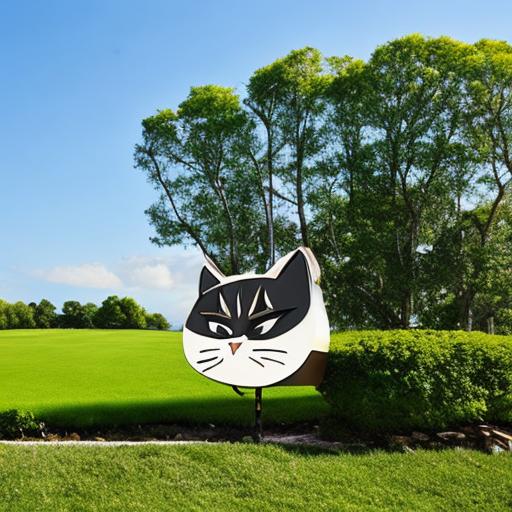}
    \includegraphics[width=0.15\linewidth]{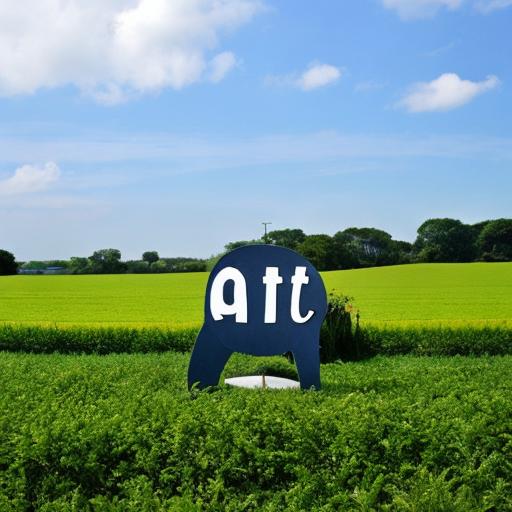}
    \includegraphics[width=0.15\linewidth]{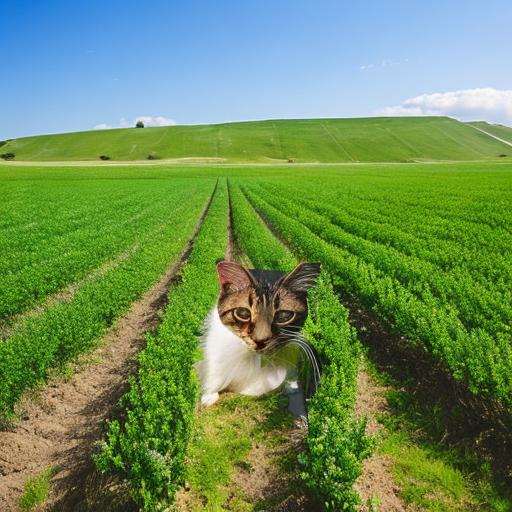}
    \includegraphics[width=0.15\linewidth]{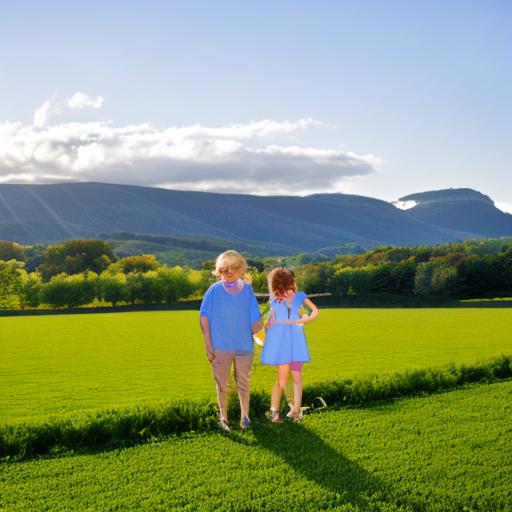}
    \includegraphics[width=0.15\linewidth]{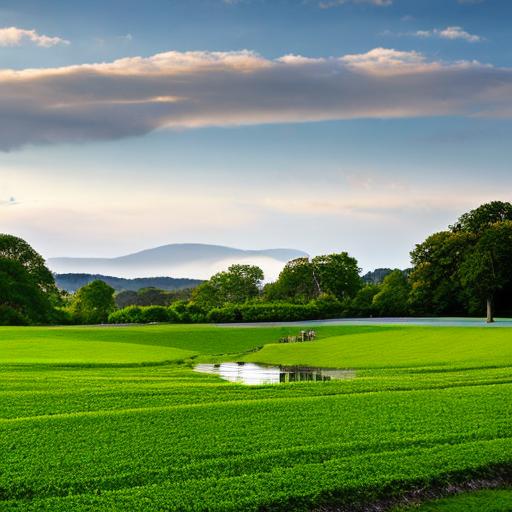}
    \includegraphics[width=0.15\linewidth]{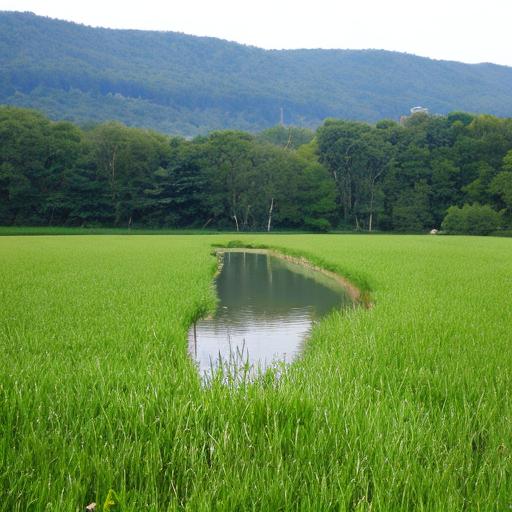}
    \includegraphics[width=0.15\linewidth]{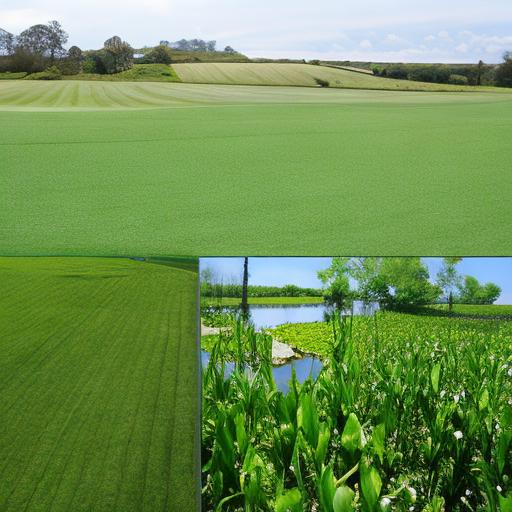}
    \includegraphics[width=0.15\linewidth]{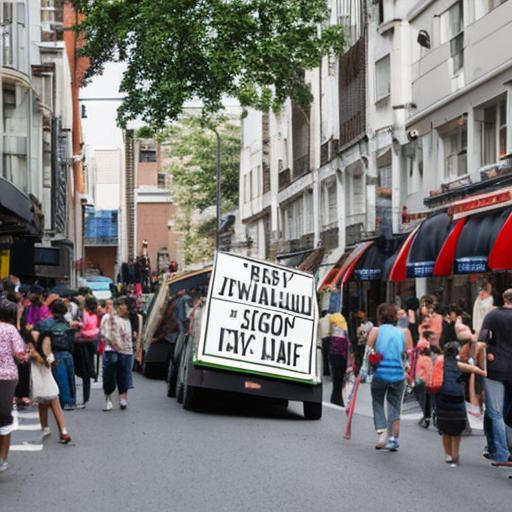}
    \includegraphics[width=0.15\linewidth]{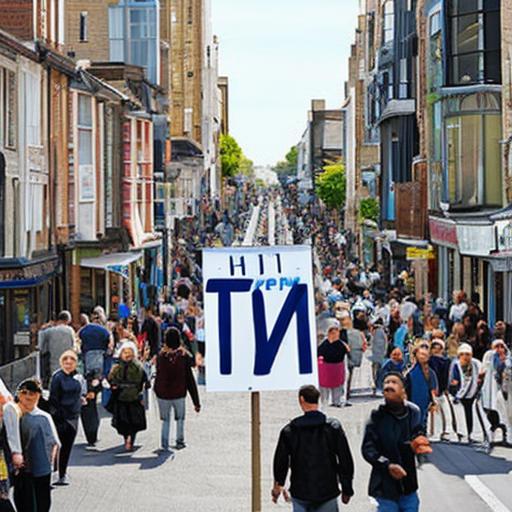}
    \includegraphics[width=0.15\linewidth]{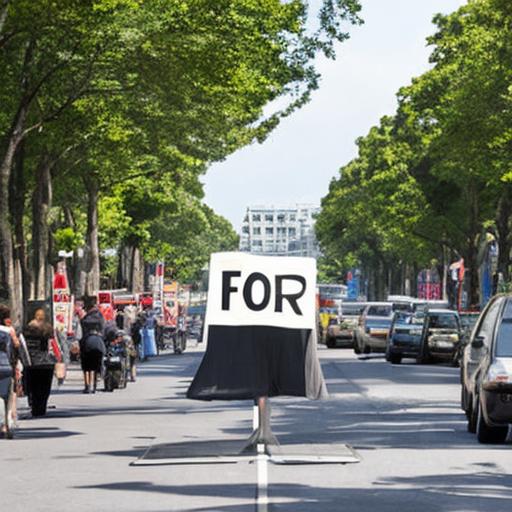}
    \includegraphics[width=0.15\linewidth]{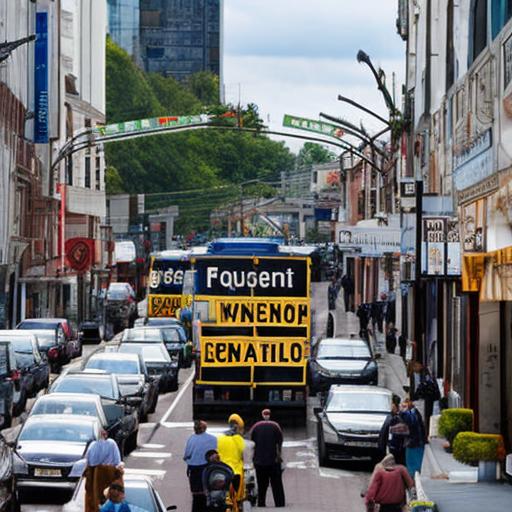}
    \includegraphics[width=0.15\linewidth]{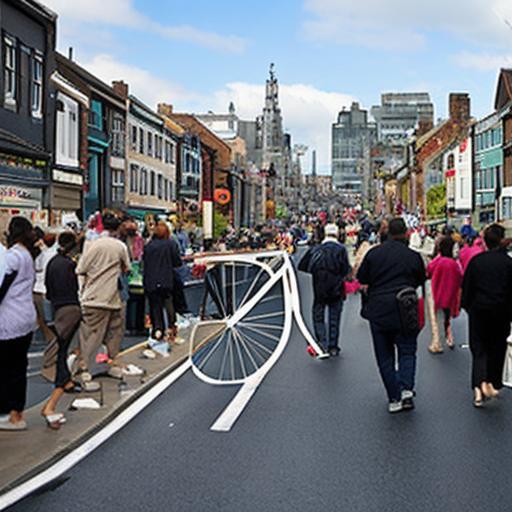}
    \includegraphics[width=0.15\linewidth]{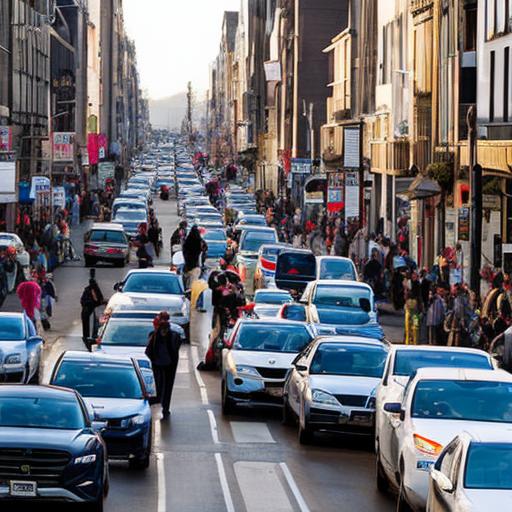}
    \includegraphics[width=0.15\linewidth]{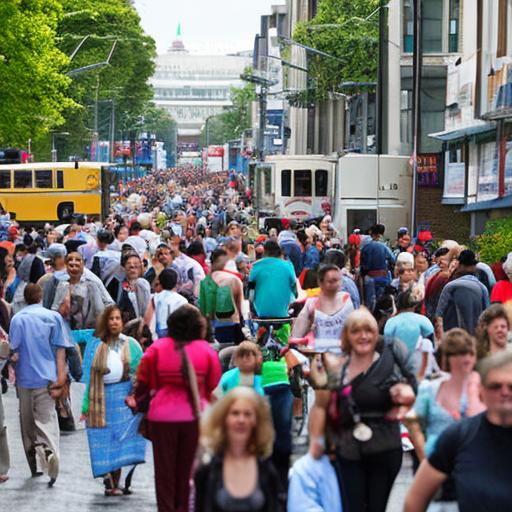}
    \includegraphics[width=0.15\linewidth]{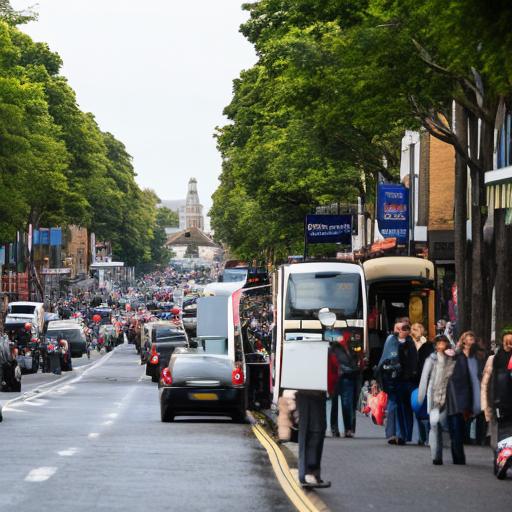}
    \caption{Example images from our DSynView dataset}
    \label{fig:sample_data}
\end{figure}

\begin{figure}[!ht]
    \centering
    \includegraphics[width=1\linewidth]{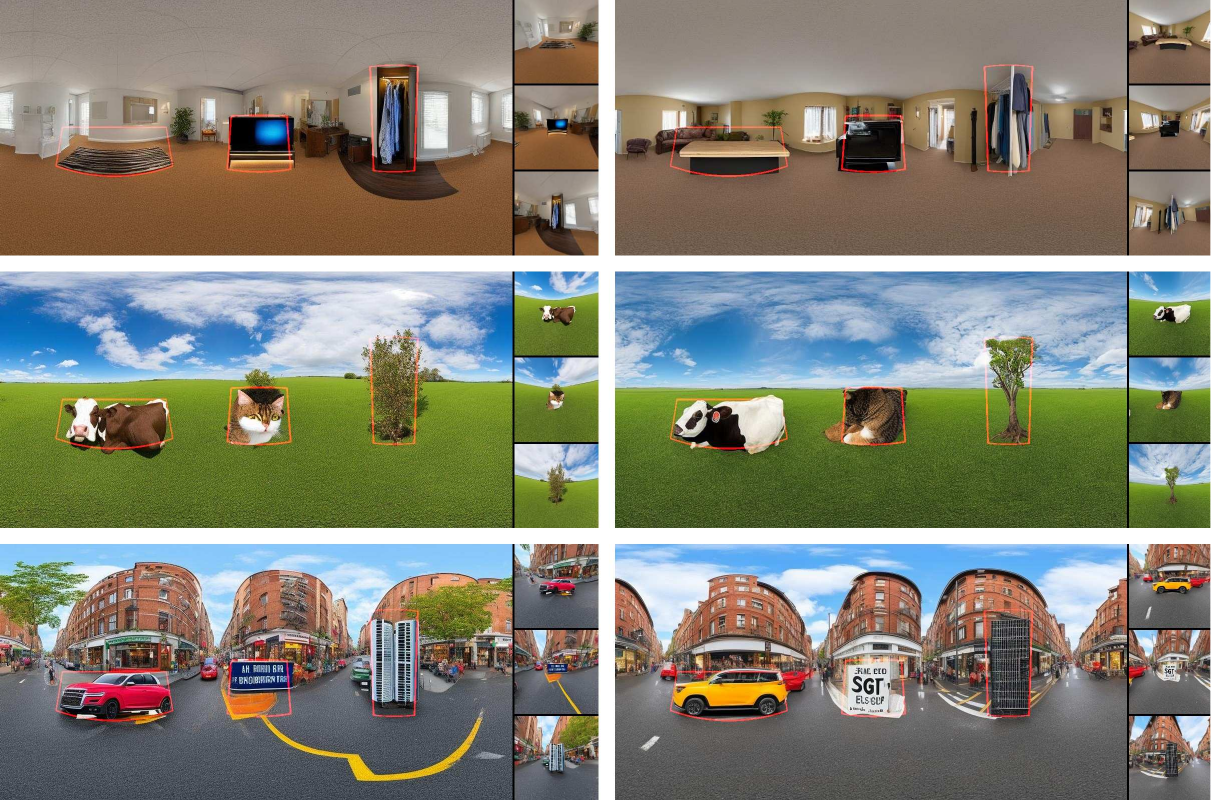}
    \caption{Image-quality and -diversity of MSTD; showing a decent quality but also similar-looking backgrounds for different seeds.}
    \label{fig:stitch_quality_diversity}
\end{figure}
\begin{figure}[!ht]
    \centering
    \includegraphics[width=1\linewidth]{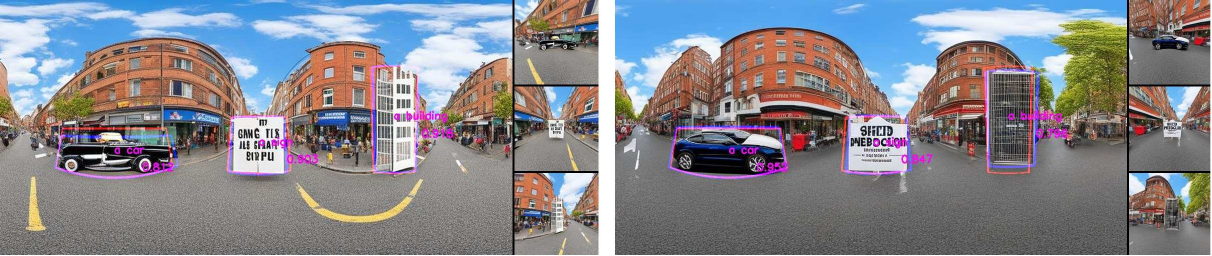}
    \caption{Influence of stitching-operation. Left: stitching. Right: No stitching}
    \label{fig:stitch_nostitch}
\end{figure}
\begin{figure}[!ht]
    \centering
    \includegraphics[width=1\linewidth]{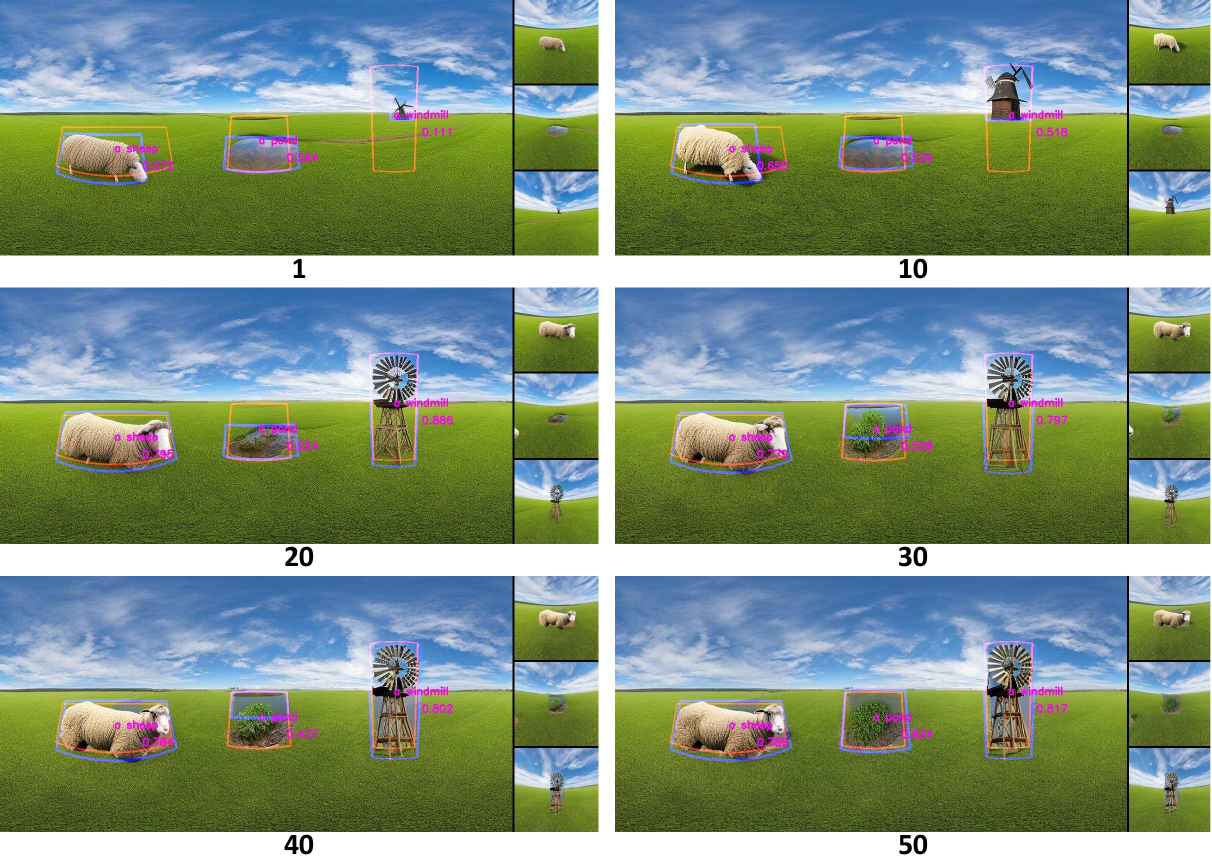}
    \caption{Influence of bootstrapping, visualizing a smoother blending of objects and background at lower values and a higher mask-adherence at higher values.}
    \label{fig:stitch_bootstrap}
\end{figure}
\begin{figure}[!ht]
    \centering
    \includegraphics[width=1\linewidth]{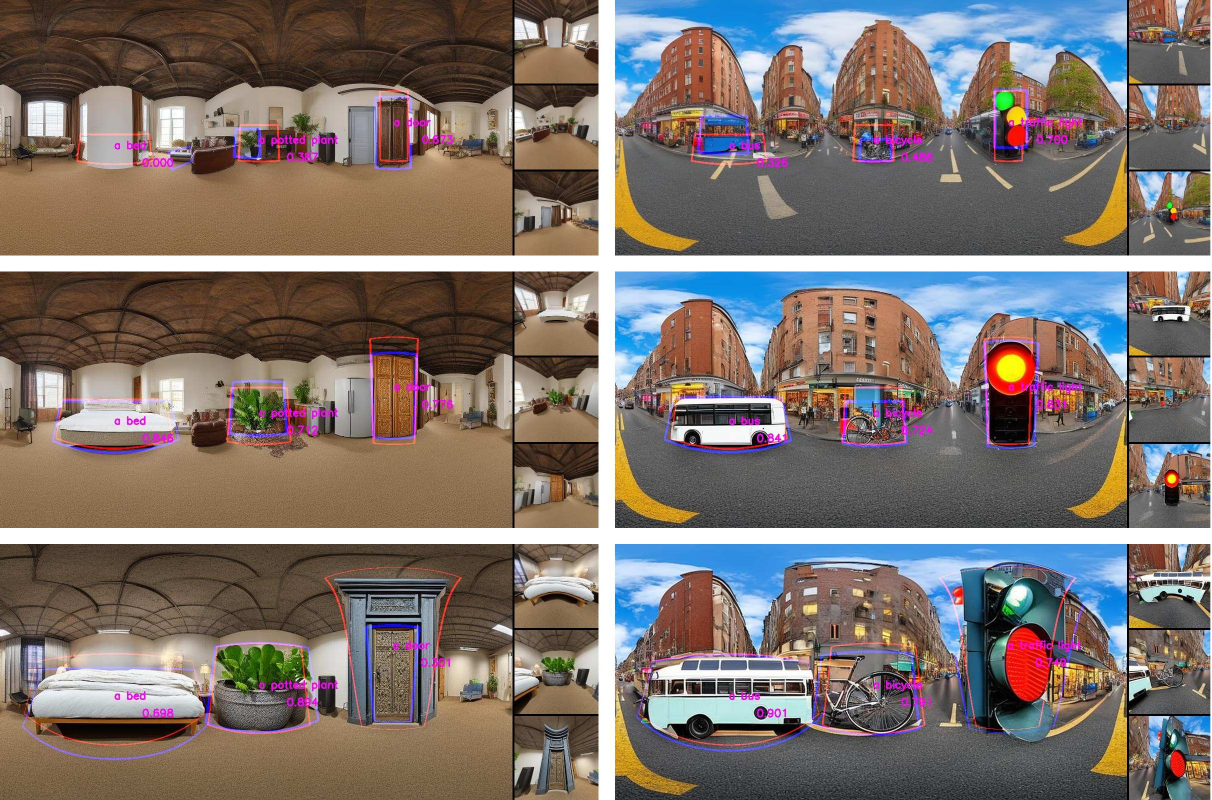}
    \caption{Influence of mask size. Top: small. Middle: middle. Bottom: large}
    \label{fig:stitch_size}
\end{figure}
\begin{figure}[!ht]
    \centering
    \includegraphics[width=1\linewidth]{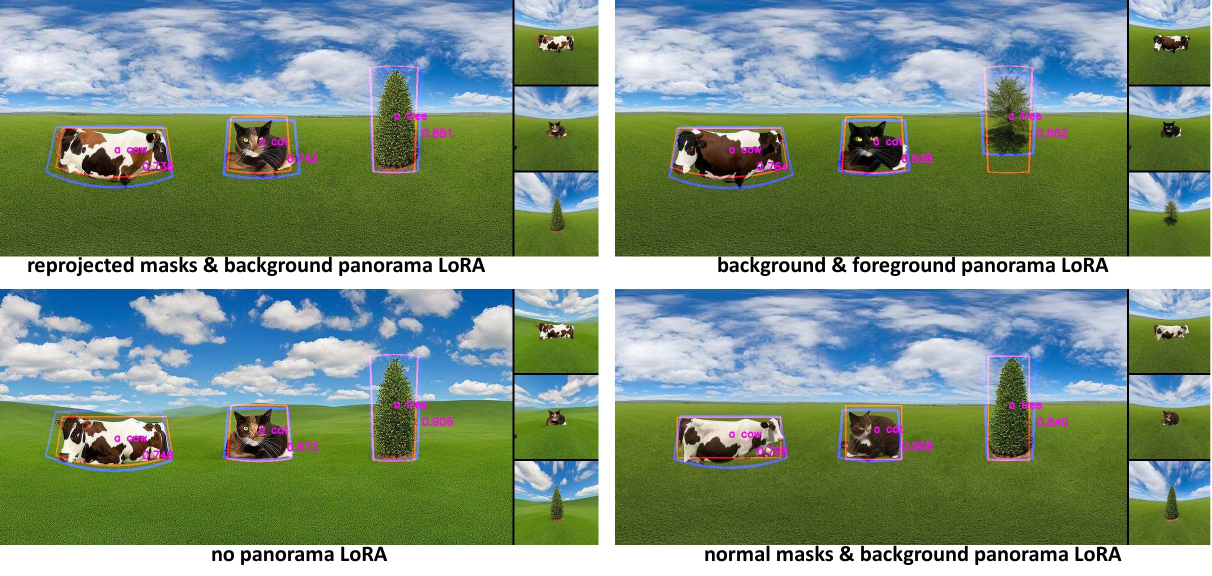}
    \caption{Influence of spherical parameters. The top-left setting usually yields the best results, while the bottom-left is insufficient for spherical images, as the missing distortions in the ERP image show.}
    \label{fig:stitch_pano}
\end{figure}
\begin{figure}[!ht]
    \centering
    \includegraphics[width=1\linewidth]{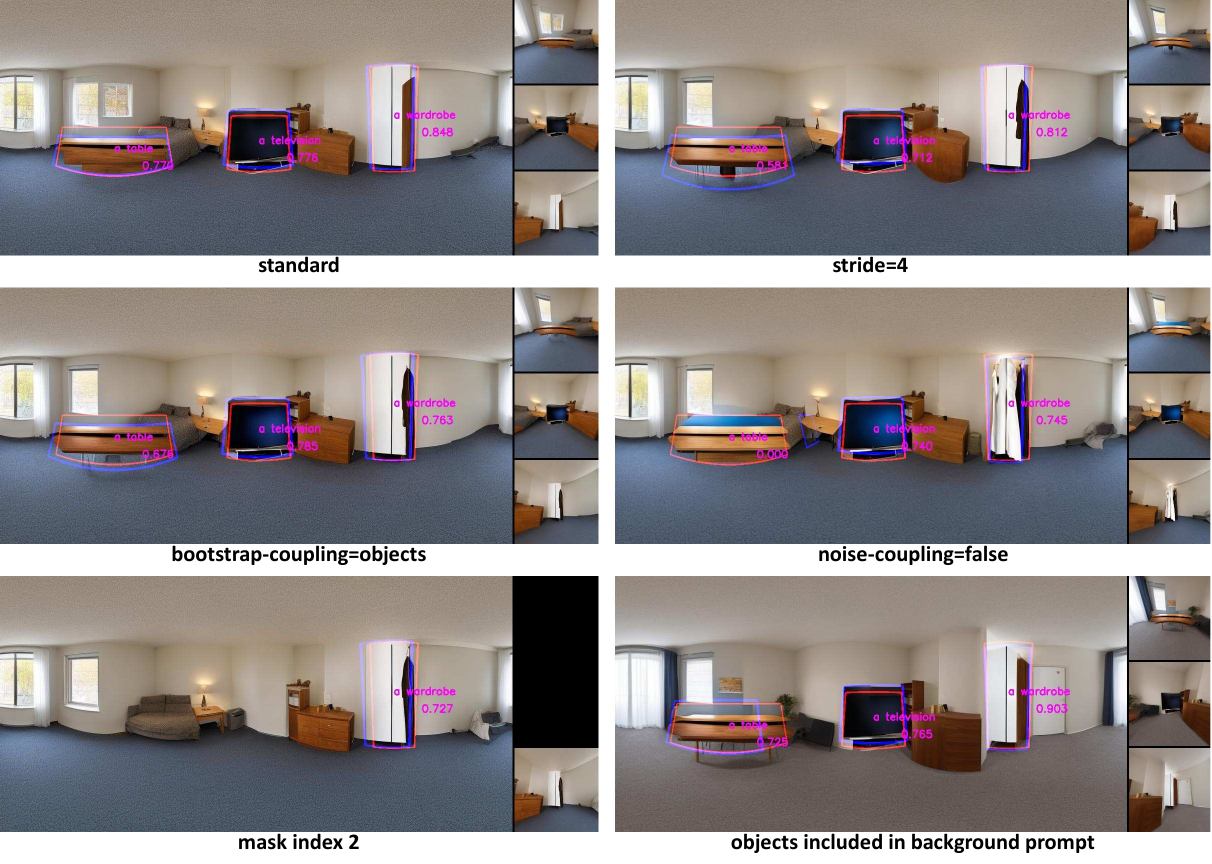}
    \caption{Influence of other parameters. The comparison between the top-left and bottom-left images shows that adding more objects doesn't influence the background's or other object's look.}
    \label{fig:stitch_other}
\end{figure}

\subsection{More Qualitative Results with MultiPanFusion} \label{sec:qualitative_mstd}
\begin{figure}[!ht]
    \centering
    \includegraphics[width=1\linewidth]{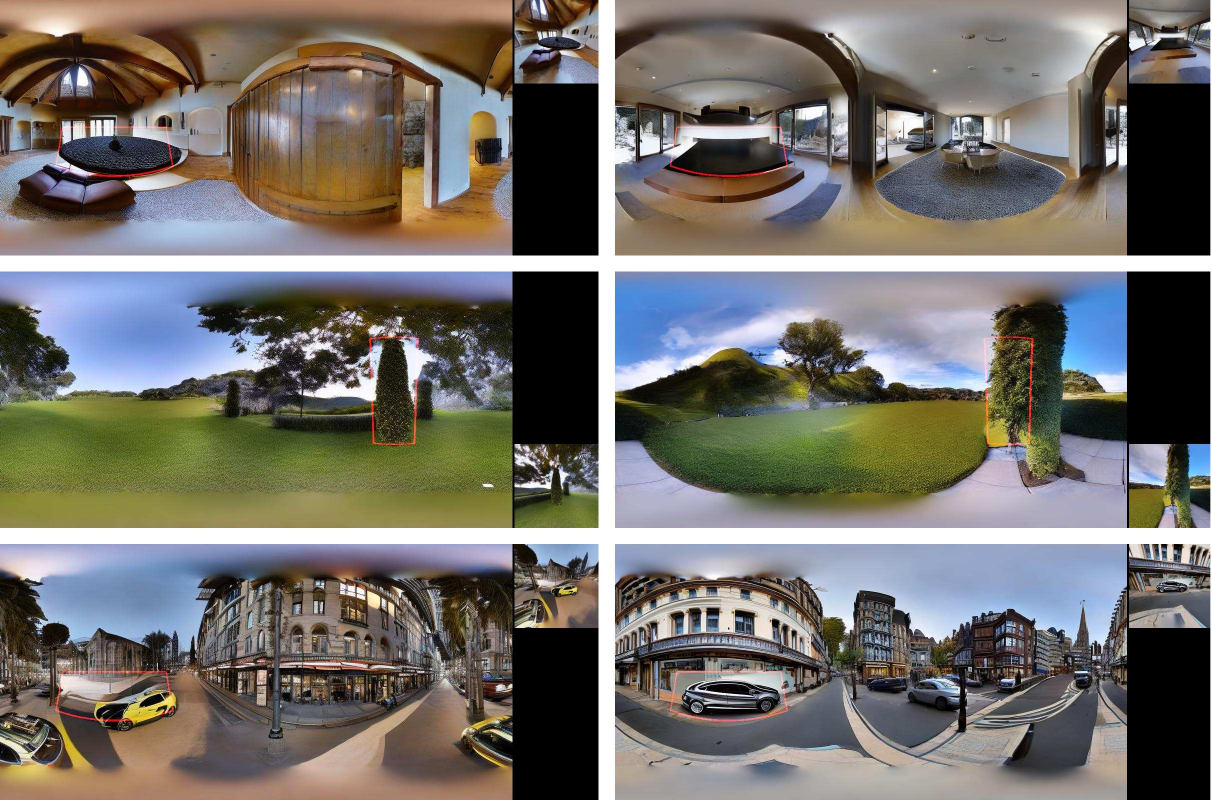}
    \caption{Image-quality and -diversity of MPF. Contrarily to MSTD, images are filled with rich and varied content here but show some quality-reducing artifacts, like the blurry area behind the car.}
    \label{fig:panfusion_quality_diversity}
\end{figure}
\begin{figure}[!ht]
    \centering
    \includegraphics[width=1\linewidth]{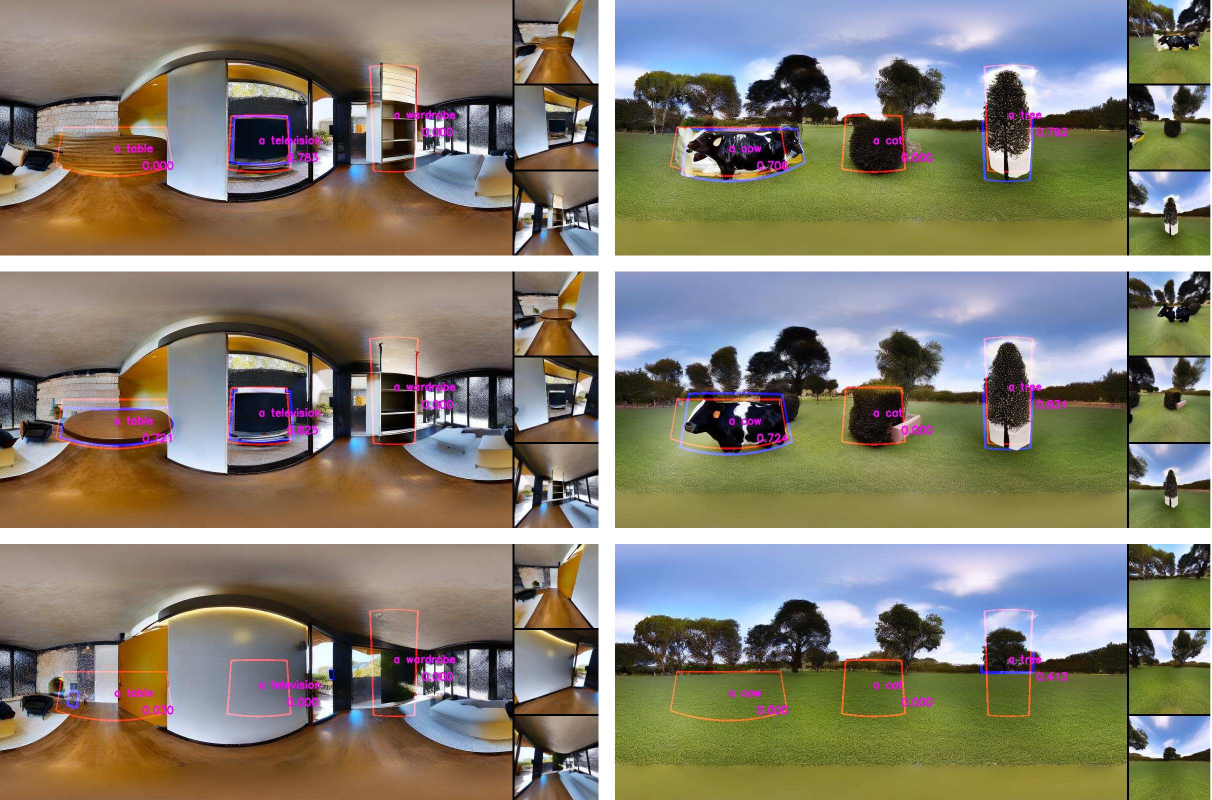}
    \caption{Top: MD applied at the panorama branch only. Middle: MD applied at both branches. Bottom: MD applied at the perspective branch only, which fails to synthesize objects.}
    \label{fig:panfusion_branches}
\end{figure}
\begin{figure}[!ht]
    \centering
    \includegraphics[width=1\linewidth]{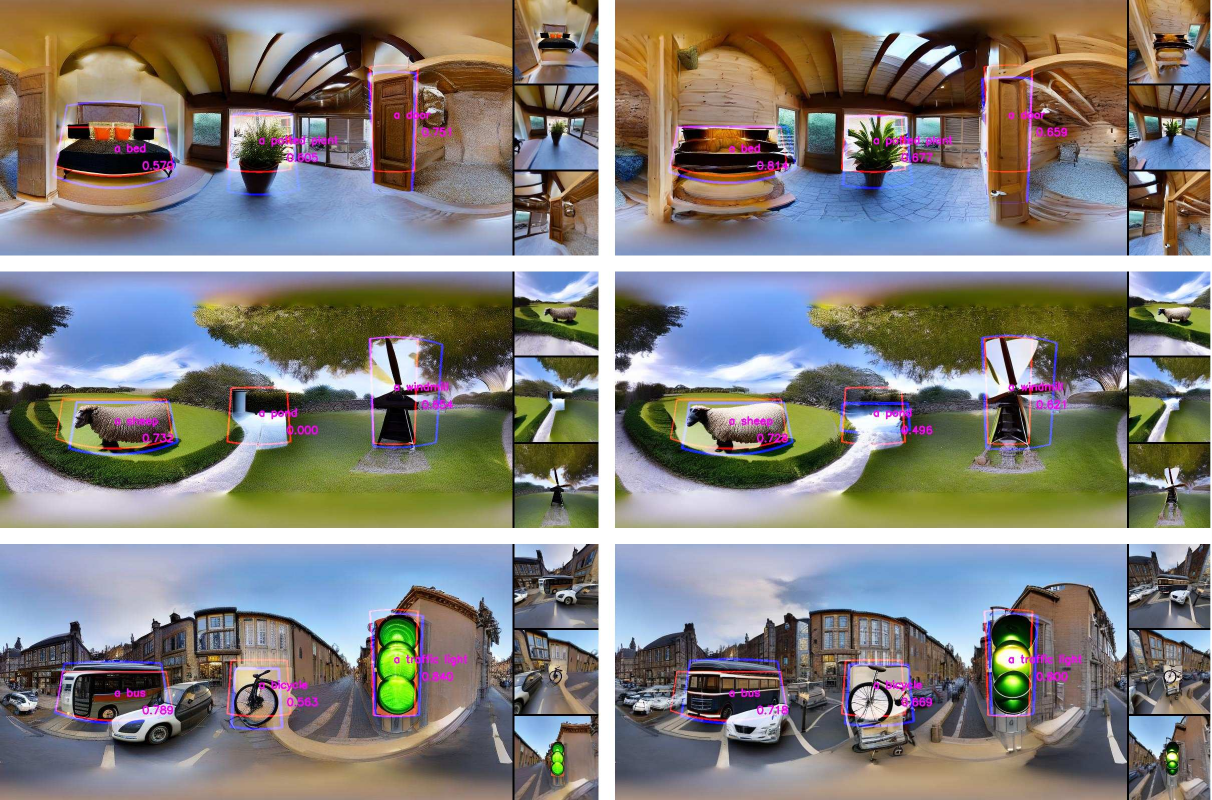}
    \caption{Left: EPPA employed at all paths. Right: EPPA employed only at the background during bootstrapping. The pond can only be recognized in the right image. Such cases cause the IoU to increase in this setting.}
    \label{fig:panfusion_eppa}
\end{figure}
\begin{figure}[!ht]
    \centering
    \includegraphics[width=1\linewidth]{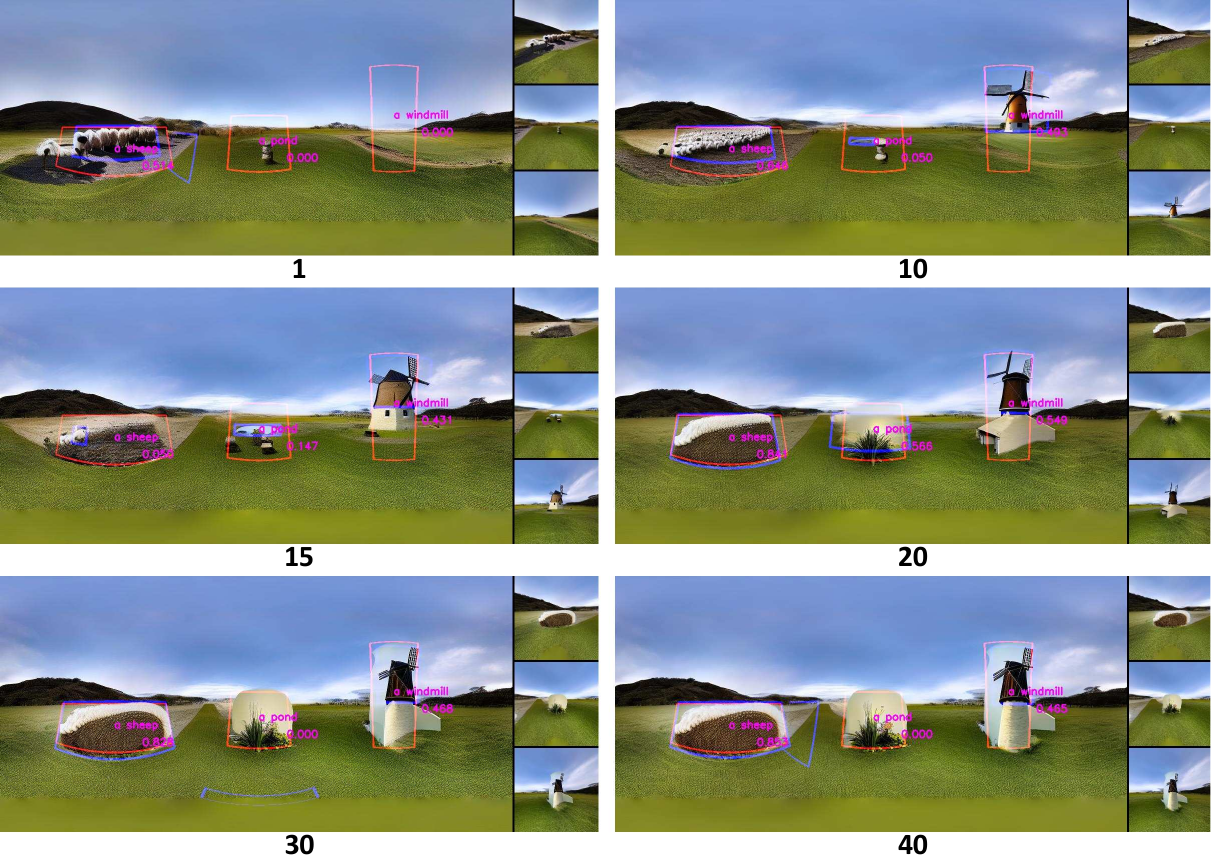}
    \caption{Influence of bootstrapping on MPF, showing similar effects as with MSTD.}
    \label{fig:panfusion_bootstrap}
\end{figure}
\begin{figure}[!ht]
    \centering
    \includegraphics[width=1\linewidth]{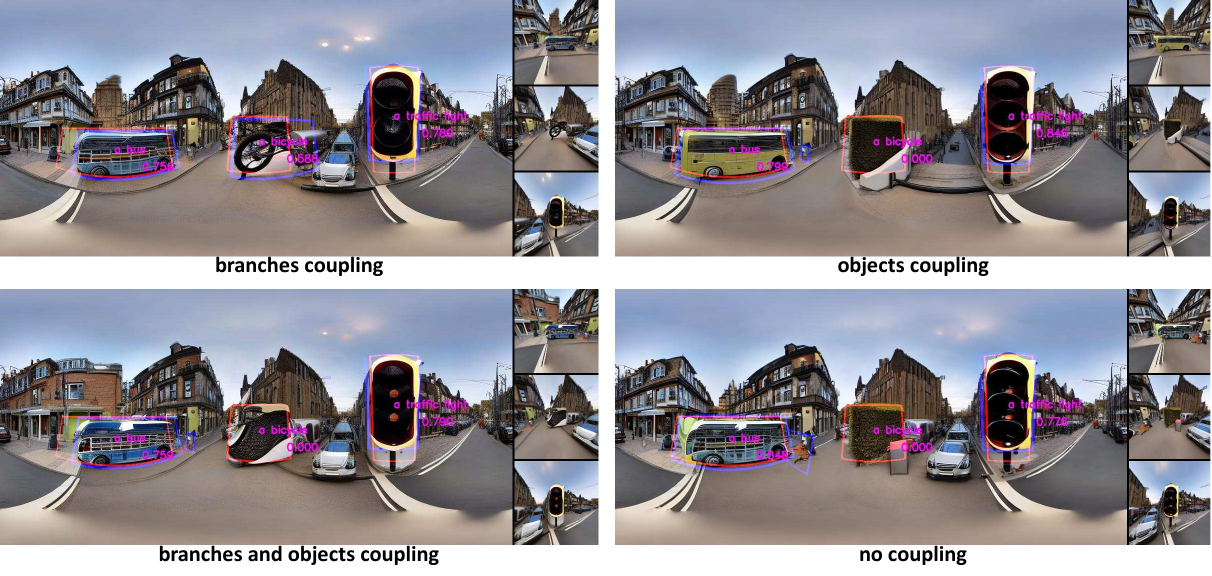}
    \caption{Influence of bootstrap-coupling. The largest effect can be seen at the middle object, where the bicycle only becomes visible with branches coupling}
    \label{fig:panfusion_coupling}
\end{figure}

\begin{figure}[!ht]
    \centering
    \includegraphics[width=0.15\linewidth]{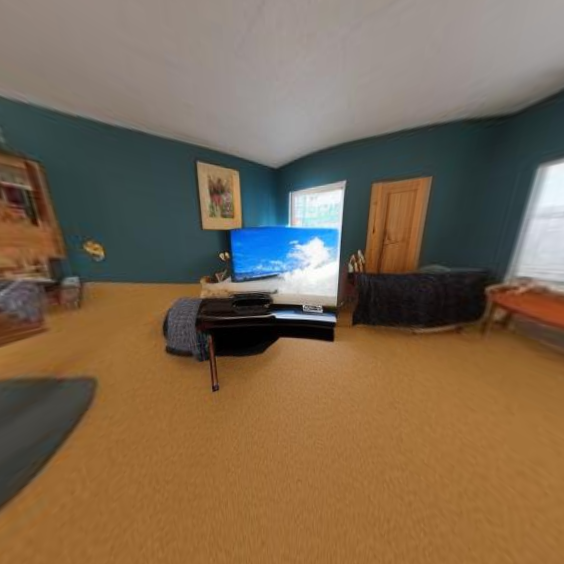}
    \includegraphics[width=0.15\linewidth]{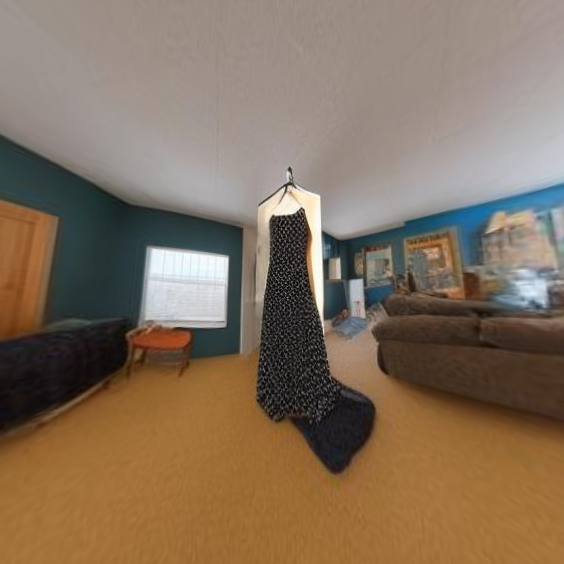}
    \includegraphics[width=0.15\linewidth]{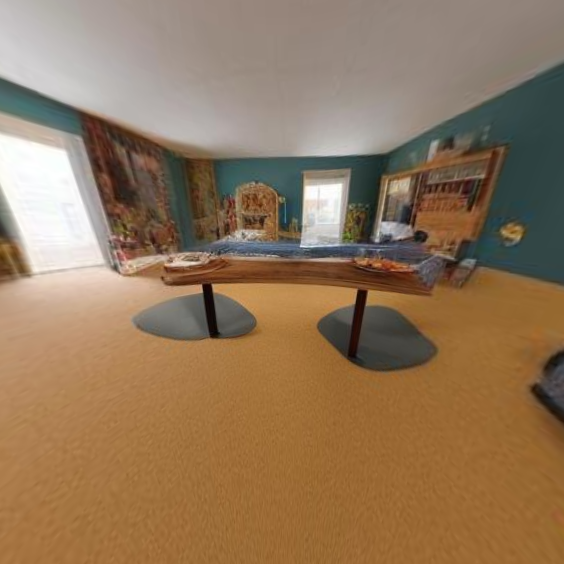}
    \includegraphics[width=0.15\linewidth]{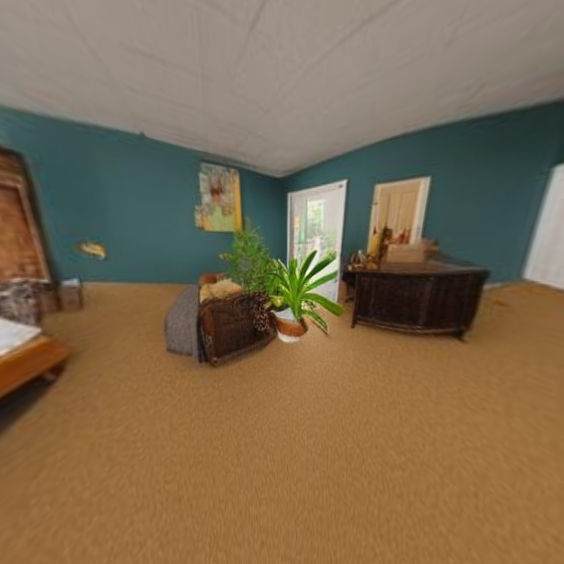}
    \includegraphics[width=0.15\linewidth]{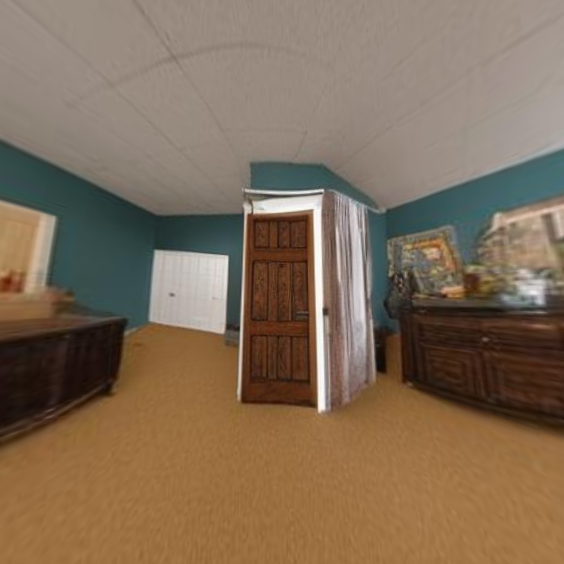}
    \includegraphics[width=0.15\linewidth]{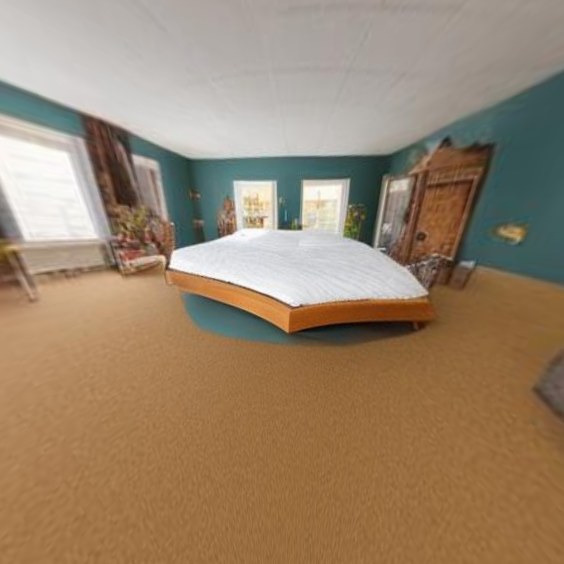}
    \hfill
    \includegraphics[width=0.15\linewidth]{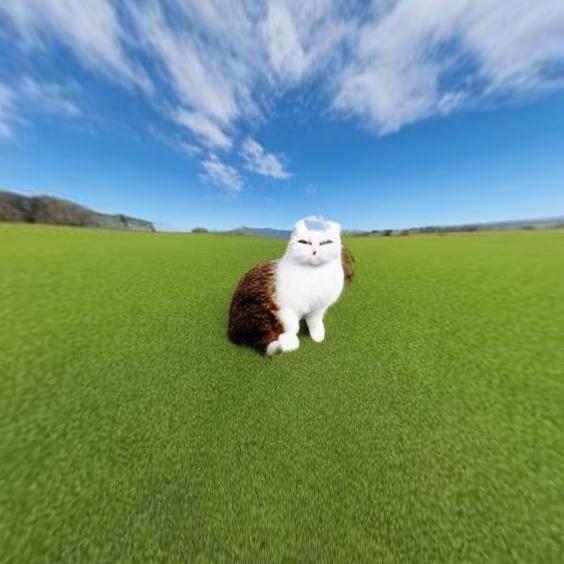}
    \includegraphics[width=0.15\linewidth]{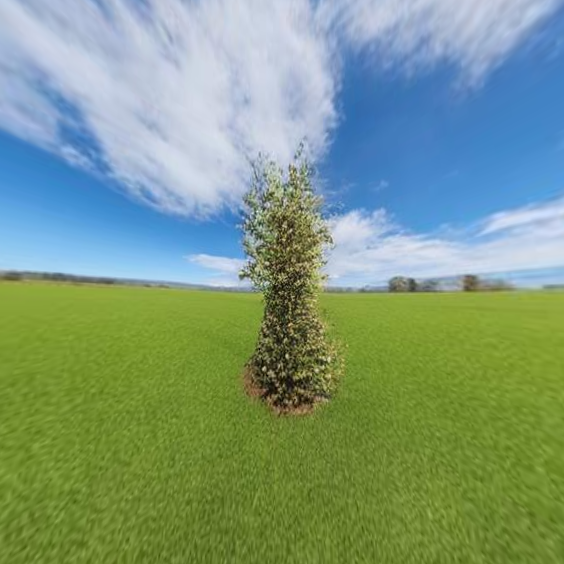}
    \includegraphics[width=0.15\linewidth]{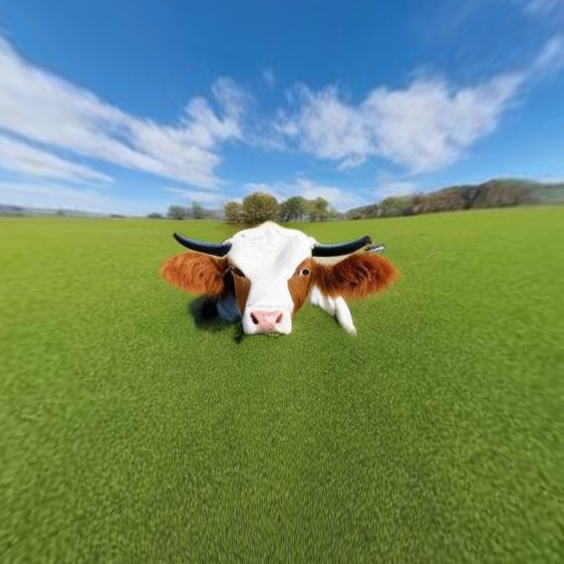}
    \includegraphics[width=0.15\linewidth]{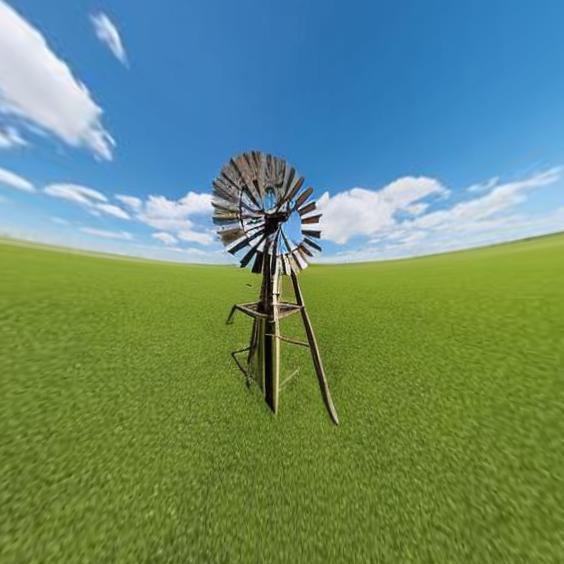}
    \includegraphics[width=0.15\linewidth]{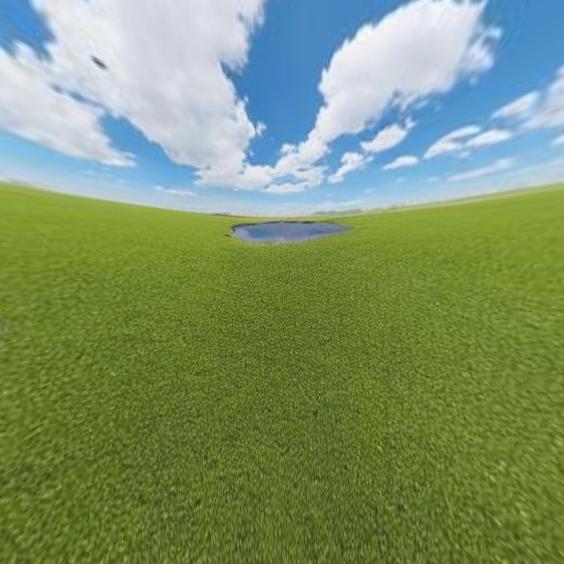}
    \includegraphics[width=0.15\linewidth]{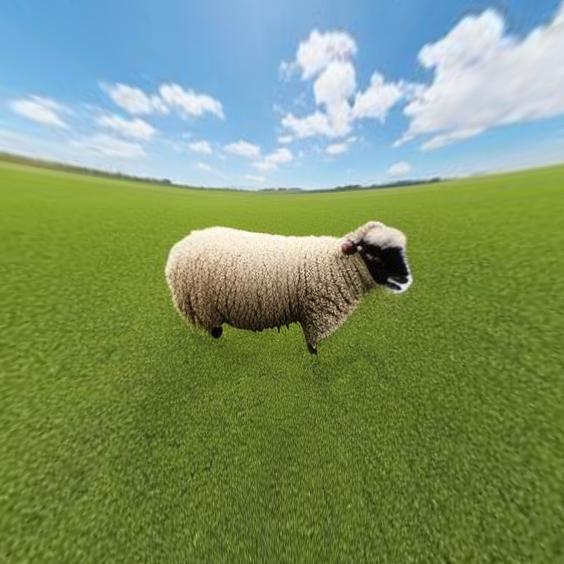}
    \includegraphics[width=0.15\linewidth]{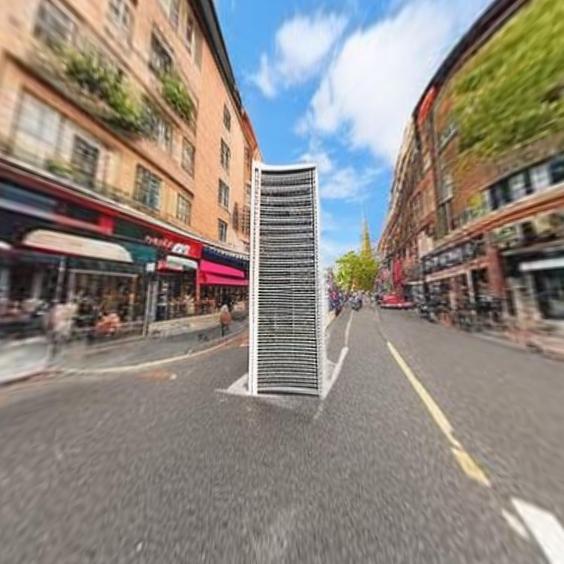}
    \includegraphics[width=0.15\linewidth]{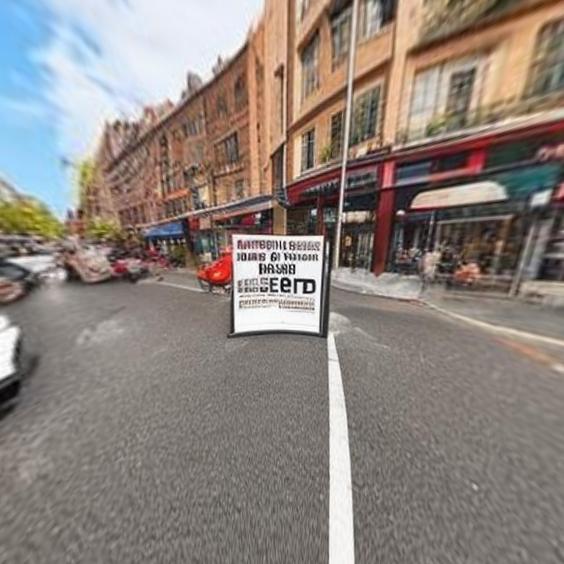}
    \includegraphics[width=0.15\linewidth]{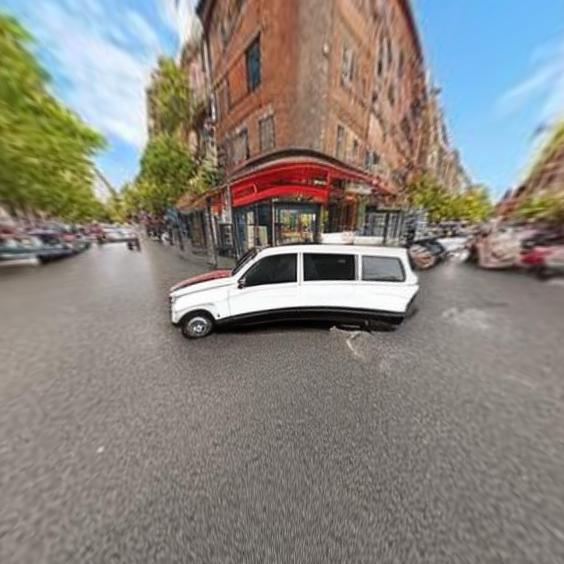}
    \includegraphics[width=0.15\linewidth]{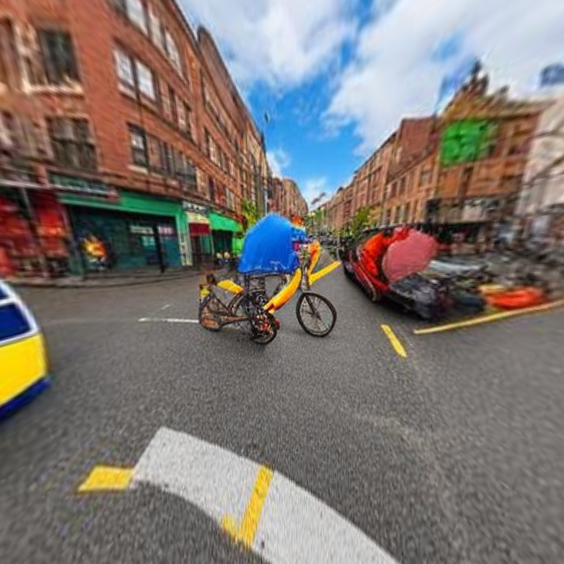}
    \includegraphics[width=0.15\linewidth]{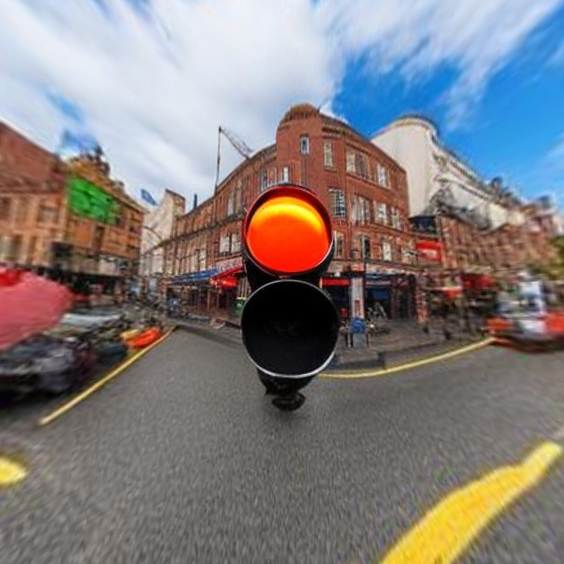}
    \includegraphics[width=0.15\linewidth]{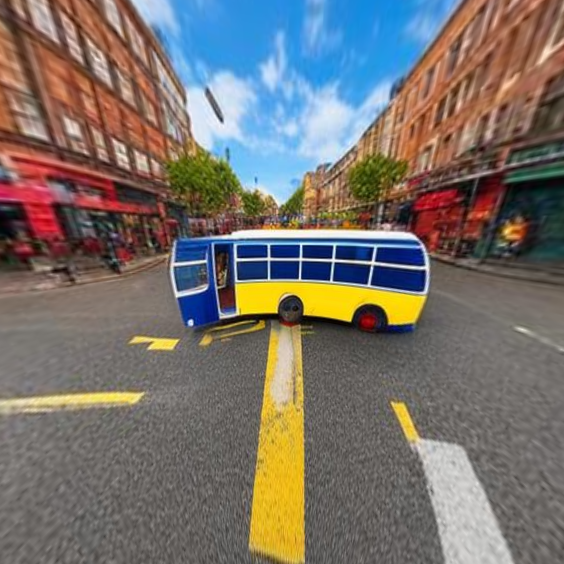}
    \includegraphics[width=0.15\linewidth]{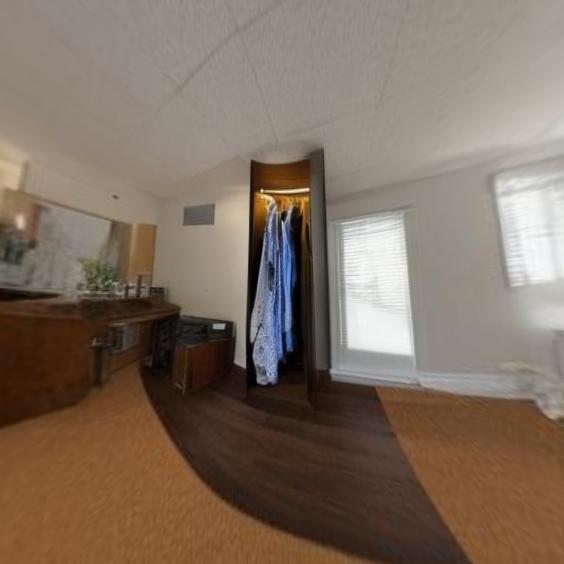}
    \includegraphics[width=0.15\linewidth]{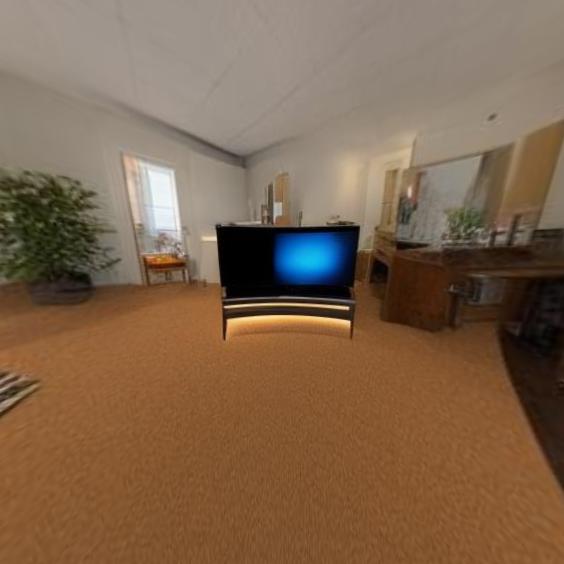}
    \includegraphics[width=0.15\linewidth]{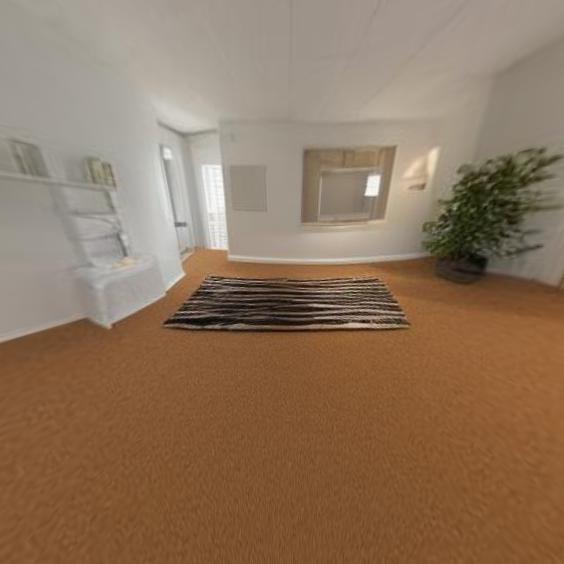}
    \includegraphics[width=0.15\linewidth]{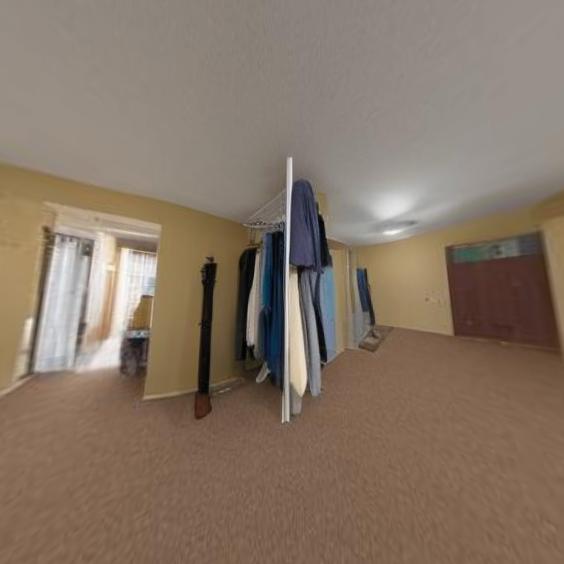}
    \includegraphics[width=0.15\linewidth]{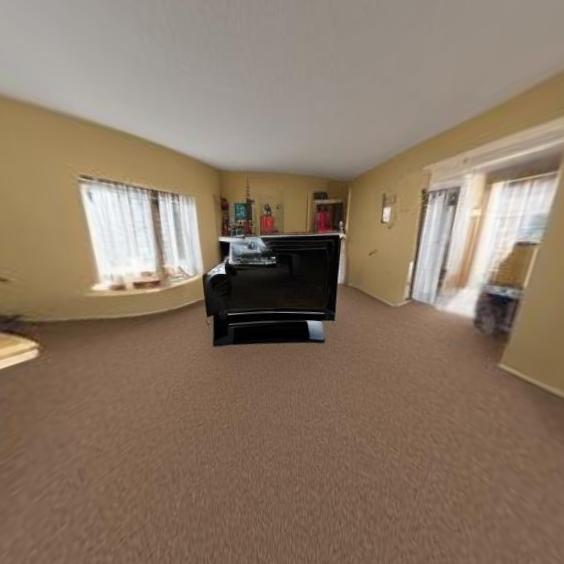}
    \includegraphics[width=0.15\linewidth]{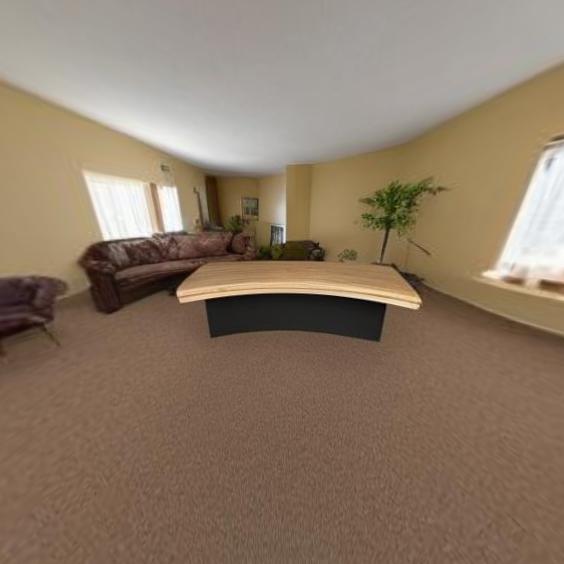}
    \includegraphics[width=0.15\linewidth]{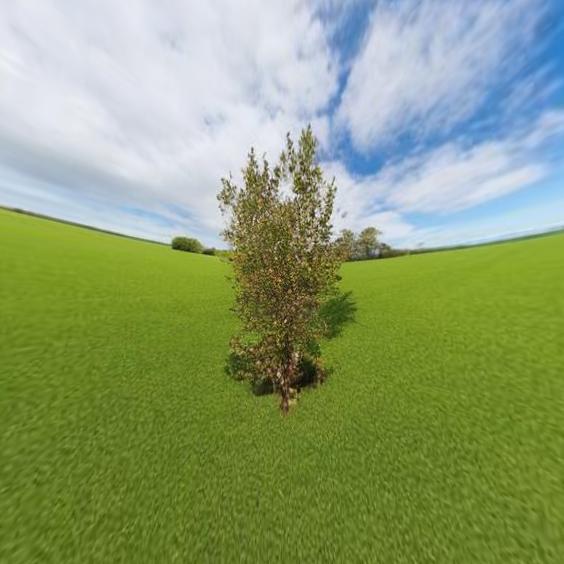}
    \includegraphics[width=0.15\linewidth]{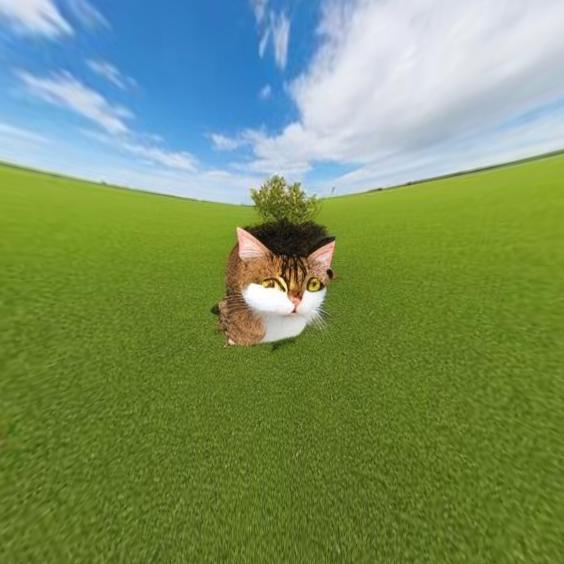}
    \includegraphics[width=0.15\linewidth]{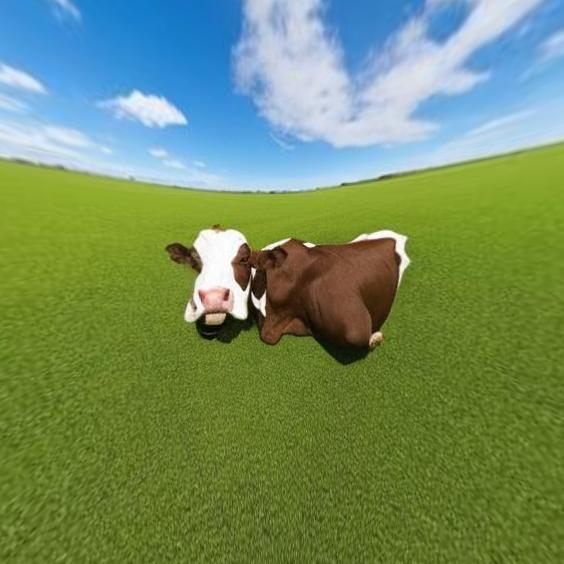}
    \includegraphics[width=0.15\linewidth]{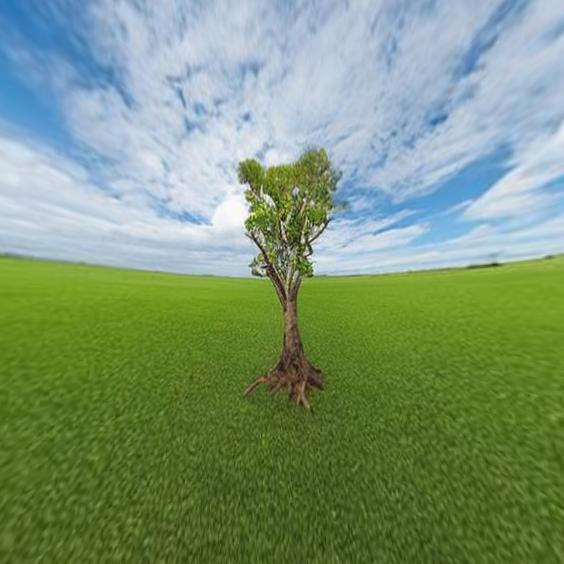}
    \includegraphics[width=0.15\linewidth]{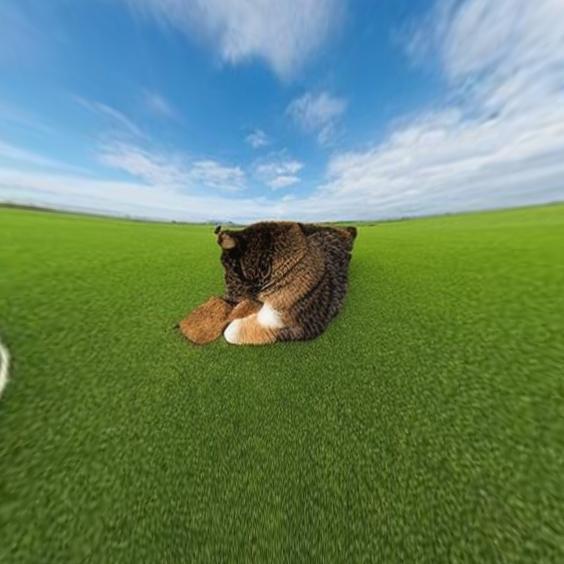}
    \includegraphics[width=0.15\linewidth]{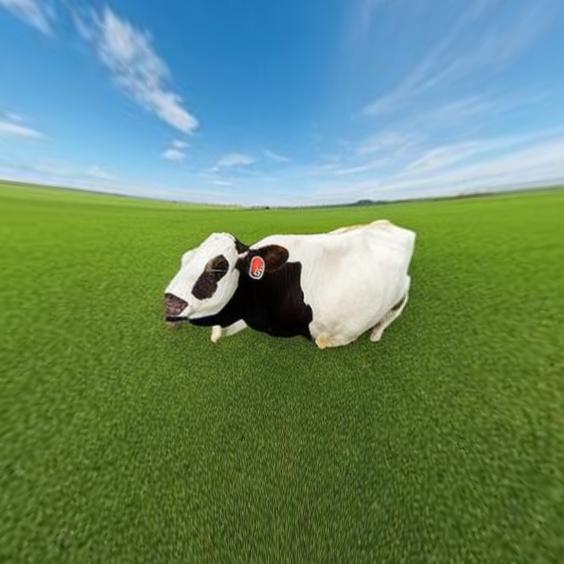}
    \includegraphics[width=0.15\linewidth]{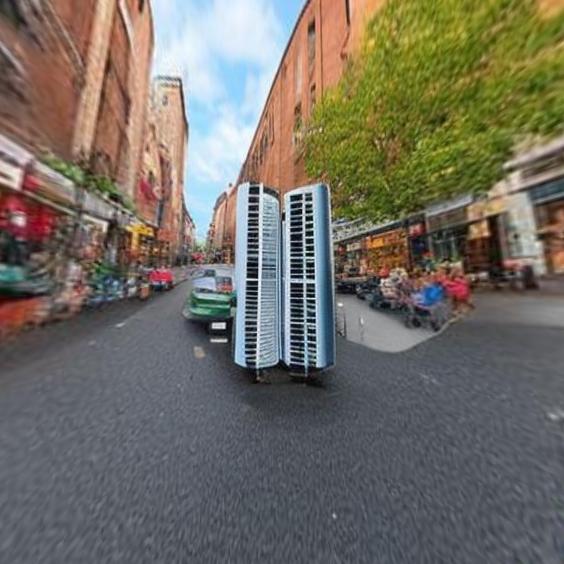}
    \includegraphics[width=0.15\linewidth]{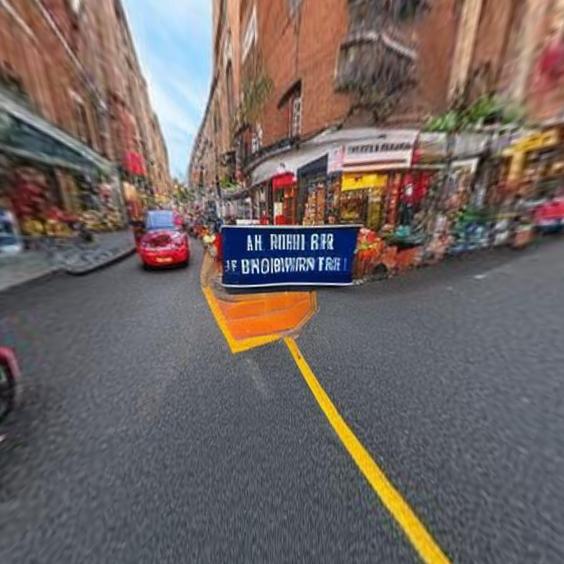}
    \includegraphics[width=0.15\linewidth]{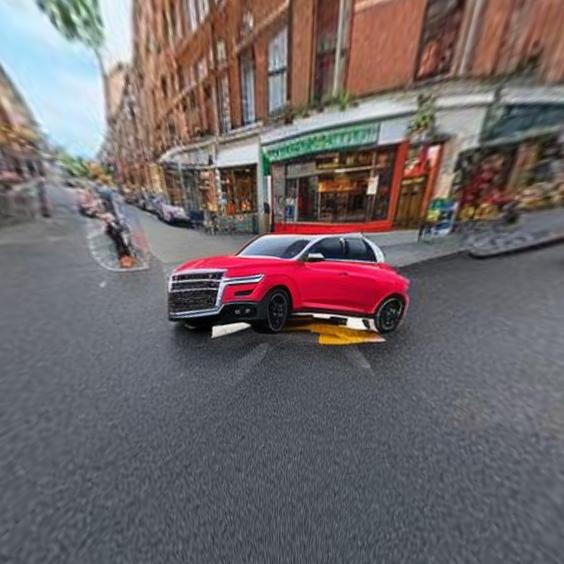}
    \includegraphics[width=0.15\linewidth]{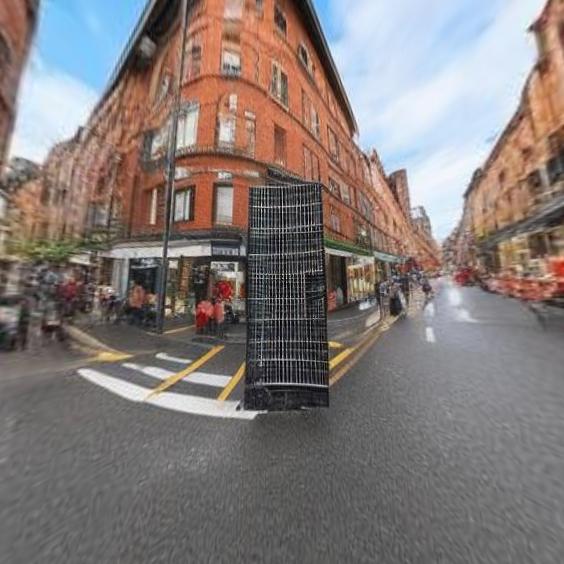}
    \includegraphics[width=0.15\linewidth]{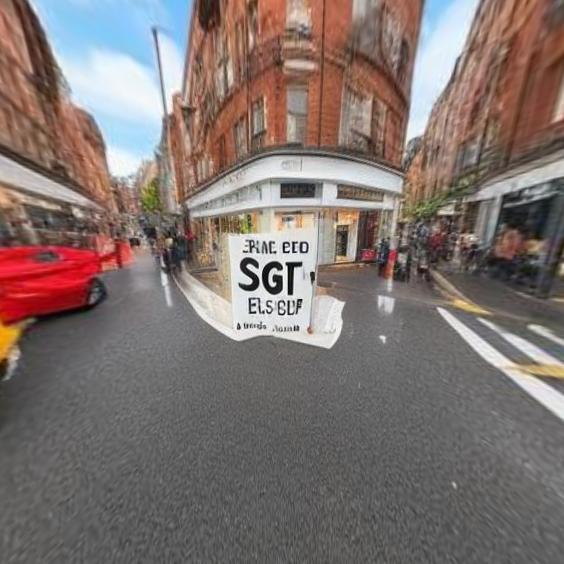}
    \includegraphics[width=0.15\linewidth]{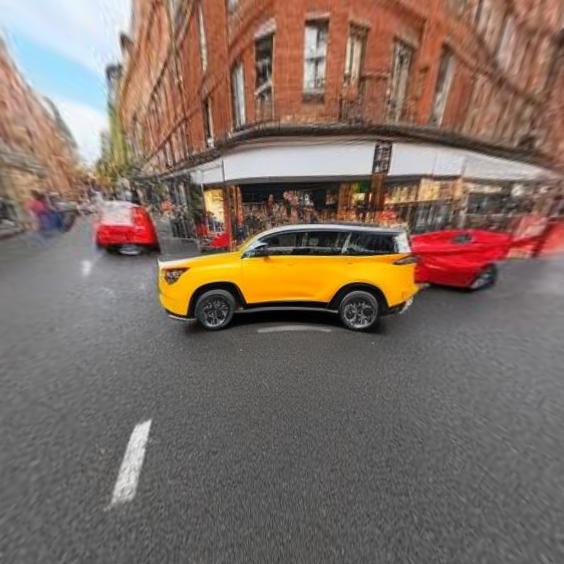}
    \includegraphics[width=0.15\linewidth]{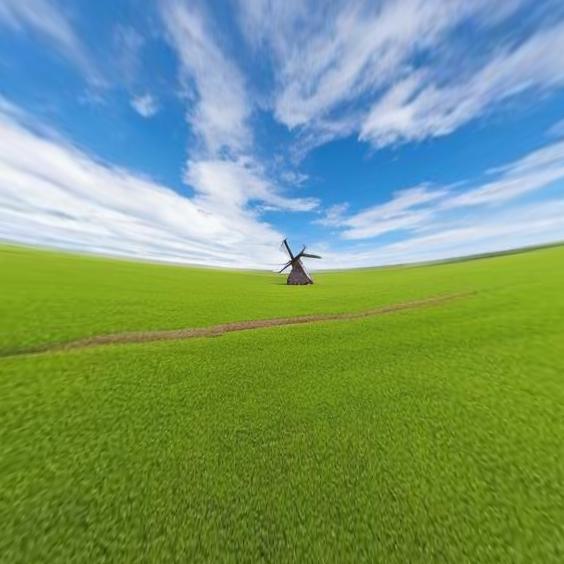}
    \includegraphics[width=0.15\linewidth]{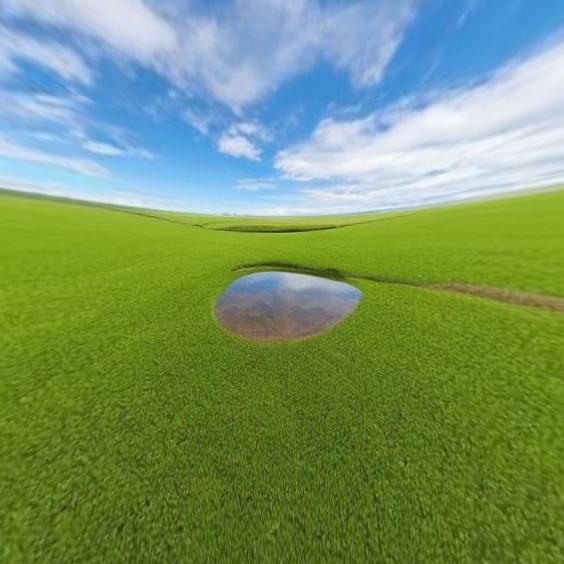}
    \includegraphics[width=0.15\linewidth]{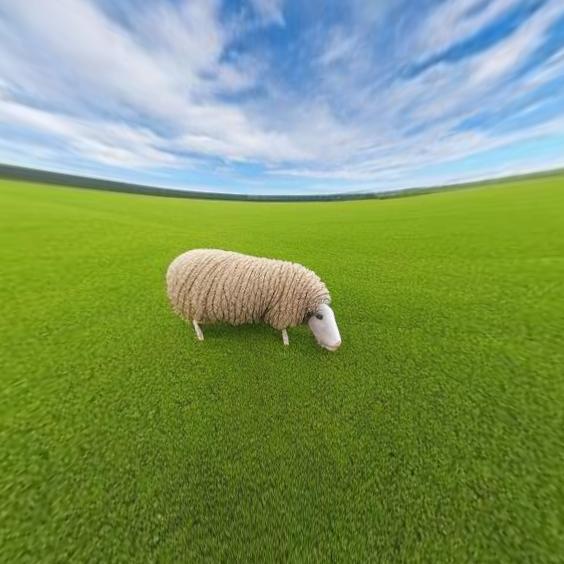}
    \includegraphics[width=0.15\linewidth]{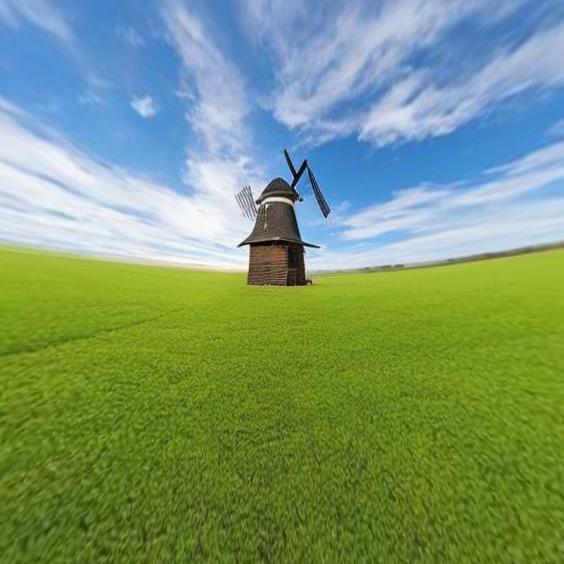}
    \includegraphics[width=0.15\linewidth]{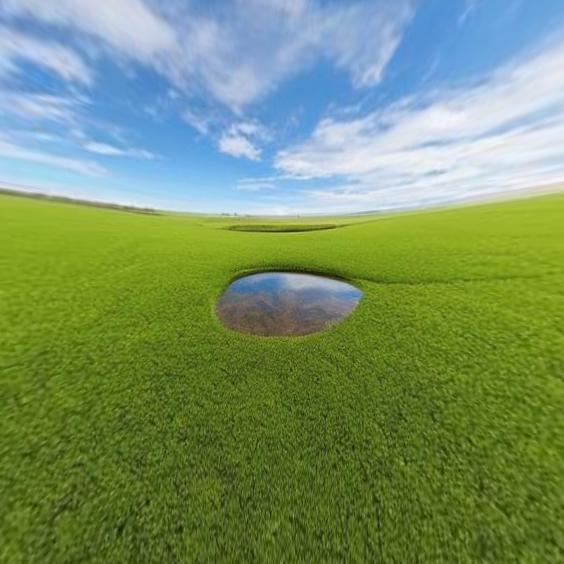}
    \includegraphics[width=0.15\linewidth]{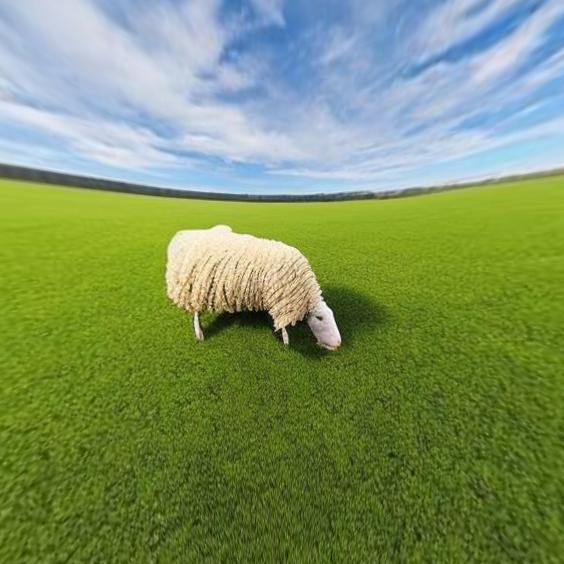}
    \includegraphics[width=0.15\linewidth]{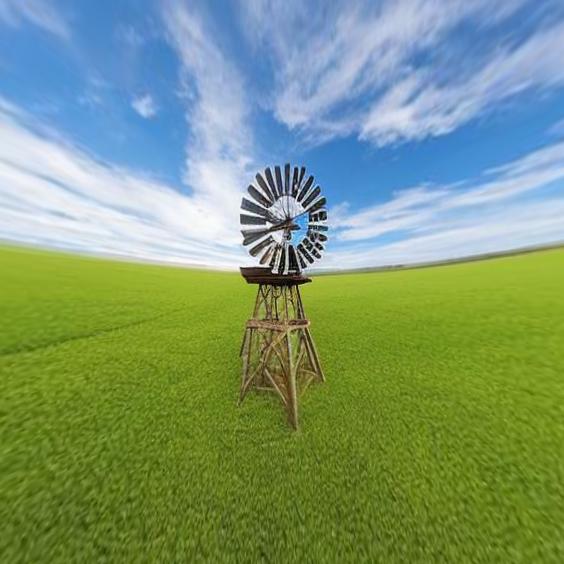}
    \includegraphics[width=0.15\linewidth]{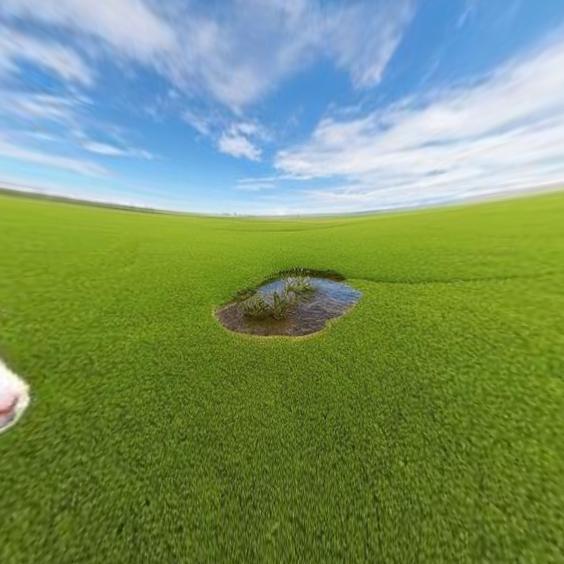}
    \includegraphics[width=0.15\linewidth]{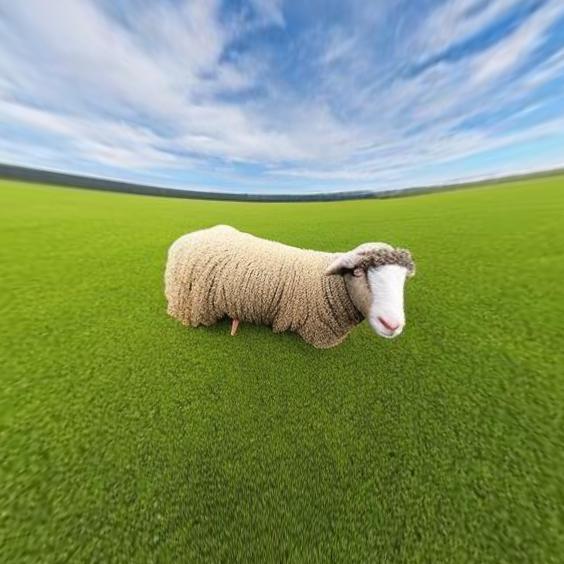}
    \includegraphics[width=0.15\linewidth]{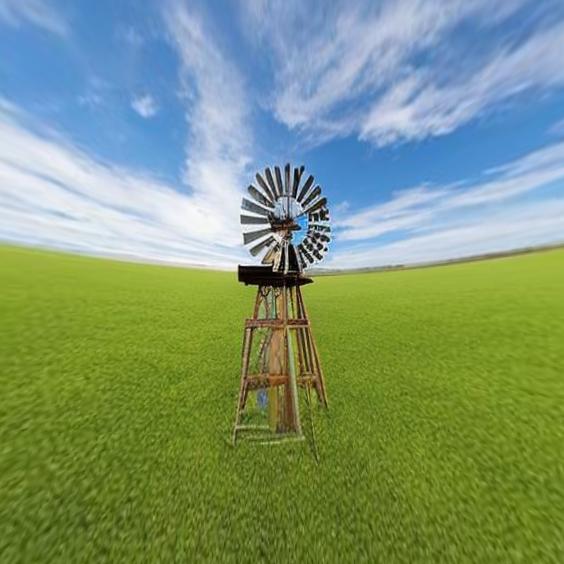}
    \includegraphics[width=0.15\linewidth]{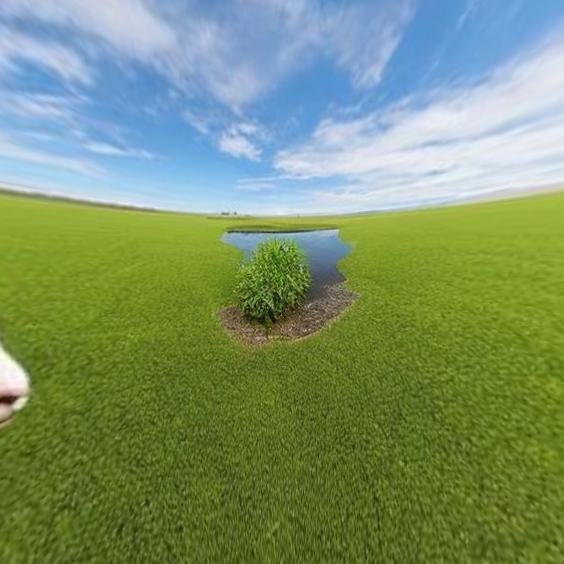}
    \includegraphics[width=0.15\linewidth]{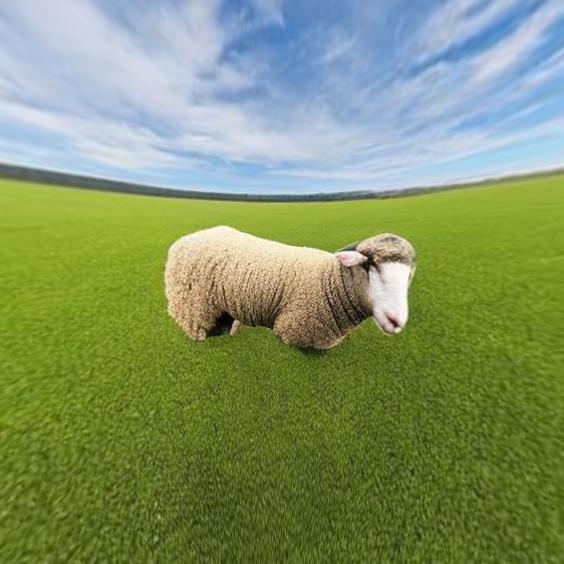}
    \includegraphics[width=0.15\linewidth]{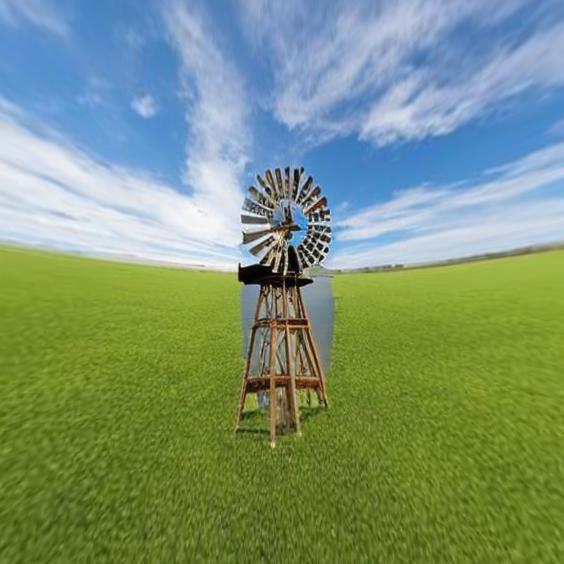}
    \includegraphics[width=0.15\linewidth]{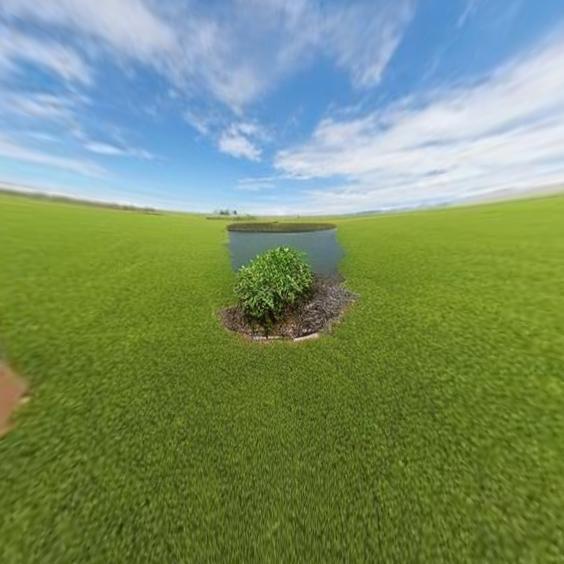}
    \includegraphics[width=0.15\linewidth]{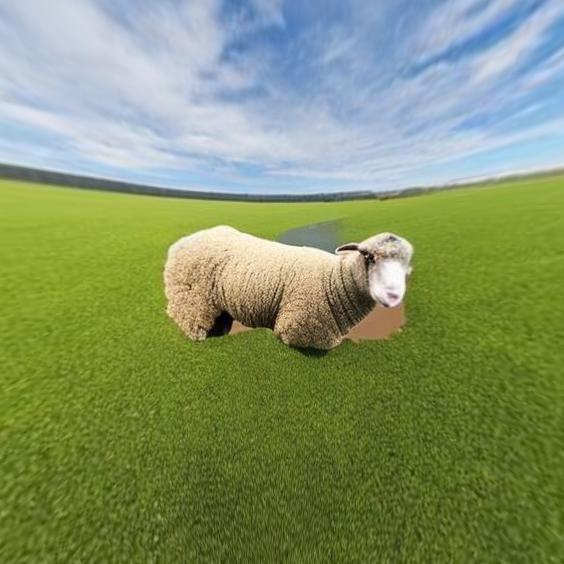}
    \includegraphics[width=0.15\linewidth]{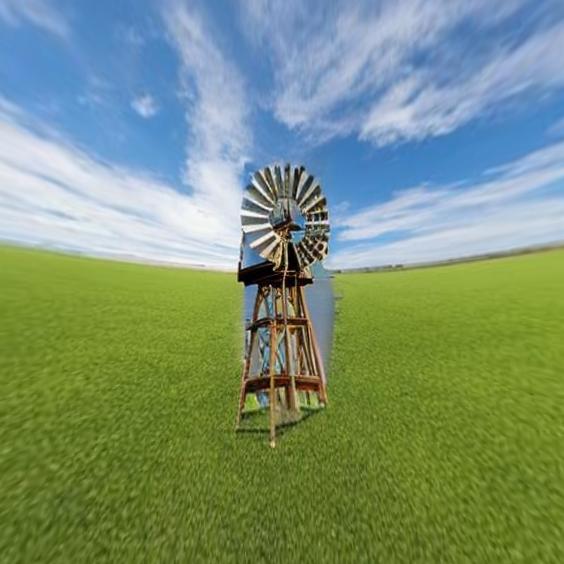}
    \includegraphics[width=0.15\linewidth]{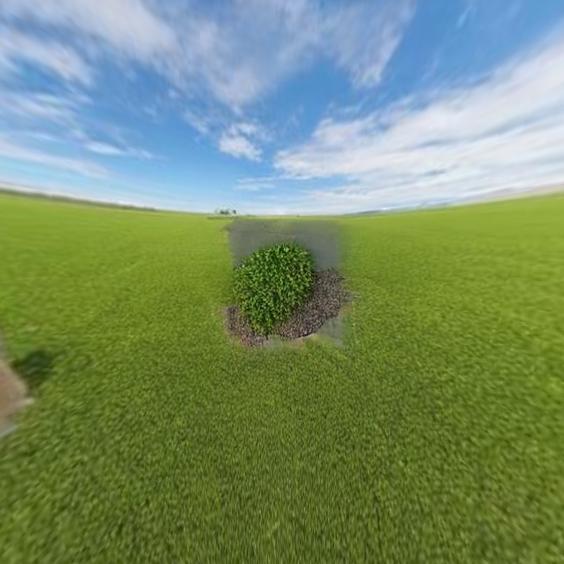}
    \includegraphics[width=0.15\linewidth]{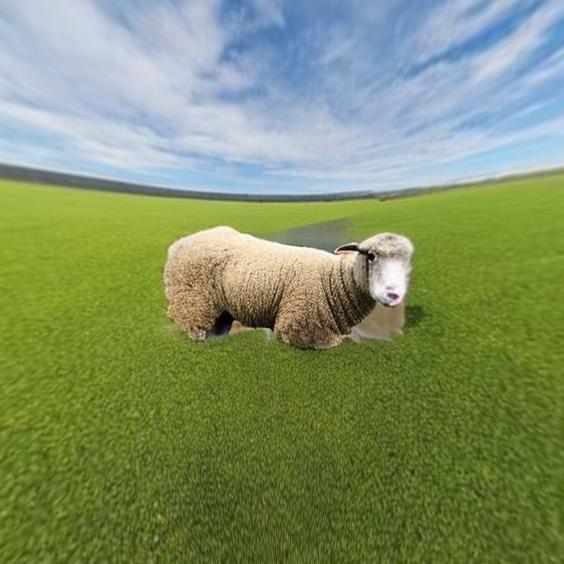}
    \includegraphics[width=0.15\linewidth]{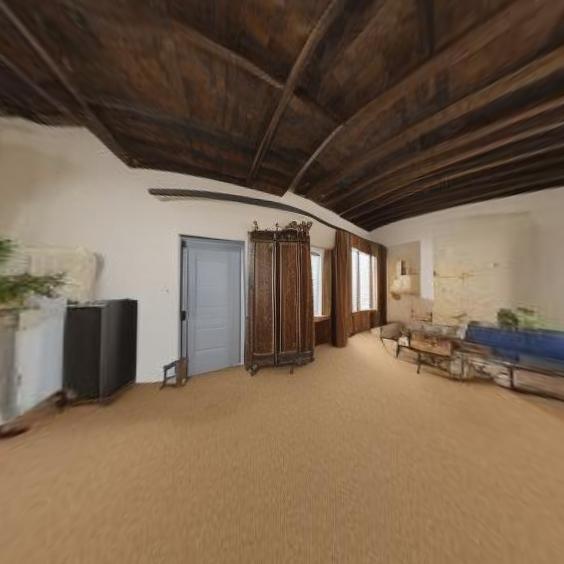}
    \includegraphics[width=0.15\linewidth]{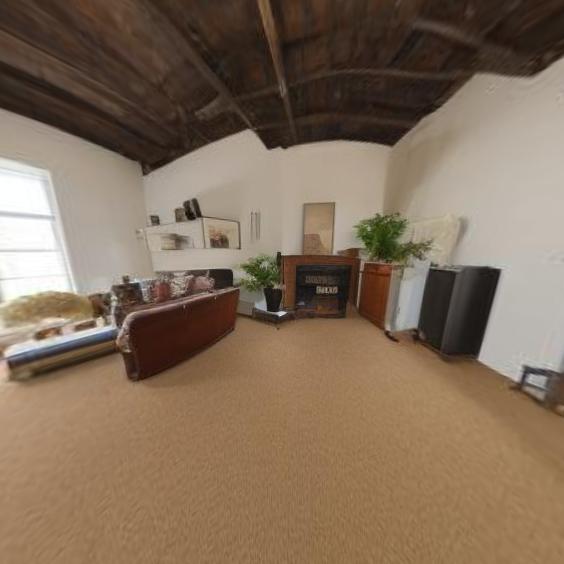}
    \includegraphics[width=0.15\linewidth]{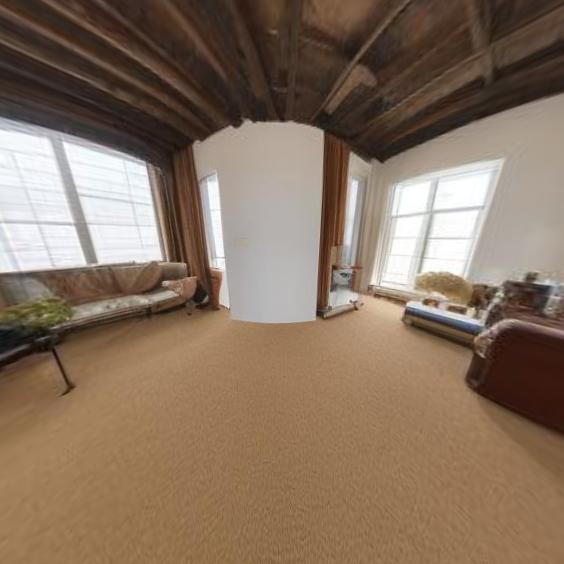}
    \includegraphics[width=0.15\linewidth]{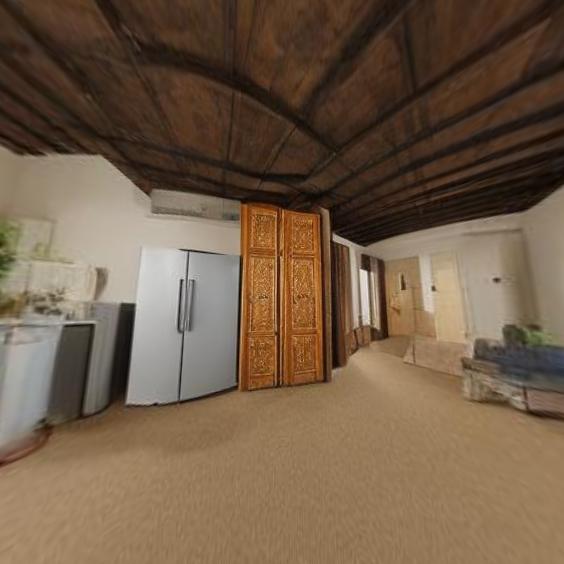}
    \includegraphics[width=0.15\linewidth]{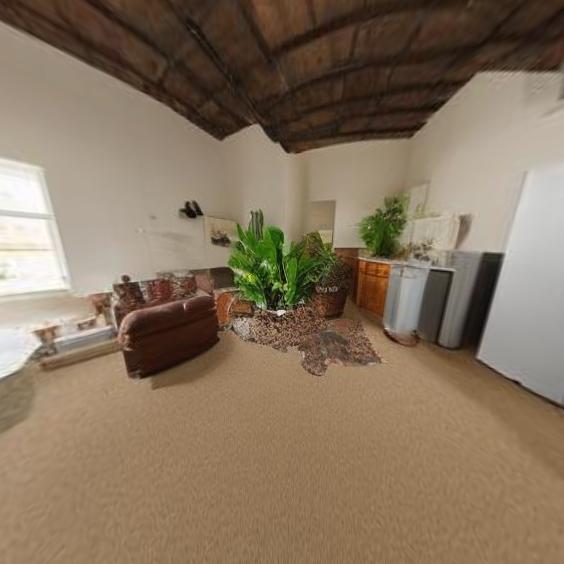}
    \includegraphics[width=0.15\linewidth]{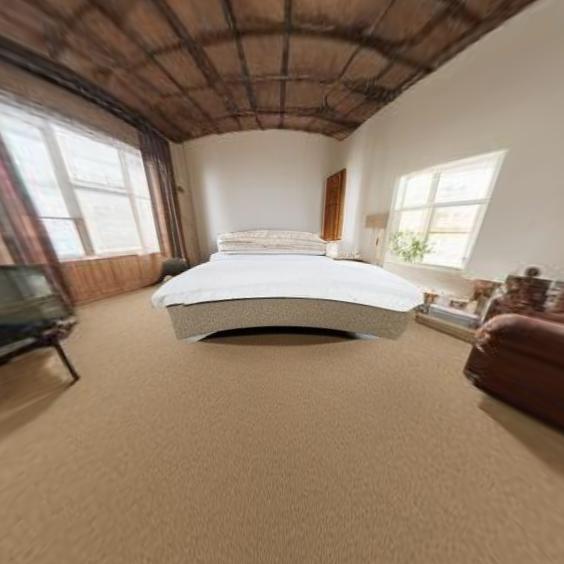}
    \includegraphics[width=0.15\linewidth]{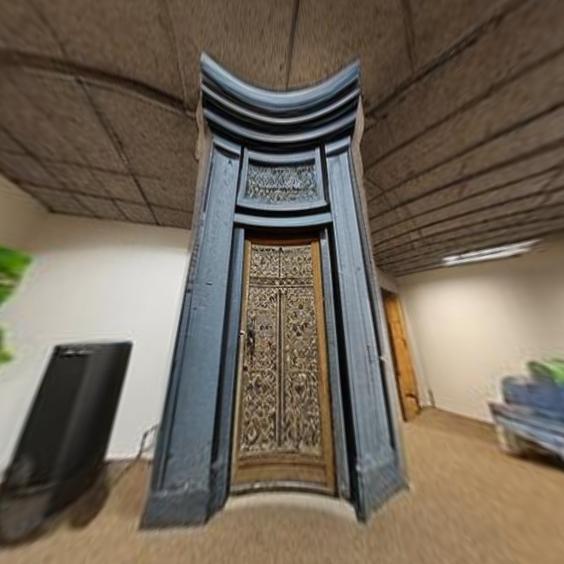}
    \includegraphics[width=0.15\linewidth]{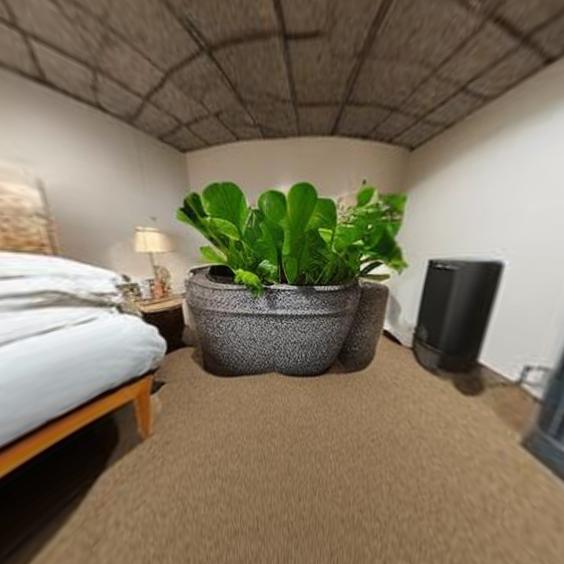}
    \includegraphics[width=0.15\linewidth]{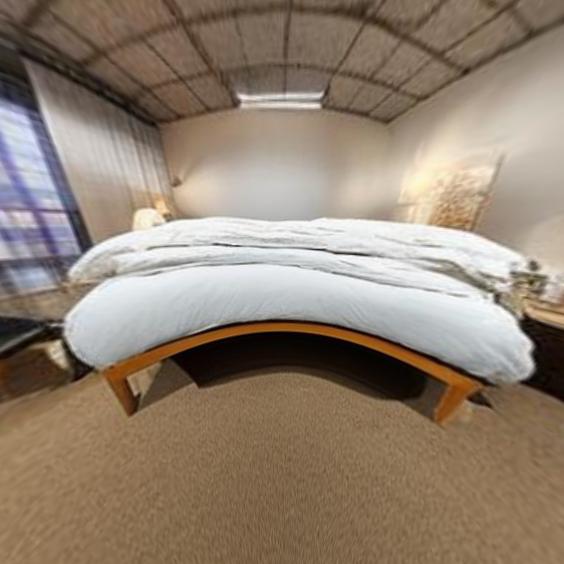}
    \includegraphics[width=0.15\linewidth]{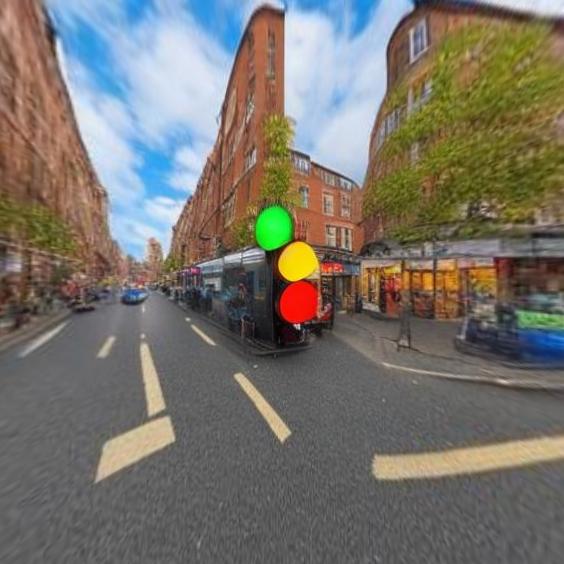}
    \includegraphics[width=0.15\linewidth]{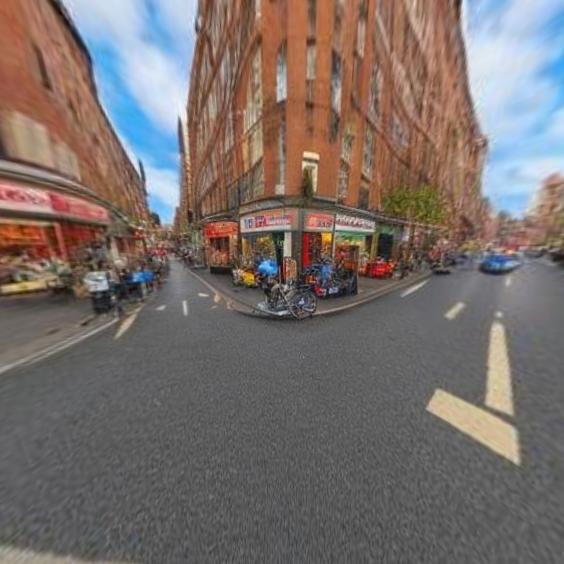}
    \includegraphics[width=0.15\linewidth]{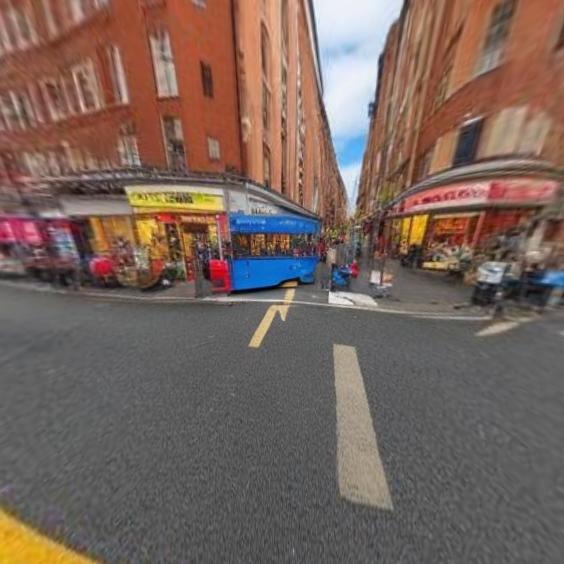}
    \includegraphics[width=0.15\linewidth]{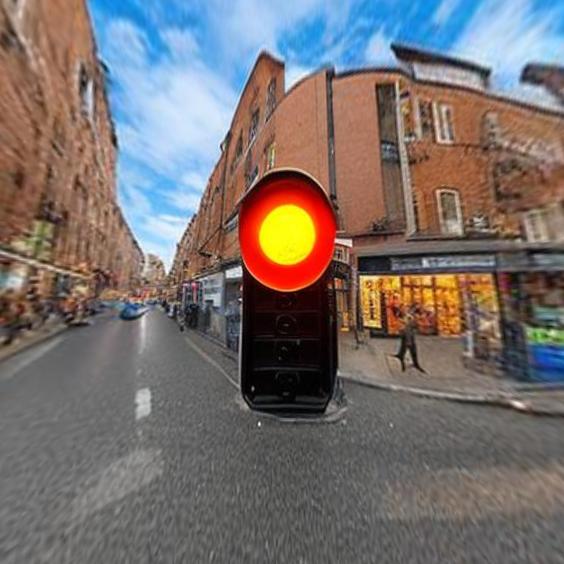}
    \includegraphics[width=0.15\linewidth]{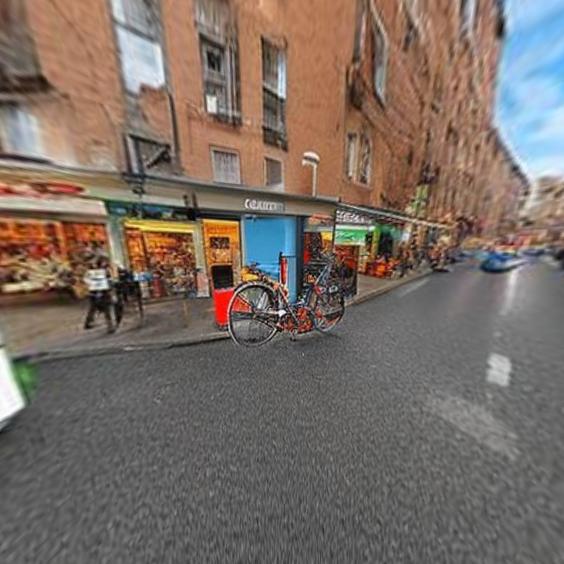}
    \includegraphics[width=0.15\linewidth]{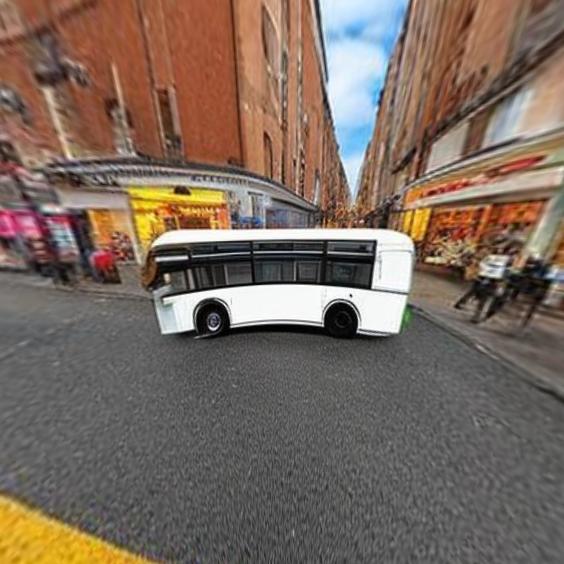}
    \includegraphics[width=0.15\linewidth]{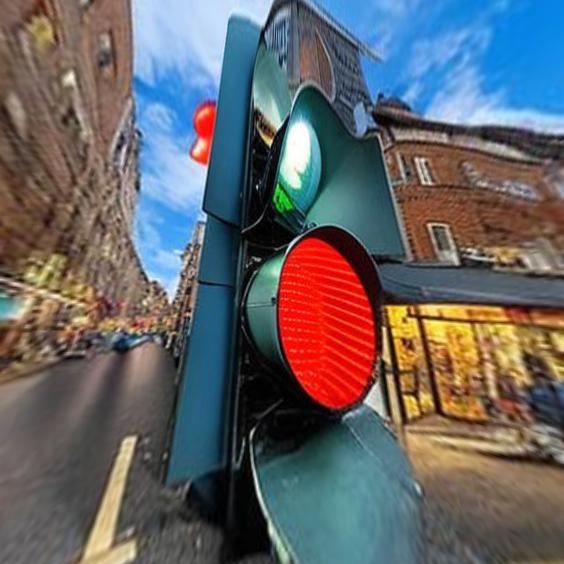}
    \includegraphics[width=0.15\linewidth]{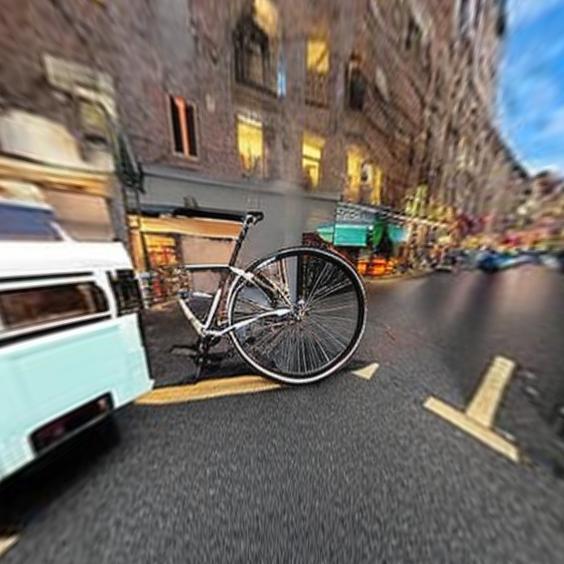}
    \includegraphics[width=0.15\linewidth]{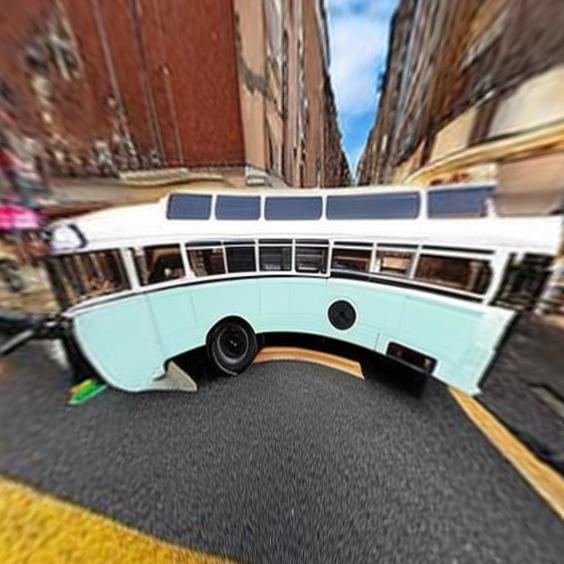}
    \includegraphics[width=0.15\linewidth]{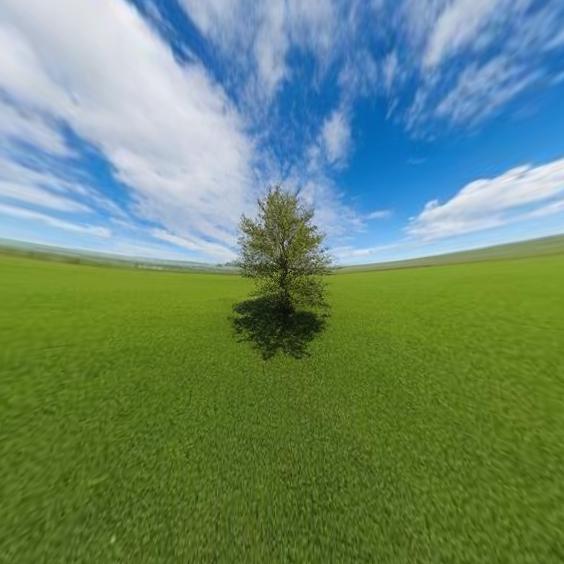}
    \includegraphics[width=0.15\linewidth]{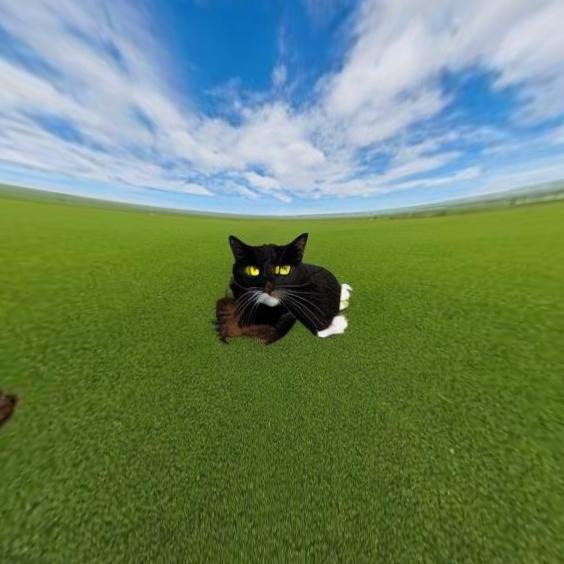}
    \includegraphics[width=0.15\linewidth]{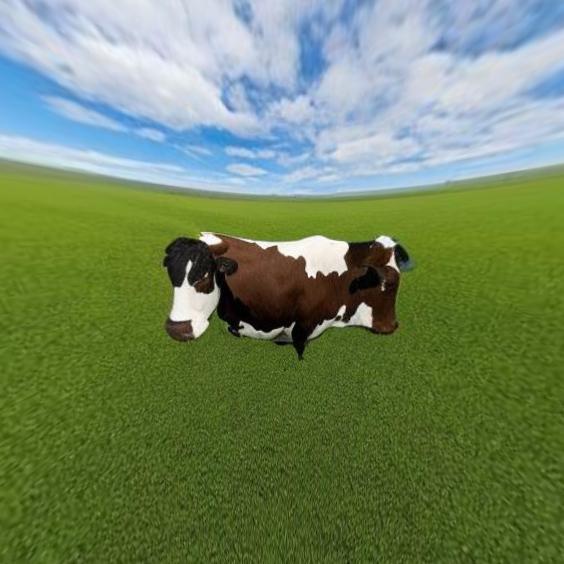}
    \includegraphics[width=0.15\linewidth]{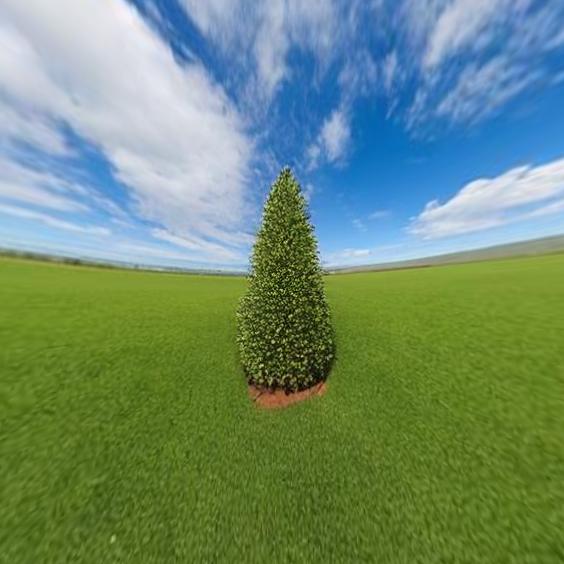}
    \includegraphics[width=0.15\linewidth]{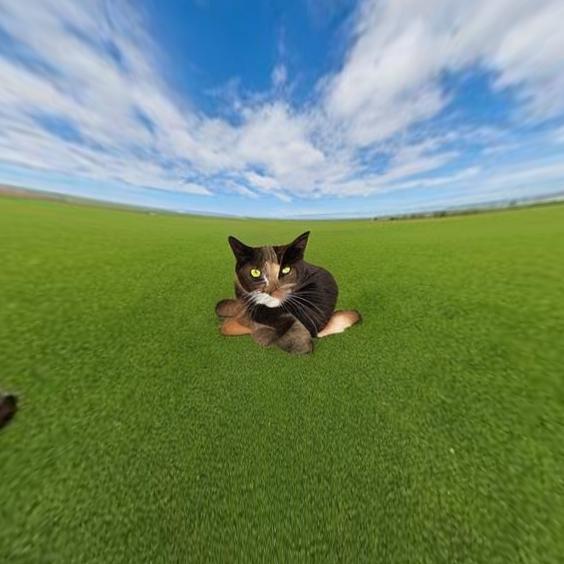}
    \includegraphics[width=0.15\linewidth]{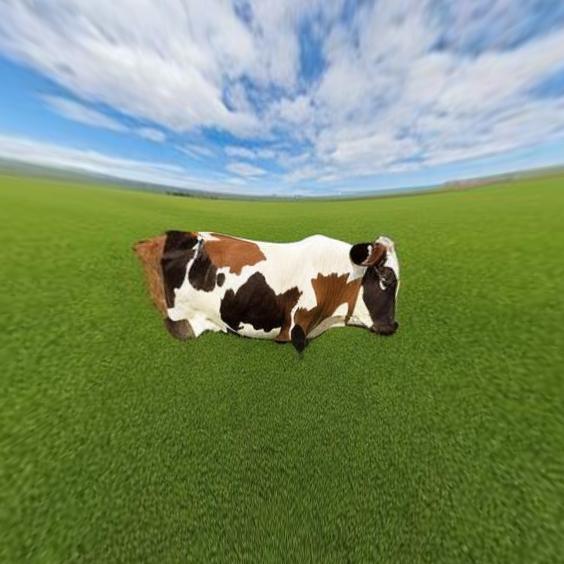}
    \includegraphics[width=0.15\linewidth]{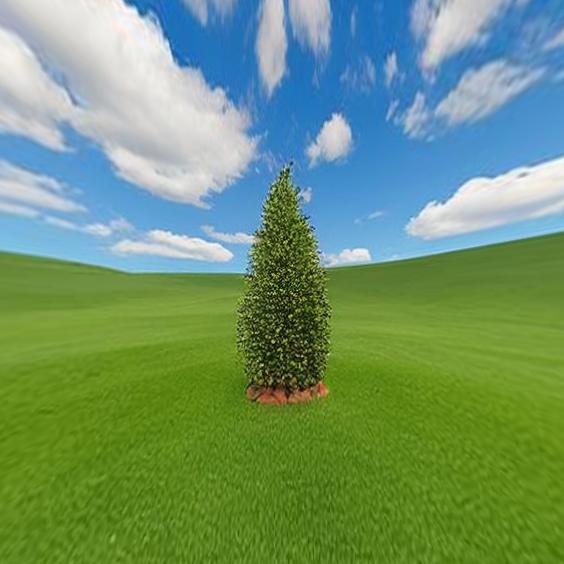}
    \includegraphics[width=0.15\linewidth]{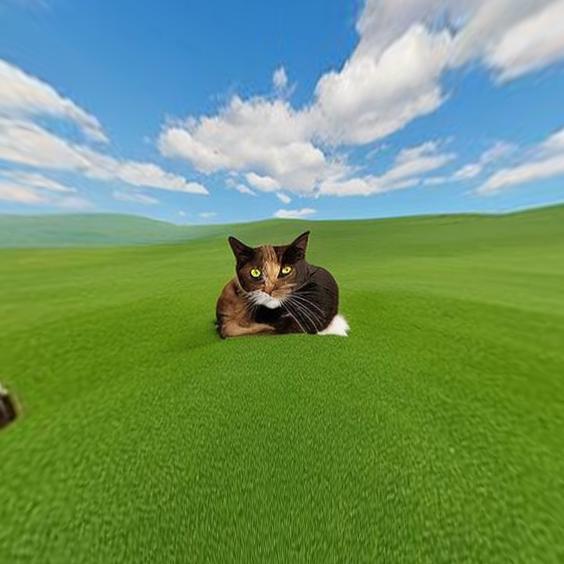}
    \includegraphics[width=0.15\linewidth]{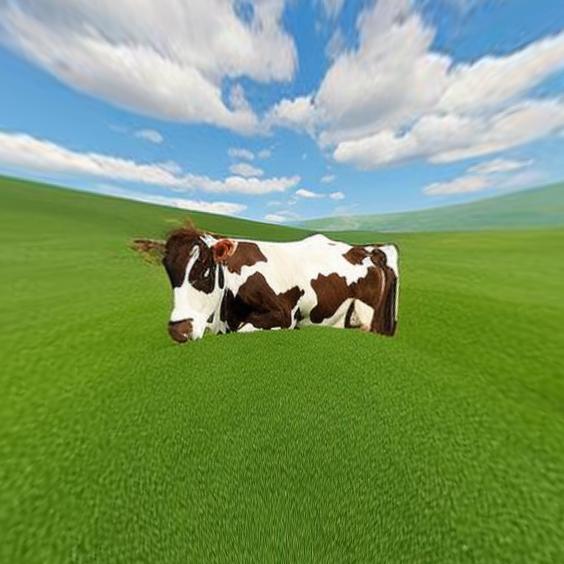}
    \includegraphics[width=0.15\linewidth]{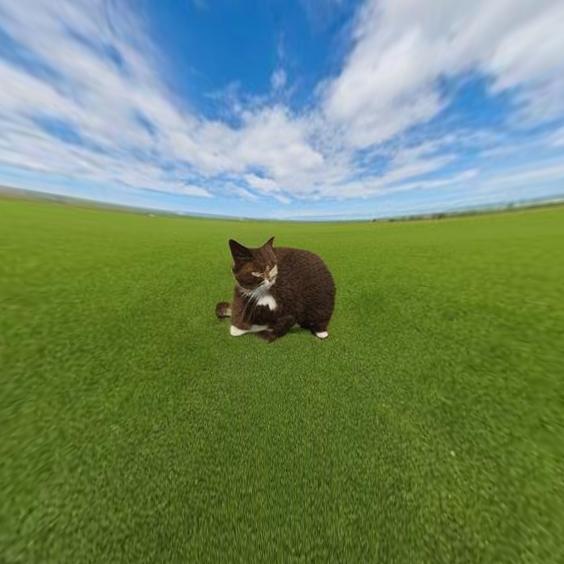}
    \includegraphics[width=0.15\linewidth]{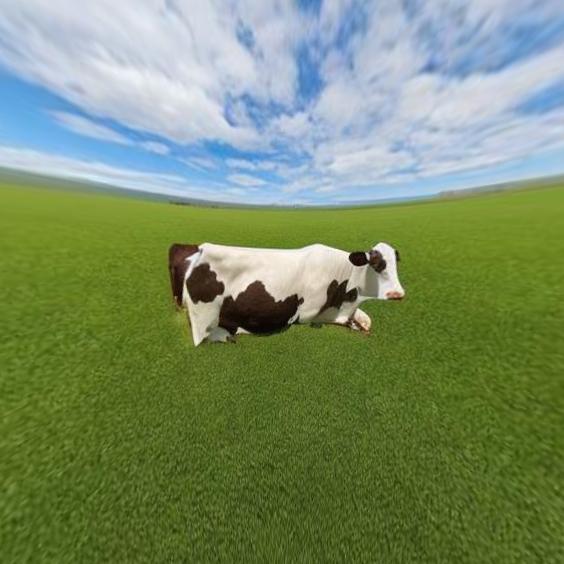}
    \includegraphics[width=0.15\linewidth]{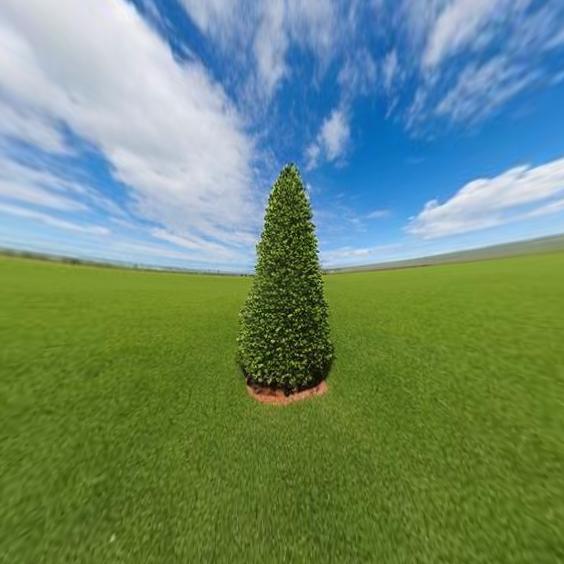}
    \includegraphics[width=0.15\linewidth]{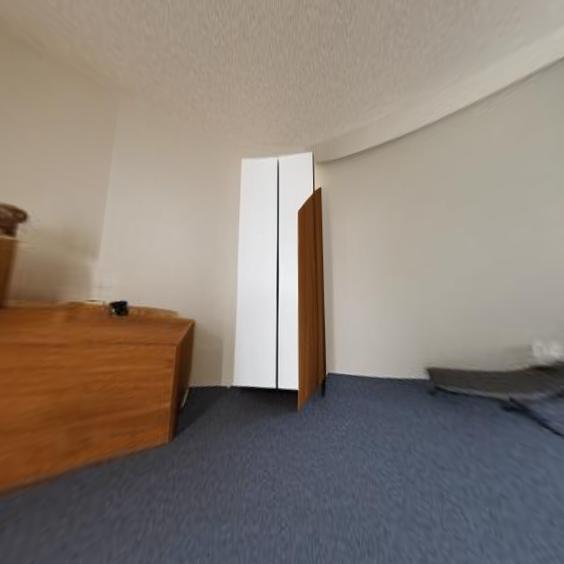}
    \includegraphics[width=0.15\linewidth]{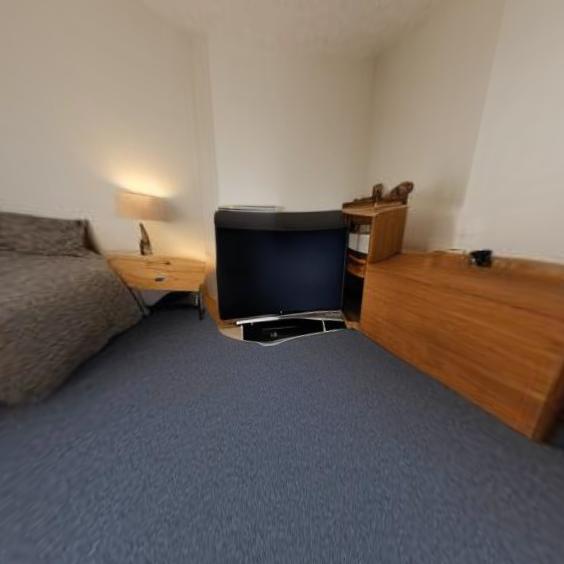}
    \includegraphics[width=0.15\linewidth]{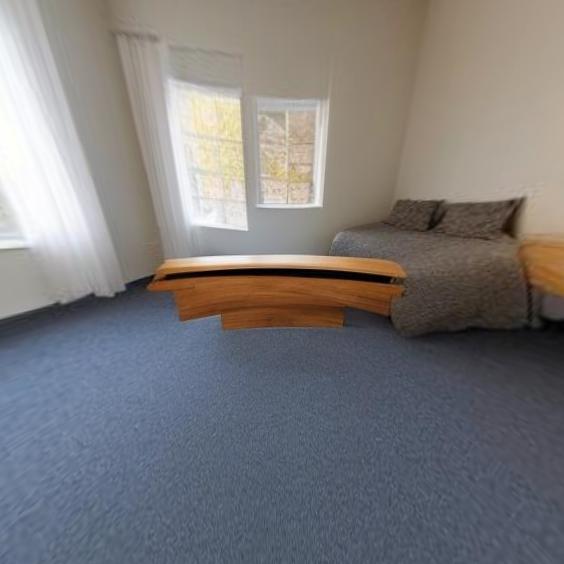}
    \includegraphics[width=0.15\linewidth]{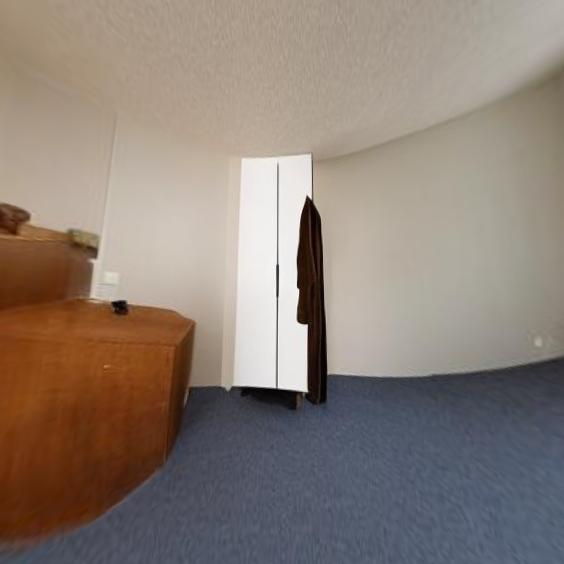}
    \includegraphics[width=0.15\linewidth]{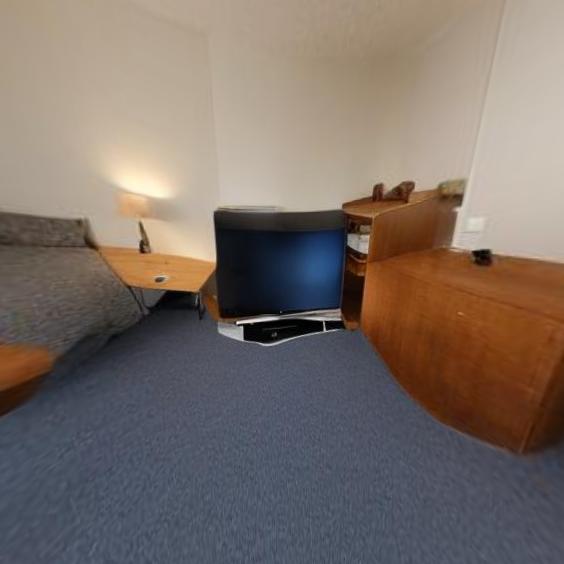}
    \includegraphics[width=0.15\linewidth]{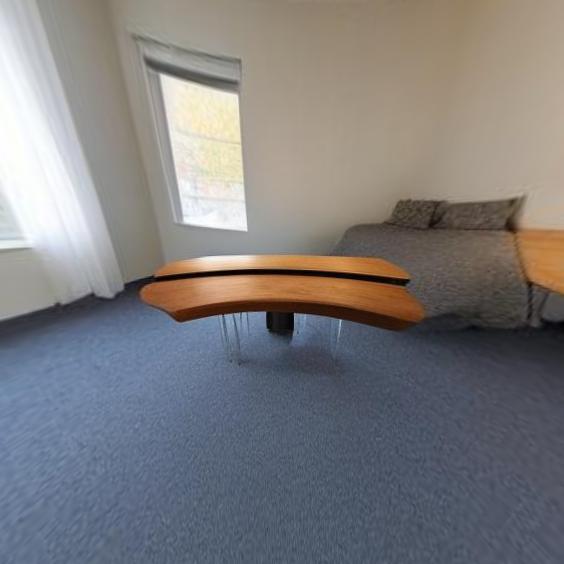}
    \includegraphics[width=0.15\linewidth]{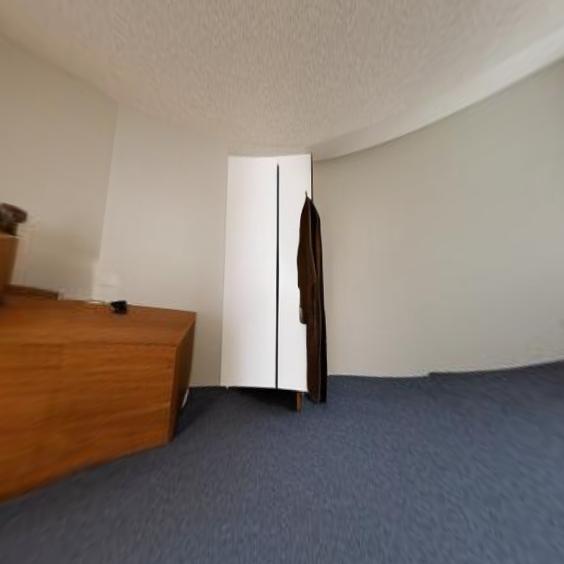}
    \includegraphics[width=0.15\linewidth]{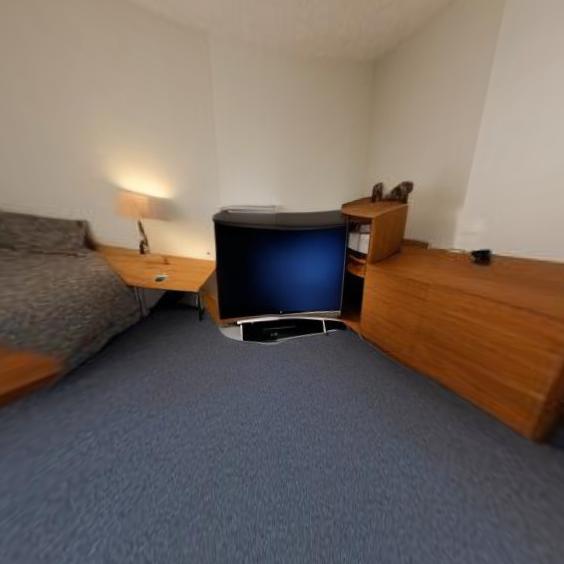}
    \includegraphics[width=0.15\linewidth]{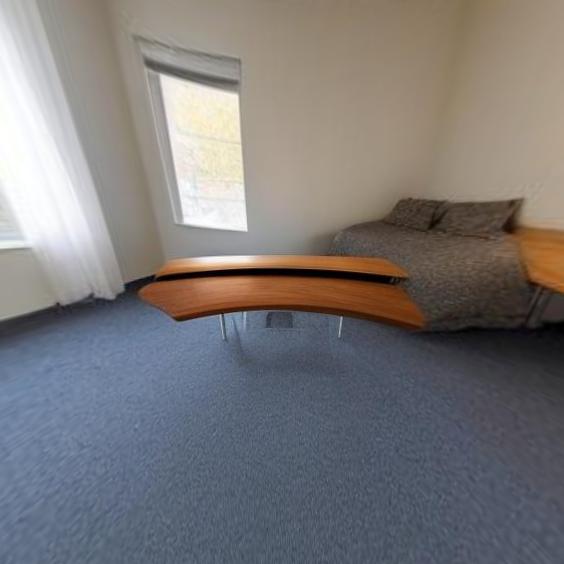}
    \includegraphics[width=0.15\linewidth]{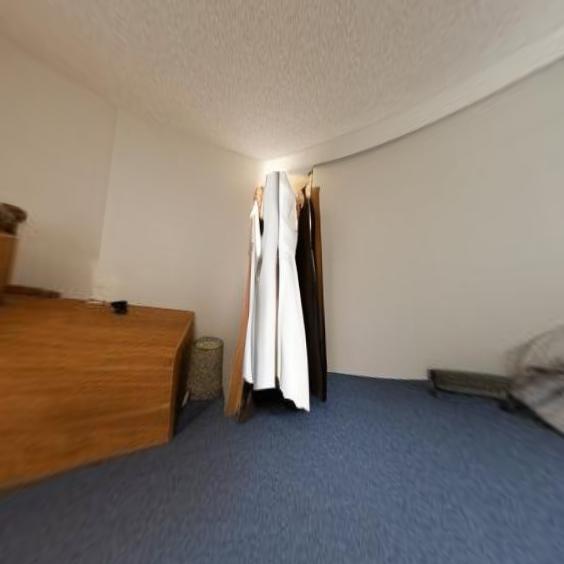}
    \includegraphics[width=0.15\linewidth]{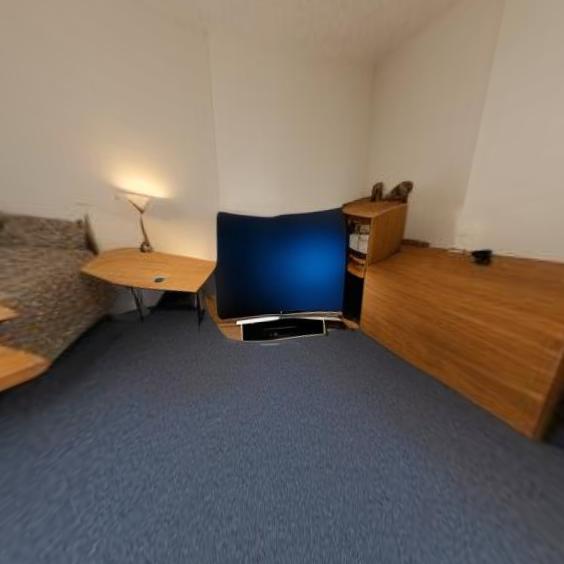}
    \includegraphics[width=0.15\linewidth]{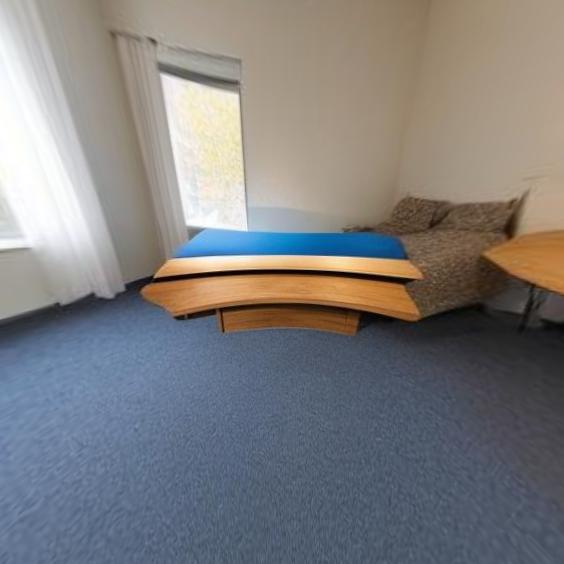}
    \includegraphics[width=0.15\linewidth]{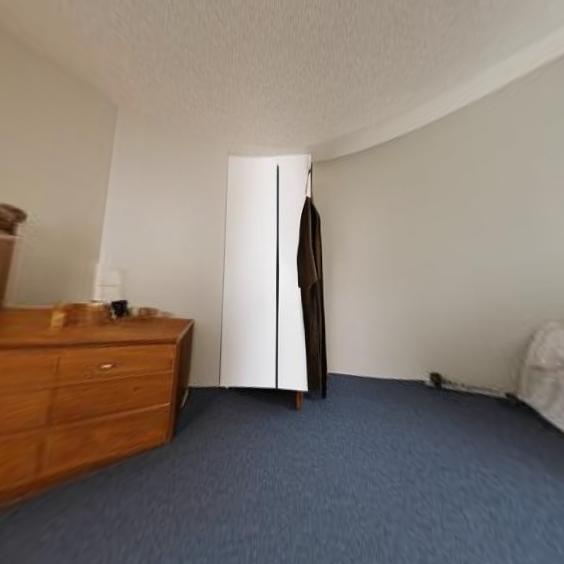}
    \includegraphics[width=0.15\linewidth]{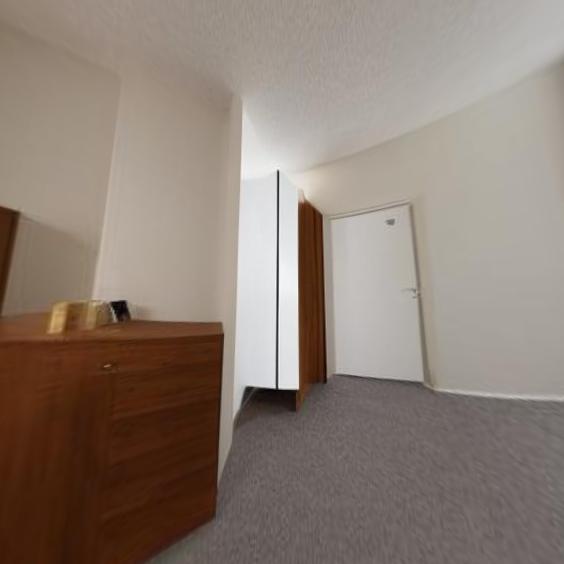}
    \includegraphics[width=0.15\linewidth]{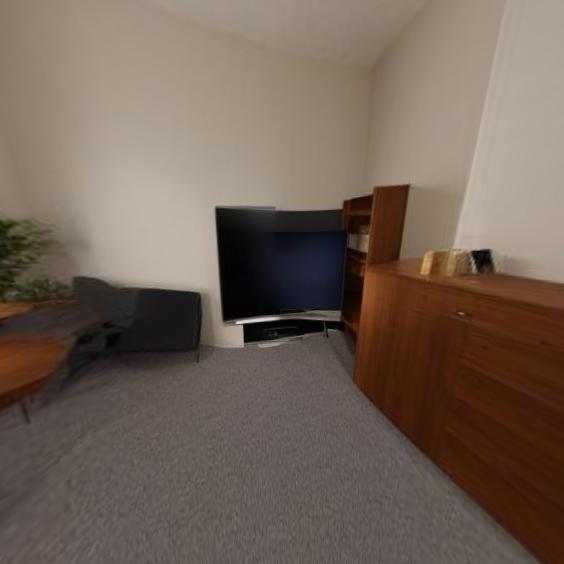}
    \includegraphics[width=0.15\linewidth]{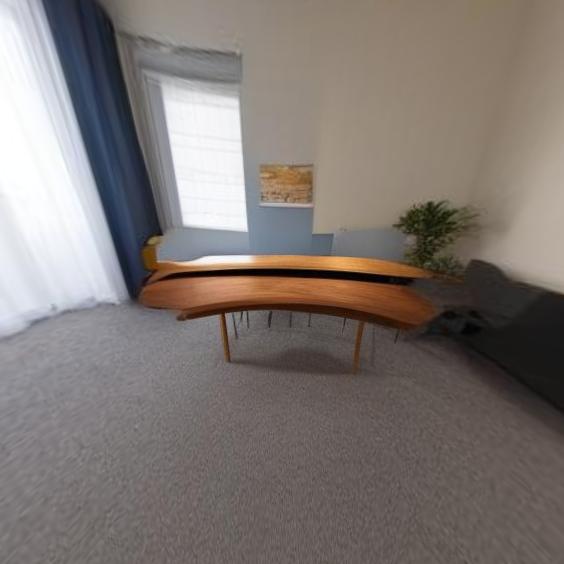}
    \caption{Perspective Images of MultiStitchDiffusion's qualitative results}
    \label{fig:tangents_mstd}
\end{figure}

\begin{figure}[!ht]
    \centering
    \includegraphics[width=0.15\linewidth]{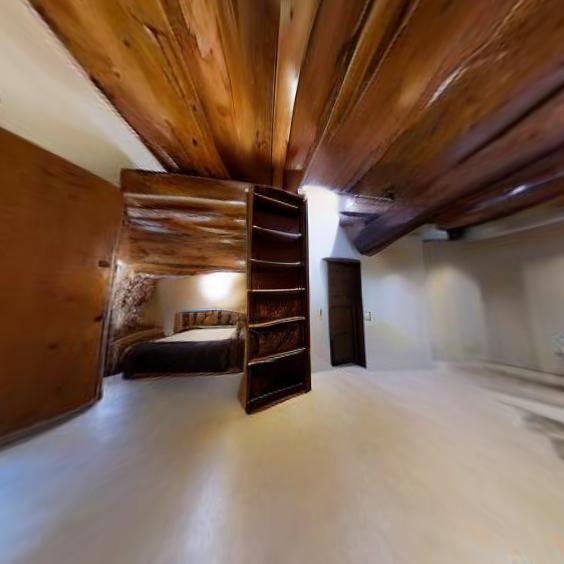}
    \includegraphics[width=0.15\linewidth]{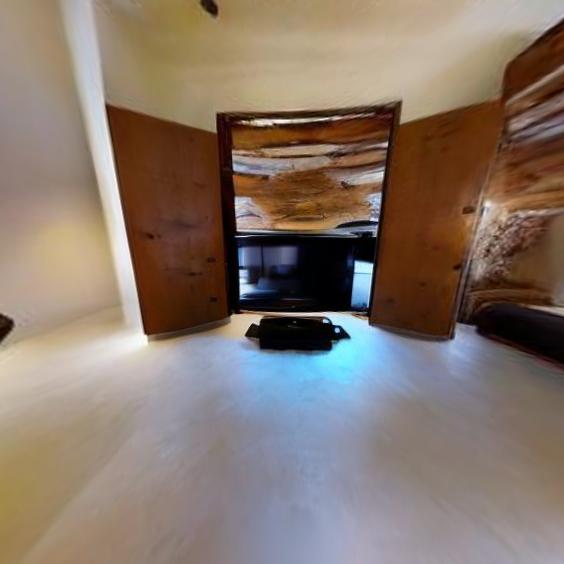}
    \includegraphics[width=0.15\linewidth]{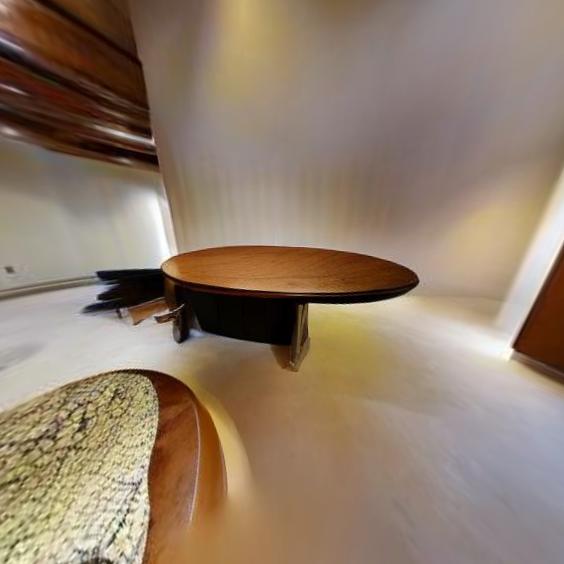}
    \includegraphics[width=0.15\linewidth]{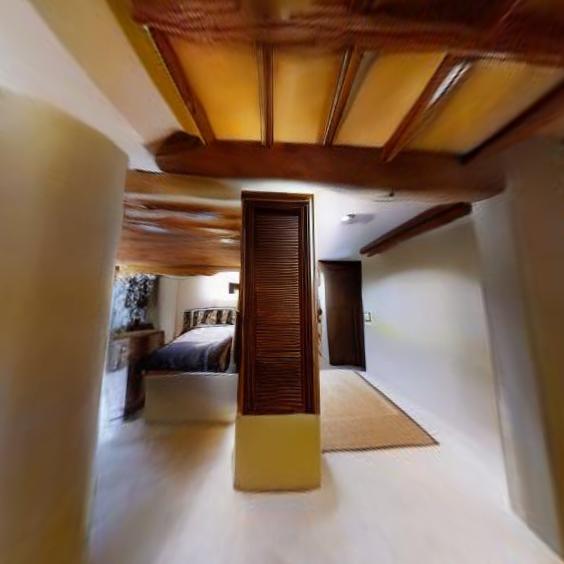}
    \includegraphics[width=0.15\linewidth]{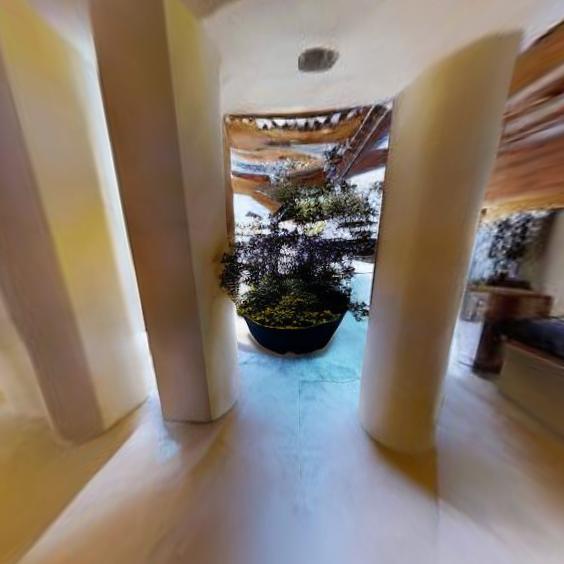}
    \includegraphics[width=0.15\linewidth]{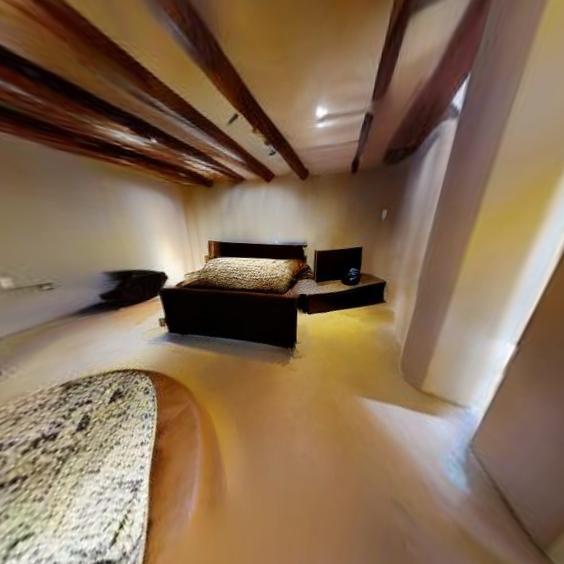}
    \hfill
    \includegraphics[width=0.15\linewidth]{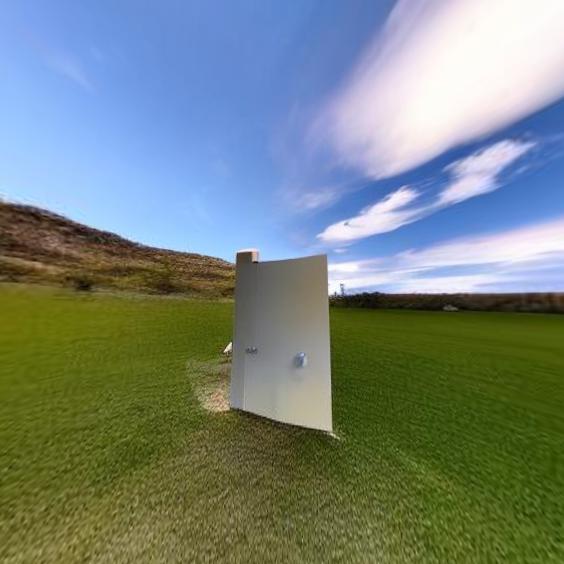}
    \includegraphics[width=0.15\linewidth]{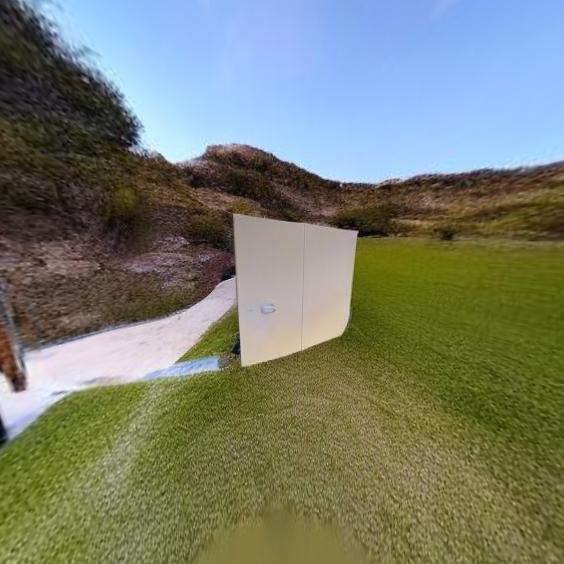}
    \includegraphics[width=0.15\linewidth]{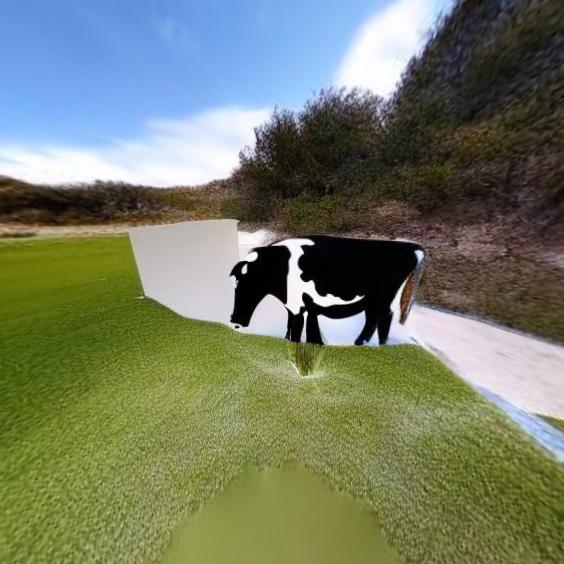}
    \includegraphics[width=0.15\linewidth]{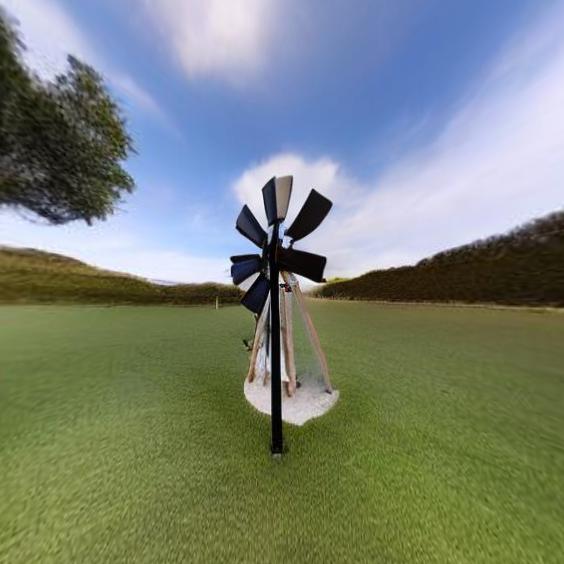}
    \includegraphics[width=0.15\linewidth]{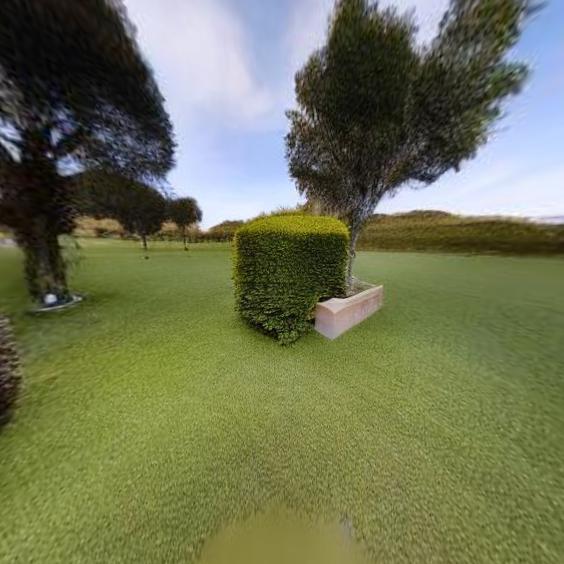}
    \includegraphics[width=0.15\linewidth]{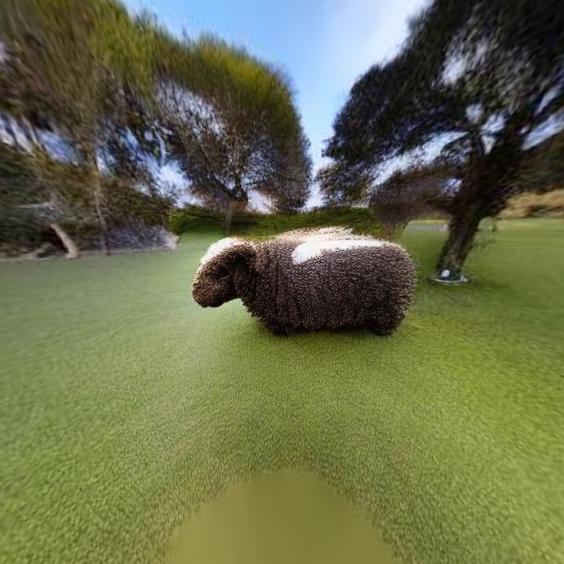}
    \includegraphics[width=0.15\linewidth]{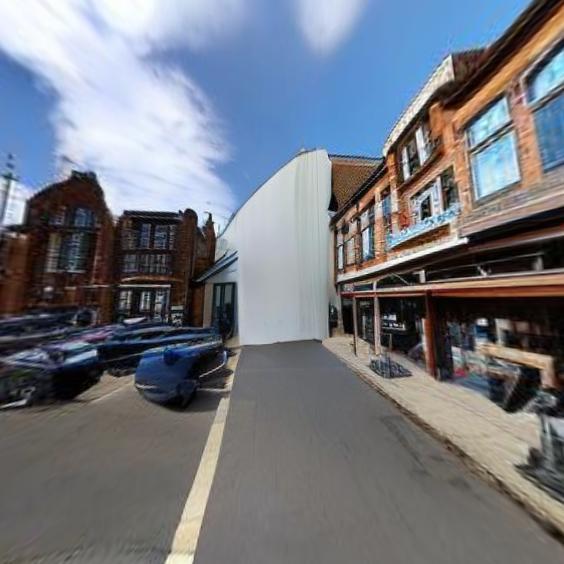}
    \includegraphics[width=0.15\linewidth]{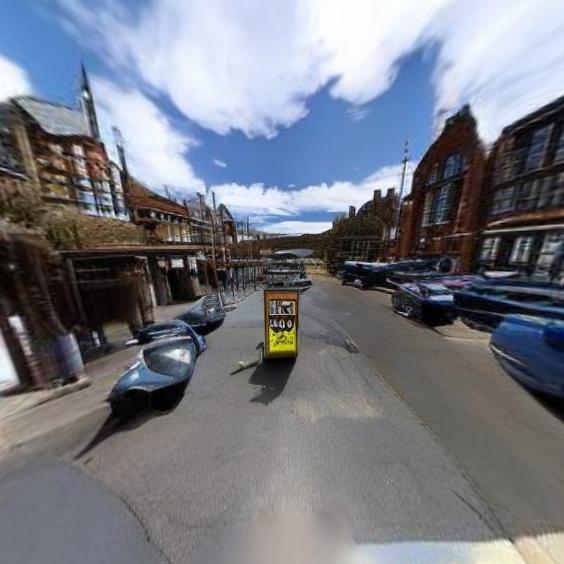}
    \includegraphics[width=0.15\linewidth]{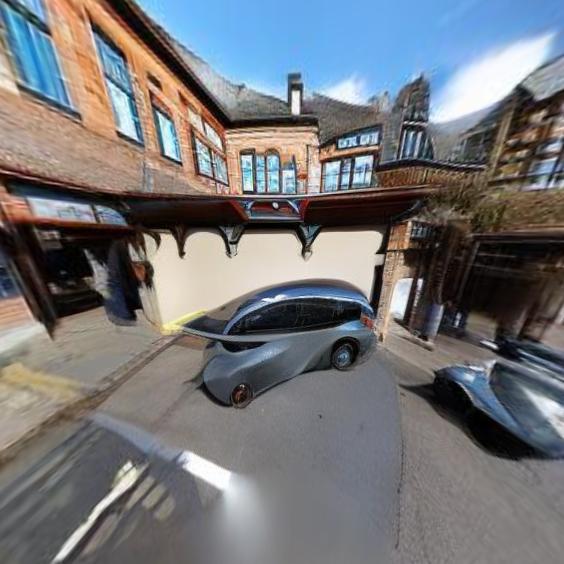}
    \includegraphics[width=0.15\linewidth]{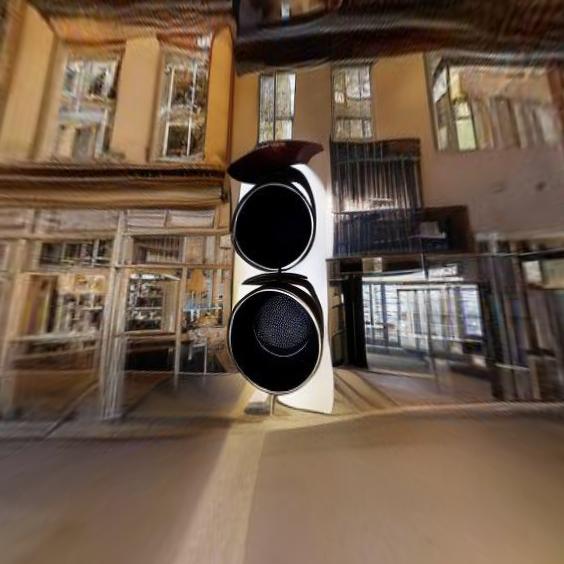}
    \includegraphics[width=0.15\linewidth]{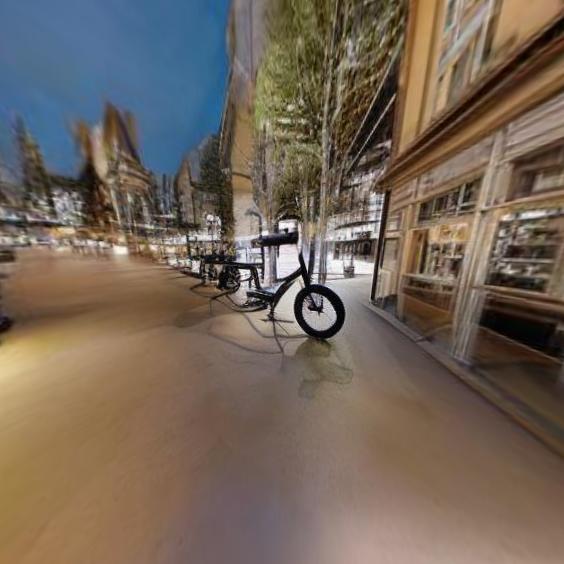}
    \includegraphics[width=0.15\linewidth]{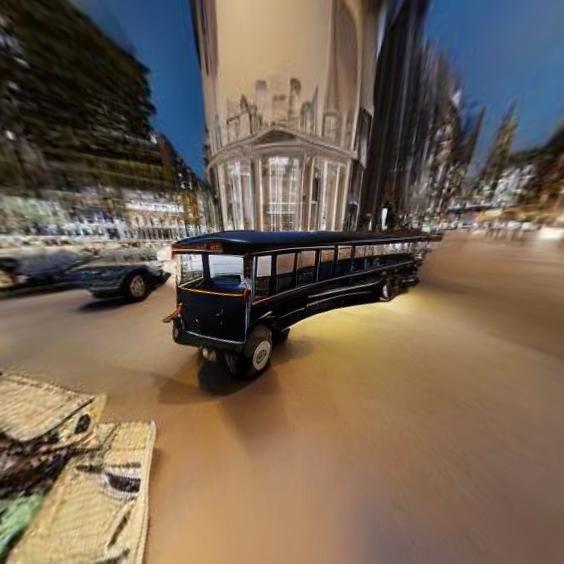}
    \includegraphics[width=0.15\linewidth]{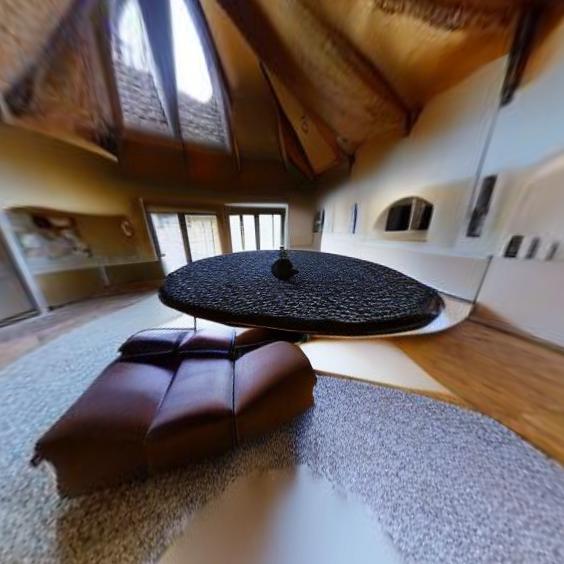}
    \includegraphics[width=0.15\linewidth]{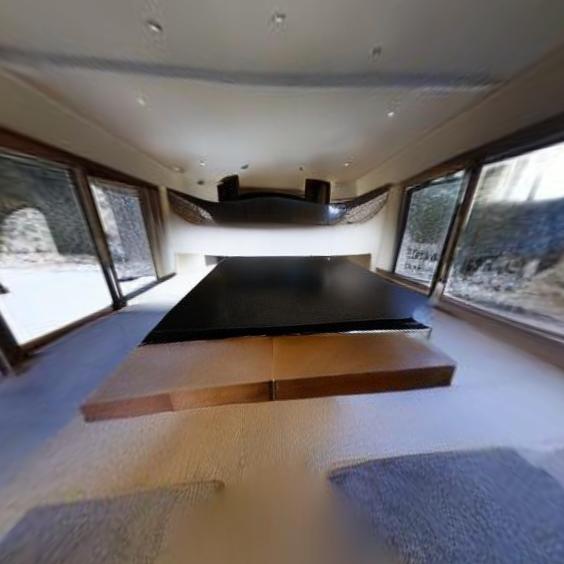}
    \includegraphics[width=0.15\linewidth]{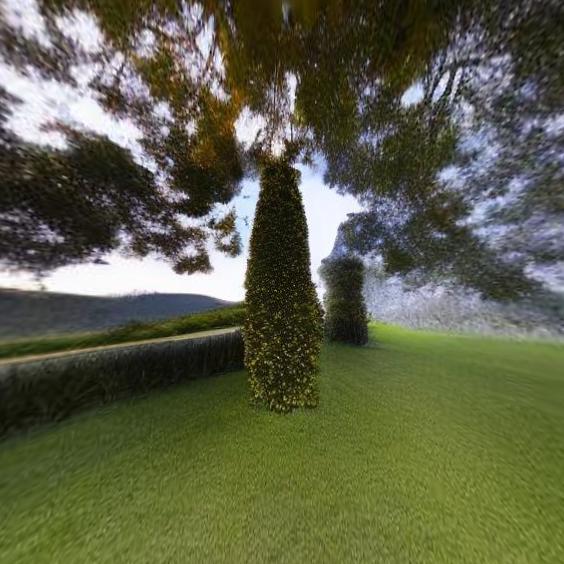}
    \includegraphics[width=0.15\linewidth]{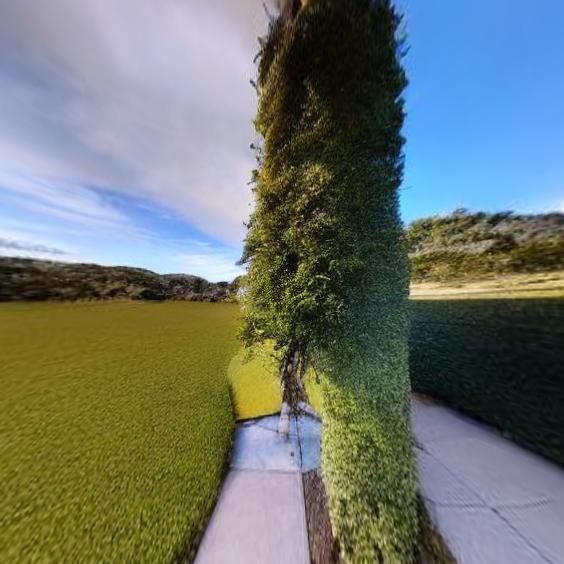}
    \includegraphics[width=0.15\linewidth]{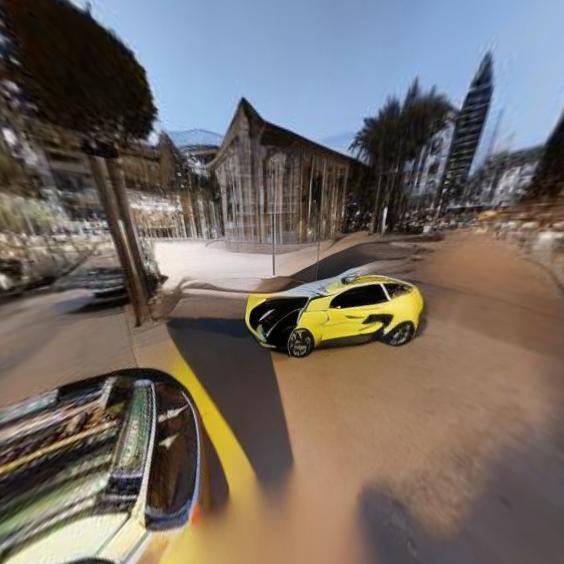}
    \includegraphics[width=0.15\linewidth]{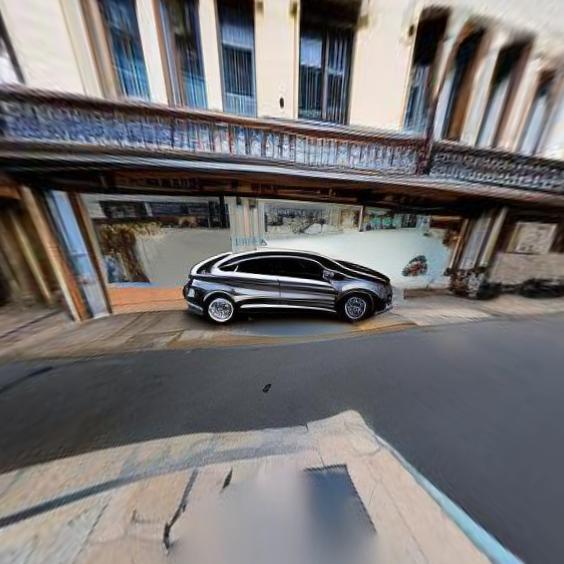}
    \includegraphics[width=0.15\linewidth]{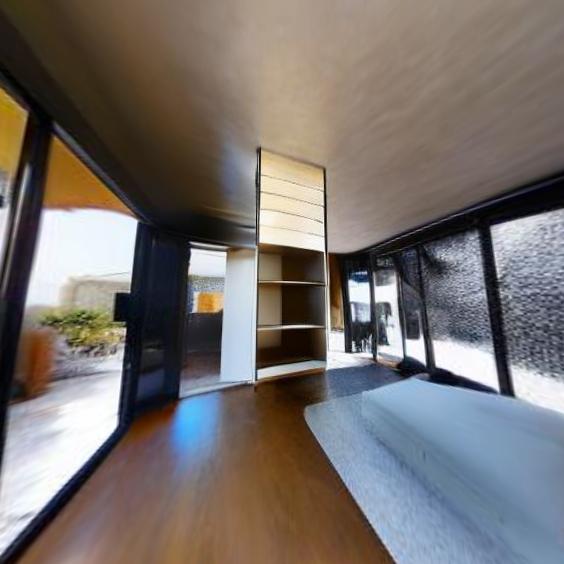}
    \includegraphics[width=0.15\linewidth]{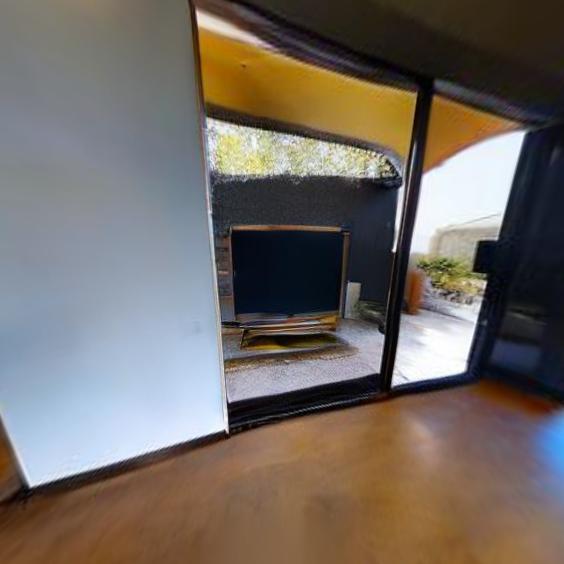}
    \includegraphics[width=0.15\linewidth]{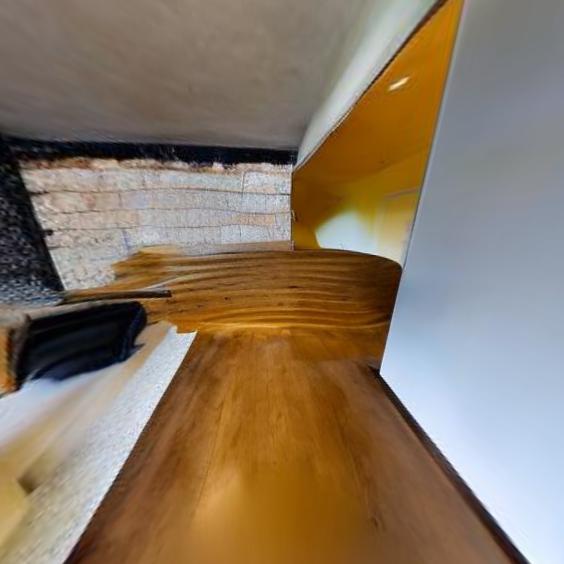}
    \includegraphics[width=0.15\linewidth]{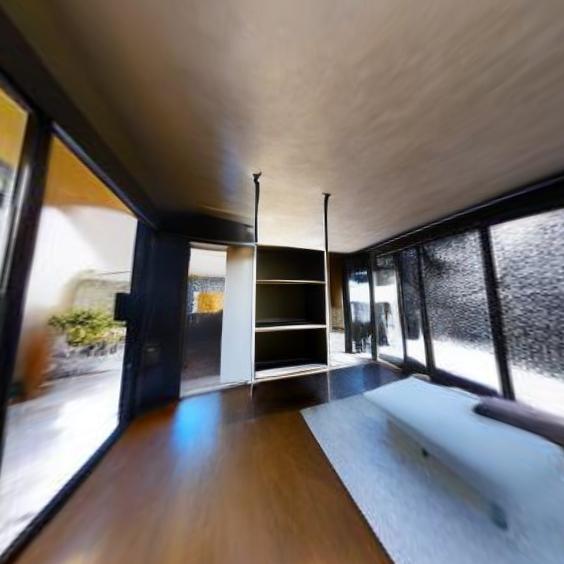}
    \includegraphics[width=0.15\linewidth]{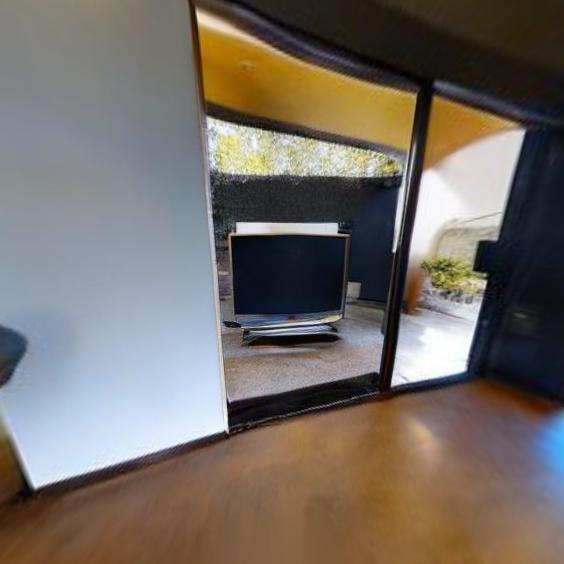}
    \includegraphics[width=0.15\linewidth]{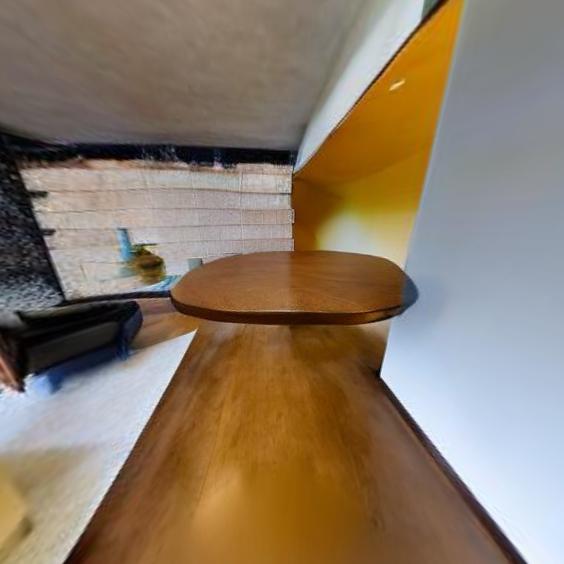}
    \includegraphics[width=0.15\linewidth]{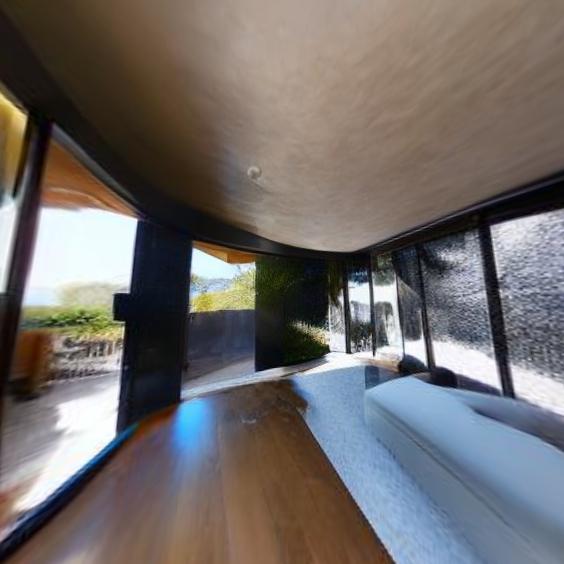}
    \includegraphics[width=0.15\linewidth]{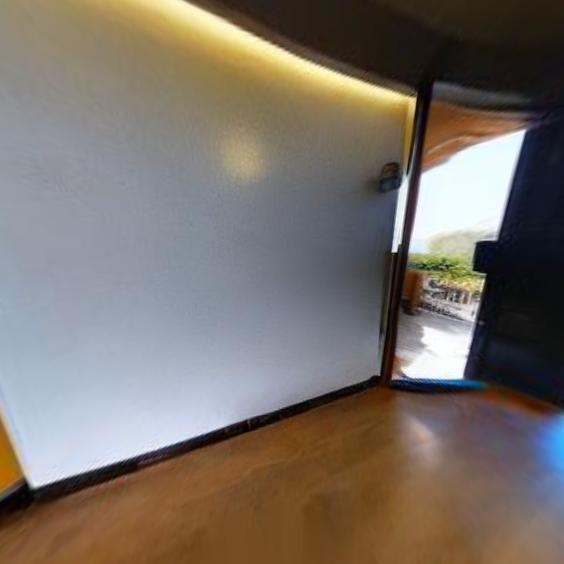}
    \includegraphics[width=0.15\linewidth]{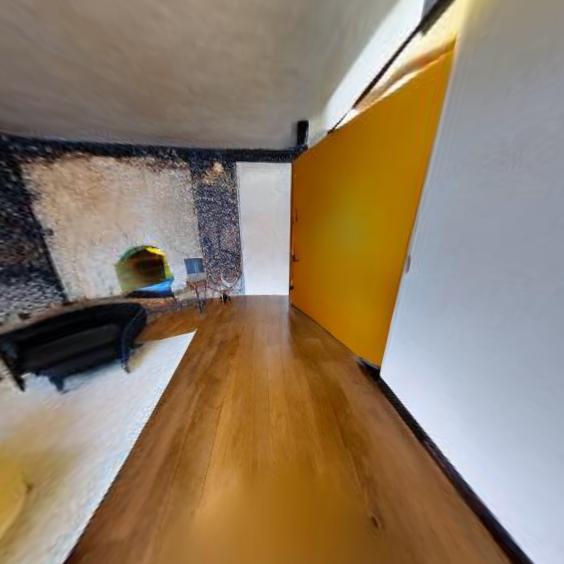}
    \includegraphics[width=0.15\linewidth]{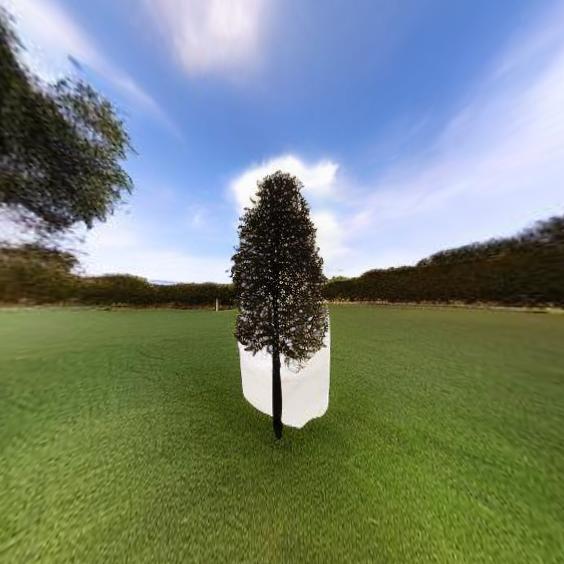}
    \includegraphics[width=0.15\linewidth]{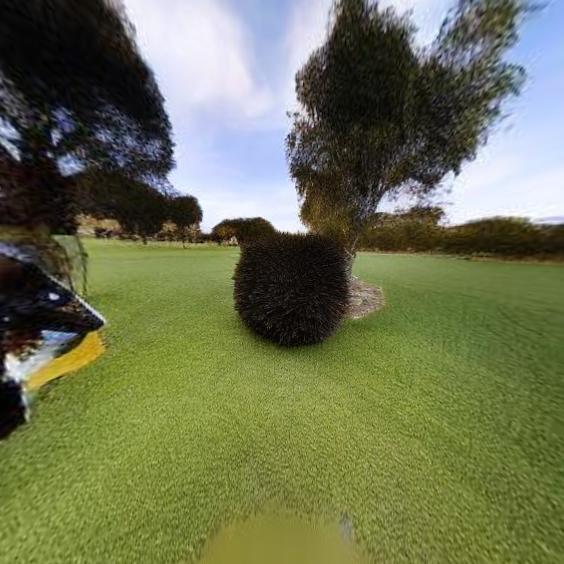}
    \includegraphics[width=0.15\linewidth]{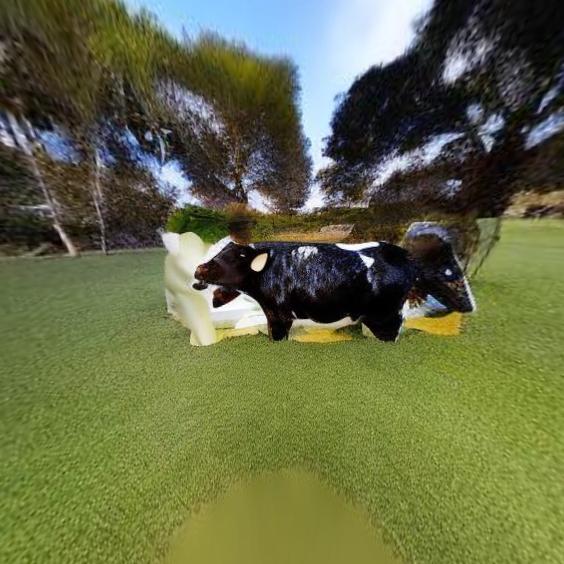}
    \includegraphics[width=0.15\linewidth]{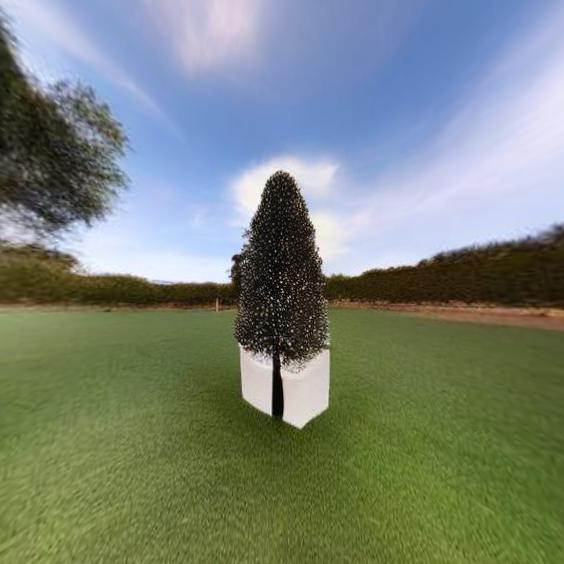}
    \includegraphics[width=0.15\linewidth]{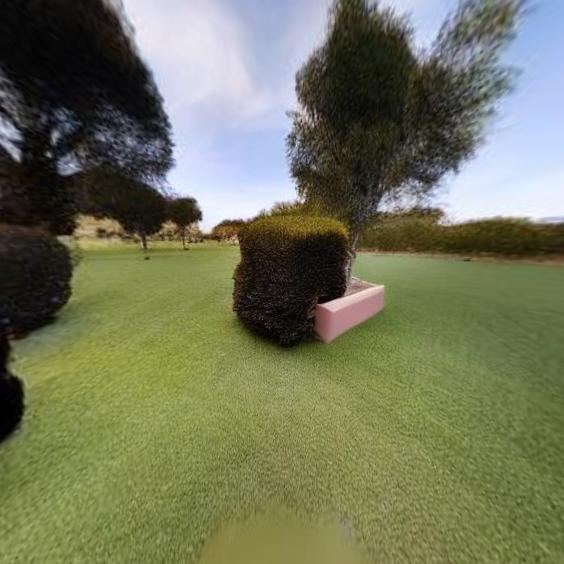}
    \includegraphics[width=0.15\linewidth]{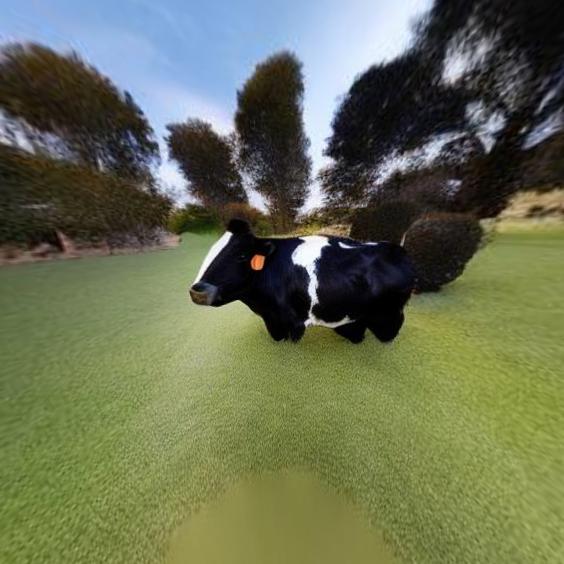}
    \includegraphics[width=0.15\linewidth]{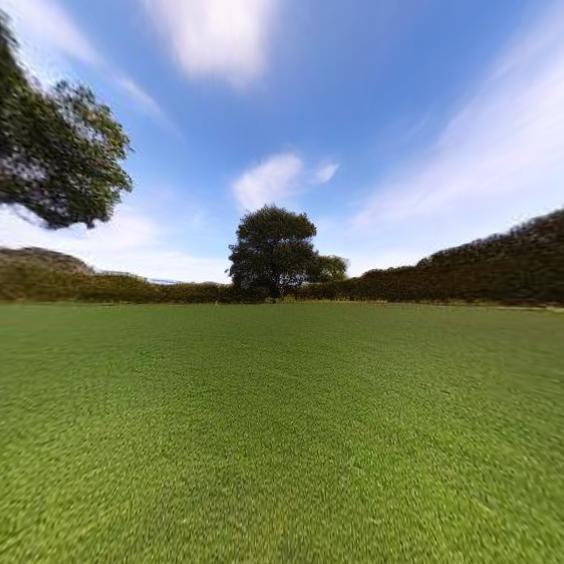}
    \includegraphics[width=0.15\linewidth]{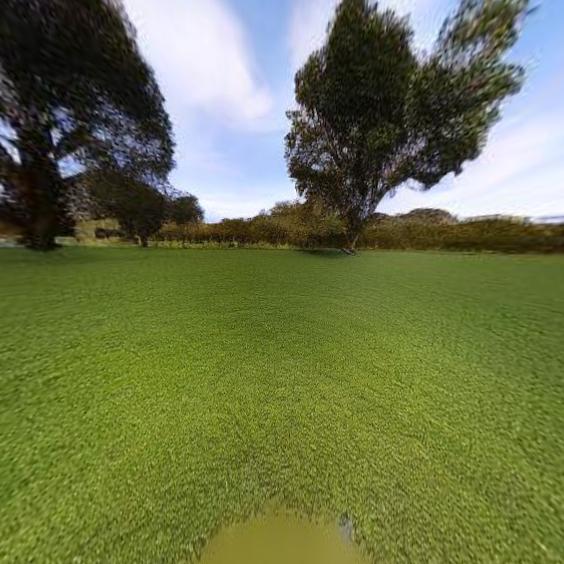}
    \includegraphics[width=0.15\linewidth]{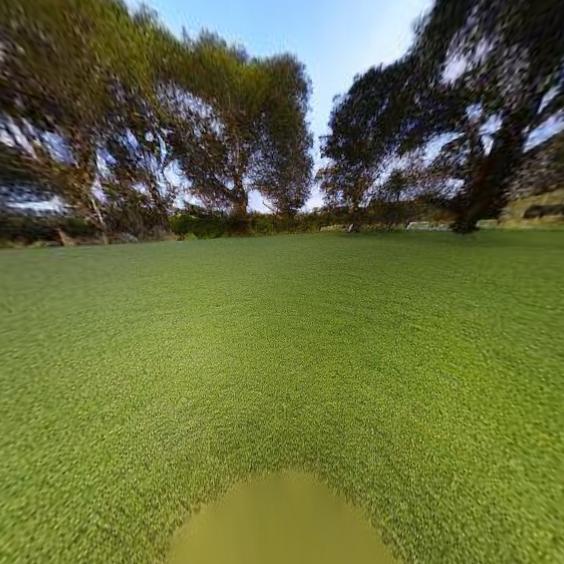}
    \includegraphics[width=0.15\linewidth]{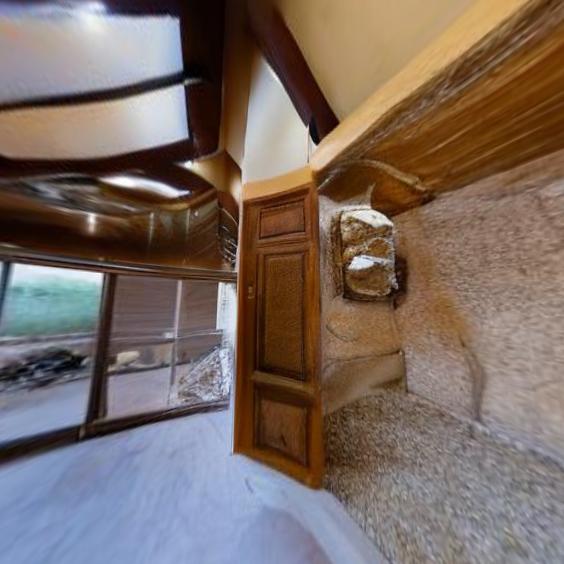}
    \includegraphics[width=0.15\linewidth]{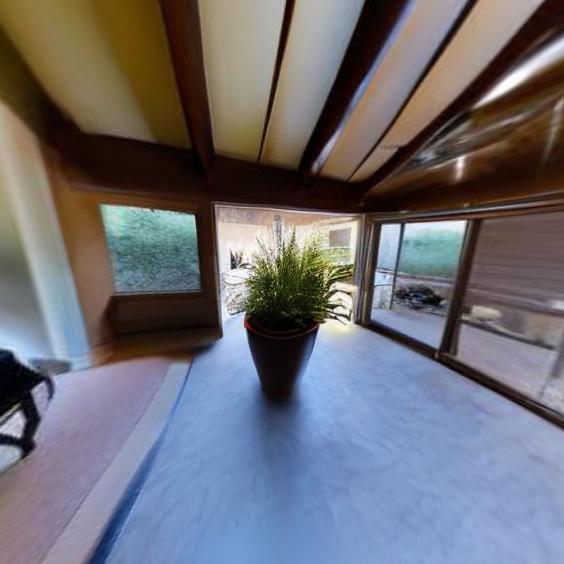}
    \includegraphics[width=0.15\linewidth]{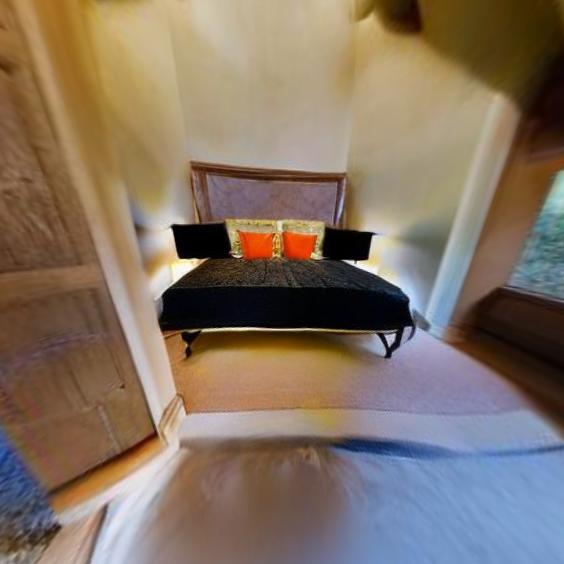}
    \includegraphics[width=0.15\linewidth]{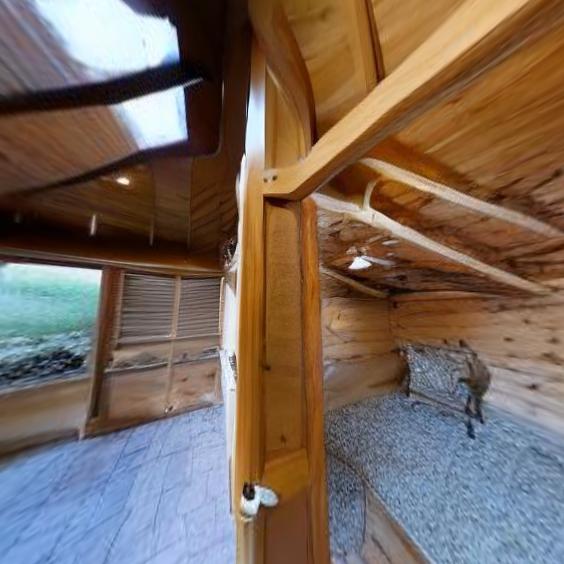}
    \includegraphics[width=0.15\linewidth]{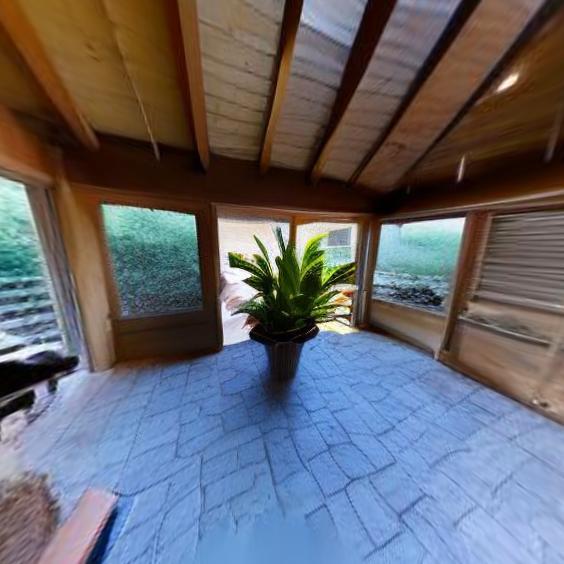}
    \includegraphics[width=0.15\linewidth]{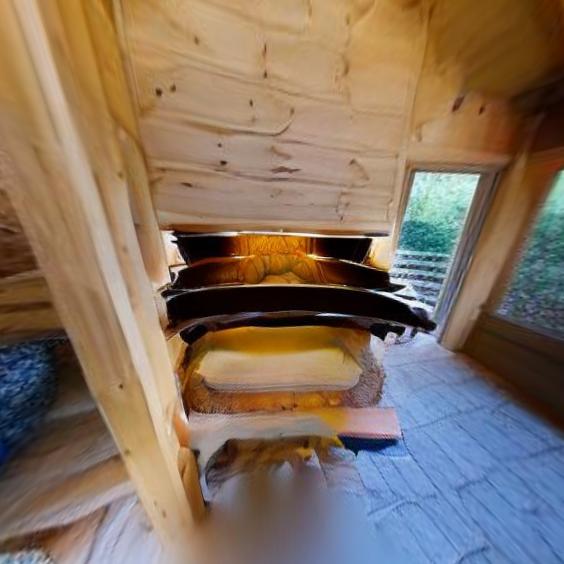}
    \includegraphics[width=0.15\linewidth]{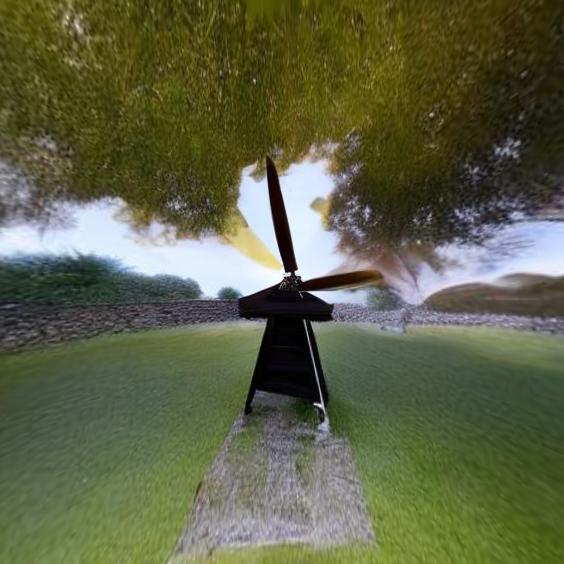}
    \includegraphics[width=0.15\linewidth]{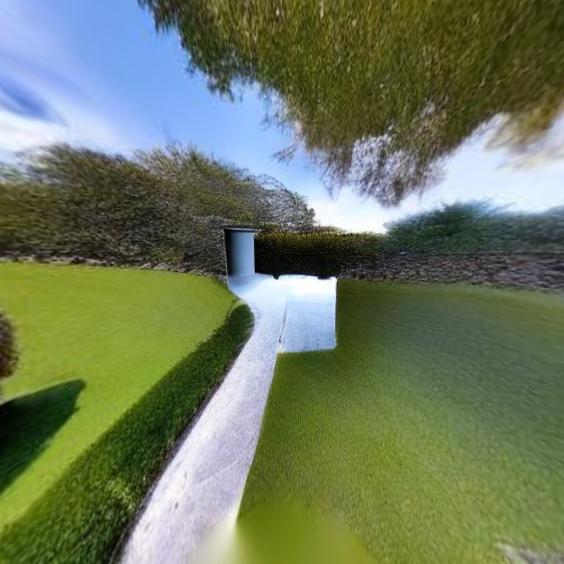}
    \includegraphics[width=0.15\linewidth]{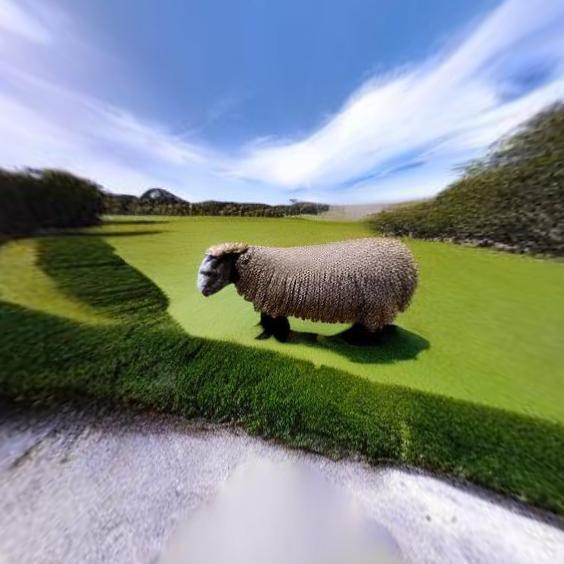}
    \includegraphics[width=0.15\linewidth]{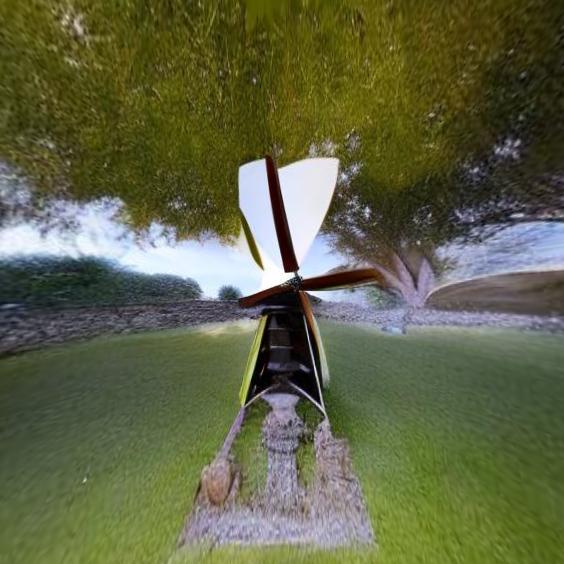}
    \includegraphics[width=0.15\linewidth]{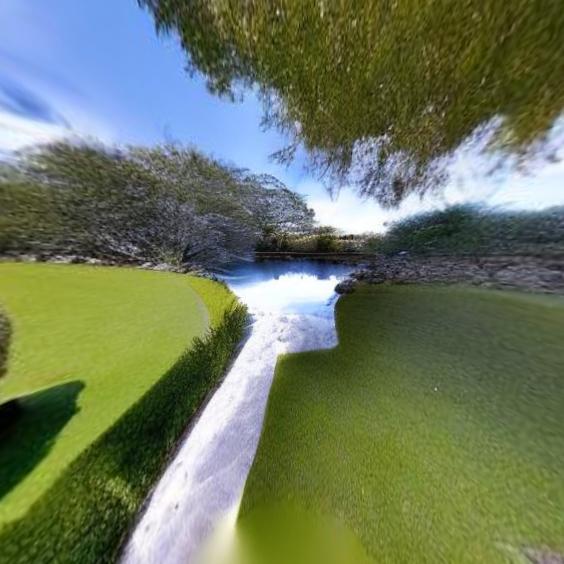}
    \includegraphics[width=0.15\linewidth]{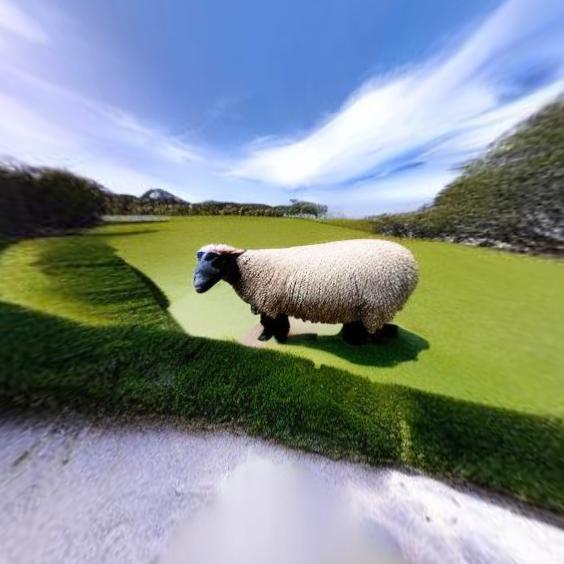}
    \includegraphics[width=0.15\linewidth]{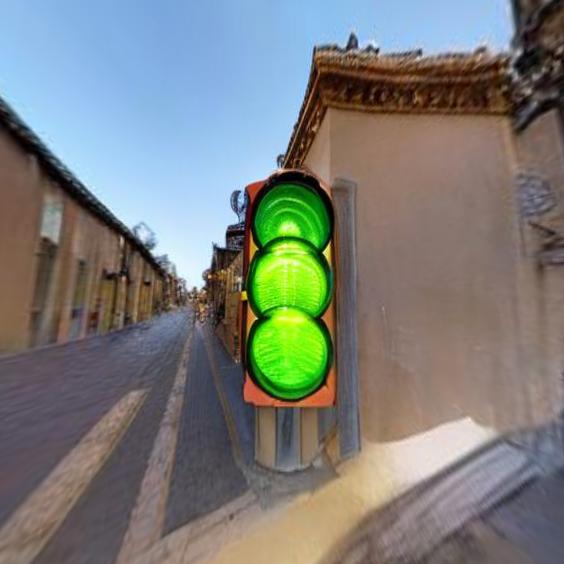}
    \includegraphics[width=0.15\linewidth]{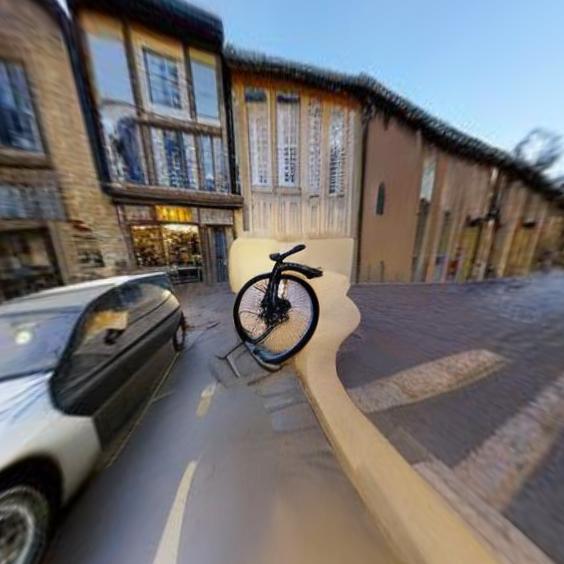}
    \includegraphics[width=0.15\linewidth]{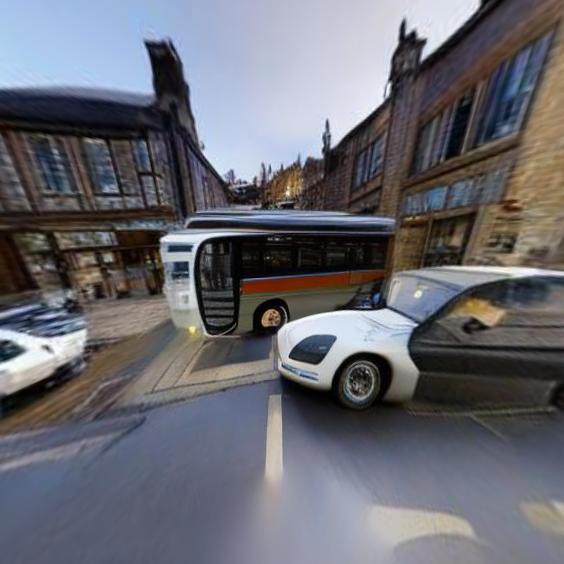}
    \includegraphics[width=0.15\linewidth]{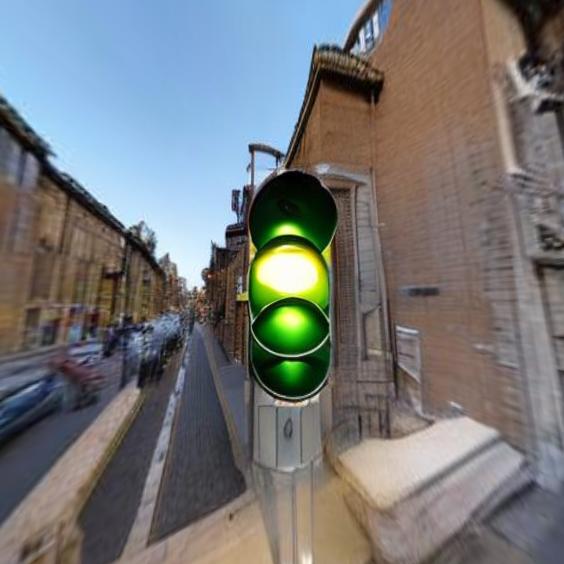}
    \includegraphics[width=0.15\linewidth]{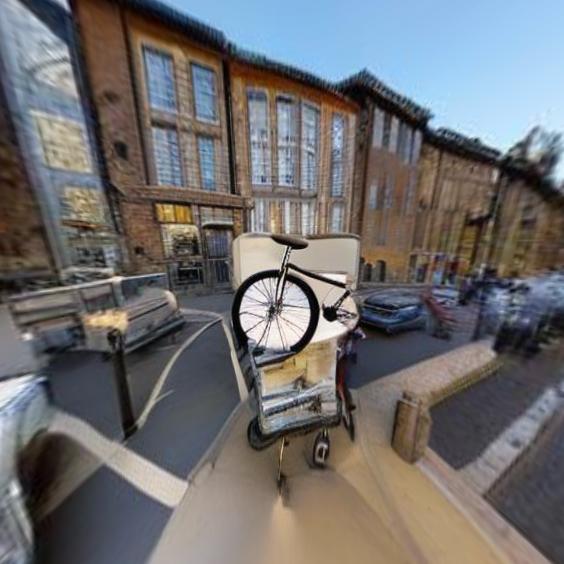}
    \includegraphics[width=0.15\linewidth]{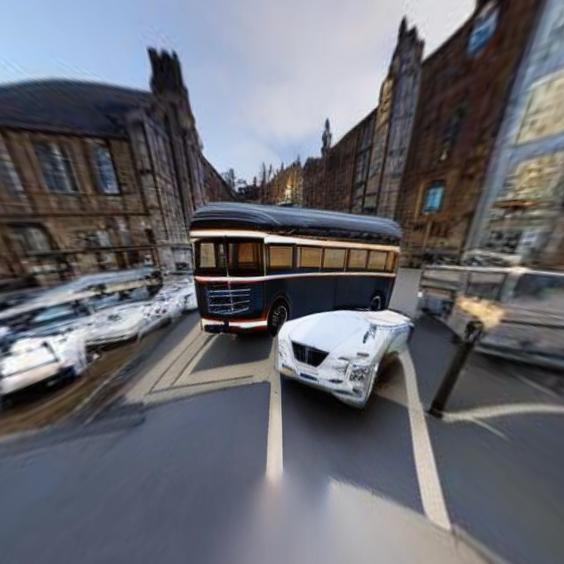}
    \includegraphics[width=0.15\linewidth]{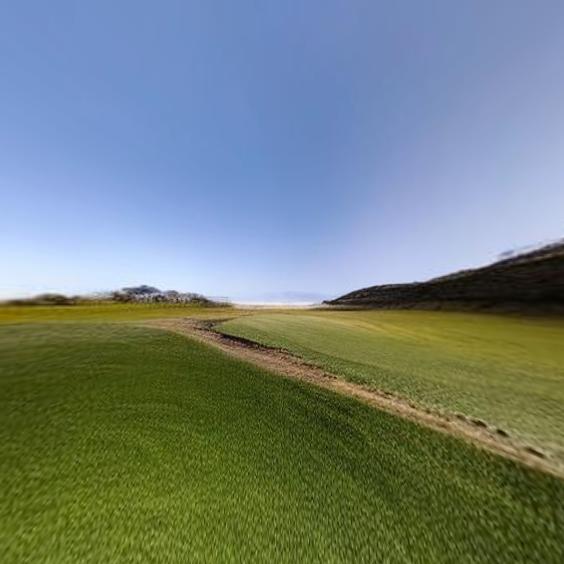}
    \includegraphics[width=0.15\linewidth]{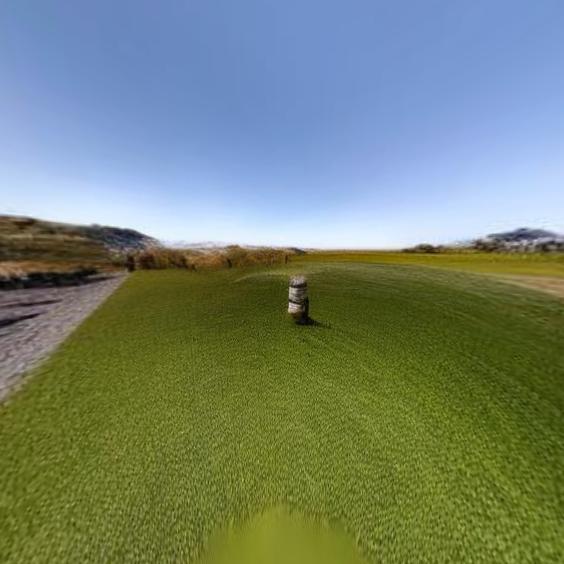}
    \includegraphics[width=0.15\linewidth]{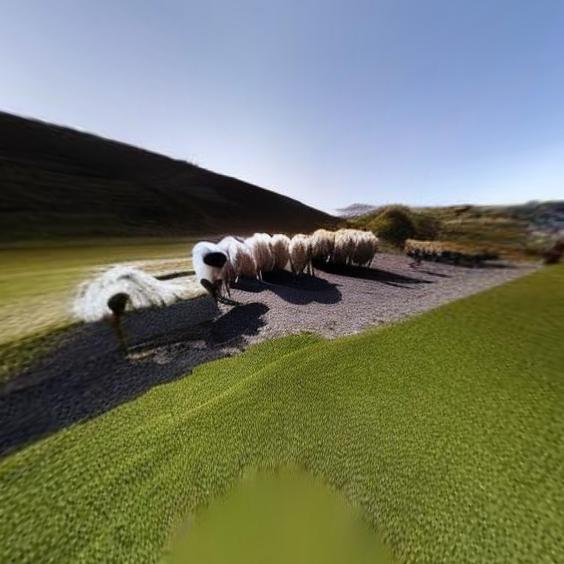}
    \includegraphics[width=0.15\linewidth]{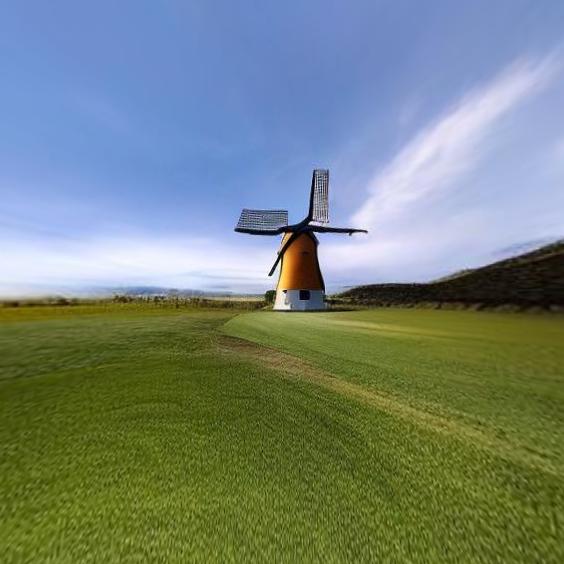}
    \includegraphics[width=0.15\linewidth]{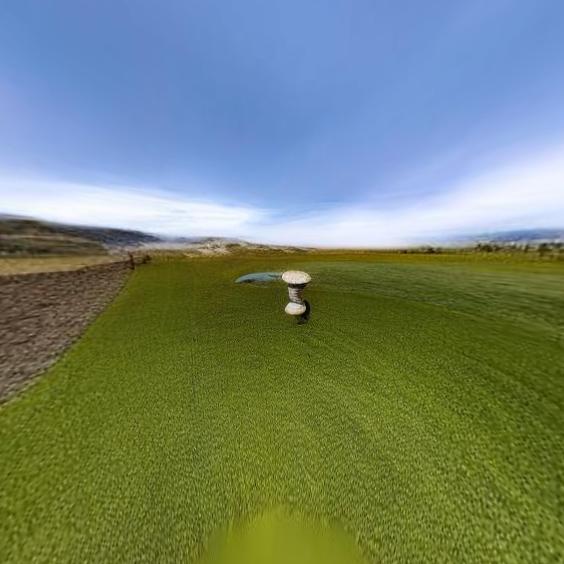}
    \includegraphics[width=0.15\linewidth]{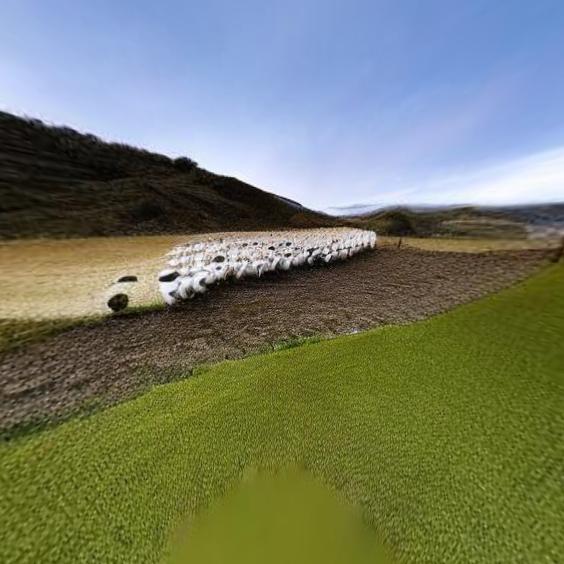}
    \includegraphics[width=0.15\linewidth]{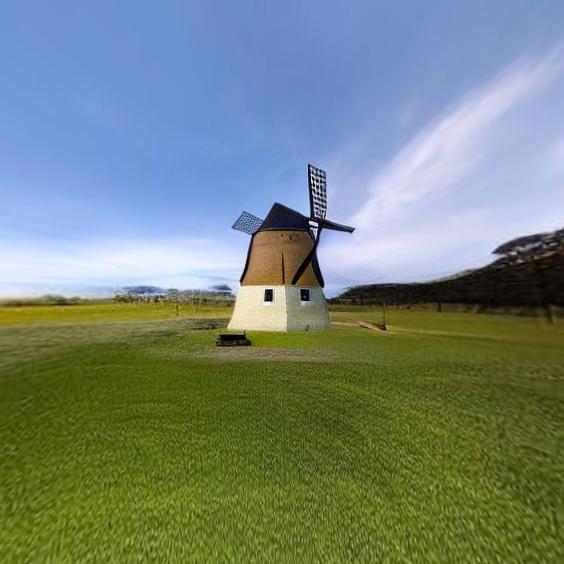}
    \includegraphics[width=0.15\linewidth]{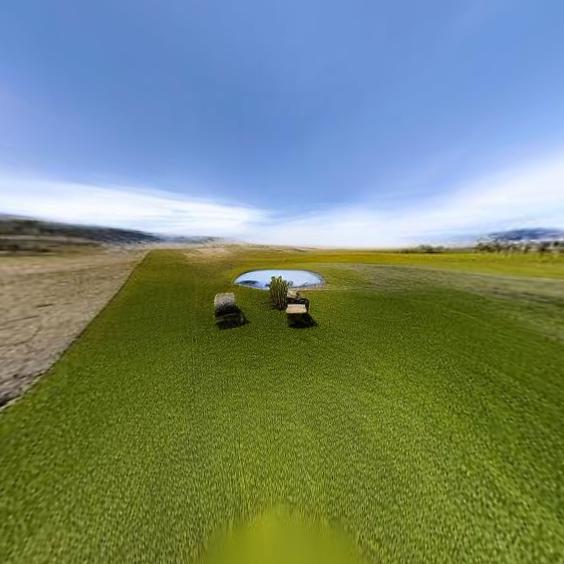}
    \includegraphics[width=0.15\linewidth]{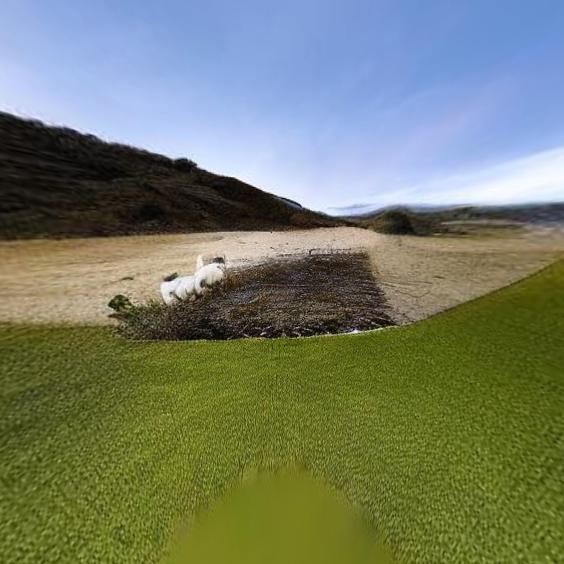}
    \includegraphics[width=0.15\linewidth]{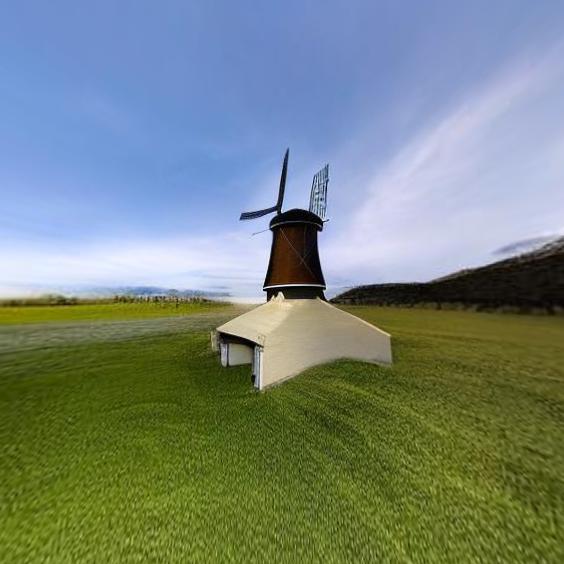}
    \includegraphics[width=0.15\linewidth]{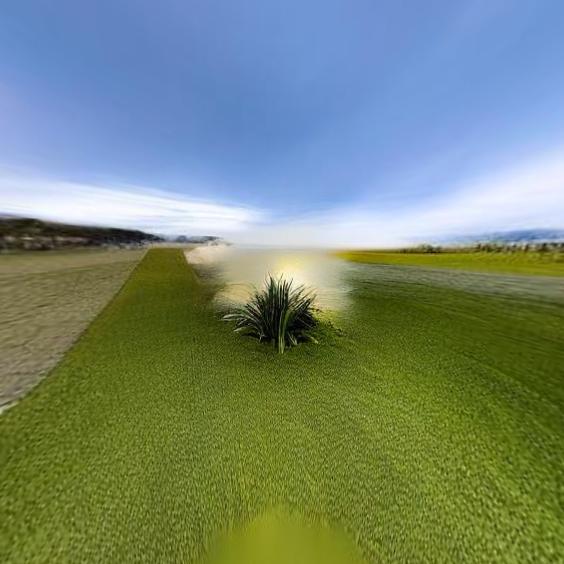}
    \includegraphics[width=0.15\linewidth]{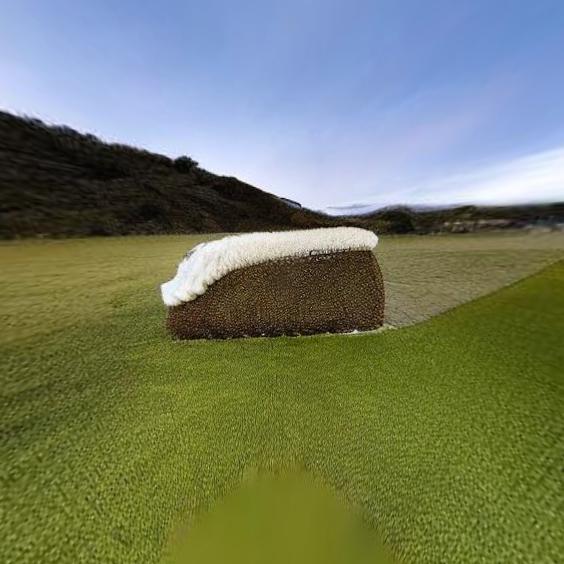}
    \includegraphics[width=0.15\linewidth]{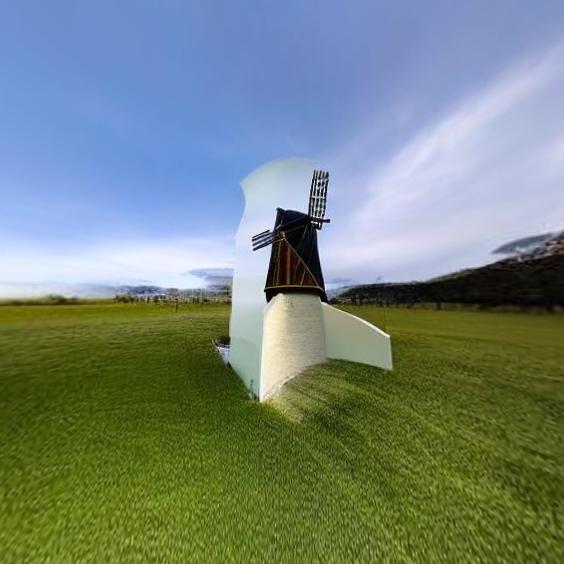}
    \includegraphics[width=0.15\linewidth]{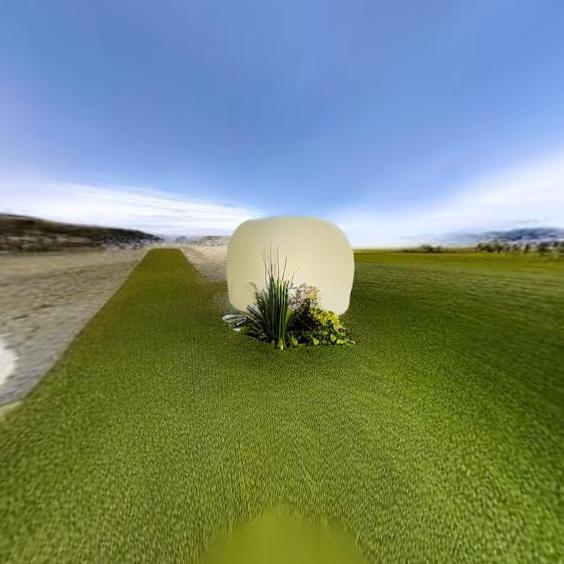}
    \includegraphics[width=0.15\linewidth]{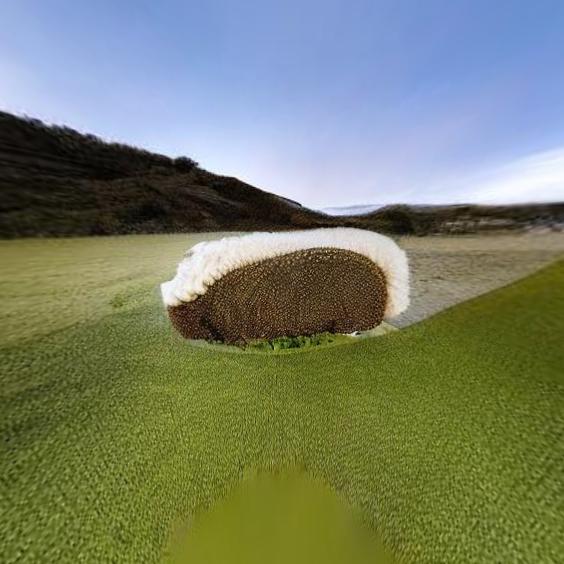}
    \includegraphics[width=0.15\linewidth]{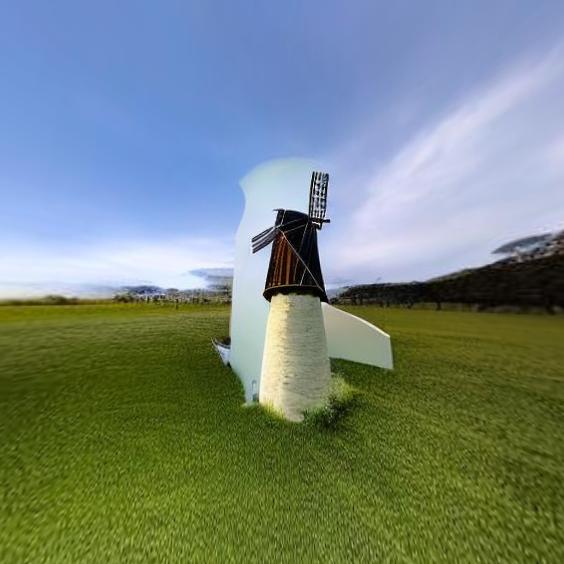}
    \includegraphics[width=0.15\linewidth]{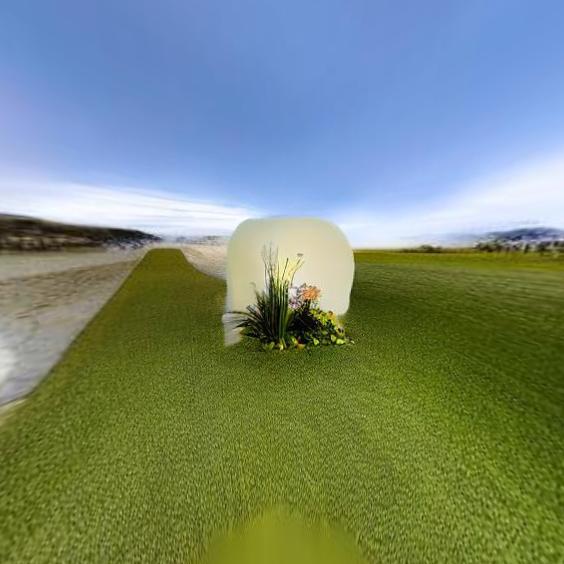}
    \includegraphics[width=0.15\linewidth]{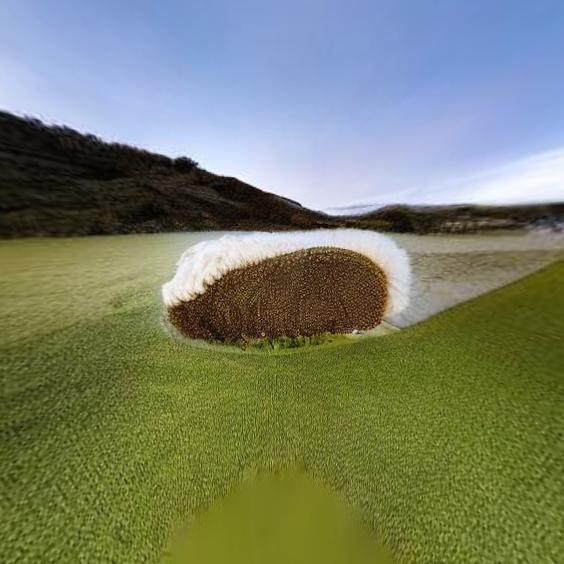}
    \includegraphics[width=0.15\linewidth]{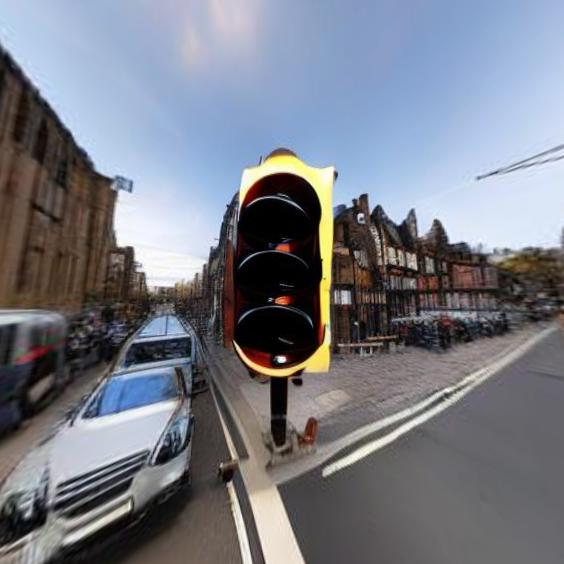}
    \includegraphics[width=0.15\linewidth]{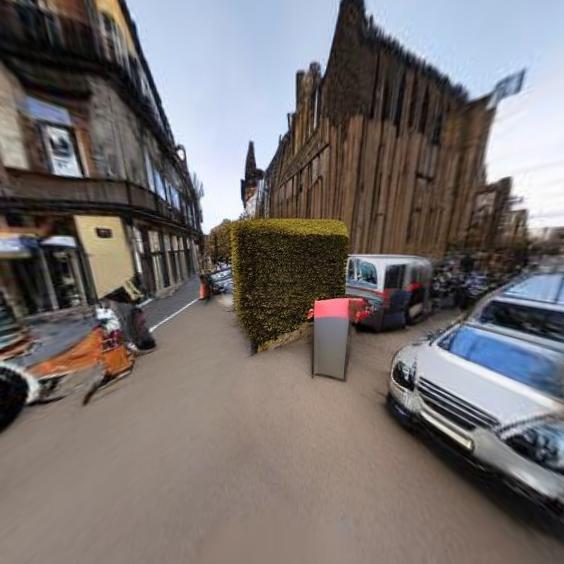}
    \includegraphics[width=0.15\linewidth]{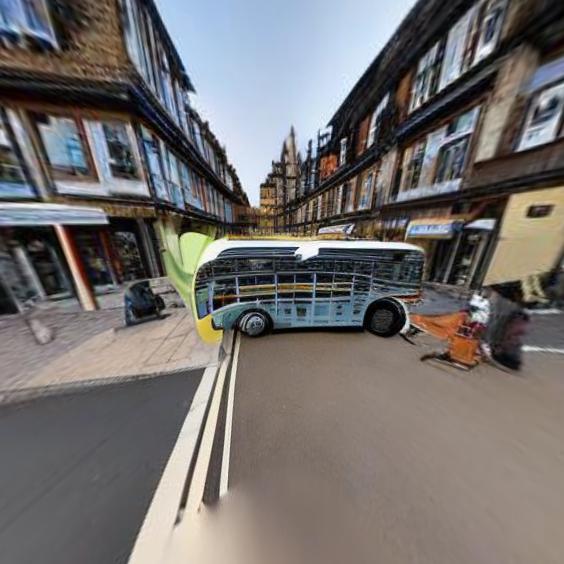}
    \includegraphics[width=0.15\linewidth]{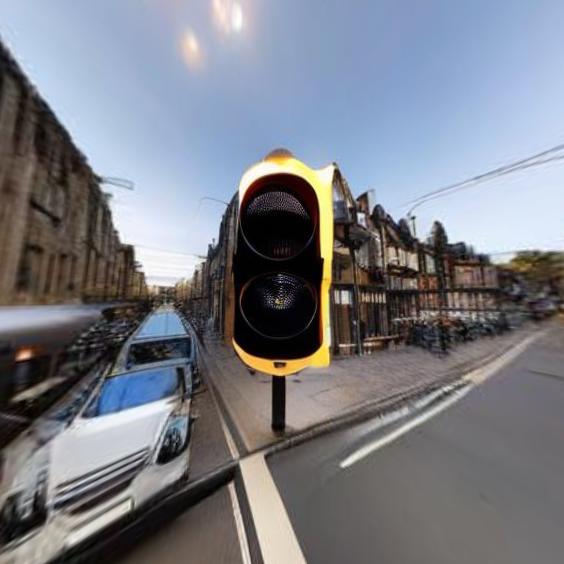}
    \includegraphics[width=0.15\linewidth]{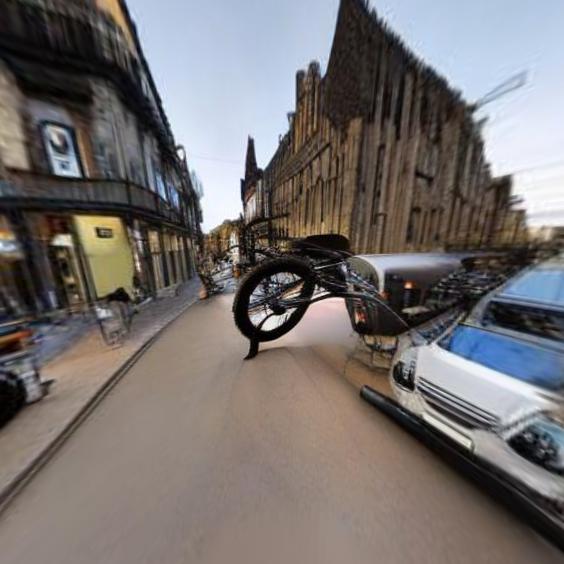}
    \includegraphics[width=0.15\linewidth]{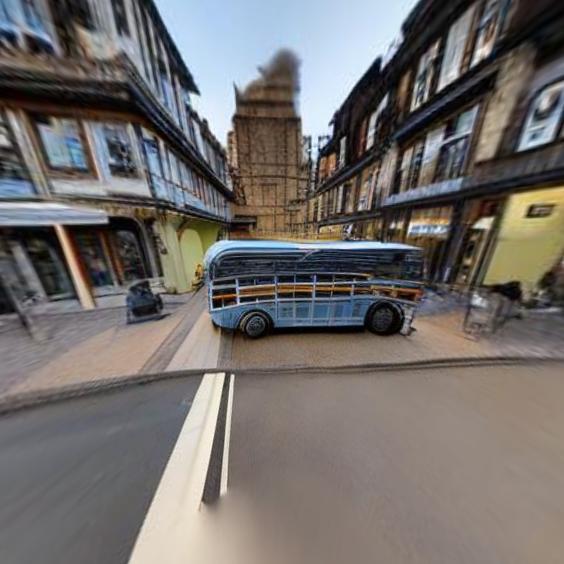}
    \includegraphics[width=0.15\linewidth]{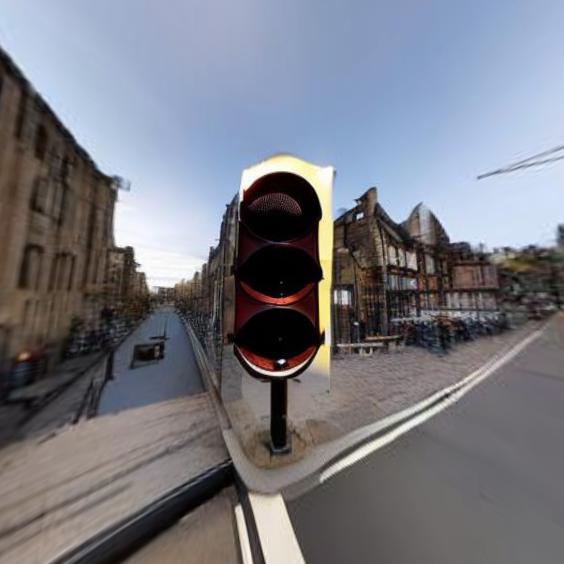}
    \includegraphics[width=0.15\linewidth]{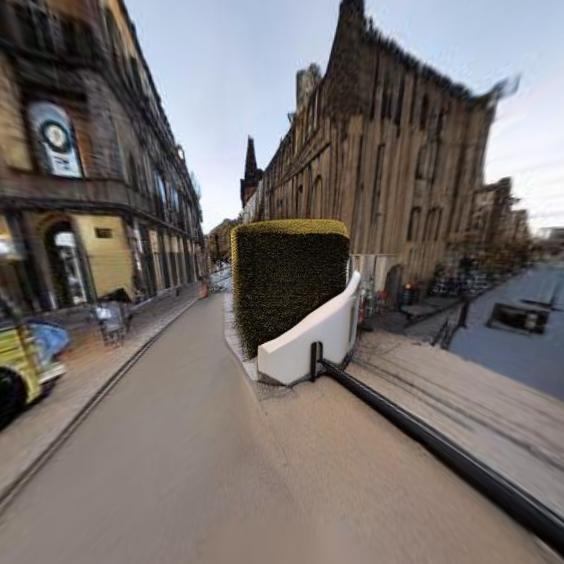}
    \includegraphics[width=0.15\linewidth]{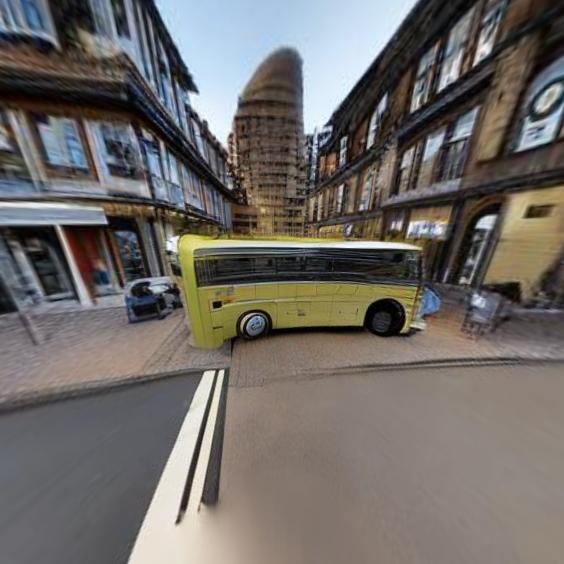}
    \includegraphics[width=0.15\linewidth]{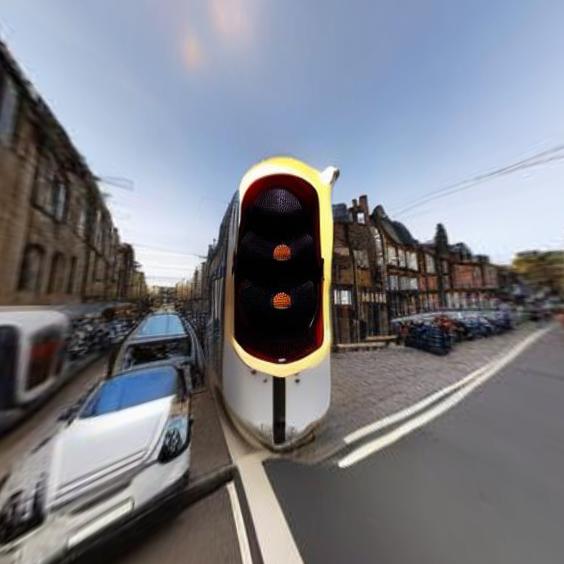}
    \includegraphics[width=0.15\linewidth]{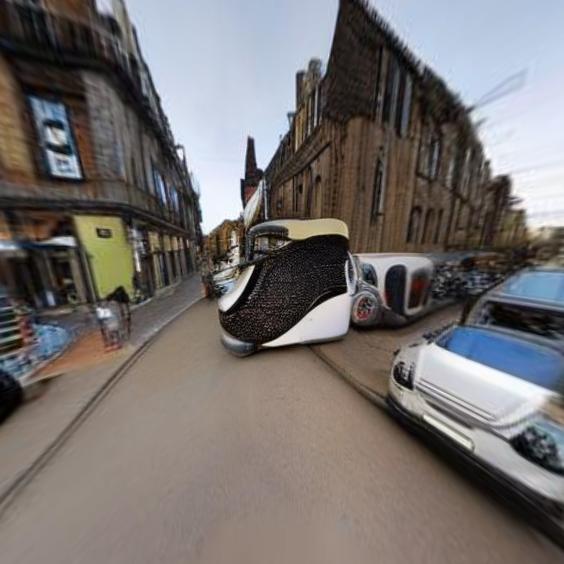}
    \includegraphics[width=0.15\linewidth]{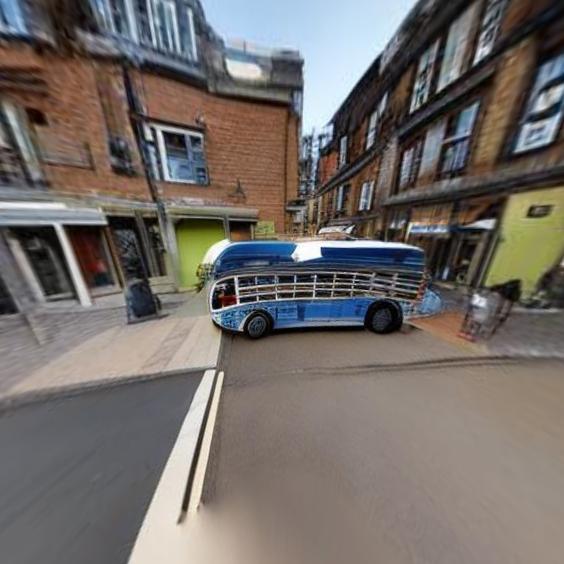}
    \includegraphics[width=0.15\linewidth]{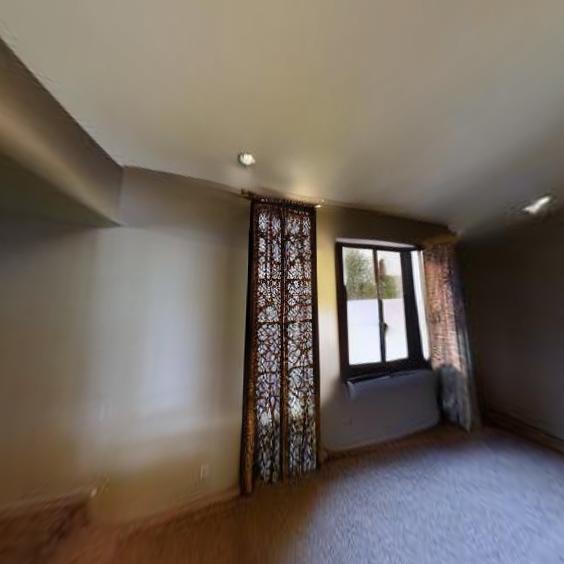}
    \includegraphics[width=0.15\linewidth]{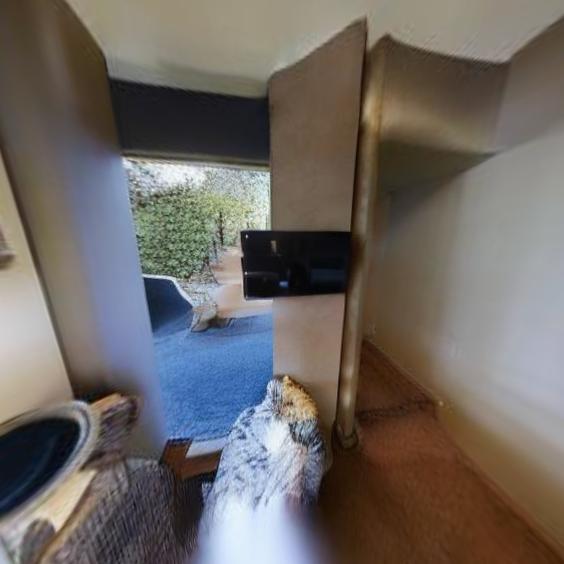}
    \includegraphics[width=0.15\linewidth]{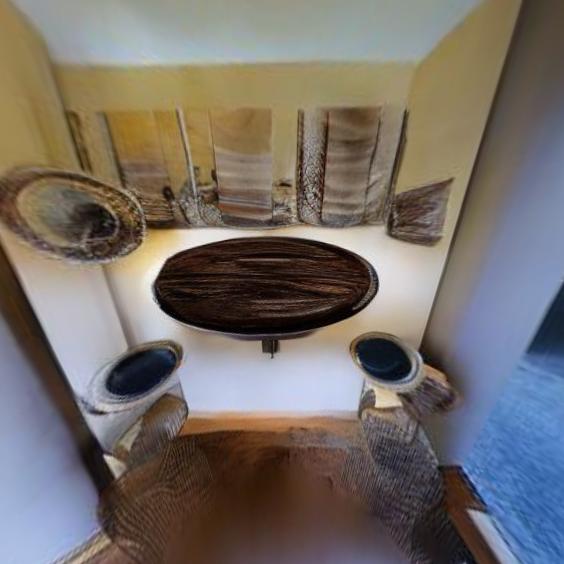}
    \includegraphics[width=0.15\linewidth]{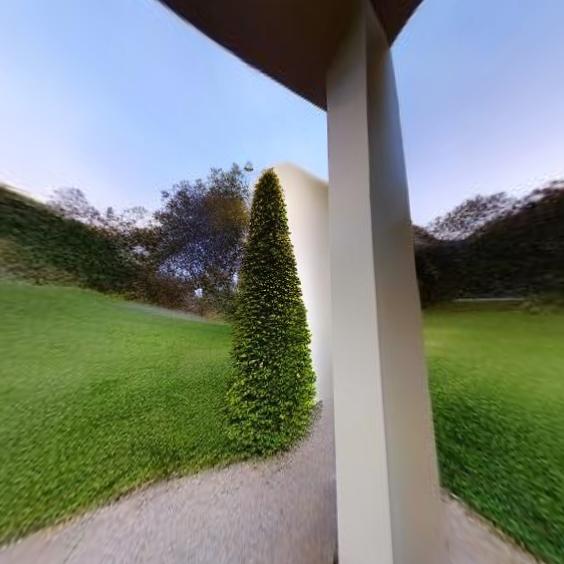}
    \includegraphics[width=0.15\linewidth]{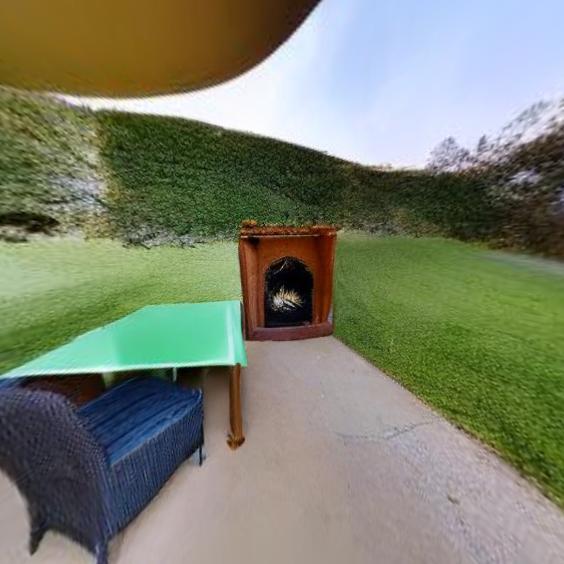}
    \includegraphics[width=0.15\linewidth]{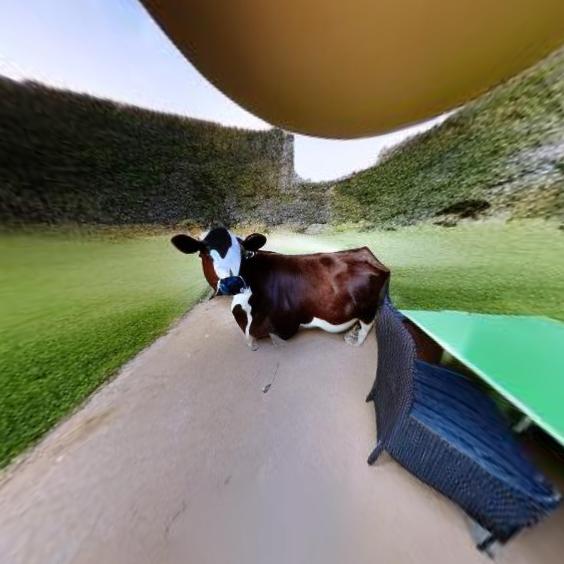}
    \includegraphics[width=0.15\linewidth]{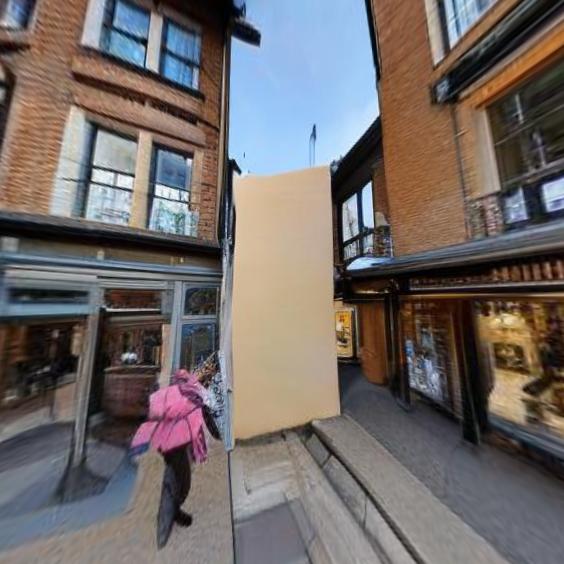}
    \includegraphics[width=0.15\linewidth]{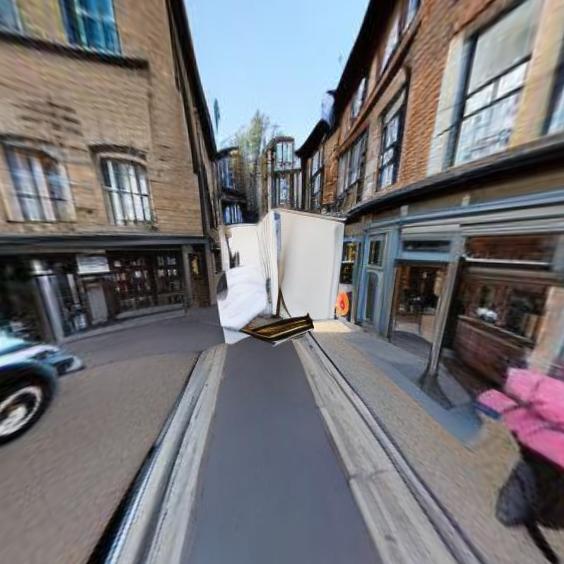}
    \includegraphics[width=0.15\linewidth]{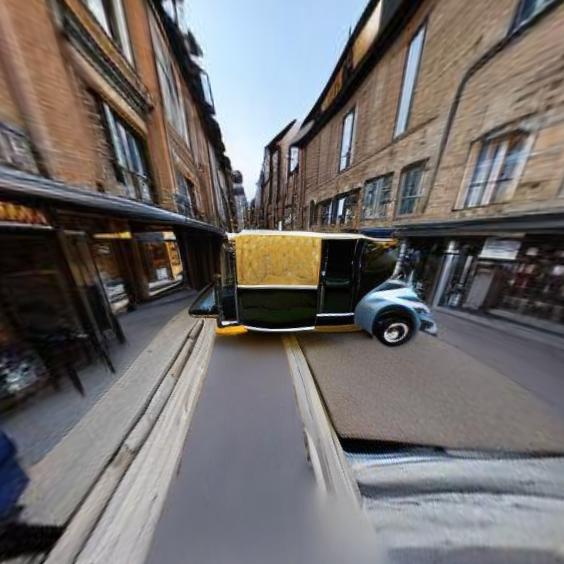}
    \includegraphics[width=0.15\linewidth]{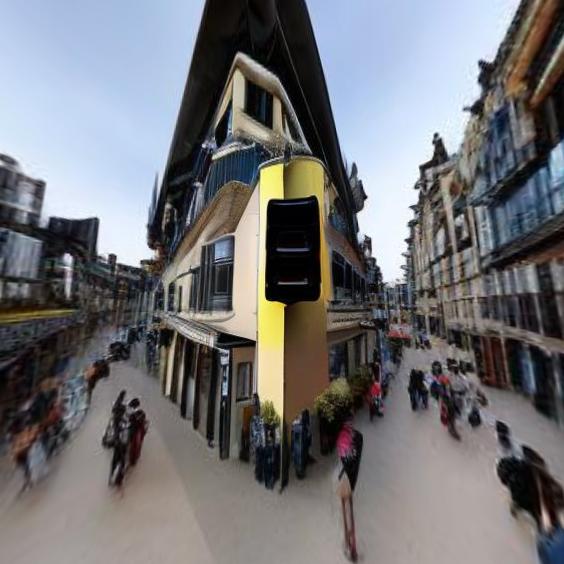}
    \includegraphics[width=0.15\linewidth]{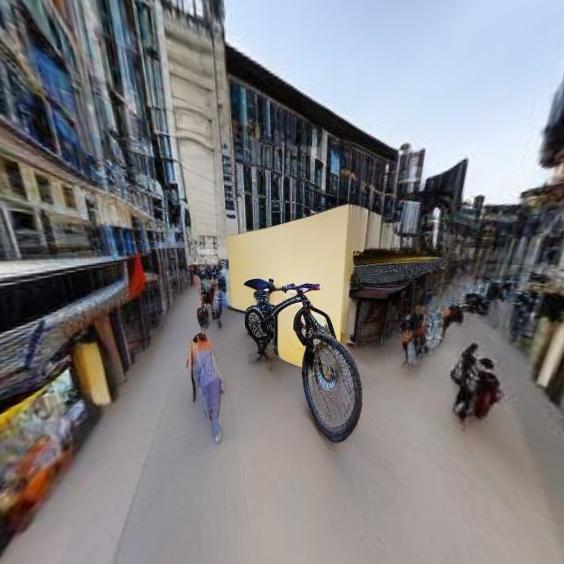}
    \includegraphics[width=0.15\linewidth]{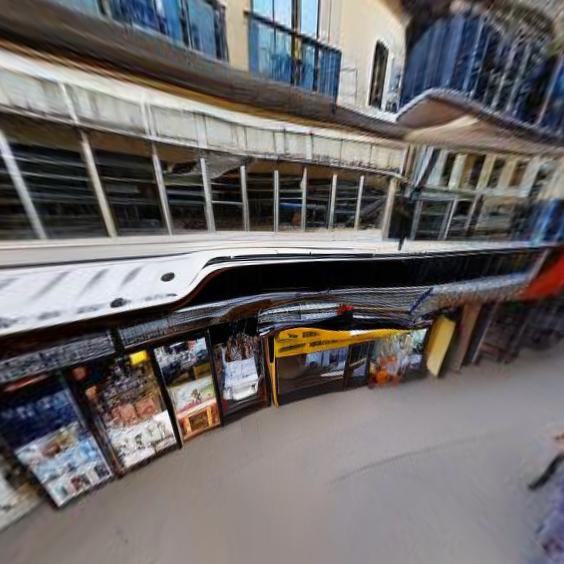}
    \caption{Perspective Images of MultiPanFusion's qualitative results}
    \label{fig:tangents_mpf}
\end{figure}

\begin{figure}[!ht]
    \centering
    \includegraphics[width=0.48\linewidth]{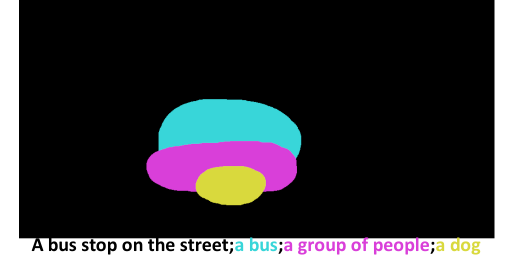}
    \includegraphics[width=0.48\linewidth]{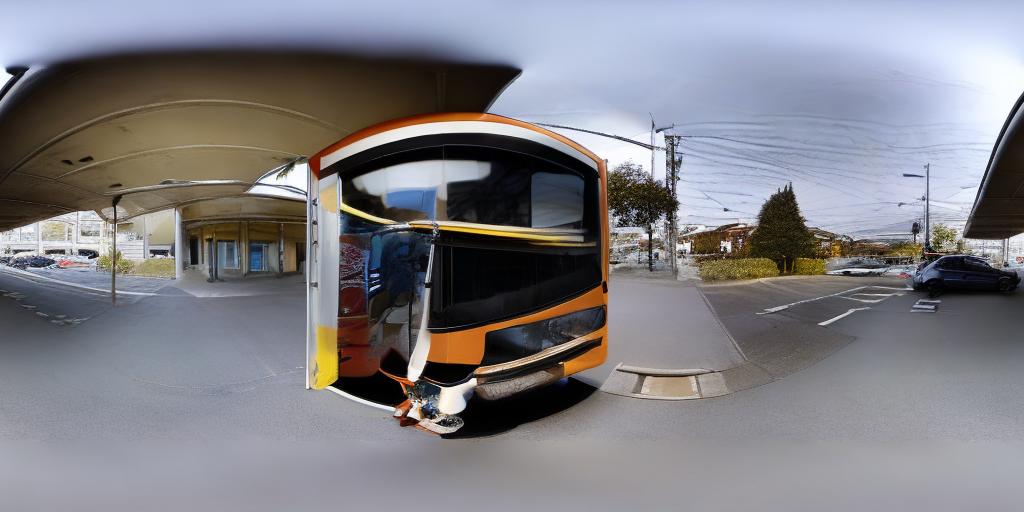}
    \hfill
    \includegraphics[width=0.48\linewidth]{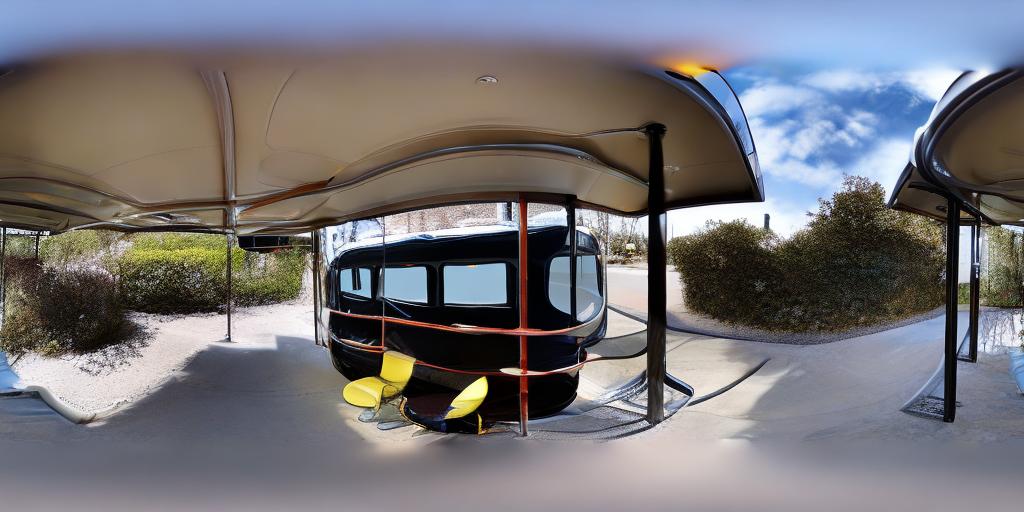}
    \includegraphics[width=0.48\linewidth]{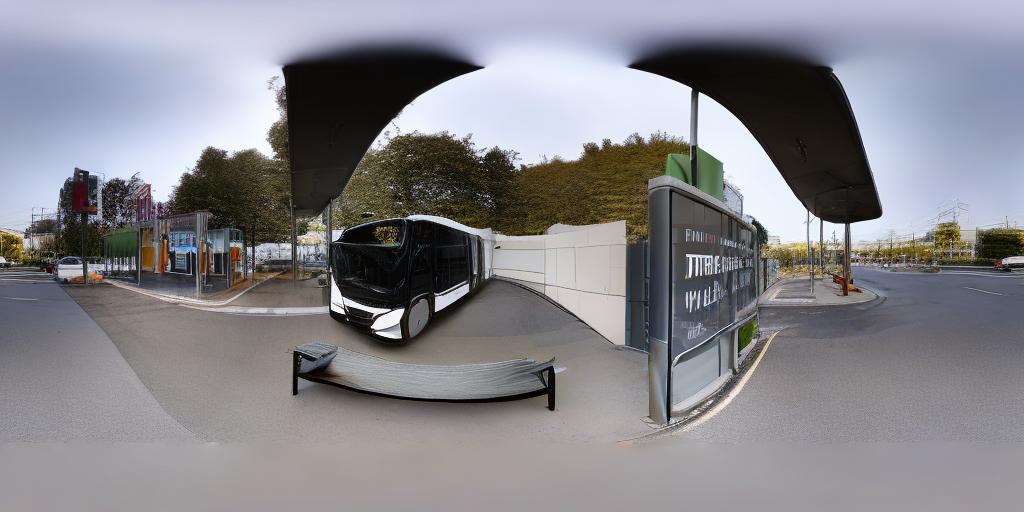}
    \includegraphics[width=0.48\linewidth]{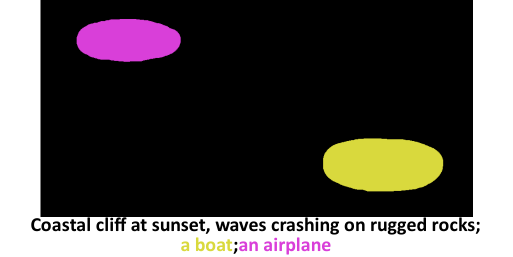}
    \includegraphics[width=0.48\linewidth]{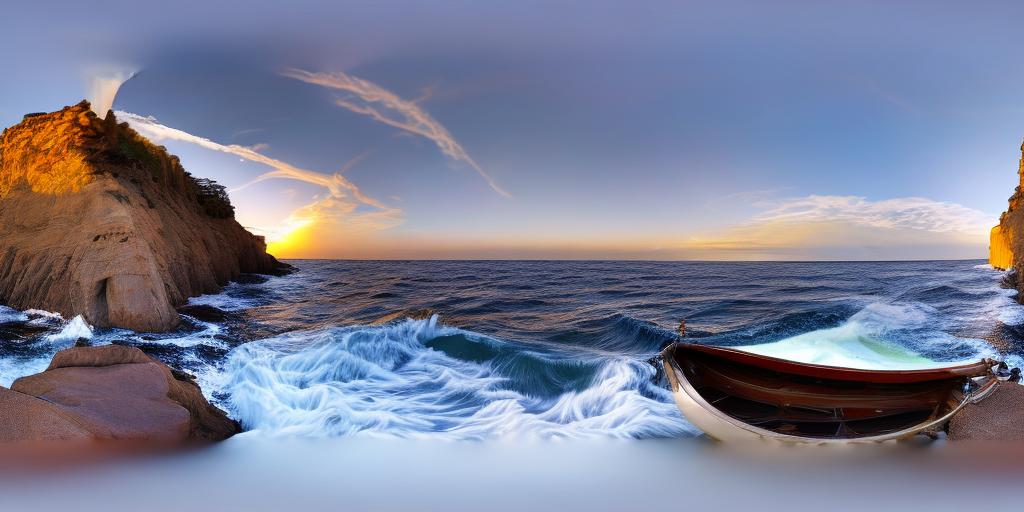}
    \includegraphics[width=0.48\linewidth]{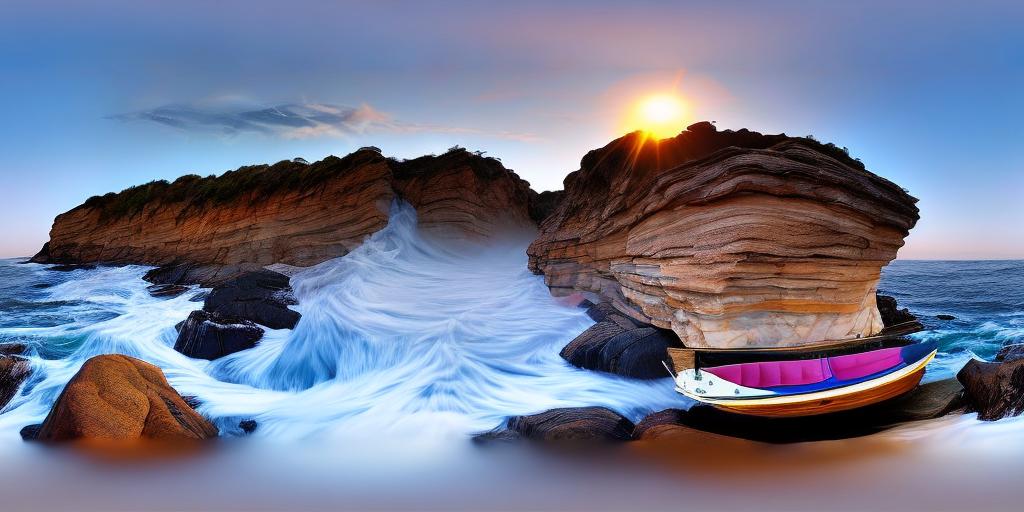}
    \includegraphics[width=0.48\linewidth]{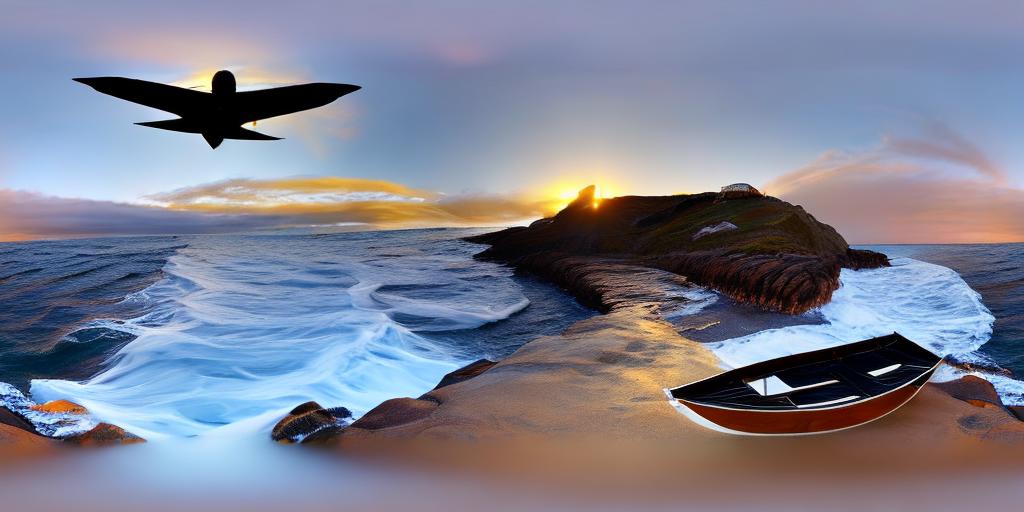}
    \includegraphics[width=0.48\linewidth]{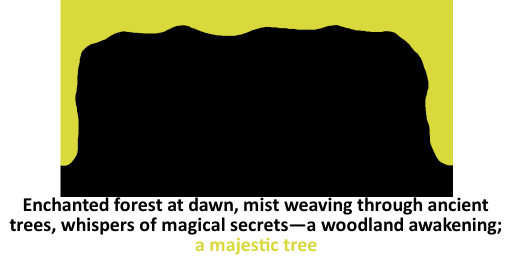}
    \includegraphics[width=0.48\linewidth]{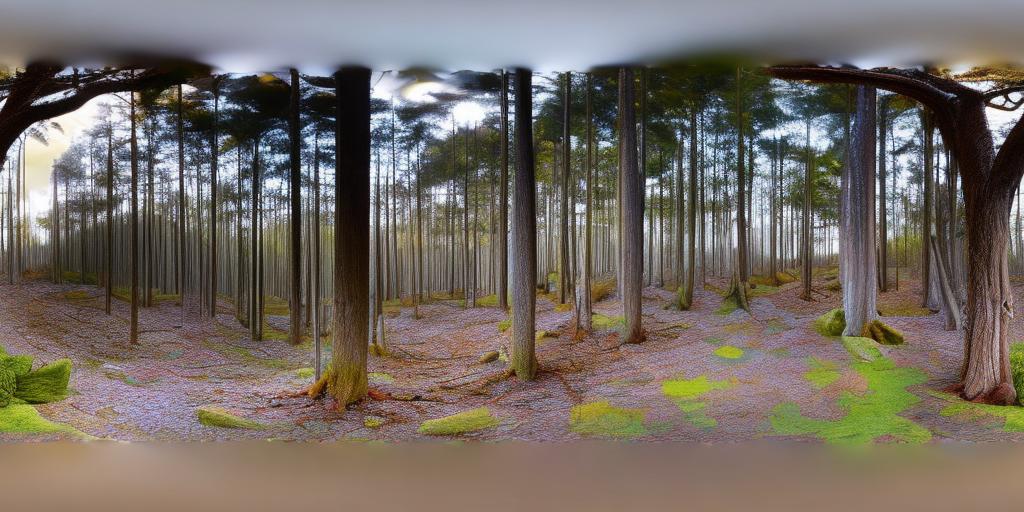}
    \includegraphics[width=0.48\linewidth]{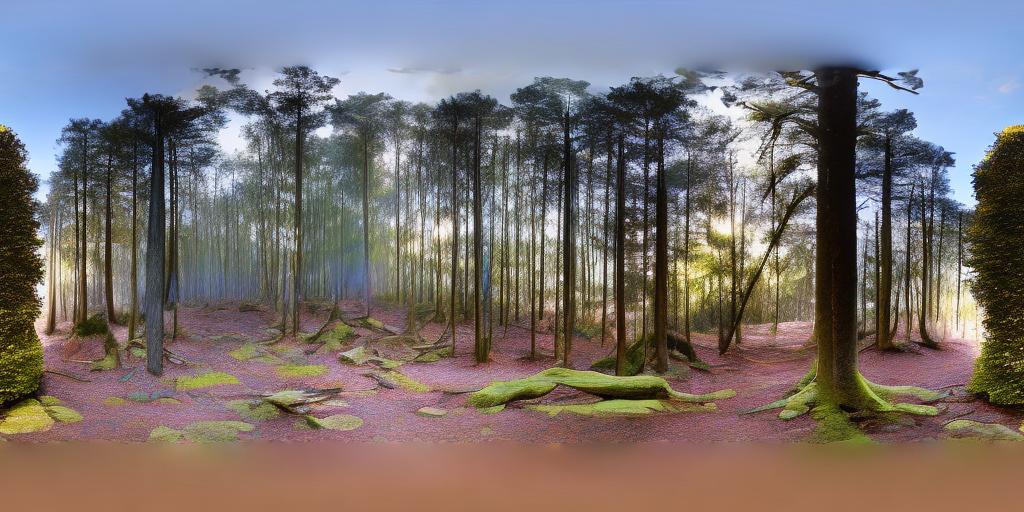}
    \includegraphics[width=0.48\linewidth]{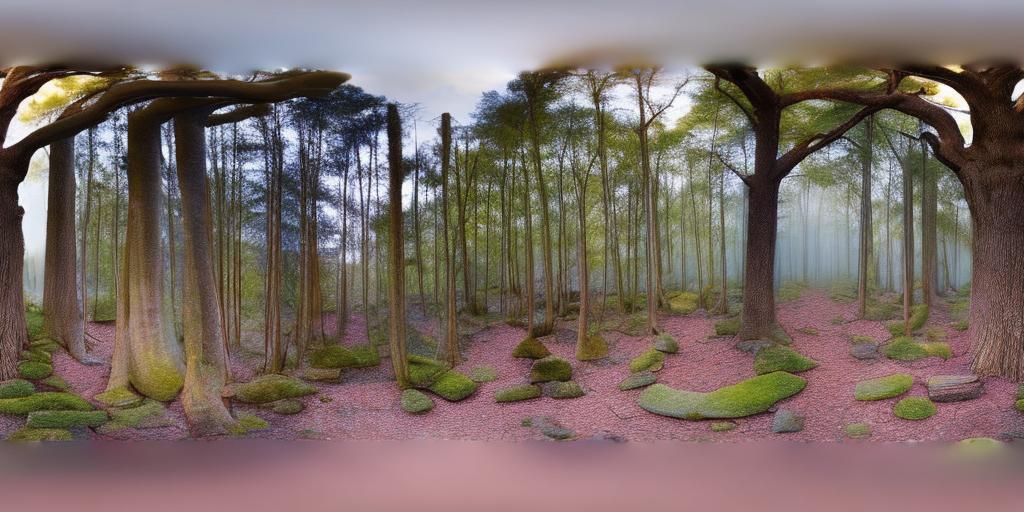}
    \includegraphics[width=0.48\linewidth]{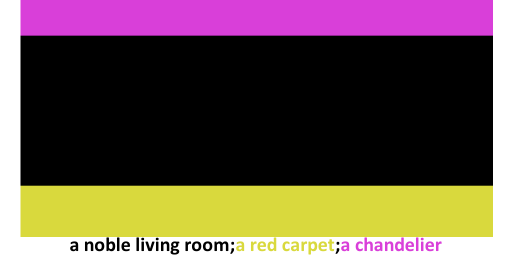}
    \includegraphics[width=0.48\linewidth]{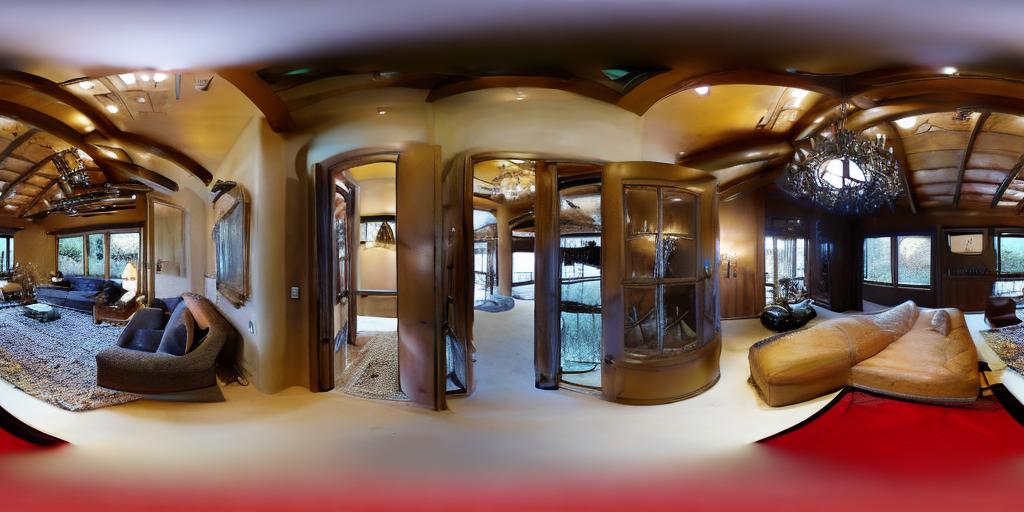}
    \includegraphics[width=0.48\linewidth]{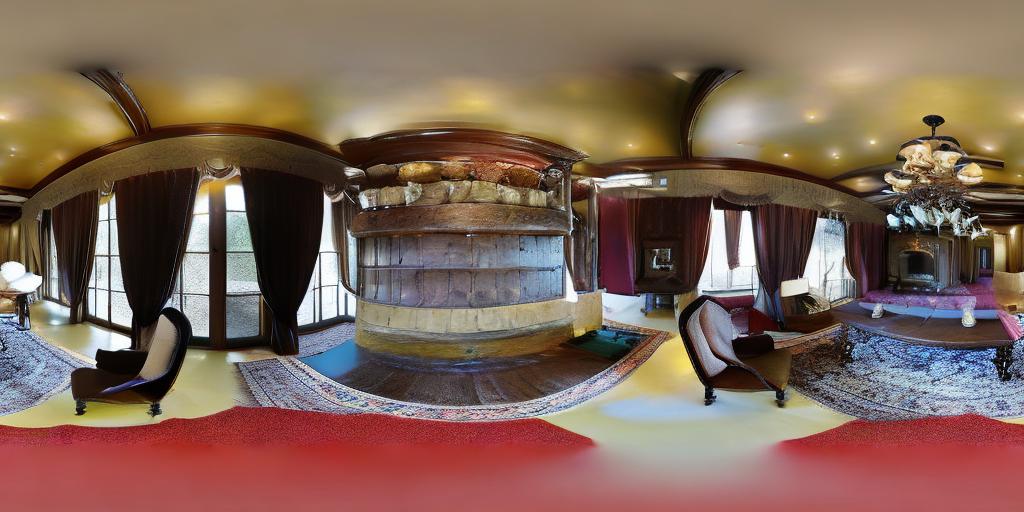}
    \includegraphics[width=0.48\linewidth]{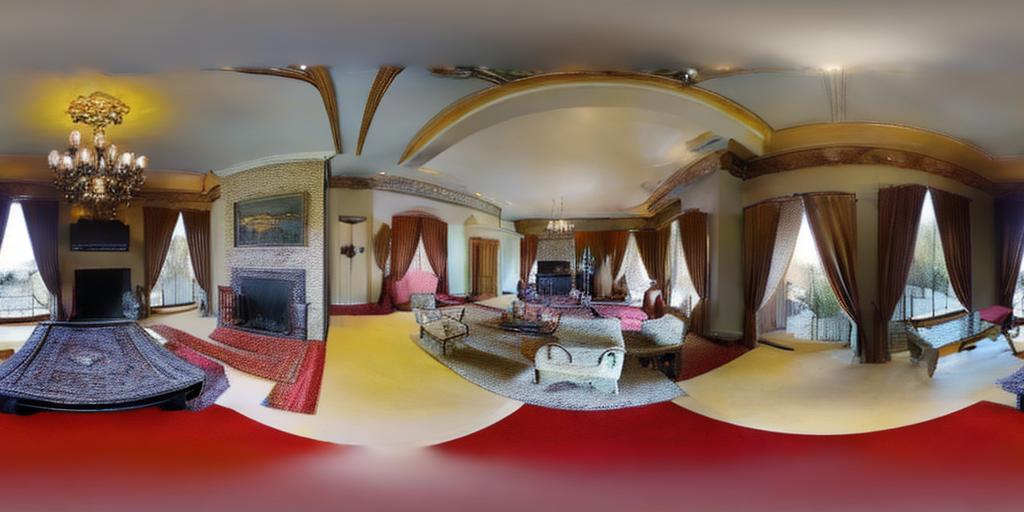}
    \caption{Corner cases of MultiPanFusion with free-hand masks. The model has problems synthesizing objects near the top and bottom poles. Elements at the horizontal image borders are unproblematic. Overlapping masks result in object-neglect and -distortion.}
    \label{fig:corner_cases_mpf}
\end{figure}

\begin{table*}[t]
    \centering
    \small
    \begin{tabular}{@{}ll|ccccc|ccccc@{}}
    \toprule
    & & \multicolumn{5}{c}{without StitchDiffusion process} & \multicolumn{5}{c}{with StitchDiffusion process} \\
    & & IoU$\uparrow$ & CS$\uparrow$ & IR$\uparrow$ & FID$\downarrow$ & CMMD$\downarrow$ & IoU$\uparrow$ & CS$\uparrow$ & IR$\uparrow$ & FID$\downarrow$ & CMMD$\downarrow$ \\
    \midrule
    bootstrap &
    1 & 0.36 & 27.90 & -0.77 & 66.81 & 1.71 & 0.37 & 28.02 & -0.73 & 68.11 & 1.73 \\
     & 5 & 0.41 & 28.14 & -0.68 & 63.50 & 1.68 & 0.41 & 28.28 & -0.62 & 64.64 & 1.71 \\
     & 10 & 0.49 & 28.59 & -0.53 & 60.17 & 1.65 & 0.49 & 28.74 & -0.48 & 61.23 & 1.68 \\
     & 15 & 0.59 & 28.91 & -0.47 & 59.53 & 1.64 & 0.59 & 29.08 & -0.41 & 59.69 & 1.65 \\
     & 20 & 0.66 & 29.15 & -0.42 & 60.99 & 1.60 & 0.67 & 29.25 & -0.37 & 61.12 & 1.63 \\
     & 25 & 0.70 & 29.30 & -0.41 & 62.23 & 1.59 & 0.70 & 29.37 & -0.37 & 62.94 & 1.60 \\
     & 30 & 0.72 & 29.33 & -0.42 & 63.29 & 1.59 & 0.72 & 29.38 & -0.39 & 63.63 & 1.59 \\
     & 35 & 0.73 & 29.34 & -0.43 & 64.08 & 1.59 & 0.73 & 29.40 & -0.41 & 63.97 & 1.61 \\
     & 40 & 0.74 & 29.33 & -0.45 & 65.03 & 1.59 & 0.73 & 29.39 & -0.43 & 65.28 & 1.61 \\
     & 45 & 0.74 & 29.32 & -0.47 & 66.53 & 1.60 & 0.74 & 29.36 & -0.45 & 66.58 & 1.63 \\
     & 50 & 0.74 & 29.27 & -0.49 & 68.24 & 1.67 & 0.74 & 29.31 & -0.47 & 68.05 & 1.68 \\
    \midrule
    stride &
    4 & 0.66 & 29.15 & -0.44 & 62.47 & 1.63 & 0.68 & 29.27 & -0.36 & 60.96 & 1.62 \\
     & 8 & 0.67 & 29.18 & -0.42 & 61.70 & 1.62 & 0.67 & 29.28 & -0.37 & 61.13 & 1.62 \\
     & 16 & 0.66 & 29.15 & -0.42 & 60.99 & 1.60 & 0.67 & 29.25 & -0.37 & 61.12 & 1.63 \\
     & 32 & 0.67 & 29.15 & -0.41 & 60.90 & 1.61 & 0.67 & 29.24 & -0.37 & 61.32 & 1.62 \\
    \midrule
    size &
    0 & 0.45 & 27.92 & -0.92 & 71.51 & 1.85 & 0.46 & 28.01 & -0.89 & 72.79 & 1.91 \\
     & 1 & 0.66 & 29.15 & -0.42 & 60.99 & 1.60 & 0.67 & 29.25 & -0.37 & 61.12 & 1.63 \\
     & 2 & \textbf{0.75} & 29.00 & -0.48 & 66.14 & \textbf{1.39} & \textbf{0.76} & 29.17 & -0.45 & 65.25 & \textbf{1.40} \\
    \midrule
    pano &
    None & 0.57 & 28.84 & -0.56 & 82.50 & 1.50 & 0.57 & 28.91 & -0.48 & 82.61 & 1.51 \\
     & BG & 0.66 & 29.15 & -0.42 & 60.99 & 1.60 & 0.67 & 29.25 & -0.37 & 61.12 & 1.63 \\
     & All & 0.56 & 28.66 & -0.47 & 67.78 & 1.70 & 0.55 & 28.74 & -0.43 & 68.42 & 1.72 \\
    \midrule
    proj &
    False & 0.62 & 29.04 & -0.49 & 61.35 & 1.61 & 0.63 & 29.16 & -0.43 & 61.40 & 1.62 \\
     & True & 0.66 & 29.15 & -0.42 & 60.99 & 1.60 & 0.67 & 29.25 & -0.37 & 61.12 & 1.63 \\
    \midrule
    mask idx &
    0 & 0.73 & \textbf{29.94} & -0.58 & 75.89 & 1.64 & 0.73 & \textbf{29.82} & -0.60 & 77.30 & 1.71 \\
     & 1 & 0.57 & 28.64 & -0.27 & 90.56 & 1.79 & 0.59 & 28.82 & \textbf{-0.16} & 91.10 & 1.77 \\
     & 2 & 0.69 & 29.06 & \textbf{-0.21} & 81.71 & 1.50 & 0.69 & 29.17 & -0.18 & 80.86 & 1.48 \\
     & 01 & 0.65 & 29.17 & -0.49 & 68.23 & 1.71 & 0.66 & 29.23 & -0.42 & 68.67 & 1.75 \\
     & 02 & 0.72 & 29.53 & -0.42 & 63.40 & 1.52 & 0.71 & 29.55 & -0.40 & 64.34 & 1.54 \\
     & 12 & 0.64 & 28.83 & -0.27 & 69.63 & 1.63 & 0.64 & 29.01 & -0.20 & 69.20 & 1.60 \\
     & 012 & 0.66 & 29.15 & -0.42 & 60.99 & 1.60 & 0.67 & 29.25 & -0.37 & 61.12 & 1.63 \\
    \midrule
    bootstrap &
    None & 0.66 & 29.15 & -0.42 & 60.99 & 1.60 & 0.67 & 29.25 & -0.37 & 61.12 & 1.63 \\
    coupling & Objects & 0.67 & 29.20 & -0.41 & 61.04 & 1.62 & 0.67 & 29.30 & -0.36 & 61.06 & 1.63 \\
    \midrule
    noise &
    False & 0.67 & 29.21 & -0.41 & 61.42 & 1.61 & 0.68 & 29.32 & -0.36 & 61.84 & 1.63 \\
    coupling & True & 0.66 & 29.15 & -0.42 & 60.99 & 1.60 & 0.67 & 29.25 & -0.37 & 61.12 & 1.63 \\
    \midrule
    global &
    False & 0.66 & 29.15 & -0.42 & 60.99 & 1.60 & 0.67 & 29.25 & -0.37 & 61.12 & 1.63 \\
    prompt & True & 0.58 & 29.23 & -0.41 & \textbf{56.38} & 1.57 & 0.59 & 29.36 & -0.35 & \textbf{57.03} & 1.59 \\
    \bottomrule
    \end{tabular}
    \caption{Complete results with MultiDiffusion and MultiStitchDiffusion}
    \label{tab:md_mstd}
\end{table*}
\newpage
\begin{table*}[t]
    \centering
    \small
    \begin{tabular}{@{}ll|ccccc|ccccc@{}}
    \toprule
    & & \multicolumn{5}{c}{MPF (md\_both)} & \multicolumn{5}{c}{MPF (md\_pano)} \\
    & & IoU$\uparrow$ & CS$\uparrow$ & IR$\uparrow$ & FID$\downarrow$ & CMMD$\downarrow$ & IoU$\uparrow$ & CS$\uparrow$ & IR$\uparrow$ & FID$\downarrow$ & CMMD$\downarrow$ \\
    \midrule
    bootstrap &
    1 & 0.16 & 25.18 & -1.51 & 90.99 & 2.31 & 0.13 & 25.01 & -1.57 & 93.00 & 2.33 \\
     & 5 & 0.19 & 25.38 & -1.46 & 90.13 & 2.29 & 0.18 & 25.22 & -1.51 & 91.73 & 2.31 \\
     & 10 & 0.27 & 25.69 & -1.36 & 87.13 & 2.23 & 0.26 & 25.61 & -1.38 & 87.59 & 2.24 \\
     & 15 & 0.36 & 25.97 & -1.29 & 84.72 & 2.15 & 0.36 & 25.95 & -1.26 & 84.85 & 2.15 \\
     & 20 & 0.44 & 26.14 & -1.22 & 84.82 & 2.09 & 0.45 & 26.16 & -1.19 & 84.60 & 2.09 \\
     & 25 & 0.49 & 26.25 & -1.20 & 87.12 & 2.04 & 0.50 & 26.34 & -1.16 & 85.26 & 2.06 \\
     & 30 & 0.52 & 26.33 & -1.19 & 88.66 & 2.01 & 0.53 & 26.42 & -1.14 & 85.50 & 2.03 \\
     & 35 & 0.53 & 26.36 & -1.19 & 90.03 & 2.00 & 0.54 & 26.48 & -1.13 & 86.26 & 2.03 \\
     & 40 & 0.54 & 26.39 & -1.19 & 91.23 & 2.00 & 0.55 & 26.49 & -1.13 & 87.28 & 2.02 \\
     & 45 & 0.55 & 26.43 & -1.19 & 92.15 & 2.02 & 0.56 & 26.48 & -1.13 & 88.47 & 2.03 \\
     & 50 & \textbf{0.56} & 26.43 & -1.20 & 93.76 & 2.06 & \textbf{0.57} & 26.42 & -1.14 & 90.43 & 2.05 \\
    \midrule
    size &
    0 & 0.21 & 25.18 & -1.55 & 90.67 & 2.30 & 0.23 & 25.15 & -1.54 & 90.28 & 2.29 \\
     & 1 & 0.44 & 26.14 & -1.22 & 84.82 & 2.09 & 0.45 & 26.16 & -1.19 & 84.60 & 2.09 \\
     & 2 & 0.55 & 25.99 & -1.32 & 100.59 & \textbf{1.92} & 0.54 & 26.22 & -1.25 & 95.89 & 1.93 \\
    \midrule
    pano &
    BG & 0.43 & 26.45 & -1.23 & 81.90 & 2.07 & 0.49 & 26.89 & -1.07 & 76.57 & 2.03 \\
     & All & 0.44 & 26.14 & -1.22 & 84.82 & 2.09 & 0.45 & 26.16 & -1.19 & 84.60 & 2.09 \\
    \midrule
    proj &
    False & 0.39 & 25.93 & -1.29 & 85.31 & 2.11 & 0.39 & 25.97 & -1.26 & 83.77 & 2.10 \\
     & True & 0.44 & 26.14 & -1.22 & 84.82 & 2.09 & 0.45 & 26.16 & -1.19 & 84.60 & 2.09 \\
    \midrule
    mask\_idx &
    0 & 0.53 & \textbf{27.11} & -1.39 & 129.58 & 2.20 & 0.53 & \textbf{27.02} & -1.41 & 132.93 & 2.23 \\
     & 1 & 0.35 & 25.55 & -1.18 & 113.10 & 2.22 & 0.34 & 25.36 & -1.18 & 111.46 & 2.21 \\
     & 2 & 0.48 & 25.98 & \textbf{-1.05} & 96.15 & 1.98 & 0.47 & 25.91 & \textbf{-1.06} & 97.75 & 1.97 \\
     & 01 & 0.43 & 26.28 & -1.30 & 102.09 & 2.20 & 0.43 & 26.18 & -1.29 & 101.13 & 2.19 \\
     & 02 & 0.50 & 26.55 & -1.22 & 90.48 & 2.06 & 0.50 & 26.53 & -1.20 & 92.74 & 2.03 \\
     & 12 & 0.40 & 25.68 & -1.13 & 86.82 & 2.08 & 0.41 & 25.65 & -1.11 & 84.95 & 2.05 \\
     & 012 & 0.44 & 26.14 & -1.22 & 84.82 & 2.09 & 0.45 & 26.16 & -1.19 & 84.60 & 2.09 \\
    \midrule
    bootstrap &
    None & 0.43 & 26.04 & -1.26 & 85.73 & 2.11 & 0.45 & 26.16 & -1.19 & 84.63 & 2.09 \\
    coupling & Branches & 0.44 & 26.14 & -1.22 & 84.82 & 2.09 & 0.45 & 26.16 & -1.19 & 84.60 & 2.09 \\
     & Objects & 0.44 & 26.01 & -1.25 & 86.53 & 2.10 & 0.44 & 26.02 & -1.22 & 84.79 & 2.08 \\
     & All & 0.44 & 26.09 & -1.23 & 84.95 & 2.08 & 0.44 & 26.03 & -1.22 & 84.85 & 2.08 \\
    \midrule
    noise &
    False & 0.46 & 26.22 & -1.19 & 84.49 & 2.10 & 0.46 & 26.24 & -1.15 & 81.78 & 2.10 \\
    coupling & True & 0.44 & 26.14 & -1.22 & 84.82 & 2.09 & 0.45 & 26.16 & -1.19 & 84.60 & 2.09 \\
    \midrule
    global &
    False & 0.44 & 26.14 & -1.22 & 84.82 & 2.09 & 0.45 & 26.16 & -1.19 & 84.60 & 2.09 \\
    prompt & True & 0.41 & 26.43 & -1.32 & \textbf{77.20} & 2.08 & 0.43 & 26.63 & -1.25 & \textbf{74.84} & 2.05 \\
    \midrule
    fg\_eppa &
    False & 0.52 & 26.63 & -1.10 & 82.21 & 2.06 & 0.53 & 26.70 & -1.09 & 80.61 & \textbf{1.97} \\
     & True & 0.44 & 26.14 & -1.22 & 84.82 & 2.09 & 0.45 & 26.16 & -1.19 & 84.60 & 2.09 \\
    \bottomrule
    \end{tabular}
    \caption{Results with MultiPanFusion (md\_both and md\_pano)}
    \label{tab:md_pano}
\end{table*}
\newpage
\begin{table*}[t]
    \centering
    \small
    \begin{tabular}{@{}ll|ccccc|ccccc@{}}
    \toprule
    & & \multicolumn{5}{c}{MPF (md\_both)} & \multicolumn{5}{c}{MPF (md\_pers)} \\
    & & IoU$\uparrow$ & CS$\uparrow$ & IR$\uparrow$ & FID$\downarrow$ & CMMD$\downarrow$ & IoU$\uparrow$ & CS$\uparrow$ & IR$\uparrow$ & FID$\downarrow$ & CMMD$\downarrow$ \\
    \midrule
    bootstrap &
    1 & 0.16 & 25.18 & -1.51 & 90.99 & 2.31 & 0.05 & 24.53 & -1.77 & 104.68 & 2.37 \\
     & 5 & 0.19 & 25.38 & -1.46 & 90.13 & 2.29 & 0.05 & 24.51 & -1.77 & 105.25 & 2.37 \\
     & 10 & 0.27 & 25.69 & -1.36 & 87.13 & 2.23 & 0.05 & 24.49 & -1.77 & 105.44 & 2.38 \\
     & 15 & 0.36 & 25.97 & -1.29 & 84.72 & 2.15 & 0.05 & 24.46 & -1.78 & 106.40 & 2.40 \\
     & 20 & 0.44 & 26.14 & -1.22 & 84.82 & 2.09 & 0.05 & 24.42 & -1.79 & 107.00 & 2.41 \\
     & 25 & 0.49 & 26.25 & -1.20 & 87.12 & 2.04 & 0.05 & 24.39 & -1.79 & 107.73 & 2.42 \\
     & 30 & 0.52 & 26.33 & -1.19 & 88.66 & 2.01 & 0.05 & 24.36 & -1.80 & 108.24 & 2.43 \\
     & 35 & 0.53 & 26.36 & -1.19 & 90.03 & 2.00 & 0.05 & 24.35 & -1.80 & 108.81 & 2.44 \\
     & 40 & 0.54 & 26.39 & -1.19 & 91.23 & 2.00 & 0.05 & 24.36 & -1.80 & 109.04 & 2.45 \\
     & 45 & 0.55 & 26.43 & -1.19 & 92.15 & 2.02 & 0.05 & 24.35 & -1.80 & 109.27 & 2.45 \\
     & 50 & \textbf{0.56} & 26.43 & -1.20 & 93.76 & 2.06 & 0.05 & 24.34 & -1.80 & 109.51 & 2.45 \\
    \midrule
    size &
    0 & 0.21 & 25.18 & -1.55 & 90.67 & 2.30 & 0.04 & 24.50 & -1.76 & 107.02 & 2.44 \\
     & 1 & 0.44 & 26.14 & -1.22 & 84.82 & 2.09 & 0.05 & 24.42 & -1.79 & 107.00 & 2.41 \\
     & 2 & 0.55 & 25.99 & -1.32 & 100.59 & \textbf{1.92} & 0.06 & 24.37 & -1.83 & 122.58 & 2.46 \\
    \midrule
    pano &
    BG & 0.43 & 26.45 & -1.23 & 81.90 & 2.07 & 0.05 & 24.36 & -1.81 & 108.05 & 2.42 \\
     & All & 0.44 & 26.14 & -1.22 & 84.82 & 2.09 & 0.05 & 24.42 & -1.79 & 107.00 & 2.41 \\
    \midrule
    proj &
    False & 0.39 & 25.93 & -1.29 & 85.31 & 2.11 & 0.05 & 24.42 & -1.78 & 105.76 & 2.40 \\
     & True & 0.44 & 26.14 & -1.22 & 84.82 & 2.09 & 0.05 & 24.42 & -1.79 & 107.00 & 2.41 \\
    \midrule
    mask\_idx &
    0 & 0.53 & \textbf{27.11} & -1.39 & 129.58 & 2.20 & 0.05 & 24.50 & -1.96 & 162.83 & 2.68 \\
     & 1 & 0.35 & 25.55 & -1.18 & 113.10 & 2.22 & 0.02 & 24.29 & -1.85 & 134.22 & 2.45 \\
     & 2 & 0.48 & 25.98 & \textbf{-1.05} & 96.15 & 1.98 & 0.08 & 24.50 & \textbf{-1.55} & 135.32 & 2.38 \\
     & 01 & 0.43 & 26.28 & -1.30 & 102.09 & 2.20 & 0.03 & 24.39 & -1.91 & 123.44 & 2.49 \\
     & 02 & 0.50 & 26.55 & -1.22 & 90.48 & 2.06 & 0.06 & 24.50 & -1.75 & 120.81 & 2.43 \\
     & 12 & 0.40 & 25.68 & -1.13 & 86.82 & 2.08 & 0.05 & 24.41 & -1.71 & 109.92 & 2.36 \\
     & 012 & 0.44 & 26.14 & -1.22 & 84.82 & 2.09 & 0.05 & 24.42 & -1.79 & 107.00 & 2.41 \\
    \midrule
    bootstrap &
    None & 0.43 & 26.04 & -1.26 & 85.73 & 2.11 & 0.05 & 24.41 & -1.79 & 107.09 & 2.41 \\
    coupling & Branches & 0.44 & 26.14 & -1.22 & 84.82 & 2.09 & 0.05 & 24.42 & -1.79 & 107.00 & 2.41 \\
     & Objects & 0.44 & 26.01 & -1.25 & 86.53 & 2.10 & 0.05 & 24.42 & -1.79 & 106.97 & 2.41 \\
     & All & 0.44 & 26.09 & -1.23 & 84.95 & 2.08 & 0.05 & 24.42 & -1.79 & 107.02 & 2.41 \\
    \midrule
    noise &
    False & 0.46 & 26.22 & -1.19 & 84.49 & 2.10 & 0.05 & 24.41 & -1.78 & 106.23 & 2.42 \\
    coupling & True & 0.44 & 26.14 & -1.22 & 84.82 & 2.09 & 0.05 & 24.42 & -1.79 & 107.00 & 2.41 \\
    \midrule
    global &
    False & 0.44 & 26.14 & -1.22 & 84.82 & 2.09 & 0.05 & 24.42 & -1.79 & 107.00 & 2.41 \\
    prompt & True & 0.41 & 26.43 & -1.32 & \textbf{77.20} & 2.08 & \textbf{0.08} & \textbf{25.38} & -1.60 & \textbf{81.94} & \textbf{2.29} \\
    \midrule
    fg\_eppa &
    False & 0.52 & 26.63 & -1.10 & 82.21 & 2.06 & 0.05 & 24.44 & -1.79 & 106.68 & 2.42 \\
     & True & 0.44 & 26.14 & -1.22 & 84.82 & 2.09 & 0.05 & 24.42 & -1.79 & 107.00 & 2.41 \\
    \bottomrule
    \end{tabular}
    \caption{Results with MultiPanFusion (md\_both and md\_pers)}
    \label{tab:md_pers}
\end{table*}

\begin{figure}[!ht]
    \centering
    \includegraphics[width=0.48\linewidth]{fig/plots/iou_bootstrapping.pdf}
    \includegraphics[width=0.48\linewidth]{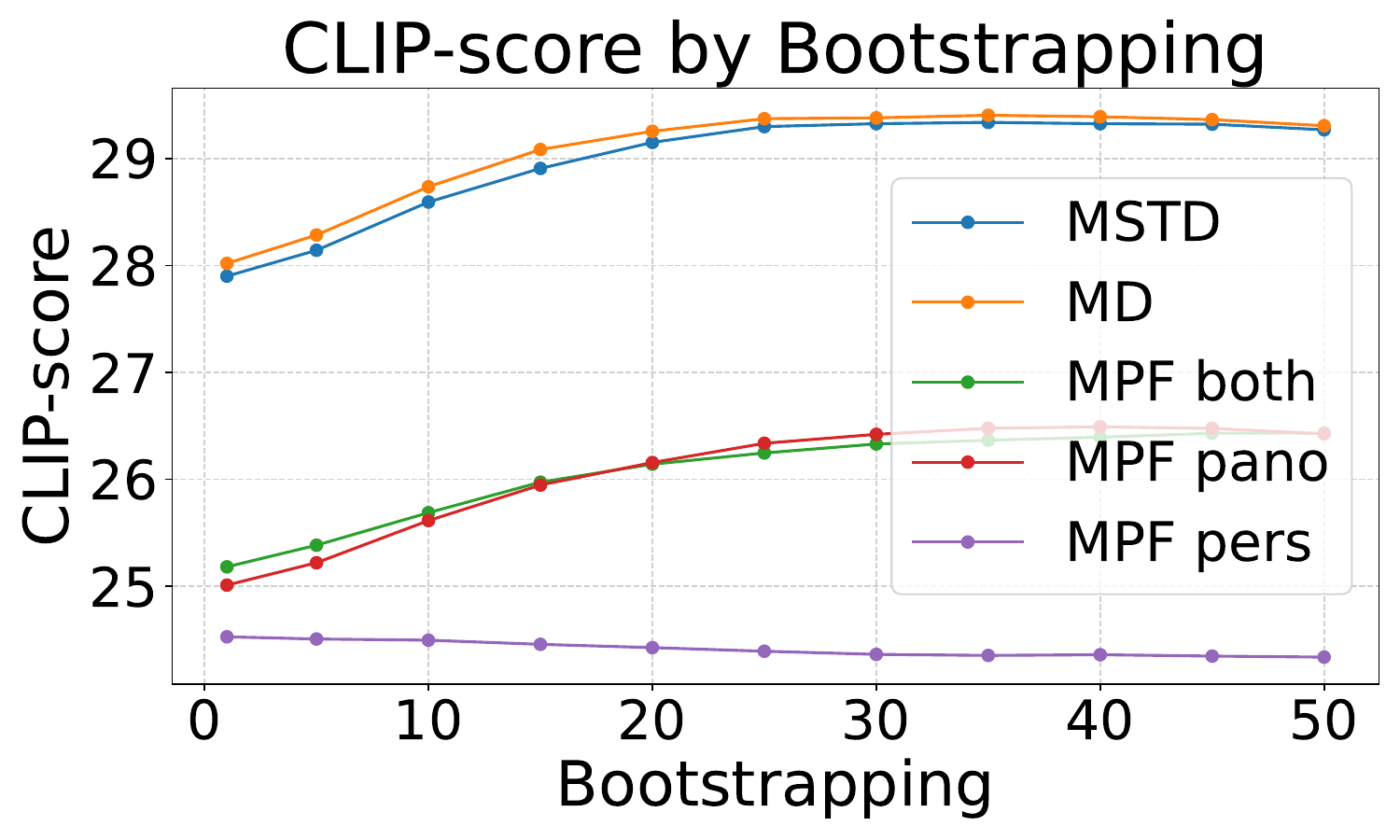}
    \hfill
    \includegraphics[width=0.48\linewidth]{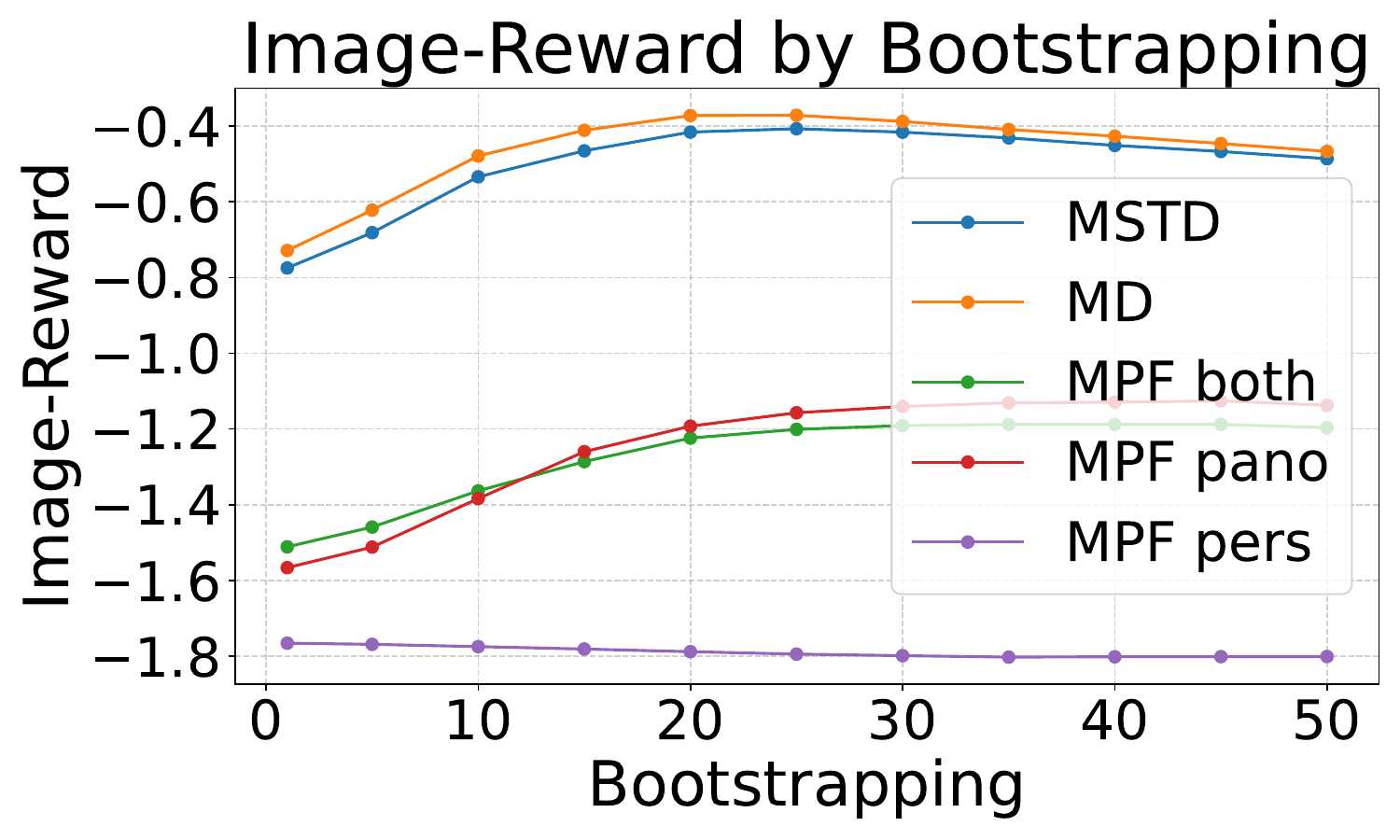}
    \includegraphics[width=0.48\linewidth]{fig/plots/fid_bootstrapping.pdf}
    \includegraphics[width=0.48\linewidth]{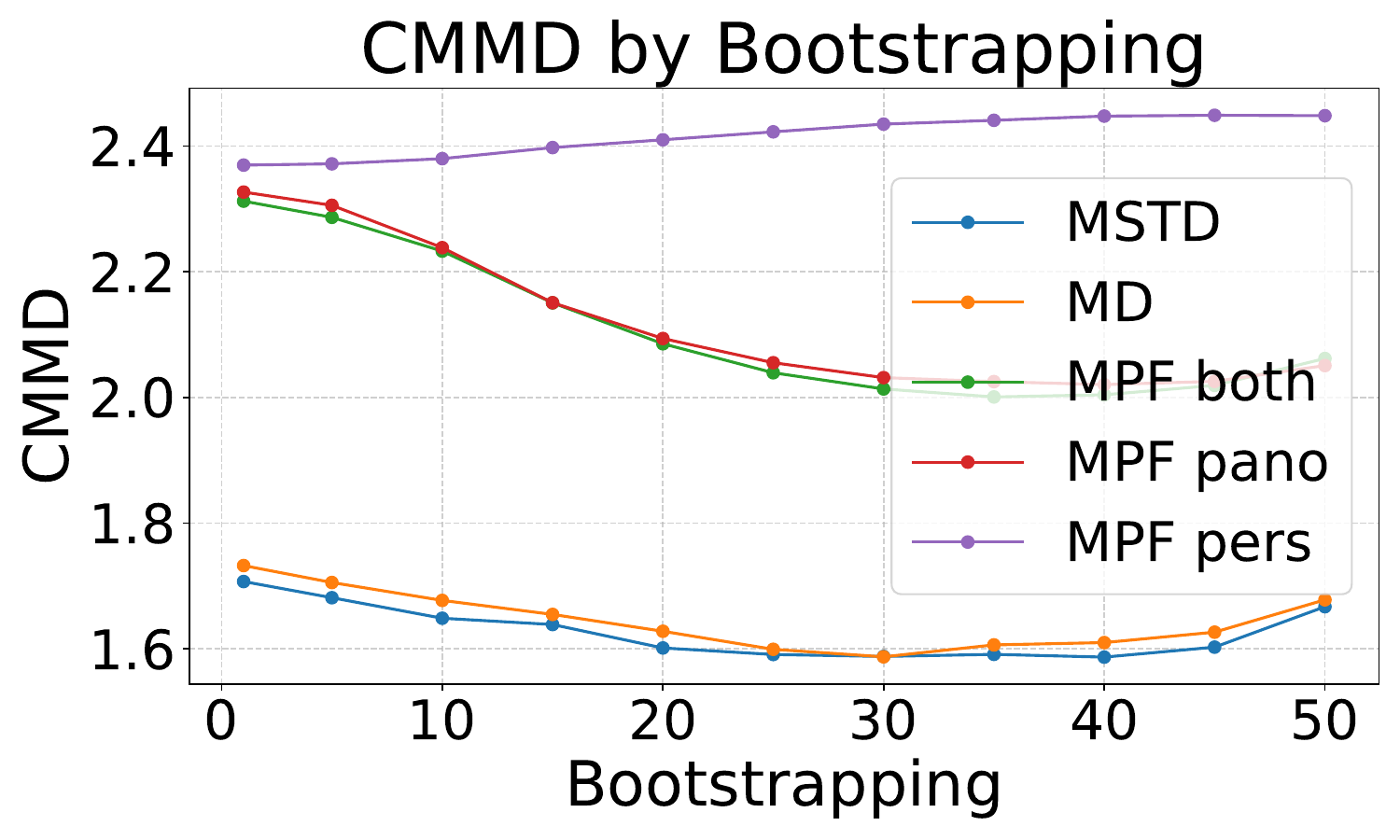}
    \caption{The influence of bootstrapping on our metrics for every approach, showing a functional relationship which is non-monotonous at FID.}
    \label{fig:plots_bootstrapping}
\end{figure}
\begin{figure}[!ht]
    \centering
    \includegraphics[width=0.48\linewidth]{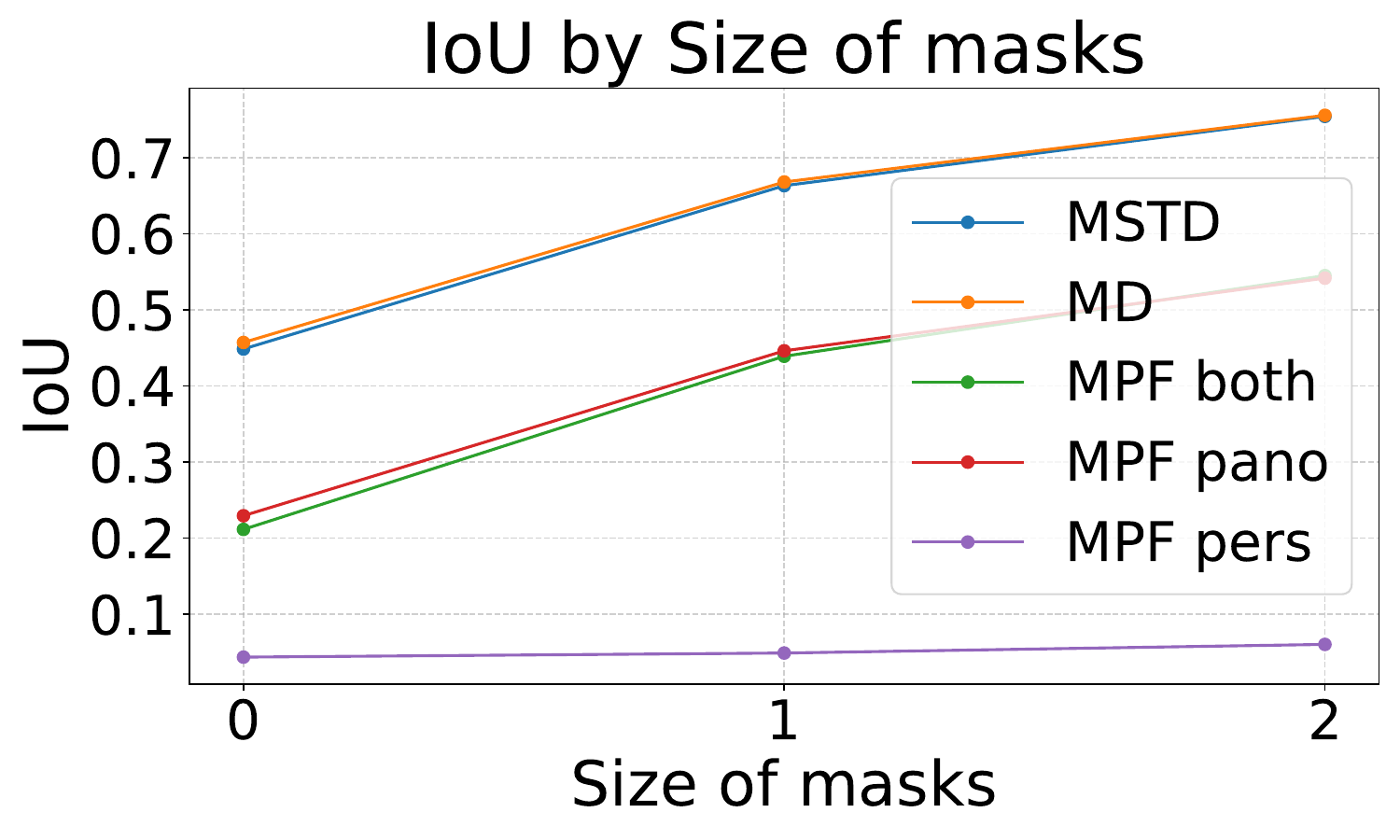}
    \includegraphics[width=0.48\linewidth]{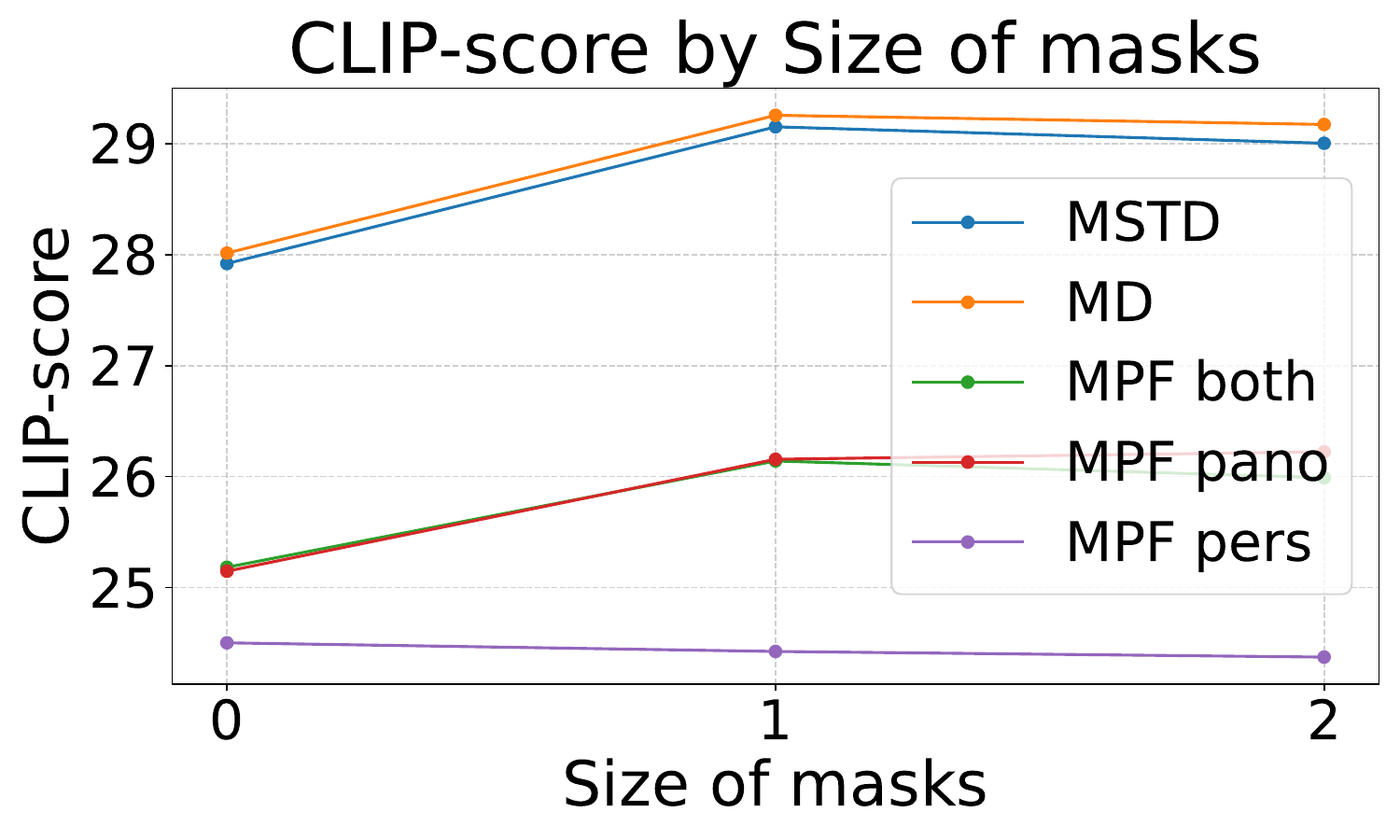}
    \hfill
    \includegraphics[width=0.48\linewidth]{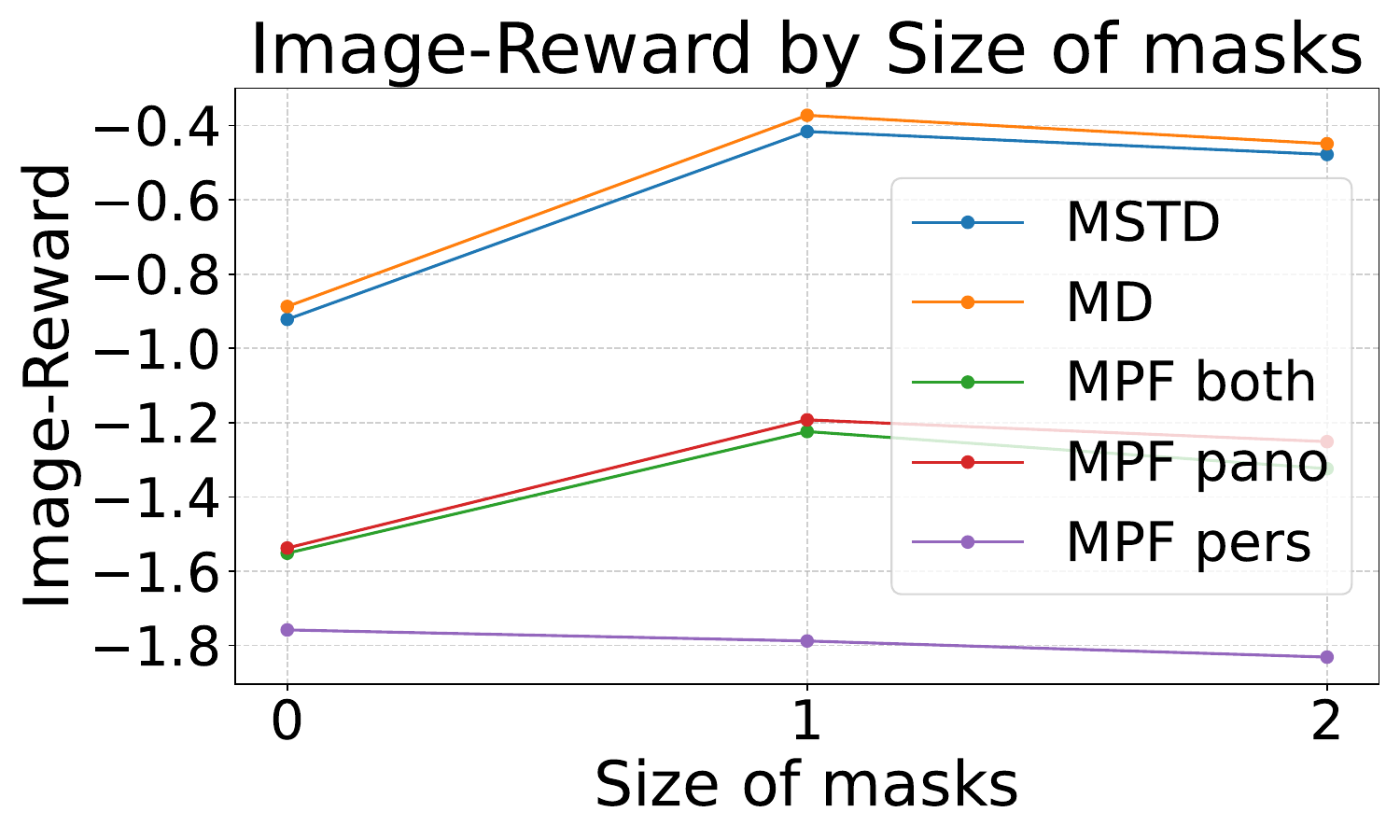}
    \includegraphics[width=0.48\linewidth]{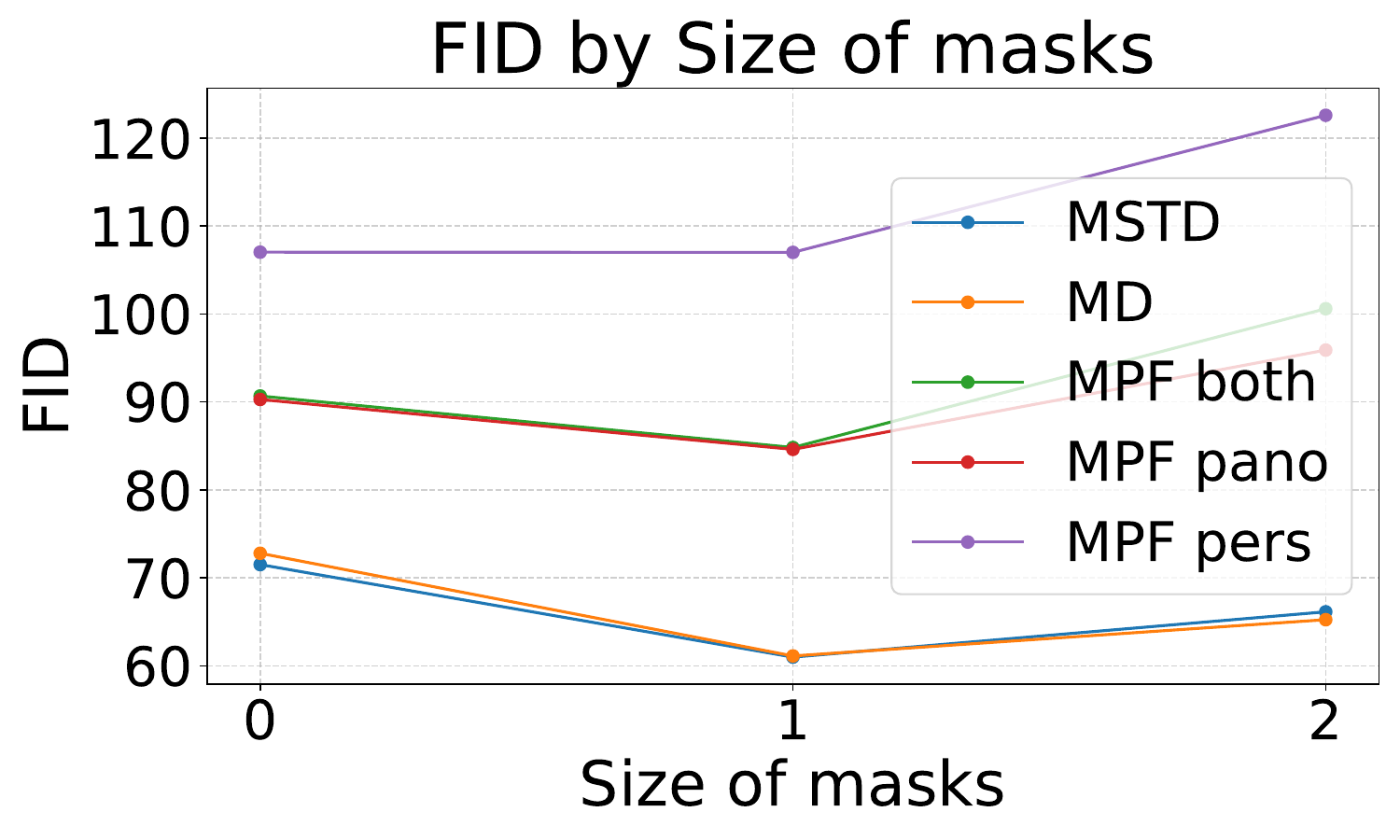}
    \includegraphics[width=0.48\linewidth]{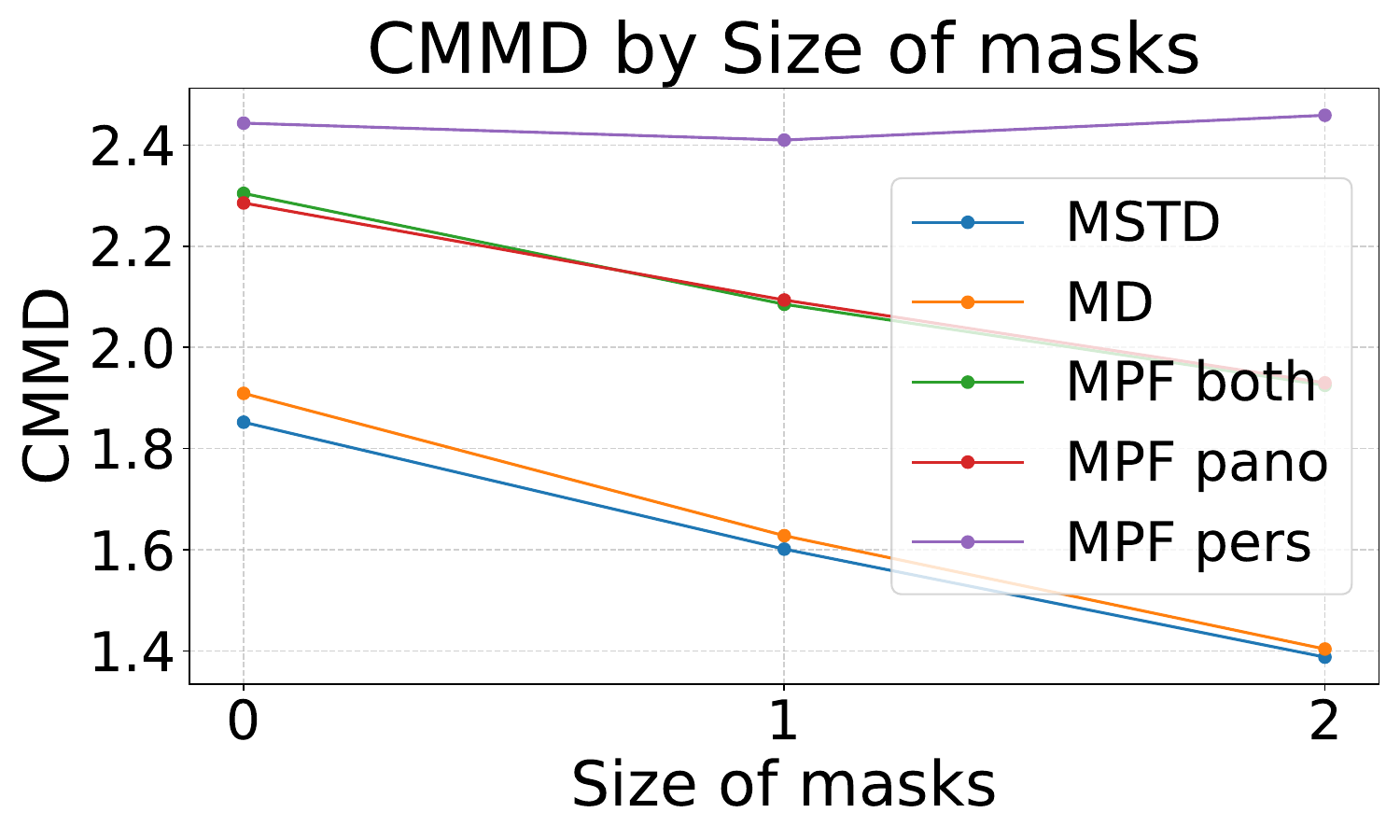}
    \caption{The influence of mask size on our metrics for every approach. The plots show a similar effect as with bootstrapping.}
    \label{fig:plots_masksize}
\end{figure}

\end{document}